%% file: main.tex
\pdfoutput=1
\documentclass[11pt,a4paper]{article}
\usepackage[margin=25mm]{geometry}
\usepackage[numbers,sort&compress]{natbib}
\usepackage[T1]{fontenc}
\usepackage[utf8]{inputenc}
\usepackage{amsmath,amssymb,amsthm,mathtools,bm}
\usepackage{graphicx}
\usepackage{pgfplots}
\pgfplotsset{compat=1.18}
\usepgfplotslibrary{groupplots}
\usepackage{flafter}
\usepackage{float}
\usepackage{placeins}
\usepackage{rotating}
\usepackage{algorithm}
\usepackage{algpseudocode}
\usepackage{booktabs,tabularx,array}
\usepackage[expansion=false]{microtype}
\usepackage{ragged2e,xurl}

\usepackage{enumitem}
\usepackage[colorlinks=true,linkcolor=blue,citecolor=blue,urlcolor=blue]{hyperref}
\setlist[itemize]{leftmargin=*,itemsep=2pt,topsep=4pt}
\newtheorem{proposition}{Proposition}
\newtheorem{theorem}{Theorem}
\newtheorem{definition}{Definition}
\newtheorem{remark}{Remark}
\newtheorem{corollary}{Corollary}
\newcommand{\R}{\mathbb R}
\newcommand{\E}{\mathbb E}
\newcommand{\cH}{\mathcal H}
\newcommand{\cK}{\mathcal K}
\newcommand{\Bred}{\mathcal B_{\mathrm{red}}}
\newcommand{\grp}{\mathsf G}
\newcommand{\Gram}{G_{\mathrm{Gram}}}
\DeclareMathOperator{\Span}{span}
\DeclareMathOperator{\ran}{ran}
\DeclareMathOperator{\diag}{diag}
\DeclareMathOperator{\clip}{clip}

\title{Hiding a Gap Does Not Close It:\\ The Price of Removing the Mean Group Contrast from Learned Models of Learning Dynamics}
\author{Sosth\`ene Bidojesussi Hounzinhin\textsuperscript{1}\thanks{Corresponding author. Email: \href{mailto:Sosthene.Hounzinhin@um6p.ma}{\nolinkurl{Sosthene.Hounzinhin@um6p.ma}}. ORCID: \href{https://orcid.org/0009-0008-8579-3236}{0009-0008-8579-3236}.}\\[3pt]
	Ahmed Ratnani\textsuperscript{1}\\[3pt]
	Abdeladim El Akri\textsuperscript{1}\\[6pt]
	\small \textsuperscript{1}Vanguard Center, Mohammed VI Polytechnic University (UM6P), Rabat, Morocco}
\date{}
\hypersetup{pdftitle={Hiding a Gap Does Not Close It: The Price of Removing the Mean Group Contrast from Learned Models of Learning Dynamics},pdfauthor={Sosthène Bidojesussi Hounzinhin; Ahmed Ratnani; Abdeladim El Akri}}

\begin{document}
	\maketitle

	\begin{abstract}
		Adaptive tutors use models learned from student data, and a common fairness remedy erases the group signal from these models. We turn the cost of this remedy into an audit tool. Building on mean-parity regression, we give the minimum extra error, relative to the best predictor, of removing the mean group contrast from a predictor that cannot use the group, \(p_0p_1\Delta^2/R^2\), and the exact cost of any correction built from an estimated group model, which, for the best predictor, is never below this minimum. A three-stage audit (fit on training units, announce the cost on validation units, test on independent units) estimates this cost before the correction is used. In a controlled study, the price can be large even when the state predicts the group poorly. On three real data sets, erasing the group from a learned model of learning dynamics predicts almost no gap where one is observed. In a selected knowledge-tracing network, erasing the hidden state leaves most of the gap; a correction of the prediction removes it at an extra Brier score of 0.021 on average. On Eedi, with the network selected on separate units, the audit accepted the announced cost in all five splits, the measured cost fell inside the announced interval in all five, and the gap reduction transferred to new classes in four; on new ASSISTments schools the audit abstains. In a tutoring simulator, the objective and the help budget matter more than the erasure, and floors on each group's knowledge separate narrowing a gap from levelling down. Hiding a gap in a model does not close it.
	\end{abstract}

	\noindent\textbf{Keywords:} adaptive tutoring; mean parity; levelling down; concept erasure; knowledge tracing; Koopman operator; extended dynamic mode decomposition; model predictive control; educational data.

	\section{Introduction}
	When an adaptive tutor decides which exercise a student sees next, it also shapes what that student will be able to learn tomorrow. Such tutors are built from historical learning data, and these data are not neutral: they record students who learned in unequal conditions, in schools with fewer resources or in poorer areas. A model learned from these data can carry the same differences into its recommendations. Fairness must therefore be part of the design, and because the decisions follow one another, the check cannot look at one prediction alone; it must look at whole learning paths \citep{liu2018delayed,damour2020fairness}.

	A popular remedy is to remove group information from what the model has learned \citep{zemel2013learning,bolukbasi2016man,ravfogel2020null,belrose2023leace}. In a model of how students change over time, one estimates the direction of the model's state that separates the two groups and removes it from the rule that moves the state forward, and also from the effect of the tutor's actions and from the constant term, so that it cannot come back at the next step. The reduced model then predicts a small \emph{gap}, the difference between the average outcomes of the two groups, but only because it was built to do so. This is not a hypothetical practice in student modelling. An audit of deep knowledge-tracing models on the Eedi data applies reweighting and adversarial debiasing to remove a socioeconomic bias \citep{minh2026accuracy}; grade-prediction models have been trained adversarially, or with race removed at prediction time, to narrow racial gaps \citep{jiang2021towards}; and fairness-aware cognitive diagnosis removes the paths through which sensitive attributes reach the estimate of a student's proficiency \citep{zhang2024path}. Our question is what such a removal costs, when this cost can be known in advance, and what should be done instead.

	\paragraph{Why equal predictions are a contestable goal for a tutor.} Predicting what a student needs and deciding how much help to give are two different tasks. A prediction is useful to a tutor when it describes the student as the student is, including a gap between groups when one exists, because this gap is what the help must address. If two groups do not perform equally, a model that predicts equal means is wrong, on average, for at least one of them. Fairness concerns the decision: who receives help, and whether any group loses. Removing the group contrast from the predictions is therefore not an obvious fairness goal for this function, even though it is what the methods above do.

	\paragraph{What is at stake in a tutor.} A learner model feeds three decisions of a tutor: whether a student receives help, which exercise comes next, and when a skill is considered mastered. Erasing a real gap from the predictions moves them in one direction for each group: on average, the students of the group behind are predicted to succeed more often than they do, and the others less often. A tutor that offers help when the predicted success is low then offers less help to the students behind; one that picks exercises at a target difficulty gives them exercises that are too hard; and one that stops practice when the predicted mastery passes a threshold stops it too early for them. The extra error per prediction can look small in an aggregate measure (an extra Brier score of about 0.02 in our knowledge-tracing study), but it is not random: it goes in one direction for each group. This is why we ask what the removal costs before it is used, and why we judge a fairness change by the decisions and the learning it produces, not by the predicted gap.

	\paragraph{What we find.} In our learning data, the direction that separates the groups is largely the direction of learning itself: the groups differ in their level of knowledge, so the signal that tells them apart is the signal that predicts what a student will do next. Removing it removes predictive information, and we can say exactly how much. Among all predictors that use only the state and whose predicted gap is removed, the smallest extra error is \(p_0p_1\Delta^2/R^2\), where \(p_0,p_1\) are the group shares, \(\Delta\) is the gap predicted by the best model and \(R^2\) is the share of the group's variance that the state explains (Proposition~\ref{prop:general-price}). Equivalently, it is the variance of the best prediction times its squared correlation with the group signal, so it never exceeds the cost of discarding the prediction (Corollary~\ref{cor:price-bound}). The structure of this minimum is known from mean-parity regression \citep{calders2013controlling,wei2023mean,petrulionyte2026directly}. Our contribution is to turn it into an audit tool: the exact cost of the corrections that one can actually build from an estimated group model, a way to estimate this cost before the correction is used, and a diagnostic that says when the data do not support the estimate.

	\paragraph{Contributions.}
	\begin{enumerate}[leftmargin=*,itemsep=3pt,topsep=4pt]
		\item \textbf{An audit tool for the cost of removing the mean group contrast.} The minimum extra error, in closed form and with its bound, is a special case of mean-parity regression (Proposition~\ref{prop:general-price}). We add the exact cost of any correction built from an estimated group score: it removes the gap whether or not the score is calibrated, and, for the best predictor, its quadratic part is never below the minimum (Proposition~\ref{prop:built}). A three-stage procedure, fit on training units, announce and diagnose on validation units, test on independent units, estimates this cost before the correction is used (Section~\ref{sec:price-estimation}). Its point estimate is the change of error measured on validation units; what the audit adds is a quadratic part that needs no outcomes and explains the cost, an interval that includes the refit of the group model, a diagnostic that abstains, and a check on independent units that the gap reduction transfers. A controlled study with a known truth (Section~\ref{sec:controlled}) shows that the price can be large even when the state predicts the group poorly, because it is set by the alignment of the prediction with the group signal; that the diagnostic abstains when there is nothing to remove and flags almost every case, in the scenarios studied, in which the group signal changes between training and test; and that a calibrated but misspecified group model makes the estimate conservative for the minimum.
		\item \textbf{Evidence on real data that erasure hides gaps without closing them.} On three data sets (ASSISTments, Eedi, OULAD), removing the group direction from a learned model of learning dynamics predicts almost no gap where one is observed, and rank-matched controls, including LEACE, show that the loss comes from the group direction (Section~\ref{sec:results-prediction}). In a selected knowledge-tracing network, erasing the hidden state leaves most of the gap in the predictions, because the output also uses the skill of the next item; at the same remaining gap, a correction of the output built from a group model has the lowest observed mean cost of the four methods tested (clearly lower than LEACE on the hidden state; not separable from the two others), With the network selected on units kept apart from those that announce the cost, the three-stage audit accepts the estimate in all five Eedi splits, the cost measured on new classes lies inside the announced interval in all five, with intervals one to two times as wide as the cost, and the gap reduction transfers in four; the audit abstains on every split with new ASSISTments schools, where the group model carries no signal (Section~\ref{sec:results-next-item}).
		\item \textbf{Decisions, as a secondary result.} In a tutoring simulator, with a reference controller that uses the known equations, the objective and the help budget matter more than the erasure, and floors on each group's knowledge are what separates narrowing a gap from levelling down; a simple top-up for the group behind does as well (Section~\ref{sec:results-policy}). We turn these results into an audit protocol with three levels, structure, prediction and policy, and practical recommendations (Section~\ref{sec:general-diagnostic-protocol}).
	\end{enumerate}
	Table~\ref{tab:claims} links each claim to its evidence and to its main limit.

	\begin{table}[t]
		\centering\small
		\caption{Each claim of the paper, the evidence for it, and its main limit.}
		\label{tab:claims}
		\begin{tabularx}{\linewidth}{@{}>{\RaggedRight\arraybackslash}p{3.6cm}>{\RaggedRight\arraybackslash}X>{\RaggedRight\arraybackslash}X@{}}
			\toprule
			Claim & Evidence & Main limit \\
			\midrule
			The minimum cost of removing the mean contrast has an explicit value and a bound, and, for the best predictor, a correction built from any group model costs at least this minimum & Proofs (Propositions~\ref{prop:general-price} and~\ref{prop:built}) & Population \(f^*\) and \(q\); for a learned predictor the cost of the correction can be below the minimum (Eq.~\eqref{eq:quad-min}); mean parity, not independence; fixed law of the states \\
			\addlinespace
			The cost of the built correction can be announced before use, and a diagnostic screens the estimate & Controlled study with a known truth (Section~\ref{sec:controlled}); on Eedi, with the network selected on separate units: accepted in five splits, measured cost in the announced interval in five, gap reduction transfers in four (Section~\ref{sec:results-next-item}) & Conservative for the minimum when the group model misses part of the group signal; approximate intervals: nominal 95\%, observed coverage 90--100\% with 1,000 bootstrap draws (both learners and sample sizes), so not a 95\% guarantee; on Eedi, five splits that share units, and intervals one to two times as wide as the cost; intervals conditional on the fitted predictor; thresholds are conventions \\
			\addlinespace
			Removing the group direction hides observed gaps and costs accuracy & Three data sets, cluster-bootstrap intervals; score delays, reweighting, held-out presentations (Section~\ref{sec:results-realdata}) & Performance proxies inflate error ratios (1.5--1.7 on weeks with a new score); retrospective school groups \\
			\addlinespace
			The loss comes from the group direction, and linear removal costs what the formula says & Rank-matched controls, LEACE, school-level shuffle; formula within 0--13\% on held-out students (Section~\ref{sec:results-rank-matched}) & Linear removal, one feature dictionary; an iterated forecast adds a cross term \\
			\addlinespace
			Erasing the hidden state of a selected knowledge-tracing network does not remove the gap from its predictions; at equal remaining gap, the built output correction has the lowest observed mean cost & Four methods, five splits, paired cluster intervals; intensity also set on validation (Section~\ref{sec:results-next-item}) & Clearly cheaper only than LEACE on the hidden state; the audit abstains on new ASSISTments schools; for the retrained network the comparison is an empirical reference \\
			\addlinespace
			In control, the objective and the help budget matter more than the erasure & Reference that uses the known equations, approximately equal budgets (Section~\ref{sec:results-oracle}) & Noise-free plans by grid search; one simulator family; learned models stay below the reference at \(\gamma=2\) \\
			\addlinespace
			Floors avoid levelling down; extra help to the group behind narrows the knowledge gap & Factorial comparison at equal model and budget (Table~\ref{tab:knowledge-controller}) & Simulator only; floors protect predicted group means one step ahead \\
			\bottomrule
		\end{tabularx}
	\end{table}

	\paragraph{Scope.}
	We do not propose a new definition of fairness: no result in this paper proves that a system is fair. We do not evaluate a tutor used by real students. The simulator tests outcomes under a policy; the three observational studies test predictions of performance proxies and of next-item correctness. The ASSISTments groups are school-performance groups built after the fact, so we read that study as a test of the mechanism; Eedi and OULAD give attributes recorded outside the data. An exploratory analysis of 30 randomized experiments, which tests a different decision (choosing between two existing conditions), is reported as a complementary illustration in Appendix~\ref{app:rct}. No experiment uses a language model.

	\paragraph{Position in the literature.} Fairness over time has been studied with known models of the process or at one level only, prediction or policy \citep{liu2018delayed,damour2020fairness,wen2021algorithms,rateike2024designing}; in education, equally accurate predictions can still produce inequitable tutoring policies \citep{tschiatschek2022equity,doroudi2019fairer}, and a socioeconomic bias of knowledge-tracing models on the Eedi data resists reweighting and adversarial debiasing \citep{minh2026accuracy}. Concept erasure acts on a static representation \citep{ravfogel2020null,belrose2023leace}; in a model of dynamics the representation is pushed forward at every step, so we remove a subspace built from the group contrast and from the update rule. Mean-parity regression \citep{calders2013controlling,wei2023mean,petrulionyte2026directly} and full demographic parity \citep{chzhen2020fair,legouic2020projection} give the structure of our price; Table~\ref{tab:mean-parity} states the difference precisely. Floors on group outcomes \citep{mittelstadt2023unfairness,gultchin2023pragmatic} are an answer to levelling down \citep{zietlow2022leveling,maheshwari2023fair}; Proposition~\ref{prop:no-levelling} says, for one step, when they still allow a gap to narrow. Appendix~\ref{app:related} gives the full comparison.

	\section{Problem setup and reference laws}
	\label{sec:problem}
	\subsection{The physical system and the target gap}
	Let \(x_t\in\mathcal X\) denote the state of a tutoring session at step \(t\), and let \(u_t\) denote a control action, such as the choice of a hint. We represent group membership by a binary label \(\grp\in\{0,1\}\), fixed throughout the session. The label need not be a sensitive attribute: our mathematical results apply to any binary grouping, and we use \(S\) only when the attribute has an external sensitive-status justification. This distinction matters for ASSISTments, where \(\grp_{\mathrm{school}}\) identifies schools by their own performance and has no demographic interpretation. (The group label \(\grp\) is not the Gram matrix of the features, which we write \(\Gram\) in Section~\ref{sec:method}.)

	A transition kernel \(\mathsf T(dy\mid x,s,u)\) describes how the state changes in response to the current state, group and action. Under a fixed feedback policy \(u_t=\pi(\Psi(x_t))\), the Markov state must include group membership whenever it directly affects the transition. The resulting closed-loop Koopman operator is
	\begin{equation}
		(\cK^\pi f)(x,s)=\int f(y,s)\,\mathsf T(dy\mid x,s,\pi(\Psi(x))).
		\label{eq:closedloop}
	\end{equation}
	If the policy or environment changes over time, the state must also include a clock. The feature map \(\Psi\) need not encode the full Markov state; when it does not, the fitted model remains a predictor over the observed features, but it cannot be interpreted as a complete model of the underlying process.
	
	For a measurable final outcome \(D_T\), a joint law \(\nu\) and a fixed horizon \(T\), the gap in the population is
	\begin{equation}
		\Delta_D(\pi,\nu,T)=\E_{\pi,\nu}[D_T\mid \grp=1]-\E_{\pi,\nu}[D_T\mid \grp=0].
		\label{eq:population}
	\end{equation}
	We require both groups to have positive probability. This population gap is a difference in group means: a zero value indicates equal average outcomes, not identical outcome distributions. For a finite cohort \(b\) containing both groups, the corresponding observed gap is
	\begin{equation}
		\delta_b=\frac{1}{n_{1,b}}\sum_{i:\grp_{b,i}=1}D_{T,b,i}
		-\frac{1}{n_{0,b}}\sum_{i:\grp_{b,i}=0}D_{T,b,i}.
		\label{eq:cohort}
	\end{equation}
	Throughout, we distinguish the empirical cohort gap \(\delta_b\), the population quantity \(\Delta_D\), and the model-predicted gap \(\widehat\delta_b\); agreement between the latter two is an empirical question, not an identity.
	
	\begin{definition}[Levelling down]
		\label{def:levelling-down}
		A change of policy \emph{levels down} if it narrows the absolute gap \(|\delta_T|\) while lowering the mean outcome of at least one group. In this paper, the outcome protected by the floors is mean knowledge: a narrower gap obtained while one group's mean knowledge falls counts as levelling down.
	\end{definition}

	\subsection{Group contrasts}
	To connect group gaps to the operator representation, fix a joint law \(\nu\) of \((X,\grp)\), let \(p_s=\Pr_\nu(\grp=s)>0\), and write \(\mu_\nu=\mathcal L_\nu(X)\). Define
	\begin{equation}
		Y_{\grp}^\nu=\frac{\grp-p_1}{p_0p_1},\qquad
		q_{\grp}^\nu(x)=\E_\nu[Y_{\grp}^\nu\mid X=x].
		\label{eq:score}
	\end{equation}
	Both \(Y_{\grp}^\nu\) and its conditional expectation \(q_{\grp}^\nu\) have mean zero. Conditional Jensen's inequality gives
	\begin{equation}
		\|q_{\grp}^\nu\|_{L^2(\mu_\nu)}^2\leq\E[(Y_{\grp}^\nu)^2]=\frac{1}{p_0p_1}.
	\end{equation}
	Consequently, for any square-integrable feature \(f\), the difference between its group averages is \(\langle q_{\grp}^\nu,f\rangle_{\mu_\nu}\). Thus the conditional contrast \(q_{\grp}^\nu\) represents each group gap as an inner product. If \(X\) and \(\grp\) are independent, this conditional contrast is zero even though the label score itself is not.
	
	The contrast depends on the state distribution under which it is estimated. In particular, the initial contrast generally differs from the contrast under the state law pooled over an exploratory trajectory of length \(H\):
	\begin{equation}
		\nu_{\mathrm{occ}}=\frac1H\sum_{t=0}^{H-1}\mathcal L(x_t,\grp).
		\label{eq:occupation}
	\end{equation}
	A contrast fitted under \(\nu_{\mathrm{occ}}\) therefore describes group differences under exploratory control, not necessarily at time zero or under the eventual deployment policy. Pooling observations over time also does not make observations from the same trajectory independent.
	
	When group membership is included in the state, the initial contrast has the simple form \(q_0(x,s)=(s-p_1)/(p_0p_1)\). The tower property then expresses the population gap after \(T\) steps as
	\begin{equation}
		\Delta_D(\pi,\nu,T)=\langle q_0,(\cK^\pi)^T f_D\rangle_{\nu},
		\label{eq:physical-identity}
	\end{equation}
	where \(f_D\) maps a state to the outcome of interest. This identity relates the population gap to the true closed-loop operator; it does not identify that operator with a fitted finite-dimensional matrix. It also exposes a substantive tension in group-direction removal: because membership is fixed, \(\cK^\pi q_0=q_0\). Removing this direction from a model can therefore suppress a genuine invariant feature of the data-generating process, rather than merely eliminating an irrelevant nuisance.
	
	Appendix~\ref{app:function-space} states the function space and the conditions under which the operator statements hold.

	\section{Metric-aware operator compression}
	\label{sec:method}
	\subsection{Coefficient operators and feature values}
	Let \(\Psi(x)\in\R^N\) be a real-valued feature dictionary and write \(f_c(x)=\Psi(x)^\top c\) for the function with coefficient vector \(c\). The distinction between a feature value and its coefficient representation is central to the projection used here. To make this distinction explicit, begin with an uncontrolled regression without an intercept. Given \(M\) state pairs, set \(Z=[\Psi(x_1),\ldots,\Psi(x_M)]\) and \(Z_+=[\Psi(y_1),\ldots,\Psi(y_M)]\). The least-squares value propagator and its dual coefficient operator are
	\begin{equation}
		A_0=Z_+Z^\dagger,\qquad
		\Gram=\frac{ZZ^\top}{M},\qquad
		C_+=\frac{ZZ_+^\top}{M},\qquad
		K_0=\Gram^\dagger C_+=A_0^\top.
		\label{eq:duality}
	\end{equation}
	Here \(\dagger\) denotes the Moore--Penrose pseudoinverse. For this regression, \(K_0=A_0^\top\) exactly, since \((ZZ^\top)^\dagger Z=(Z^\dagger)^\top\). The equality provides the bridge between the two representations: \(A_0\) propagates feature values, while \(K_0\) propagates function coefficients. Although related by transposition, these matrices act on different objects and cannot be substituted for one another.
	
	The experiments include both controls and an intercept, so they use a joint affine regression rather than the simpler fit above. With \(U=[u_1,\ldots,u_M]\), define
	\begin{equation}
		X=\begin{bmatrix}Z\\ U\\ \mathbf 1^\top\end{bmatrix},
		\qquad [A\;B\;a]=Z_+X^\dagger,
		\qquad K:=A^\top.
		\label{eq:controlled-fit}
	\end{equation}
	This joint fit fixes the operator used throughout the controlled experiments: it is the transpose of the state block \(A\), and it is this \(K\) that generates the cyclic subspace, projector and structural residuals. The moment formula \(\Gram^\dagger C_+\) from Eq.~\eqref{eq:duality} is not generally equivalent when controls and an intercept are included, because the state coefficients are estimated jointly with those terms. For ASSISTments, which has no control input, we retain the same convention and include only the intercept in \(X\). Consequently, the operator used to construct \(P\) is exactly the one corresponding to the state dynamics later projected and evaluated.
	
	The projection geometry is defined on coefficient space by a symmetric positive-definite metric \(\Gamma\). Its associated weighted adjoint and matrix norm are
	\begin{equation}
		K^{*\Gamma}=\Gamma^{-1}K^\top\Gamma,\qquad
		\|L\|_{\Gamma,F}=\|\Gamma^{1/2}L\Gamma^{-1/2}\|_F,
		\label{eq:geometry}
	\end{equation}
	and the spectral norm \(\|L\|_{\Gamma,2}\) is defined analogously. In the implementation, \(\Gamma=\Gram+\eta I\), where \(\eta=10^{-3}\lambda_{\max}(\Gram)\). This ridge regularizes the projection metric; it does not alter the fitted dynamics themselves.
	The relative ridge \(10^{-3}\) is a numerical default, not a value selected by cross-validation; Appendix~\ref{app:construction} bounds the condition number it implies, and Appendix~\ref{sec:results-ridge} tests its effect.

	\subsection{Numerical selection of the removed directions}
	Ideally, one would remove the smallest subspace that contains the group contrast and that the dynamics and their adjoint leave invariant; Appendix~\ref{sec:theory} defines this exact closure.
	This exact closure can be too large to use as a reduced model, so the implementation constructs a rank-limited approximation from the pooled sample of current states. After standardizing the features, we arrange feature vectors as columns of \(Z_c\in\R^{N\times M}\) and collect group scores in \(\mathbf y\in\R^M\). The fitted contrast coefficients are
	\begin{equation}
		\widehat c_{\grp}^{\mathrm{occ}}=(Z_c^\top)^\dagger\mathbf y\in\R^N.
	\end{equation}
	Using the same observations for feature centering and contrast regression makes the fitted contrast mean zero up to numerical precision. We then work in whitened coordinates, \(W=\Gamma^{1/2}K\Gamma^{-1/2}\) and \(b=\Gamma^{1/2}\widehat c_{\grp}^{\mathrm{occ}}\), where the weighted adjoint becomes an ordinary transpose. Candidate directions are generated by applying \(W\) and \(W^\top\) repeatedly to \(b\). The contrast direction itself is always retained in the candidate basis; remaining directions are selected from the leading singular vectors, subject to the requested rank. This procedure approximates the exact closure while keeping the model finite-dimensional.
	
	Let \(Q_r\) denote the resulting \(\Gamma\)-orthonormal basis, so \(Q_r^\top\Gamma Q_r=I_r\). The projection onto the selected contrast subspace and its complement are
	\begin{equation}
		R=Q_rQ_r^\top\Gamma,\qquad P=I-R,\qquad F=PKP.
		\label{eq:projector}
	\end{equation}
	These projectors define the compressed operator \(F=PKP\), but a chosen rank alone does not establish that the selected space is truly reducing. We therefore report both implementation checks and the residual mixing between the two subspaces. The structural table in Appendix~\ref{app:structural} summarizes requested and numerical ranks, closure depth, singular-value cutoff, seed residual, duality and idempotency errors for 650 simulator fits and 19 ASSISTments fits, and the appendix also reports the Eedi, OULAD and group-aware fits; it also reports leakage through the control and intercept terms. Automated tests fail if an identity exceeds \(10^{-9}\). The distance from exact invariance is measured by
	\begin{equation}
		r_1=\|PKR\|_{\Gamma,F},\qquad r_2=\|RKP\|_{\Gamma,F},\qquad \epsilon_q=\|P\widehat c_{\grp}^{\mathrm{occ}}\|_\Gamma.
		\label{eq:diagnostics}
	\end{equation}
	Here \(r_1\) measures how strongly the fitted dynamics send vectors from the removed subspace into the retained one, and \(r_2\) measures the reverse transfer. They are cross-block residuals: both are zero exactly when the two subspaces do not mix under \(K\). Thus, smaller values indicate a closer approximation to a reducing subspace, not better prediction by themselves. These operator-specific residuals are diagnostics of the chosen compression, not standard fairness metrics. The seed residual \(\epsilon_q\) measures the part of the fitted group contrast left in the retained space after projection; it should be near zero if that contrast was removed.

	\subsection{Affine dynamics and control}
	The joint regression in Eq.~\eqref{eq:controlled-fit} yields the controlled feature dynamics. To compare model forecasts with physical outcomes, we also fit an affine decoder on the same training split:
	\begin{equation}
		z^+=Az+Bu+a,\qquad \widehat x=Cz+c.
		\label{eq:affine}
	\end{equation}
	The coefficient-space projector \(P\) induces a dual projector \(P_v=P^\top\) in value space. This is orthogonal under the dual metric \(\Gamma^{-1}\), although generally not under \(\Gamma\). Consistency therefore requires projecting every component of the affine update with this dual operator:
	\begin{equation}
		A_P=P^\top AP^\top,\qquad B_P=P^\top B,\qquad a_P=P^\top a.
		\label{eq:affine-projected}
	\end{equation}
	Projecting only the state matrix would leave the removed direction able to re-enter through either the control response or the intercept. Applying the same dual projection to all three terms prevents this leakage by construction.

	Appendix~\ref{app:construction} gives the augmented form of this projection and explains why the decoder is not projected again. The compressed model is always used from the newly observed state (Algorithm~\ref{alg:tutoring-inference} in Appendix~\ref{sec:theory}): with \(u_t\equiv0\) this gives the observational replay of the real data, and with an MPC policy the simulator controller of Section~\ref{sec:protocol}. Leaving \(B\) or \(a\) unprojected would let the removed direction come back, which the structural audit of Table~\ref{tab:protocol} checks.

	\paragraph{What the construction guarantees, and what it does not.} For a fixed projector, \(F=PKP\) is the closest operator to \(K\) that keeps only the selected subspace, its error splits into the three cross blocks \(RKR\), \(PKR\) and \(RKP\), and the error of a \(T\)-step forecast is bounded by these blocks times an amplification factor; the compressed fits also converge when the fitted operators and projectors converge. Appendix~\ref{sec:theory} states these results with their proofs, together with the terms that a guarantee for a real population would need and the two different limits of EDMD. They are standard tools that make the assumptions explicit; none of them says anything about outcomes for students. We have not compared the computed bounds with observed forecast errors, so they remain theoretical properties, not validated indicators.

	\section{The price of erasure}
	\label{sec:price}
	An erased model must still predict. This section gives the smallest extra error that any predictor pays when it uses only the state and is required to remove all or part of the gap it predicts between groups. Let \(z\) be the information that the predictor may use (here the state of a student, or its features), \(y\) a target with \(\E y^2<\infty\), and \(\grp\in\{0,1\}\) a group with shares \(p_0,p_1>0\). Write \(f^*(z)=\E[y\mid z]\) for the best predictor and \(q(z)=\Pr(\grp=1\mid z)\) for the probability of the group given the state. For any predictor \(f\) of \(z\), its \emph{predicted gap} is \(\Delta(f)=\E[f\mid\grp=1]-\E[f\mid\grp=0]\). Put \(\Delta^*=\Delta(f^*)\) and
	\begin{equation}
		R^2_{\grp}=\frac{\operatorname{Var}(q)}{p_0p_1}\in[0,1],
	\end{equation}
	the share of the variance of the group that the state explains, linearly or not.

	\begin{proposition}[The price of erasure]
		\label{prop:general-price}
		Assume \(\operatorname{Var}(q)>0\). For \(\tau\in\mathbb R\), among all square-integrable predictors \(f(z)\) whose predicted gap is \(\tau\Delta^*\), the smallest extra mean squared error is
		\begin{equation}
			\min_f\;\operatorname{MSE}(f)-\operatorname{MSE}(f^*)=(1-\tau)^2\,\frac{p_0p_1\,(\Delta^*)^2}{R^2_{\grp}},
			\label{eq:general-price}
		\end{equation}
		reached by \(f_\tau=f^*-(1-\tau)\,\dfrac{\operatorname{Cov}(f^*,q)}{\operatorname{Var}(q)}\,(q-p_1)\).
	\end{proposition}
	\begin{proof}
		For any \(f\), \(\operatorname{Cov}(f,\grp)=\E[f(z)(\grp-p_1)]=\E[f(z)(q(z)-p_1)]=\operatorname{Cov}(f,q)\), and the predicted gap is \(\operatorname{Cov}(f,\grp)/(p_0p_1)\). So the constraint is one linear condition on \(f-f^*\) in the space of square-integrable functions of \(z\): \(\langle f-f^*,q-p_1\rangle=-(1-\tau)\operatorname{Cov}(f^*,q)\). Since \(y-f^*\) is orthogonal to every function of \(z\), \(\operatorname{MSE}(f)=\operatorname{MSE}(f^*)+\|f-f^*\|^2\). The smallest norm under one linear condition is reached by a multiple of \(q-p_1\), and it equals \((1-\tau)^2\operatorname{Cov}(f^*,q)^2/\operatorname{Var}(q)\). Finally, \(\operatorname{Cov}(f^*,q)=p_0p_1\Delta^*\) and \(\operatorname{Var}(q)=p_0p_1R^2_{\grp}\).
	\end{proof}

	\begin{corollary}[Two forms of the price, and a bound]
		\label{cor:price-bound}
		The price of Eq.~\eqref{eq:general-price} has the two equivalent forms
		\begin{equation}
			\frac{p_0p_1\,(\Delta^*)^2}{R^2_{\grp}}=\frac{\operatorname{Cov}(f^*,q)^2}{\operatorname{Var}(q)}=\operatorname{Var}(f^*)\,\operatorname{corr}(f^*,q)^2,
			\label{eq:price-forms}
		\end{equation}
		so the extra error is at most \((1-\tau)^2\operatorname{Var}(f^*)\). For \(\tau=0\), erasing the gap never costs more than replacing the best prediction by its mean, which also has a zero predicted gap; the two costs are equal exactly when \(f^*\) is an affine function of \(q\). The cost is quadratic in \(1-\tau\): halving the predicted gap costs a quarter of the full price.
	\end{corollary}
	\begin{proof}
		The first equality uses \(\operatorname{Cov}(f^*,q)=p_0p_1\Delta^*\) and \(\operatorname{Var}(q)=p_0p_1R^2_{\grp}\); the second is the definition of the correlation. The bound follows from \(\operatorname{corr}^2\le1\), with equality exactly when \(f^*\) is affine in \(q\); the constant predictor \(\E y\) has a zero predicted gap and an extra error of \(\operatorname{Var}(f^*)\).
	\end{proof}

	\paragraph{How to read the price.} The price is large when the best prediction is strongly aligned with the group signal; equivalently, when the predicted gap is large relative to how well the state explains the group. The two forms describe the same quantity, and their parts cannot be varied one at a time: as \(R^2_{\grp}\to0\), \(q\) becomes constant and the predicted gap \(\Delta^*\) goes to zero as well. The price does not depend on how the erasure is done; better engineering cannot lower it. It can be estimated before any intervention, from a standard model of the target and a model of the group. For a learner model, the price is the extra error that any model of the student's state must accept to predict equal mean success for the two groups; when the groups differ in what they know, it is large, and it is paid mostly as an overestimate of the students behind.

	\paragraph{Degenerate cases.} If \(\operatorname{Var}(q)=0\), the state carries no information about the group: every predictor of the state has a zero predicted gap, \(\Delta^*=0\), and nothing has to be removed. If \(\operatorname{Var}(f^*)=0\), the best predictor is constant, \(\Delta^*=0\) and the price is zero; the correlation form is then undefined, but the first form and the bound still hold.

	\paragraph{What the constraint means.} The constraint is mean parity, \(\E[f\mid\grp=1]=\E[f\mid\grp=0]\). It does not make \(f(z)\) independent of the group: two groups can have equal mean predictions and different distributions of predictions. When we speak of erasing or removing the group signal, we mean removing this mean contrast. The price is also computed under the law of the states that generated the data; a policy that changes this law changes the price.

	\paragraph{Theory and estimates.} Proposition~\ref{prop:general-price} concerns the population objects \(f^*=\E[y\mid z]\) and \(q=\Pr(\grp=1\mid z)\). In our experiments both are estimated. For linear predictors on the training data the identity of Proposition~\ref{prop:erasure-price} holds exactly, by construction. Elsewhere, for held-out students, nonlinear models or a neural network, a close match between the estimated formula and the measured cost shows that the two are compatible; it does not test the assumptions of the proposition. For any predictor \(f\) and change \(d\),
	\begin{equation}
		\operatorname{MSE}(f+d)-\operatorname{MSE}(f)=\E[d^2]+2\,\E[(f-y)\,d],
		\label{eq:cross-term}
	\end{equation}
	and the second term vanishes only when \(f\) is optimal in a space that contains \(d\). We report this term wherever the predictor is not the best one.

	\begin{proposition}[The linear case]
		\label{prop:erasure-price}
		Let \(z\) be centered features with covariance \(S\), possibly singular (in our dictionary \(p_t\) is binary, so \(p_t^2=p_t\)), and write \(S^+\) for its Moore--Penrose inverse. Let \(\hat y=\beta^\top z\) with \(\beta=S^{+}\operatorname{Cov}(z,y)\) be the best linear predictor of \(y\), and \(\hat g=\gamma^\top z\) with \(\gamma=S^{+}\operatorname{Cov}(z,\grp)\) the best linear predictor of the group, and write \(\rho=\operatorname{corr}(\hat y,\hat g)\). Assume \(\operatorname{Var}(\hat g)>0\) and \(\operatorname{Var}(\hat y)>0\). For \(\tau\in[0,1]\), among the linear predictors \(w^\top z\) whose predicted gap is \(\tau\) times the predicted gap \(\hat\Delta\) of \(\hat y\), the smallest mean squared error is
		\begin{equation}
			\operatorname{MSE}(\hat y)+(1-\tau)^2\operatorname{Var}(\hat y)\,\rho^2=\operatorname{MSE}(\hat y)+(1-\tau)^2\,\frac{p_0p_1\hat\Delta^2}{R^2},
			\label{eq:erasure-price}
		\end{equation}
		with \(R^2=\operatorname{Var}(\hat g)/(p_0p_1)\), reached by \(w_\tau=\beta-(1-\tau)\,\gamma\,(\beta^\top S\gamma)/(\gamma^\top S\gamma)\). For \(\tau=0\), these are exactly the linear predictors of LEACE-erased features \citep{belrose2023leace}.
	\end{proposition}
	\begin{proof}
		The predicted gap of \(w^\top z\) is \(\operatorname{Cov}(w^\top z,\grp)/(p_0p_1)=w^\top S\gamma/(p_0p_1)\), so the condition is linear: \(w^\top S\gamma=\tau\,\beta^\top S\gamma\). Because the residual \(y-\hat y\) is uncorrelated with \(z\), \(\operatorname{MSE}(w)=\operatorname{MSE}(\hat y)+(w-\beta)^\top S(w-\beta)\); directions in the null space of \(S\) change neither the predictions nor the error, so the minimization can be done in the range of \(S\), where \(S^+\) is an inverse. In the inner product of \(S\), the smallest change that meets the condition is a multiple of \(\gamma\), and its squared size is \((1-\tau)^2(\beta^\top S\gamma)^2/(\gamma^\top S\gamma)=(1-\tau)^2\operatorname{Var}(\hat y)\rho^2\); the second form follows as in Corollary~\ref{cor:price-bound}. For \(\tau=0\), LEACE-erased features \(Ez\) satisfy \(\operatorname{Cov}(Ez,\grp)=0\), so every linear function of them has a zero predicted gap; conversely every \(w\) with \(w^\top S\gamma=0\) satisfies \(w^\top Ez=w^\top z\).
	\end{proof}
	Proposition~\ref{prop:erasure-price} is Proposition~\ref{prop:general-price} with \(f\) and \(q\) restricted to linear functions. The frontier in \(\tau\) is the analytic version of the soft projections \(P_\alpha\) of Appendix~\ref{sec:results-assist}.

	\paragraph{The price over a forecast horizon.} With target \(k_{t+h}\) and information \(x_t\), Proposition~\ref{prop:general-price} applies to the best predictor at each horizon, \(\E[k_{t+h}\mid x_t]\), and Proposition~\ref{prop:erasure-price} to the best linear predictor of \(k_{t+h}\) fitted directly for that horizon. Their prices are
	\begin{equation}
		\frac{p_0p_1(\Delta^*_h)^2}{R^2_{\grp}}\qquad\text{and}\qquad\frac{p_0p_1\Delta_h^2}{R^2},
		\label{eq:horizon-price}
	\end{equation}
	where \(\Delta^*_h\) and \(\Delta_h\) are the gaps that these predictors give \(h\) steps ahead. The share \(R^2_{\grp}\) (or \(R^2\)) depends on the law of the information, so it is the same for every horizon only if the starting points are the same; we therefore compare horizons on a common set of starting points. A forecast made by iterating a one-step model \(h\) times is not, in general, the best predictor of \(k_{t+h}\), so Eq.~\eqref{eq:horizon-price} does not give the cost of removing the group from it: by Eq.~\eqref{eq:cross-term} its cost includes the term \(2\E[(f-y)d]\), which Section~\ref{sec:results-rank-matched} measures.

	\paragraph{Relation to mean-parity regression.} The structure of this result is known, and Table~\ref{tab:mean-parity} places it. \citet{calders2013controlling} fit linear regressions under constraints on group mean differences, so the linear case is close to their work. \citet{wei2023mean} study regression under mean parity in a reproducing kernel Hilbert space with a discrete sensitive attribute; their regressor takes the attribute as an input, they build the space of admissible functions by a projection, give the optimal fair regressor in closed form, and show that relaxing the constraint by a factor \(\alpha\) changes the risk quadratically, as \(\tau\) does here, with an upper bound on the cost. \citet{petrulionyte2026directly} consider, as we do, a predictor of the non-sensitive features only, without access to the attribute at prediction time; they allow continuous and multidimensional attributes, penalize the variance of the conditional mean prediction given the attribute, and train with a functional bilevel algorithm, without a closed form for the cost. For full demographic parity, \citet{chzhen2020fair} and \citet{legouic2020projection} characterize the best regressor, which may use the attribute, through Wasserstein barycentres. Proposition~\ref{prop:general-price} is the special case of a binary group and an exact constraint on the mean gap, for a predictor of the state alone. In this case the constraint is a single orthogonality condition to \(q-p_1=\E[\grp-p_1\mid z]\), so the population optimum and its cost have a closed form. What we add is narrow: the explicit minimum cost, in two forms that can be estimated from a standard model and a group model, and its bound by \(\operatorname{Var}(f^*)\); the exact cost of a correction built from any group score (Proposition~\ref{prop:built}, Section~\ref{sec:price-estimation}), which needs no calibrated group model and, for the best predictor, is never below the minimum; and a three-stage audit that estimates this cost before the correction is used, applied to learned models of learning. Section~\ref{sec:results-rank-matched} compares it with measured costs on three real data sets, and Section~\ref{sec:results-next-item} with a knowledge-tracing network.

	\begin{table}[t]
		\centering\small
		\caption{Work on mean parity and related constraints in regression.}
		\label{tab:mean-parity}
		\begin{tabularx}{\linewidth}{@{}>{\RaggedRight\arraybackslash}p{2.6cm}>{\RaggedRight\arraybackslash}X>{\RaggedRight\arraybackslash}p{2.0cm}>{\RaggedRight\arraybackslash}X>{\RaggedRight\arraybackslash}X@{}}
			\toprule
			Work & Predictor & Attribute & Constraint & What is given\\
			\midrule
			\citet{calders2013controlling} & linear & discrete & mean differences & fitted linear models\\
			\addlinespace
			\citet{wei2023mean} & RKHS, uses the attribute & discrete & exact mean parity, and a relaxation & closed-form regressor; quadratic trade-off; upper bound on the cost\\
			\addlinespace
			\citet{chzhen2020fair}; \citet{legouic2020projection} & any, uses the attribute & discrete & full demographic parity & optimal regressor (Wasserstein barycentre)\\
			\addlinespace
			\citet{petrulionyte2026directly} & neural, features only & continuous, multidimensional & penalty on the variance of the conditional mean & training algorithm (functional bilevel)\\
			\addlinespace
			This paper & any function of the state & binary & exact fraction \(\tau\) of the mean gap & minimum cost in closed form, two forms, bound; exact cost of a correction built from any group score (Prop.~\ref{prop:built}); three-stage audit of learned dynamic models\\
			\bottomrule
		\end{tabularx}
	\end{table}

	\subsection{From the minimum price to a correction one can build}
	\label{sec:price-estimation}
	In practice \(q\) is unknown, and any correction must be built from an estimated group score. The next result gives the exact cost of such a correction.

	\begin{proposition}[The price of a correction built from a group score]
		\label{prop:built}
		Let \(s(z)\) be a square-integrable score with \(\operatorname{Cov}(s,\grp)\neq0\), let \(f\) be a predictor of \(z\) and \(\tau\in[0,1]\), and set
		\[
		f_s=f-(1-\tau)\,c_s\,(s-\E s),\qquad c_s=\frac{\operatorname{Cov}(f,\grp)}{\operatorname{Cov}(s,\grp)}.
		\]
		Then the predicted gap of \(f_s\) is \(\tau\,\Delta(f)\), whether or not \(s\) is calibrated. If \(f=f^*\), the extra error of \(f_s\) is
		\begin{equation}
			\operatorname{MSE}(f_s)-\operatorname{MSE}(f^*)=(1-\tau)^2\,\frac{p_0p_1(\Delta^*)^2}{\rho_s^2},\qquad \rho_s^2=\operatorname{corr}(s,\grp)^2\le R^2_{\grp},
			\label{eq:built-price}
		\end{equation}
		with equality if and only if \(s\) is an affine function of \(q\). When \(f=f^*\), the cost of the built correction is therefore at least the minimum of Proposition~\ref{prop:general-price}, and the two differ by the factor \(R^2_{\grp}/\rho_s^2\ge1\).
	\end{proposition}
	\begin{proof}
		As in Proposition~\ref{prop:general-price}, \(\operatorname{Cov}(h,\grp)=\operatorname{Cov}(h,q)\) for every function \(h\) of \(z\), and the predicted gap of \(h\) is \(\operatorname{Cov}(h,\grp)/(p_0p_1)\). So \(\operatorname{Cov}(f_s,\grp)=\operatorname{Cov}(f,\grp)-(1-\tau)c_s\operatorname{Cov}(s,\grp)=\tau\operatorname{Cov}(f,\grp)\). If \(f=f^*\), the change \(d=f_s-f^*\) is a function of \(z\), so the extra error is \(\E[d^2]=(1-\tau)^2c_s^2\operatorname{Var}(s)=(1-\tau)^2\operatorname{Cov}(f^*,\grp)^2\operatorname{Var}(s)/\operatorname{Cov}(s,\grp)^2\), which is Eq.~\eqref{eq:built-price} because \(\operatorname{Cov}(f^*,\grp)=p_0p_1\Delta^*\) and \(\operatorname{Var}(\grp)=p_0p_1\). Finally, \(R^2_{\grp}=\operatorname{Var}(q)/(p_0p_1)=\operatorname{corr}(q,\grp)^2\) because \(\operatorname{Cov}(q,\grp)=\operatorname{Var}(q)\), and by Cauchy--Schwarz \(\operatorname{Cov}(s,\grp)^2=\operatorname{Cov}(s,q)^2\le\operatorname{Var}(s)\operatorname{Var}(q)\), with equality exactly when \(s-\E s\) is proportional to \(q-p_1\).
	\end{proof}
	Three consequences matter for an audit. First, a correction built from an estimated score removes the gap by construction, and its quadratic part can be computed from three moments, \(\operatorname{Cov}(f,\grp)\), \(\operatorname{Cov}(s,\grp)\) and \(\operatorname{Var}(s)\); calibration of the score is not needed. Second, the announced cost is the cost of \emph{this} correction; for the best predictor it is an upper bound of the minimum, and a score that misses part of the group signal overstates the minimum. Calibration does not prevent this. With \(z=(X_1,X_2)\) uniform on \(\{-1,1\}^2\), \(y=X_1\) and \(q=0.5+0.2X_1+0.2X_1X_2\), the score \(s=0.5+0.2X_1\) is perfectly calibrated (\(\E[\grp\mid s]=s\)) but misses the interaction: the built correction costs 1, while the minimum is 0.5. A high built cost therefore does not prove a high minimum; it may come from a poor score.

	Third, the two parts of the proposition need different assumptions. The first, that \(f_s\) keeps the fraction \(\tau\) of the gap, holds for any predictor \(f\). The second, that the extra error equals \((1-\tau)^2p_0p_1(\Delta^*)^2/\rho_s^2\), needs \(f=f^*\). For any predictor \(f\) we define the \emph{quadratic part}
	\begin{equation}
		\pi_s=c_s^2\operatorname{Var}(s)=\frac{p_0p_1\Delta(f)^2}{\rho_s^2}.
		\label{eq:quad-part}
	\end{equation}
	With \(d=f_s-f\), \(\E[d^2]=(1-\tau)^2\pi_s\), so \(\pi_s=\E[d^2]/(1-\tau)^2\) when \(\tau\neq1\). By Eq.~\eqref{eq:cross-term}, the change of the squared error is \(\operatorname{MSE}(f_s)-\operatorname{MSE}(f)=(1-\tau)^2\pi_s+2\E[(f-y)d]\). The cross term vanishes when \(f=f^*\); otherwise it can have either sign. We call \(\pi_s+2\E[(f-y)d]\) (at \(\tau=0\)) the \emph{total cost}, and we report both.

	\paragraph{What the minimum says about a learned predictor.} When \(f\neq f^*\), three statements must be kept apart (all at \(\tau=0\)). (i) The corrected predictor is still compared with the best one: \(f_s\) is a function of \(z\) with a zero predicted gap, so by Proposition~\ref{prop:general-price}, \(\operatorname{MSE}(f_s)-\operatorname{MSE}(f^*)\ge p_0p_1(\Delta^*)^2/R^2_{\grp}\). (ii) The quadratic part has its own lower bound. The argument of Proposition~\ref{prop:general-price}, with \(f\) in place of \(f^*\), shows that every change \(d(z)\) that removes the predicted gap of \(f\) has \(\E[d^2]\ge p_0p_1\Delta(f)^2/R^2_{\grp}\); hence
	\begin{equation}
		\pi_s\;\ge\;\pi_{\min}(f)=\frac{p_0p_1\,\Delta(f)^2}{R^2_{\grp}},
		\label{eq:quad-min}
	\end{equation}
	with equality if and only if \(s\) is an affine function of \(q\), when \(\Delta(f)\neq0\); when \(\Delta(f)=0\), \(c_s=0\) and both sides are zero for every score. This bound uses the gap \(\Delta(f)\) of the learned predictor, not the gap \(\Delta^*\) of the best one. (iii) Therefore, neither the quadratic part nor the total cost of a correction of a learned predictor needs to be above the minimum of Proposition~\ref{prop:general-price}. For example, if \(f=f^*/2\) and \(s=q\), then \(\Delta(f)=\Delta^*/2\) and \(\pi_s=\pi_{\min}(f)=p_0p_1(\Delta^*)^2/(4R^2_{\grp})\), a quarter of the minimum: a predictor that already predicts less of the gap needs a smaller change to remove it. The minimum of Proposition~\ref{prop:general-price} bounds the excess risk of the corrected predictor over the best predictor (statement i), not the change of error relative to the learned predictor, which is what the audit announces. Throughout the paper, ``exact'' and ``at least the minimum'' refer to the population result with \(f=f^*\); for a learned predictor, the quadratic part is compared only with \(\pi_{\min}(f)\), and the total cost is reported as measured.

	\paragraph{The target of the audit.} The audit estimates the cost of the correction that is actually fitted: with \(f\), \(s\), \(\widehat c_s\) and the centre of \(s\) fixed as fitted, the change of the expected squared error on new units from the same population, in its quadratic part and in total. In the controlled study we compute this target on a large independent sample for every replicate, and we report how often the announced interval contains it (Section~\ref{sec:controlled}). The minimum of Proposition~\ref{prop:general-price} is a second, population target; it equals the first only when the score is well specified, \(f\) is close to \(f^*\) and the samples are large.

	\paragraph{What the audit adds to a direct measurement.} The announced total cost is not a new estimator of the change of error. On the validation units, \(\widehat\pi_s\) is the mean of \(d^2\), so \(\widehat\pi_s+2\widehat\E_{\mathrm{val}}[(f-y)d]\) is exactly the change of the mean squared error measured by applying the correction to these units; a careful analyst can obtain the same number without our formulas. What matters is where this number is computed and what comes with it. Computed on the units that fit the models, the same quantity is biased: the first version of our procedure, which did this, announced \(0.79\) for a population cost of \(1.01\) (Section~\ref{sec:controlled}). Around the number, the audit adds four things. (i) The quadratic part needs no outcomes: it uses only \(f\), \(s\) and the group, so it can be computed on units whose outcomes are not yet observed, and it does not depend on how well \(f\) fits. (ii) Its form \(p_0p_1\Delta(f)^2/\rho_s^2\) explains a cost: a high value comes either from a large predicted gap or from a weak group score. By Eq.~\eqref{eq:quad-min}, a group score closer to \(q\) lowers the cost towards \(\pi_{\min}(f)\), so comparing scores built from the same information is a sensitivity check: if a richer score lowers the cost, part of the cost came from the weaker score; if it does not, both scores may still miss the same part of the group signal, so the comparison does not show that \(\pi_{\min}(f)\) is reached or that the remaining cost cannot be lowered. (iii) Its interval refits the group model, so it includes the variation of the score, and the diagnostic D1--D3 below says when the number is not supported by the data. (iv) The test stage checks, on independent units, that the gap reduction transfers, which a measurement on the units that built the correction cannot do. The contribution of the audit is this decomposition and these checks, not a more accurate point estimate.

	\paragraph{A three-stage audit.} We therefore estimate the cost of the correction one can build, and we keep the units that fit the models, the units that announce the price and the units that test it apart.
	\begin{enumerate}[leftmargin=*,itemsep=2pt,topsep=3pt]
		\item \textbf{Fit} (training units): a predictor \(f\) and a group score \(s\) that uses the \emph{same} information \(z\).
		\item \textbf{Announce and diagnose} (validation units): estimate \(\widehat c_s\), \(\widehat\Delta\) and \(\widehat\rho_s^2\); announce the quadratic part \(\widehat\pi_s=\widehat p_0\widehat p_1\widehat\Delta^2/\widehat\rho_s^2\) and the total cost \(\widehat\pi_s+2\widehat\E_{\mathrm{val}}[(f-y)d]\); run the diagnostic.
		\item \textbf{Test} (independent test units): apply the fitted correction; measure the remaining share \(\widehat\tau\) of the gap and the extra error with its two terms.
	\end{enumerate}
	Table~\ref{tab:audit-data} lists which units enter each quantity and each interval. Three intervals are kept apart: the \emph{announced} interval uses only training and validation units; the \emph{measured} interval uses only test units, with the correction fixed; the interval of the \emph{difference} between measured and announced cost resamples all three.
	\begin{table}[t]
		\centering\small
		\caption{Which units enter each quantity of the audit. Resampling is by cluster (school or class) on the real data and by unit in the controlled study.}
		\label{tab:audit-data}
		\begin{tabular}{@{}p{0.47\linewidth}p{0.47\linewidth}@{}}
			\toprule
			Quantity & Units \\
			\midrule
			\(f\) and \(s\) & training (for the network of Section~\ref{sec:results-next-item}: selection and early stopping on separate selection units) \\
			\(\widehat c_s\), centre of \(s\), announced \(\widehat\pi_s\) and total & validation \\
			Announced interval and diagnostic D1--D3 & bootstrap of training units (\(s\) refitted; \(f\) refitted when cheap) and of validation units \\
			\(\widehat\tau\), measured cost, checks T1--T2 & test; interval: test units resampled, correction fixed \\
			Measured minus announced & training, validation and test units resampled together \\
			\bottomrule
		\end{tabular}
	\end{table}
	\begin{definition}[Diagnostic of the announced price]
		\label{def:interpretable}
		On the validation units, the announced cost is \emph{accepted} if (D1) the 95\% cluster-bootstrap interval of the predicted gap of \(f\) excludes zero, (D2) the interval of \(\operatorname{Cov}(s,\grp)\) excludes zero, and (D3) the coefficient of variation of \(\widehat\pi_s\) over the bootstrap is at most 0.5; otherwise the audit abstains. For an accepted cost, the \emph{gap reduction transfers} to the test units if (T1) the interval of the predicted gap of \(f\) excludes zero there too and (T2) the correction leaves at most 20\% of it, \(|\widehat\tau|\le0.2\); otherwise the audit reports that the gap reduction does not transfer to these units, together with the measured cost. Whether the announced cost agrees with the measured cost is a third, separate question, judged by the interval of their difference (Table~\ref{tab:audit-data}).
	\end{definition}
	The diagnostic is an empirical screen, not a guarantee. On the validation units it detects three failures: there is no detectable gap to remove (D1), the score carries no detectable group information (D2), or the price changes too much between refits to guide a decision (D3). The test units detect a fourth one, a change of the group signal between the units used to build the correction and the units where it is used (T1, T2). These two checks concern the gap that the correction removes, not the accuracy of the announced cost; this needs the correction to be applied to independent units, which can be done before deployment but not before the correction is built. The diagnostic cannot detect a score that misses part of the group signal, which by Proposition~\ref{prop:built} makes the announced cost conservative as an estimate of the minimum (for the best predictor), while it remains the cost of the correction actually built. An abstention is a result: it says that the data do not support an estimate of the cost; it does not say that there is no gap. The interval, finally, is conditional on the fitted predictor when \(f\) is not refitted. For the network of Section~\ref{sec:results-next-item}, a fourth set of units, used only to select the network and to stop its training, keeps the announcement apart from the fit; if the validation units also selected the network, the announcement would not be fully independent of the fit. This costs units: half as many students announce the cost, and the intervals are wider. The five splits share units, so they are five related checks, not five independent validations.

	A first version of this procedure announced the price on the training units and required a calibrated score (four conditions R1--R4); we use it only as a comparator. The change to the three-stage audit is motivated by Proposition~\ref{prop:built} and by the controlled study; the real-data results of both versions are reported (Appendix~\ref{app:v1-procedure}).

	\section{A controlled study of the price and its estimate}
	\label{sec:controlled}
	A reader may object that imposing equal mean predictions where the outcomes differ must cost some accuracy, so that the price only confirms the obvious. What the audit adds is the size of the cost, the conditions under which it is large or small, and whether it can be known before an intervention. To show this, we use data where everything is known. The state \(z\in\R^6\) is Gaussian with a mean shift \(a\,u\) for group 1 (\(p_1=0.4\)), and the outcome is \(y=\beta^\top z+\varepsilon\) with \(\varepsilon\sim\mathcal N(0,1)\) and \(|\beta|=|u|=1\). Then \(f^*(z)=\beta^\top z\), \(q\) is exactly logistic in \(z\), and the true price follows from Proposition~\ref{prop:general-price}. Three regimes set the role of the group signal: \emph{useless} (\(u\perp\beta\): the group moves the state along a direction that the outcome does not use), \emph{partly aligned} (\(\cos(u,\beta)=0.5\)) and \emph{aligned} (\(u=\beta\): the group moves the state along the direction of learning, as in our data). The separation \(a\in\{0.2,0.7,1.5\}\) gives \(R^2_{\grp}=0.01\), \(0.11\) and \(0.38\); on the real data, the linear \(R^2\) of the group is \(0.004\) to \(0.06\). Two cells change the usefulness of the group signal between training and test (\(a=0.7\)): \emph{utility lost} (aligned in training, useless in test) and \emph{utility gained} (the reverse). A last pair of cells is the example of Section~\ref{sec:price-estimation}, where a main-effects logistic group model is calibrated but misses an interaction, and the same cell with a correct group model. We run the three-stage audit with training, validation and test sets of \(n=2{,}000\) or \(20{,}000\) units each, with least squares (LS; well specified, refitted in each bootstrap draw) or gradient boosting (GB; flexible, kept fixed), in 100 and 50 replicates, with 1,000 bootstrap draws per replicate. In each replicate we also compute the target of Section~\ref{sec:price-estimation}, the realized cost of the fitted correction, on 200{,}000 new units. Table~\ref{tab:controlled} gives the result; Appendix~\ref{app:controlled} gives every cell and both learners from a first run with 60 draws on the same data, and the stability of the bootstrap.

	\begin{table}[tbp]
		\centering\small
		\caption{Controlled study, least squares. (a) Population cost of full removal: the minimum (Proposition~\ref{prop:general-price}) and the cost of the correction built from the population limit of the fitted group score (Proposition~\ref{prop:built}); they differ only when the score is misspecified. Median announced: the first version of the procedure (v1, training units) and the quadratic part announced by the three-stage audit, \(n=20{,}000\). (b) The four results to read together, for each sample size (Definition~\ref{def:interpretable}; 1{,}000 bootstrap draws). Acc.: share of replicates accepted on validation units (one minus the share in which the audit abstains). Transf.: share accepted and whose gap reduction transfers to test units (T1--T2). Cov.: share of the accepted replicates whose nominal 95\% audit interval contains the realized total cost of the fitted correction (200{,}000 new units). Width: median width of this interval divided by the median realized total cost (over all replicates). Coverage and width of a zero cost are not shown: the audit abstains there. Each share is over 100 replicates; an observed coverage of 0.91 has a 95\% binomial (Wilson) interval of 0.84 to 0.95, and one of 1.00 an interval of 0.96 to 1.00. Table~\ref{tab:controlled-robust} gives the coverage with 60 draws and with a second interval.}
		\label{tab:controlled}
		\setlength{\tabcolsep}{3pt}
		\input{tab_controlled.tex}
	\end{table}

	\paragraph{When is erasure cheap, and when is it expensive?} The price can be large even when the state predicts the group poorly, because it is set by the alignment of the prediction with the group signal. When the group signal is useless, erasure is free: the true price is zero. When it is aligned, the price is almost the whole variance of the best prediction (1.10 out of \(\operatorname{Var}(f^*)=1.12\) at \(a=0.7\)), and it stays as large when the state explains only 1\% of the group's variance: at \(a=0.2\) it is still 1.01. A partly aligned signal gives an intermediate price (0.25 to 0.36). This is Corollary~\ref{cor:price-bound} at work, and it explains the real data, where the group explains at most 6\% of its own variance and erasure is nevertheless expensive.

	\paragraph{How accurate is the announced cost, and how useful?} We judge the announcement against its target, the realized cost of the correction that was fitted, and we read four results together (Table~\ref{tab:controlled}b): how often the audit accepts, how often the gap reduction transfers, how often the interval contains the target, and how wide the interval is. A narrow interval that misses the target and a wide interval that contains it are both of little use for a decision. When the sample is large (\(n=20{,}000\), least squares), the diagnostic accepts the estimate in every replicate where there is a gap, and the nominal 95\% audit interval contains the realized total cost in 91 to 100\% of the accepted replicates. The coverage depends on the regime: it is lowest where the interval is narrowest relative to the cost, 92\% for the aligned signal with \(a=1.5\) and 91\% for the misspecified group model, whose intervals are 6\% of the cost wide (binomial intervals 85 to 96\% and 84 to 95\% over 100 replicates); the median error of the announced total is 1 to 2\%. Where the coverage is complete, the interval can be as wide as the cost itself (partly aligned signal, \(a=0.2\)). With \(n=2{,}000\), the coverage is 96 to 100\%, and the intervals are 0.2 to 1.0 times the cost wide where the audit accepts. The intervals are therefore approximate: we do not present them as a 95\% guarantee. A basic bootstrap interval from the same draws gives 92 to 100\% (\(n=20{,}000\)), and a first run with 60 draws on the same data gave 88 to 100\% (Table~\ref{tab:controlled-robust}): part of the shortfall came from the small number of draws. With gradient boosting kept fixed, the coverage is 90 to 100\% (Appendix~\ref{app:controlled}); over both learners and both sample sizes it is 90 to 100\%. The coverage among accepted replicates is the same when the threshold of D3 is 0.25, 1 or removed; the threshold changes only how often the audit abstains. The realized quadratic part is covered in every replicate with least squares: it depends only on the fitted coefficient and the variance of the score, and the audit interval, which also refits the score, over-covers it. The cost of the population built correction, a target that does not condition on the fitted models, was covered in 86 to 96\% of the accepted replicates in the first run (Table~\ref{tab:controlled-full-ls}). With \(n=2{,}000\) and a weak group signal, a group score evaluated on held-out units has a small correlation with the group, so the announced cost is high (1.30 for a population cost of 1.01, aligned signal) and so is the realized cost of that correction (1.29); its interval is about eight times as wide as the cost, and the diagnostic accepts only 5\% of these replicates, and none for the partly aligned signal. The first version of the procedure, which computed the price on the training units, announced 0.79 for the same population cost, and at \(n=20{,}000\) its interval covered it in 41\% of the replicates that it accepted. With gradient boosting kept fixed, the interval covers the population built cost in only 30 to 50\% of the accepted replicates for the aligned signal with \(a\ge0.7\) and \(n=20{,}000\) (first run, Table~\ref{tab:controlled-full-gb}): the fitted predictor differs from \(f^*\), its cross term is about 0.03, and an interval that does not refit the predictor describes the fitted correction, not the population one. Finally, the stability of the bootstrap depends on the regime (Table~\ref{tab:controlled-stability}): with 40 or 60 draws instead of 400, the end-points of the interval move by about a tenth of its width, and with 200 draws by 3 to 4\%; where the signal is weak and the sample small, they still move by 29 to 46\% with 40 or 60 draws and by 13 to 16\% with 200. The coverage changes by at most 0.07.

	\paragraph{When the group signal changes, and when the group model is too simple.} If the usefulness of the group signal changes between training and test, the validation units, drawn like the training units, cannot see it. When the utility is lost, the cost announced on validation units is accepted in every replicate, but the test units reveal the change: there is no longer a gap to remove (T1), the correction misses its target (T2), and the transfer check passes in only 1\% of the replicates (2\% at \(n=2{,}000\)). The diagnostic therefore flags almost every such case, not all of them. When the utility is gained, there is no gap on the validation units and the audit abstains. The diagnostic cannot detect the last failure, a calibrated group model that misses part of the group signal: it accepts the estimate, and the transfer check passes, in every replicate, and the estimate (1.00) is the cost of the correction that this group model builds, 1.0, while the minimum is 0.5. With a group model that includes the interaction, the audit announces 0.500. The announced cost is accurate for the correction actually built but conservative as an estimate of the minimum, and a flexible group model brings it closer to the minimum.

	In short, erasure is cheap when the group signal does not carry the predictive signal, and costs nearly the whole predictive variance when it does, however weakly the state predicts the group. The three-stage audit announces the cost of the correction it builds, accurately for that correction; as an estimate of the minimum, it is accurate when the group model is well specified and conservative when it is not; it abstains when there is nothing to remove or too little information, and it detects a change of the group signal on independent test units.

	\section{Data, models and protocol}
	\label{sec:protocol}
	The design separates questions that need different evidence. Outcomes under a policy are evaluated in a controlled simulator, where interventions can be compared under the same noise and against a reference that uses the known equations. ASSISTments, Eedi and OULAD are observational records: they test predictions on students or schools that the models have not seen, not the effects of interventions.

	\paragraph{Choices made after seeing results.} Several choices were not fixed in advance, and we list them here once. The order of the simulator outcomes (knowledge first, the synthetic disparity second) was fixed after the first analyses had been run with the disparity gap first. The thresholds of the diagnostic of Definition~\ref{def:interpretable}, and of the first version of the estimation procedure (Appendix~\ref{app:v1-procedure}), are conventions fixed when the analysis scripts were written, after earlier results had been seen; the three-stage audit was introduced after the results of the first version were known; it is motivated by Proposition~\ref{prop:built} and by the controlled study, and the results of both versions are reported. The fourth set of units for the knowledge-tracing audit and the 1,000 bootstrap draws of the controlled study were added after the results with three sets of units and 60 draws were known; both versions are reported. The margins of Section~\ref{sec:general-diagnostic-protocol} were chosen after the results. None of these choices was preregistered; every quantity is reported so that other choices can be applied. The analysis of the randomized experiments has its own list of changes (Appendix~\ref{app:rct}).

	\paragraph{Three sources of uncertainty.} We keep three sources of variation apart. \emph{Sampling} of the tested units (schools, classes or students) is measured by a cluster bootstrap at the level of the unit, within each split; this is the uncertainty we use for claims about a split. \emph{The split} itself, which reuses the same schools or students, is described by repeating the analysis over five splits; Student-\(t\) intervals over these splits describe how much a result depends on the split, not independent samples. \emph{Retraining} enters through the replicates of the simulator and the seeds of the neural models.

	\subsection{Tutoring simulator and controllers}
	\label{sec:sim-protocol}
	The simulator has state \(x_t=(k_t,e_t,p_t,d_t)\): simulated knowledge, engagement, performance and a synthetic disparity variable. With \(\sigma\) the logistic function,
	\begin{align}
		k_{t+1}&=\clip_{[0,1]}\!\left(0.85k_t+0.15\sigma(3e_t-1.5)+0.20u_t-0.06\gamma\grp+\varepsilon^k_t\right),\\
		e_{t+1}&=\clip_{[0,1]}\!\left(0.80e_t+0.10\tanh(p_t)+0.15u_t+\varepsilon^e_t\right),\\
		p_{t+1}&=\sigma\bigl(6(k_{t+1}-0.5)\bigr),\qquad
		d_{t+1}=0.90d_t-0.05\gamma\grp+0.05u_t+\varepsilon^d_t,
		\label{eq:simulator}
	\end{align}
	with Gaussian noise of standard deviations \((0.03,0.03,0.02)\) and a group strength \(\gamma\) (\(\gamma=1\) in the main setting). Each cohort has 150 simulated students, half in each group, followed for 25 turns from the same initial distribution, so any group difference comes from the dynamics. The primary outcomes are the final mean knowledge of each group, \(\overline k_{T,0}\) and \(\overline k_{T,1}\), and the knowledge gap
	\begin{equation}
		\delta^k_T=\overline k_{T,1}-\overline k_{T,0},\qquad
		\Delta\delta^k_T=\Delta\overline k_{T,1}-\Delta\overline k_{T,0};
		\label{eq:knowledge-gap}
	\end{equation}
	the disparity gap \(\delta_T\) of \(d_T\) is a secondary outcome, because \(d_t\) is a synthetic state directly affected by control, not a measure of learning. We report both (the order is one of the choices listed below). The EDMD model uses the ten features \((k,e,p,d,k^2,e^2,d^2,ke,de,kd)\), is fitted by least squares on 150 training trajectories with random controls, and the projection removes rank 4 (Appendix~\ref{app:main-setting} walks through one replicate).

	\paragraph{Controllers.} Following \citet{korda2018mpc}, at each turn the controller encodes the observed state, forecasts three steps with its model and minimizes
	\begin{equation}
		\sum_{j=1}^{3}\left[(1-\widehat k_{t+j\mid t})
		+\beta_{\mathrm{ind}}\widehat d_{t+j\mid t}^{\,2}+\rho\,u_{t+j-1\mid t}^{\,2}\right],
		\qquad -1\leq u_{t+j-1\mid t}\leq1,
		\label{eq:mpc}
	\end{equation}
	with \(\beta_{\mathrm{ind}}=8\) and \(\rho=1\) in the main setting; only the first action is applied. Arms are compared on the same students and noise, through paired differences such as
	\begin{equation}
		\tau_\delta=\delta^{\mathrm{MPC,proj}}-\delta^{\mathrm{MPC,raw}},\qquad
		\tau_k=\overline k_T^{\mathrm{MPC,proj}}-\overline k_T^{\mathrm{MPC,raw}}.
		\label{eq:paired}
	\end{equation}
	We measure the help given by \(\mathcal U_1=(nT)^{-1}\sum_{i,t}|u_{i,t}|\) and \(\mathcal U_2=(nT)^{-1}\sum_{i,t}u_{i,t}^{2}\); matching one does not match the other, and each comparison below says which budget it holds fixed. The \emph{gap-constrained} controller replaces the individual disparity penalty by the constraint that the forecast group gap \(\widehat\delta_{t+3\mid t}\) be zero,
	\begin{equation}
		\min_{\{u_{i,t+\ell\mid t}\}}\ \sum_{i=1}^{n}\sum_{j=1}^{3}\Bigl[(1-\widehat k_{i,t+j\mid t})+\rho\,u_{i,t+j-1\mid t}^{\,2}\Bigr]
		\quad\text{subject to}\quad \widehat\delta_{t+3\mid t}=0,\qquad -1\leq u_{i,t+\ell\mid t}\leq1,
		\label{eq:gap-constrained}
	\end{equation}
	and the \emph{gap-penalty} controller adds \(\beta_{\mathrm{gap}}\widehat\delta_{t+3\mid t}^{\,2}\) instead; both use the group labels when they decide (Appendix~\ref{app:solver} gives the exact solvers).

	\paragraph{A controller with floors.}
	\label{sec:pareto-fair}
	The last controller narrows a gap while no group's predicted mean knowledge may fall. It uses a \emph{group-aware} surrogate fitted on the same data with the group as an extra regressor, \([A\;B\;a\;a_{\grp}]=Z_+X^\dagger\) with \(X=[Z;U;\mathbf 1^\top;\grp^\top]\); its reference policy is the MPC of Eq.~\eqref{eq:mpc} on this surrogate, with first actions \(u^{\mathrm{ref}}_i\). With \(v_k=C_kB\) and \(v_d=C_dB\) the one-step effects of help on predicted knowledge and disparity, the next-step predictions are affine in the first actions:
	\begin{equation}
		\widehat\delta_{t+1\mid t}(u)=\widehat\delta^{\mathrm{ref}}+v_d\sum_{i}s_i\,(u_{i}-u^{\mathrm{ref}}_{i}),\qquad
		\widehat K_{g}(u)=\widehat K^{\mathrm{ref}}_{g}+\frac{v_k}{n_g}\sum_{i:\grp_i=g}(u_{i}-u^{\mathrm{ref}}_{i}),
		\label{eq:next-step}
	\end{equation}
	with \(s_i=1/n_1\) in group 1 and \(-1/n_0\) in group 0. For a budget factor \(c\geq1\), the \emph{fair set} is
	\begin{equation}
		\mathcal F_c=\Bigl\{u\in[-1,1]^n:\ \widehat K_g(u)\geq\widehat K^{\mathrm{ref}}_g\ (g=0,1),\ \ \sum_i|u_i|\leq c\sum_i|u^{\mathrm{ref}}_i|\Bigr\}.
		\label{eq:fair-set}
	\end{equation}
	The \emph{floor-constrained MPC} first computes, by a linear program, the smallest predicted gap that \(\mathcal F_c\) allows, and then minimizes the objective of Eq.~\eqref{eq:mpc} over the plans whose first controls lie in \(\mathcal F_c\) and reach 95\% of this reduction; if no reduction is possible, it keeps the reference plan. Its target can be the disparity gap or the knowledge gap. Appendix~\ref{app:floor-controller} gives the exact dual solver and its numerical checks.

	\begin{proposition}[Narrowing the gap without levelling down]
		\label{prop:no-levelling}
		Consider one turn with \(v_k>0\), \(v_d>0\) and \(u^{\mathrm{ref}}_i\geq0\) for every student.
		\begin{enumerate}[label=(\roman*),leftmargin=*]
			\item If \(c\leq1\), every \(u\in\mathcal F_c\) gives \(\widehat\delta_{t+1\mid t}(u)=\widehat\delta^{\mathrm{ref}}\). At equal or smaller budget, any control that changes the predicted gap lowers the predicted knowledge of one group.
			\item If \(c>1\), \(\sum_iu^{\mathrm{ref}}_i>0\), \(\widehat\delta^{\mathrm{ref}}\neq0\), and the group \(g^\star\) that is behind (\(g^\star=1\) if \(\widehat\delta^{\mathrm{ref}}<0\), else \(g^\star=0\)) has a student with \(u^{\mathrm{ref}}_i<1\), then \(\mathcal F_c\) contains a control with \(|\widehat\delta_{t+1\mid t}(u)|<|\widehat\delta^{\mathrm{ref}}|\).
			\item The largest reduction over \(\mathcal F_c\) is the value of a linear program, and the smallest remaining gap is nonincreasing in \(c\).
		\end{enumerate}
	\end{proposition}
	The proof is in Appendix~\ref{app:floor-controller}. The proposition holds for one step and one kind of action with the same positive effect on both groups; it does not cover individual students or long horizons. For a tutor, it says that with the same amount of help, moving help from one group to the other cannot narrow the predicted gap without lowering the predicted knowledge of one group; narrowing it without levelling down needs extra help for the group behind.

	\paragraph{Regime study.} Around the main setting we change one factor at a time: the group strength \(\gamma\) over \(\{-1,0,0.25,0.5,1,2\}\), the rank (\(1\) to \(8\)), the closure depth (\(1\) to \(5\)), the horizon \(T\in\{5,10,25,50\}\) and \(\beta_{\mathrm{ind}}\in\{0,1,8,32\}\), and a second grid crosses \(\gamma\) and \(\beta_{\mathrm{ind}}\): 28 settings, each with the same 50 replicates. A change is called significant with a Holm correction over the 28 settings at the 5\% level.

	\subsection{Real data}
	\label{sec:data}
	\paragraph{ASSISTments 2012--2013.}
	\label{sec:assistments}
	Each student's first 26 interactions, in time order, give the states \(x_0,\ldots,x_{25}\): a rolling correctness \(k_t\), an engagement proxy, a pass indicator and a residual correctness. We use five school-disjoint 60/20/20 splits and school-cluster bootstrap intervals. The data set has no validated sensitive attribute, so the group is a \emph{group of schools defined by their own performance}: a school is in group 1 if its average success on later steps (\(t>25\), which are never evaluated) is above the median of the training schools. This label is retrospective and is not a social attribute, so we read the ASSISTments study as a test of the mechanism, and we give more weight to the two data sets with external attributes. A variant where other students of the school give the label is reported in Appendix~\ref{app:more-real}.

	\paragraph{Eedi.}
	\label{sec:eedi}
	The NeurIPS 2020 Education Challenge data \citep{wang2020diagnostic} record the answers of pupils to mathematics questions. The group is \textbf{PremiumPupil}, the pupil's eligibility for financial support. It is recorded for 34{,}386 pupils (29\%), of whom 25.8\% are disadvantaged. The \emph{target population} of this study is therefore the pupils whose status is recorded; they differ from the others (more answers, lower accuracy; standardized differences of about 0.5), and Appendix~\ref{app:more-real} reports a reweighting by the probability of being recorded. The splits are disjoint in classes (about 480 test classes), with class-cluster intervals and a strict variant without overlapping pupils.

	\paragraph{OULAD.}
	\label{sec:oulad}
	The Open University Learning Analytics Dataset \citep{kuzilek2017open} describes 27{,}814 distance-learning students with a known deprivation band. The state is weekly over 26 weeks: \(k_t\) is the mean score of the assessments submitted so far (0.5 before the first), \(e_t\) the weekly clicks, \(p_t\) the pass indicator and \(r_t\) the weekly activity. The groups are \textbf{deprivation} (home area among the 30\% most deprived, 33.5\% of students), \textbf{disability} (9.8\%) and \textbf{gender} (48.7\% women). The main splits are random within the 22 course presentations, with presentation-level intervals; held-out presentations and a later year are reported in Appendix~\ref{app:more-real}. OULAD does not record when a score became known, so Section~\ref{sec:results-prediction} also delays the scores by up to four weeks.

	Appendix~\ref{app:data} gives the full description of the three data sets, the construction of the states and the sample flow. In every study, all scalings, the operator, the metric, the contrast, the projector and every baseline are fitted on training units only and then frozen.

	\paragraph{Use of AI tools.} Parts of this work were first written in French. We used an AI assistant (Claude, Anthropic) mainly to translate these paragraphs from French into English, and to edit the language of the text. We also used it to help write and check analysis code and to format the \LaTeX{} source. The authors checked each translated paragraph against the French original and each result against the outputs of the code, and take full responsibility for the content of this paper.

	\section{Prediction on real data: erasure hides the gap}
	\label{sec:results-prediction}
	\subsection{Three data sets, one pattern}
	\label{sec:results-realdata}
	We replay the observed sequences of held-out students or schools: at each step the model forecasts the next state from the newly observed one (Algorithm~\ref{alg:tutoring-inference} with \(u_t\equiv0\)), so the test uses the policy that really generated the data. Table~\ref{tab:real-summary} summarizes the three studies; Appendix~\ref{app:more-real} gives every split, the comparison with persistence, a linear model and gradient boosting, and the calibration in each group.

	\begin{table}[t]
		\centering\small
		\caption{Prediction on held-out units, mean over five splits. Observed: final gap in \(k_{25}\) (group 1 minus group 0) and number of splits whose cluster-bootstrap interval excludes zero. Original, recalibrated, projected: one-step predictions of this gap by the original EDMD, the EDMD recalibrated for each group, and the projected EDMD. Error ratio: one-step error on \(k\) of the projected model divided by that of the original model (range over splits), over all starting points, and on a common set of early starting points (\(t=0,\dots,5\)) or, for OULAD, on the weeks in which a new score becomes known.}
		\label{tab:real-summary}
		\resizebox{\linewidth}{!}{%
		\begin{tabular}{@{}lrrrrrr@{}}
			\toprule
			& & & & & \multicolumn{2}{c}{Error ratio, projected / original}\\
			\cmidrule(l){6-7}
			Data set (group) & Observed & Original & Recal. & Projected & All starting points & Early starts / new-score weeks\\
			\midrule
			ASSISTments (school performance) & \(+0.057\) (3/5) & \(+0.052\) & \(+0.060\) & \(-0.0001\) & 7.4--8.1 & 4.4 (early)\\
			Eedi (PremiumPupil) & \(-0.085\) (5/5) & \(-0.080\) & \(-0.087\) & \(+0.0005\) & 8.4--9.0 & 4.3 (early)\\
			OULAD (deprivation) & \(-0.048\) (5/5) & \(-0.047\) & \(-0.049\) & \(+0.0013\) & 10.6--11.0 & 1.68--1.74 (new score)\\
			OULAD (disability) & \(-0.025\) (5/5) & \(-0.024\) & \(-0.026\) & \(-0.0001\) & 8.4--8.9 & 1.52--1.57 (new score)\\
			OULAD (gender) & \(-0.006\) (0/5) & \(-0.007\) & \(-0.005\) & \(-0.0040\) & 7.5--9.3 & 1.46--1.58 (new score)\\
			\bottomrule
		\end{tabular}}
	\end{table}

	The pattern is the same in the three data sets. Wherever a gap is observed, the original model predicts it approximately, a recalibration for each group predicts it without visible bias, and the projected model predicts almost none of it: on Eedi, \(-0.085\) observed and \(+0.0005\) predicted. Under the level-2 rule of Table~\ref{tab:protocol}, projection is flagged in every split of every data set. Its error is also larger, but the size of this increase depends on the starting points. Averaged over all starting points, the projected model's one-step error is seven to eleven times that of the original model; on a common set of early starting points it is three to four times larger, and 1.1 to 1.8 times at 10 to 20 steps. The reason is that the targets are rolling or cumulative proxies, which change slowly: the original model predicts them almost perfectly from one step to the next, so a small absolute error becomes a large ratio. OULAD makes this visible. On the weeks in which a new score becomes known (15\% of the weeks), the projected model's error is 1.5 to 1.7 times the original; on the other weeks, where the original model only has to carry the score forward, it is 60 to 96 times larger. The next-item target of Section~\ref{sec:results-next-item} avoids this effect.

	\paragraph{Sensitivity to the data.} The conclusion does not depend on the choices that the review of each data set questions. On ASSISTments, labelling a school from other students gives the same picture (projection \(+0.0004\) for an observed \(+0.057\)). On Eedi, a strict variant without overlapping pupils gives \(-0.080\) observed and \(+0.0016\) projected, and reweighting the recorded pupils towards all Eedi pupils gives an observed gap of \(-0.081\), \(-0.076\) for the original model and \(0.0000\) for the projected model (error ratio 9.2). On OULAD, the projected model still predicts almost no gap when whole presentations or a later year are held out (\(-0.001\) and \(-0.004\) for an observed \(-0.048\) and \(-0.046\)), and when a score becomes known one, two or four weeks after it was submitted: with a four-week delay the deprivation gap is \(-0.047\) observed, \(-0.046\) for the original model and \(+0.0006\) for the projected model, whose error ratio is 9.3--9.8 overall and 1.6--1.7 on the weeks with a new score (Appendix~\ref{app:more-real}).

	\subsection{The loss comes from the group direction: rank-matched controls}
\label{sec:results-rank-matched}

The projection removes \(r=4\) directions from the model. So a loss of accuracy may come from two sources: from removing group information, or simply from making the model smaller. To separate them, we compare the group projection with other ways of removing directions. Every arm uses the same data split, the same fitted model, the same metric \(\Gamma\) and the same rank; only the removed directions change. The controls are:
\begin{itemize}[leftmargin=*,itemsep=1pt]
	\item \emph{random}: random directions (50 draws per split);
	\item \emph{shuffled labels}: the same construction as ours, but the contrast is fitted on group labels shuffled between students (50 draws). This is our algorithm without group information;
	\item \emph{top-variance} and \emph{lowest-variance PCA}: the directions where the features vary most, or least;
	\item \emph{least-used directions}: the directions that the fitted operator uses least;
	\item \emph{LEACE} \citep{belrose2023leace}: the closed-form eraser of a linear concept. For a binary group it removes one direction.
\end{itemize}
We also run every control at rank 1, to compare LEACE at its own rank, and we measure the one-step and ten-step error on knowledge, overall and in each group, the predicted gap and the calibration in each group. Section~\ref{sec:results-policy} runs the same controls in the simulator.

\begin{figure}[t]
	\centering
	\includegraphics[width=\linewidth]{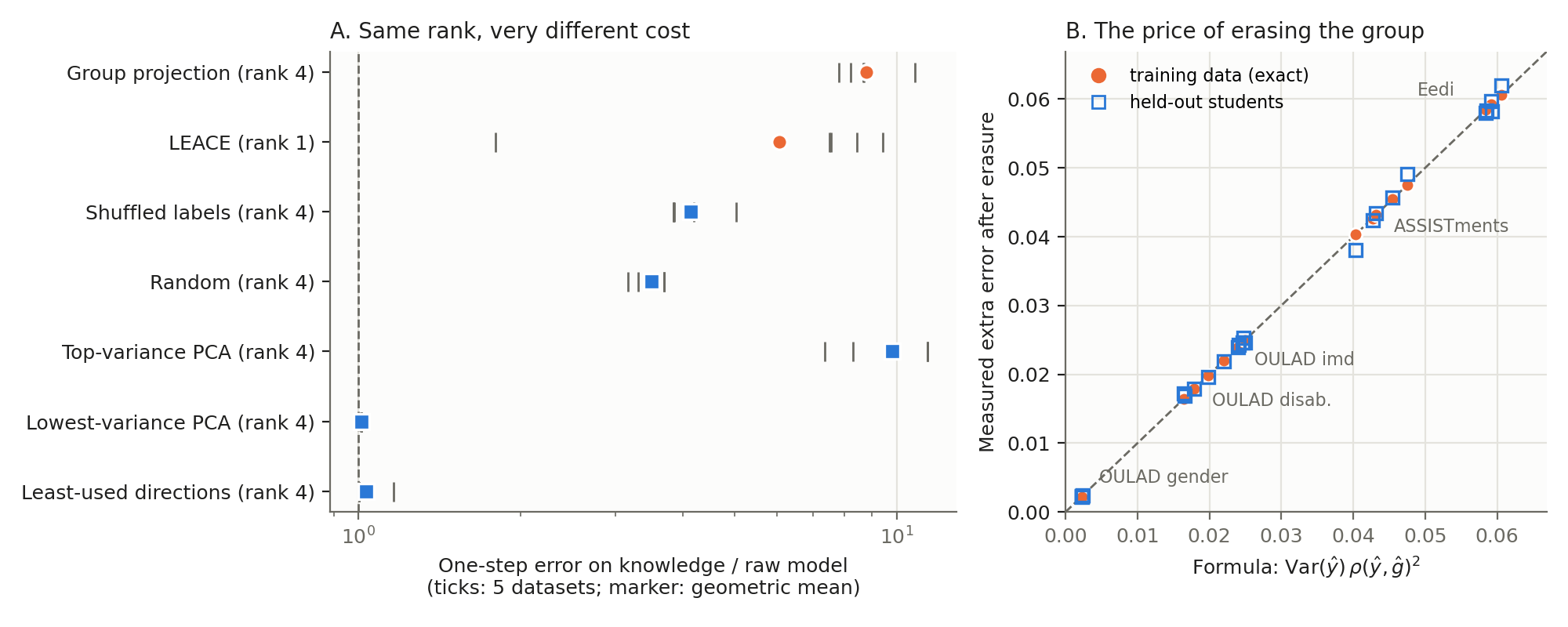}
	\caption{Rank-matched controls on the real data. A: one-step error on knowledge relative to the raw model, for each way of removing directions (ticks: five settings; marker: geometric mean). B: Proposition~\ref{prop:erasure-price} for \(\tau=0\): predicted against measured extra error after erasing the group, five splits per setting, on training data (where the identity is exact) and on held-out students.}
	\label{fig:rank-matched}
\end{figure}

\paragraph{Prediction.} Table~\ref{tab:rank-matched} and Figure~\ref{fig:rank-matched}A give the result. Compression in itself is cheap: with lowest-variance PCA or the least-used directions, the error stays at the level of the raw model in every dataset. Random directions of the same rank cost much less than the group projection. The paired difference in one-step error (group minus random) is positive in all five settings, for example \(+0.034\) \([0.032,0.037]\) on ASSISTments and \(+0.043\) \([0.042,0.044]\) on Eedi (\(t\) intervals over five splits, which describe the dependence on the split). In 21 of the 25 splits, the group projection is worse than all 50 random draws. The shuffled-label projections show that our construction itself adds some cost: it targets directions that the dynamics use a lot, so it costs more than random directions. But it still costs less than the group projection (the group projection is at or above the 84th percentile of the shuffled draws in every split). Because the ASSISTments group is a property of the school, we also shuffled the labels between schools rather than between students (50 draws per split). The school-level shuffle costs more than the student-level one (mean one-step error \(0.045\) against \(0.037\)), as expected when the shuffled label keeps the school structure, but the group projection still costs more (\(0.058\)) and is above 93\% of the school-level draws on average (78\% to 100\% across splits). The strongest result is that one direction is enough. LEACE removes a single direction, and it causes almost all the loss on ASSISTments, Eedi and OULAD (deprivation, disability): for example \(0.055\) against \(0.058\) for rank 4 and \(0.007\) for the raw model on ASSISTments. Gender is the exception, and the next result explains why.

\begin{table}[t]
	\centering\small
	\caption{One-step error on knowledge (MSE, held-out data, mean over five splits). All arms use the same fitted model and metric; ``rank'' is the number of removed directions. Random and shuffled: mean over 50 draws per split. The three OULAD rows of these two columns agree to three digits (shuffled: 0.01138, 0.01140, 0.01135) although each split uses its own labels: a shuffled label is independent of the state, so the contrast fitted on it is a random direction whose cost hardly depends on the attribute.}
	\label{tab:rank-matched}
	\begin{tabular}{lrrrrrrr}
		\toprule
		& Raw & Group & LEACE & Shuffled & Random & Top PCA & Lowest PCA \\
		Rank & -- & 4 & 1 & 4 & 4 & 4 & 4 \\
		\midrule
		ASSISTments (school) & 0.0074 & 0.0575 & 0.0553 & 0.0371 & 0.0234 & 0.0543 & 0.0074 \\
		Eedi (PremiumPupil) & 0.0080 & 0.0695 & 0.0677 & 0.0338 & 0.0266 & 0.0667 & 0.0081 \\
		OULAD (deprivation) & 0.0030 & 0.0319 & 0.0278 & 0.0114 & 0.0109 & 0.0337 & 0.0030 \\
		OULAD (disability) & 0.0030 & 0.0257 & 0.0223 & 0.0114 & 0.0109 & 0.0337 & 0.0030 \\
		OULAD (gender) & 0.0030 & 0.0242 & 0.0053 & 0.0114 & 0.0109 & 0.0337 & 0.0030 \\
		\bottomrule
	\end{tabular}
\end{table}

\paragraph{Why the group direction is expensive.} The group is hard to predict from the state: a linear model of the group from the features explains only 0.4--6\% of its variance. This does not make the group cheap to erase. What matters is how strongly the prediction is aligned with the group signal, and in these data it is strongly aligned: the best model predicts a real gap (for example \(0.075\) on ASSISTments and \(-0.072\) on Eedi) along directions that also carry knowledge. On ASSISTments, the fitted group score correlates at \(0.86\) with knowledge. Removing the group direction therefore removes most of what the model knows about knowledge. Top-variance PCA, which never sees the group, does almost the same damage, and it also brings the predicted gap to zero. Section~\ref{sec:price} gives the exact price of this loss.

\paragraph{Checks of the price.} Two different interventions appear in this subsection and should not be confused. Table~\ref{tab:rank-matched} removes directions from the dynamic model and refits nothing; Table~\ref{tab:erasure-price} computes the best predictor of one-step knowledge under the mean-parity constraint. The second is the cheapest way to remove the mean contrast, so its cost is a lower bound for the first, not the same quantity. Figure~\ref{fig:rank-matched}B and Table~\ref{tab:erasure-price} check the linear price (Proposition~\ref{prop:erasure-price}) for one-step knowledge on the five settings. On the training data it holds exactly, as it must; on held-out students it holds closely. The alignment describes the differences between settings. For gender, the prediction is weakly aligned with the group signal (\(|\rho|=0.27\)), so the price is small. For disability, the predicted gap is of similar size (\(0.029\) against \(0.024\)) but strongly aligned with the group signal (\(|\rho|=0.77\)), with a group that the state hardly explains (\(R^2=0.004\)), and the price is eight times larger. This is why LEACE is cheap for gender (0.0053 against 0.0030) and expensive elsewhere. For gender, the cost of the rank-4 projection (0.0242) comes from the three extra directions, not from the group direction.

	\begin{table}[t]
		\centering\small
		\caption{Price of erasing the group signal (Proposition~\ref{prop:erasure-price}, \(\tau=0\)), one-step knowledge, mean over five splits. \(|\rho|\): alignment between the prediction and the linear group score; \(R^2\): share of the group's variance explained by the features. Formula: \(\operatorname{Var}(\hat y)\rho^2\). Measured: extra MSE of the best group-erased linear predictor, on the training data (where the identity is exact) and on held-out students.}
		\label{tab:erasure-price}
		\resizebox{\linewidth}{!}{%
		\begin{tabular}{@{}lrrrrrr@{}}
			\toprule
			Setting & \(|\rho|\) & \(R^2\) of group & \(\operatorname{Var}(\hat y)\) & Formula & Measured (train) & Measured (held out)\\
			\midrule
			ASSISTments (school) & 0.93 & 0.030 & 0.0513 & 0.0438 & 0.0438 & 0.0437\\
			Eedi (PremiumPupil) & 0.97 & 0.017 & 0.0625 & 0.0592 & 0.0592 & 0.0593\\
			OULAD (deprivation) & 0.89 & 0.016 & 0.0310 & 0.0245 & 0.0245 & 0.0246\\
			OULAD (disability) & 0.77 & 0.004 & 0.0310 & 0.0185 & 0.0185 & 0.0187\\
			OULAD (gender) & 0.27 & 0.060 & 0.0310 & 0.0023 & 0.0023 & 0.0023\\
			\bottomrule
		\end{tabular}}
	\end{table}

\paragraph{The price over the forecast horizon.} We fit the best linear predictor of \(k_{t+h}\) directly for each horizon, on a common set of six starting points per student (\(t=0,\dots,5\)), so that the law of the information, and \(R^2\), is the same for every \(h\). On ASSISTments its price falls from \(0.054\) at \(h=1\) to \(0.023\) at \(h=5\), \(0.005\) at \(h=10\) and \(0.003\) at \(h=20\). In all five settings, the formula equals the measured extra error on the training data (up to \(2\times10^{-6}\)) and stays within 0--6\% of it on held-out students for Eedi and OULAD, and within 3--13\% for ASSISTments. The iterated EDMD forecast behaves differently. Removing the same group score from it costs between \(0.6\) and \(3.1\) times the value of the formula on held-out students (\(0.15\) to \(1.8\) for OULAD), because the cross term of Eq.~\eqref{eq:cross-term} does not vanish: on ASSISTments at \(h=1\) it is \(-0.014\), against a formula of \(0.067\). The formula therefore describes the best predictor at each horizon, not an iterated forecast. On the same common starting points, the projected EDMD model's error is \(4.4\), \(4.3\) and \(2.7\)--\(3.4\) times that of the raw model at one step on ASSISTments, Eedi and OULAD, \(1.7\)--\(2.2\) times at five steps, and \(1.1\)--\(1.8\) times at 10 to 20 steps. The ratio of seven to eleven reported in Section~\ref{sec:results-realdata} averages over all starting points; the projection is worse in every case, but by less on these early starting points and at long horizons.

\paragraph{Nonlinear erasure.} To check Proposition~\ref{prop:general-price} beyond linear predictors, we estimate \(f^*\) and \(q\) by gradient boosting on the physical state, with out-of-fold predictions on the training students, and apply \(f_0\) to held-out students. On new students, \(f_0\) does not remove the whole gap, so its cost must be compared with the frontier at the share \(\tau\) of the gap that it leaves, not with the price at \(\tau=0\). Table~\ref{tab:general-price} (Appendix~\ref{app:block-data}) makes this comparison. It checks Proposition~\ref{prop:general-price}, the minimum price with \(q\) estimated on the training students; it is not a test of the three-stage audit of Section~\ref{sec:price-estimation}, which Sections~\ref{sec:controlled} and~\ref{sec:results-next-item} evaluate. Two results stand out. First, the estimated erasure is incomplete: it leaves 12--40\% of the held-out gap on Eedi and OULAD (deprivation, disability), and it reverses the gender gap. Second, it is not efficient: for the gap it removes, it pays 1.25 to 2.6 times the nonlinear frontier on these three settings, and about as much as a linear erasure that leaves the same share of the gap (for example \(0.039\) against \(0.040\) on Eedi). Gender is the exception: the nonlinear model predicts this group much better (\(R^2_{\grp}=0.158\) against \(0.060\)), and its erasure costs a third of the linear one; it even falls below its own curve (ratio \(0.8\)), which shows that a frontier estimated on training students is not an exact bound for new students. On ASSISTments, where the test schools are new, the held-out predicted gap is close to zero in two of five splits, so the test check T1 of Definition~\ref{def:interpretable} fails and we do not interpret the comparison. By contrast, the linear erasure transfers well: on held-out students it leaves at most 3\% of the gap predicted on the training students for Eedi and OULAD, and its cost stays within 1--8\% of the formula. In short, Proposition~\ref{prop:general-price} is exact where \(q\) is known; with an estimated \(q\), nonlinear erasure on new students is neither complete nor cheaper than linear erasure, except where it predicts the group much better.

\paragraph{A link with LEACE.} When the metric is the sample covariance of the features (\(\Gamma=\Gram\), no ridge), erasing the group with LEACE and then refitting the dynamics by least squares gives exactly the projected operator \(EAE\) of the rank-1 projection. (The erased and removed parts of the features are uncorrelated in the sample, so least squares separates them.) We checked this numerically: the two models agree to \(10^{-10}\). So our projection is, at rank 1, ``LEACE then refit''; at rank 4 it adds three directions chosen from the dynamics.

	These controls change how the earlier results should be read. Two reductions of the same rank that do not use the group cost nothing, and random directions cost much less than the group projection. Wherever the best model predicts a real gap, removing the single group direction (LEACE) causes most of the loss, and Proposition~\ref{prop:erasure-price} gives its price, \(p_0p_1\hat\Delta^2/R^2\), before any experiment: it is large whenever the prediction is strongly aligned with the group signal.

	\subsection{Next-item prediction with a selected knowledge-tracing network}
	\label{sec:results-next-item}
	The targets above are rolling or cumulative proxies, which change slowly and inflate error ratios. The standard target of knowledge tracing is the correctness of the next item, so we now predict whether a student answers item \(t+1\) correctly, for \(t=0,\dots,24\), on the held-out schools or classes of the same five splits, on ASSISTments and Eedi, where item-level correctness is recorded. As the learned predictor we use deep knowledge tracing (DKT, \citealp{piech2015deep}): a recurrent network over the skill and correctness of past items, with an output weight for the skill of the next item. Its configuration is \emph{selected on the validation schools or classes} of each split, over the hidden size (32, 64, 128), the learning rate (\(10^{-3}\), \(3\times10^{-3}\)) and dropout (0, 0.2), with early stopping on the validation log-loss (Appendix~\ref{app:next-item} gives the grid and the selected configurations). We compare it with the EDMD readouts and with four ways of removing its mean group contrast:
	\begin{itemize}[leftmargin=*,itemsep=1pt,topsep=2pt]
		\item \emph{LEACE on the hidden state}, with the output layer refitted;
		\item \emph{LEACE within each next skill}: the hidden state is erased separately for each skill of the next item, so that the whole input of the output layer, hidden state and next skill, carries no linear group signal given the skill; output layer refitted;
		\item \emph{mean-parity fine-tuning}: the network is trained further with a penalty \(\lambda\) times the squared gap of its mean predictions, on all its inputs, with \(\lambda\) the smallest of \(10,10^2,10^3,10^4\) that leaves at most 10\% of the gap on the validation units;
		\item \emph{output correction}: the correction of Proposition~\ref{prop:built}, \(f_s=f-\widehat c_s\,(s-\widehat\E s)\), where the group score \(s\) is a logistic model of the group on the same information as the prediction (hidden state and next skill), fitted on the training units, and \(\widehat c_s\) and the centre are estimated on the validation units (the three-stage audit of Section~\ref{sec:price-estimation}).
	\end{itemize}
	The first three methods are fitted on training units, with early stopping on validation units; the output correction uses the validation units for its coefficient. Table~\ref{tab:next-item} gives the results with school- or class-level bootstrap intervals for the paired differences with the selected DKT. For the audit of the cost (Table~\ref{tab:next-item-audit-main}), the network is selected on separate units, as described below.

	\begin{table}[t]
		\centering\small
		\caption{Next-item correctness on held-out schools (ASSISTments) or classes (Eedi), five splits. AUC: mean over splits. \(\Delta\)AUC and \(\Delta\)Brier: mean paired difference with the selected DKT; \(k/5\downarrow\): number of splits whose 95\% cluster-bootstrap interval of \(\Delta\)AUC lies below zero. \(\widehat\tau\): median over splits of the predicted gap of the arm divided by that of the selected DKT on the same held-out units (0: gap removed; 1: unchanged). On ASSISTments the group model does not transfer to new schools and the held-out gap is close to zero in some splits, so \(\widehat\tau\) is not interpretable there (Definition~\ref{def:interpretable}). On ASSISTments the audit abstains in every split, so the output correction is shown there only to illustrate what the abstention prevents. The output correction can leave \([0,1]\): 0.3 to 1.0\% of its values on Eedi and 0.4 to 39\% on ASSISTments with the network of this table, and 0.3 to 1.0\% and 0.6 to 25\% with the network selected on separate units (Table~\ref{tab:next-item-v4}). In this table its predictions are clipped to \([0,1]\); the costs of Tables~\ref{tab:next-item-compare} and~\ref{tab:next-item-audit-main} use the unclipped predictor, and Table~\ref{tab:next-item-v4} gives both. The version fitted on training units only is reported in Appendix~\ref{app:next-item}.}
		\label{tab:next-item}
		\resizebox{\linewidth}{!}{%
			\input{tab_next_item.tex}}
	\end{table}

	\begin{table}[t]
		\centering\small
		\caption{Eedi, removals compared with the output correction of Proposition~\ref{prop:built} (mean over five splits; costs in extra Brier score, unclipped). Method: remaining share \(\widehat\tau\) of DKT's predicted gap on test classes, cost, and its cross term \(2\E[(f-y)d]\) (Eq.~\ref{eq:cross-term}). Matched on test: the output correction scaled to leave the same \(\widehat\tau\) on the test classes, and the number of splits in which the paired class-bootstrap interval of the difference (method minus output correction, 200 draws) lies above zero. This uses the test gap to set the intensity, so it is a descriptive frontier, not a procedure. Set on validation: the output correction scaled to leave the method's \(\widehat\tau\) on the \emph{validation} classes, then evaluated on test classes; this is a procedure that can be used. All methods keep the selected network, except \(^{\dagger}\)mean-parity fine-tuning, which retrains it.}
		\label{tab:next-item-compare}
		\setlength{\tabcolsep}{3pt}
		\input{tab_next_item_compare.tex}
	\end{table}

	\paragraph{The selected network is the best of the compared models, and it predicts only part of the gap.} On Eedi the selected DKT has an AUC of 0.719 (0.695 to 0.715 in the worse group), better than the EDMD readout in every split (\(+0.032\), intervals above zero in five of five); on ASSISTments it is better on average (0.676 against 0.659), significantly so in two splits. Like the EDMD model, it never sees the group, and it predicts about 58\% of the observed Eedi gap (\(-0.048\) against \(-0.082\)). The projected EDMD brings the AUC close to chance on both data sets (0.52), as before.

	\paragraph{Erasing a representation does not remove the gap from the prediction.} Two of the four methods keep the selected network and act on its representation or its output; mean-parity fine-tuning retrains it. On Eedi, LEACE on the hidden state leaves 61\% of DKT's predicted gap (57 to 69\% across splits), and erasing it within each next skill still leaves 43\%: the output layer uses the skill of the next item, the skills that pupils meet differ between groups, and the output weights are refitted on the training data. These two methods are cheap (Brier \(+0.008\) and \(+0.010\)) because they remove little. The two methods that act on the prediction itself do remove the gap: fine-tuning leaves 3\% of it and lowers the AUC by \(0.096\), and the output correction leaves 3\% (median over splits) and lowers it by \(0.077\), for an extra Brier score of \(0.021\).

	\paragraph{Which removal costs least?} Among the methods tested, the output correction has the lowest observed mean cost at the same remaining gap (Table~\ref{tab:next-item-compare}): \(0.0033\) where LEACE on the hidden state costs \(0.0077\), \(0.0079\) where LEACE within each skill costs \(0.0098\), and \(0.0200\) where fine-tuning costs \(0.0224\). The paired class-bootstrap interval of the difference lies above zero in all five splits for LEACE on the hidden state, but only in two of five for LEACE within each skill and for fine-tuning; for these two methods the data do not separate the costs. This comparison sets the intensity of the output correction from the gap measured on the test classes, so it describes a frontier and cannot be used as such. In a usable version, the intensity is set on the validation classes to match each method's validation gap: on the test classes the output correction then costs \(0.0028\), \(0.0076\) and \(0.0201\), for remaining gaps of 66, 42 and 2\%, close to the three methods' 62, 40 and 4\%. For the methods that keep the network, Proposition~\ref{prop:built} applies to the same information; fine-tuning changes the representation, so for it the comparison is an empirical reference, not a bound. The cross term of Eq.~\eqref{eq:cross-term} is small for every method (at most \(0.0015\) on average): the corrections hardly interact with DKT's errors. This does not show that DKT is close to the best predictor of its information.

	\begin{table}[tbp]
		\centering\small
		\caption{The three-stage audit of the output correction of DKT, with the network selected on separate units (four sets of units; 400 cluster-bootstrap draws; costs in extra Brier score, unclipped). Announced: total cost estimated on the announcement units, its nominal 95\% interval and the width of this interval. Measured: the cost of the same correction on the test units. Measured \(-\) announced: 95\% interval of the difference, all units resampled. Decision: the diagnostic of Definition~\ref{def:interpretable} (accepted, or abstains with the conditions that fail) and, for an accepted cost, whether the gap reduction transfers to the test units (T1--T2), with the remaining share \(\widehat\tau\) of DKT's predicted gap. The five splits share units. Table~\ref{tab:next-item-v4} compares this audit with the one that uses three sets of units.}
		\label{tab:next-item-audit-main}
		\setlength{\tabcolsep}{3pt}
		\resizebox{\linewidth}{!}{%
			\input{tab_next_item_audit_main.tex}}
	\end{table}

	\paragraph{The cost of the correction, announced before it is used.} For the audit, the units that announce the cost must not also select the network. We therefore divide the validation classes (Eedi) or schools (ASSISTments) of each split into two halves with about the same number of students. The first half, the \emph{selection} units, selects the network on the same grid, stops its training and sets the other removals; the second half, the \emph{announcement} units, only fits \(\widehat c_s\) and the centre, announces the cost and runs D1--D3 (Table~\ref{tab:audit-data}). The training and test units are unchanged, and on Eedi the selected networks have almost the same AUC as in Table~\ref{tab:next-item} (0.715 to 0.723, differences of at most 0.001). Table~\ref{tab:next-item-audit-main} gives, for each split, the results that must be read together: the announced cost and the width of its interval, the decision of the diagnostic, the measured cost, and whether the gap reduction transfers.

	On Eedi the announced total cost is \(0.018\) to \(0.023\) (quadratic part \(0.015\) to \(0.023\)), and the cost measured on the test classes is \(0.017\) to \(0.027\) (mean \(0.021\)). Three results must be kept apart. (i) The diagnostic accepts the estimate in all five splits. (ii) The measured cost lies inside the announced interval in all five, and the interval of the difference between measured and announced cost contains zero in all five; the splits share classes, so these are five related checks, not five independent validations. (iii) The gap reduction transfers to the test classes in four splits only: in the second, the correction overshoots and reverses a fifth of the gap (\(\widehat\tau=-0.21\), T2). Its cost was announced correctly, but on these classes it does not reach its parity target. The intervals are also wide: their width is 1.0 to 2.0 times the announced cost (lower ends \(0.003\) to \(0.014\), upper ends \(0.032\) to \(0.059\)), and they are asymmetric, because the price varies as \(1/\widehat\rho_s^2\) when the group model is refitted. Before use, the audit therefore gives the order of magnitude of the cost, a few hundredths of Brier score, not a precise value. A wide interval that contains the measured cost is a weak check on its own; what supports a decision is the combination of this order of magnitude with the abstention rule and the transfer check.

	When the validation classes both select the network and announce the cost (three sets of units, as in Tables~\ref{tab:next-item} and~\ref{tab:next-item-compare}), the conclusions are the same, with intervals 1.3 to 2.5 times narrower, because about twice as many pupils announce the cost (Table~\ref{tab:next-item-v4}; Table~\ref{tab:next-item-audit} in Appendix~\ref{app:next-item}): the announced cost is \(0.019\) to \(0.025\), the measured cost \(0.017\) to \(0.029\) lies inside the announced interval in all five splits, the diagnostic accepts all five, and the same split fails T2 (\(\widehat\tau=-0.25\)); in that split, the interval of the difference between measured and announced cost lies above zero (\(0.0001\) to \(0.010\)). Clipping the corrected predictions to \([0,1]\), as in Table~\ref{tab:next-item}, changes little on Eedi, where 0.3 to 1.0\% of them leave \([0,1]\): the cost falls by at most \(0.0005\), \(\widehat\tau\) moves by at most \(0.01\), and the clipped cost also lies in the announced interval in all five splits (Table~\ref{tab:next-item-v4}). Proposition~\ref{prop:built} is exact for the unclipped correction only; for the clipped one, these numbers are measured, not given by the formula. The first version of the procedure, used here as a comparator, announced lower costs (\(0.016\) to \(0.018\)) and accepted one split of five (Appendix~\ref{app:v1-procedure}). A simpler group score, the hidden state without the skill term, gives almost the same cost on Eedi (\(0.022\) to \(0.024\) announced, \(0.019\) to \(0.026\) measured). This is a sensitivity check: it shows that the cost does not depend on the skill term, not that the minimum is reached, since both scores could miss the same part of the group signal.

	On ASSISTments the audit abstains in all five splits: on new schools the group score carries no detectable group information (D2) and the price is unstable (D3), and in four splits there is no detectable gap on the validation schools (D1). The announced intervals reach 1 to 92 in Brier score, which says that the cost is not identified. This abstention matters in practice: in one split, the correction built from this group score raises the Brier score by \(0.26\) (\(0.30\) announced; \(0.16\) if its predictions are clipped to \([0,1]\), since 25\% of them leave it), while the simpler score would cost \(0.04\). Applying such a correction blindly would damage the predictions. With three sets of units the audit also abstains in every split (Table~\ref{tab:next-item-audit-assist}).

	\paragraph{Longer histories.} On Eedi students with at least 101 items, we predicted the same 25 target items from windows of 25, 50 and 100 previous items (two splits; Table~\ref{tab:next-item-long} in Appendix~\ref{app:next-item}). A longer history makes the network better (AUC 0.711 with 25 items, 0.740 with 50 and 0.746 with 100) and lets it predict more of the gap without seeing the group (in one split, \(-0.063\), \(-0.087\) and \(-0.087\) for an observed \(-0.100\)). Erasure therefore becomes more expensive, not cheaper: the cost of the output correction measured on test classes rises from \(0.031\) with 25 items to \(0.034\)--\(0.035\) with 50 or 100 (the first version of the procedure gives the same ordering; Table~\ref{tab:next-item-long}). Longer histories also make the cost easier to audit: with the three-stage procedure, the estimate is accepted and the gap reduction transfers in all six cases (two splits, three history lengths), the measured cost lies in the announced interval in all six, and the interval of the difference contains zero in all six (announced total \(0.031\) to \(0.035\), measured \(0.031\) to \(0.035\) on average; Table~\ref{tab:next-item-long-v2}). The short windows of the main study therefore understate, rather than overstate, the cost of erasure, and the audit works best where the network has the most information.

	\subsection{Correcting the dynamics only}
	\label{sec:results-dynamics-only}
	A different correction leaves the state alone and removes only the effect of the group on the dynamics: a group-aware model \(z_{t+1}=Az_t+a+(\grp-p_1)(A_{\grp}z_t+a_{\grp})\), fitted on the training students, is used with the group term set to zero. Two students in the same state then receive the same forecast. On the three data sets this keeps the one-step error of the raw model, but forecast from the start of the 25-step window it predicts almost none of the gap that develops (\(+0.003\) against an observed \(+0.057\) on ASSISTments, \(-0.007\) against \(-0.085\) on Eedi), whereas the group-aware model forecasts it (\(+0.067\) and \(-0.079\)); 91--95\% of the gap that the group-aware model forecasts comes from its group-specific dynamics (Table~\ref{tab:dynamics-only} in Appendix~\ref{app:more-real}). The correction makes a choice between two kinds of blindness explicit: erasing the group from the state hides the gap that exists now and costs accuracy, while removing it from the dynamics keeps accuracy but hides how the groups drift apart from similar states.

	\section{Consequences for decisions: simulation}
	\label{sec:results-policy}
	\subsection{Removing the group direction does not help the controller}
	\label{sec:results-regimes}
	We tested 28 settings, each with 50 replicates; every count below uses a Holm correction over the 28 settings at the 5\% level, and Table~\ref{tab:replicates50} in Appendix~\ref{app:regimes} gives every setting. On knowledge, the primary outcome, the projection lowers the final knowledge of one group in all 24 settings where the group affects the dynamics, and the mean knowledge in 22 of the 28 settings. It widens the knowledge gap \(|\delta^k_T|\) in 19 settings and narrows it in five, but these five are not improvements: with \(\beta_{\mathrm{ind}}=32\) the knowledge gap narrows because the group that is ahead loses \(0.38\) to \(0.49\), and with \(\beta_{\mathrm{ind}}=0\) it narrows by at most \(0.03\) while knowledge falls. On the disparity gap, the projection never helps: it widens \(|\delta_T|\) in 24 settings (negligibly, by at most \(0.0012\), in the three with \(\gamma=0\)) and changes nothing in the four with \(\beta_{\mathrm{ind}}=0\). The harm grows with the group strength and with \(\beta_{\mathrm{ind}}\) (at \(\gamma=1\), \(+0.024\), \(+0.136\) and \(+0.287\) for \(\beta_{\mathrm{ind}}=1\), 8 and 32), hardly depends on the rank or the depth, and holds at \(T=50\) (Appendix~\ref{app:t50}). Table~\ref{tab:regimes} (Appendix~\ref{app:block-simulator}) summarizes these results; Appendix~\ref{app:main-setting} walks through the main setting.

	\paragraph{Comparators.} Forcing the predicted gap to zero (Eq.~\eqref{eq:gap-constrained}) narrows \(|\delta_T|\) in all 24 settings with a group effect, but in 23 of them it lowers the knowledge of one group: it closes the gap by \emph{levelling down} (Definition~\ref{def:levelling-down}). It widens the knowledge gap in all 28 settings; in the main setting the knowledge gap grows from \(0.19\) to \(0.98\), and group~0 loses \(0.564\). It also usually uses more help (median 1.39 times the \(\mathcal U_1\) of the original MPC). The gap penalty behaves like the projection: the original model's long-range forecast of the final gap is never larger than \(0.008\), so a penalty with \(\beta_{\mathrm{gap}}=\beta_{\mathrm{ind}}\) gives almost no reason to act, and it narrows the gap only from \(\beta_{\mathrm{gap}}=4096\), again by lowering group~0 (Appendix~\ref{app:regimes}). Changing only the forecast afterwards sets the predicted gap to zero and changes nothing for the students.

	\paragraph{Group direction or compression?} Run inside the controller, the rank-matched controls of Section~\ref{sec:results-rank-matched} separate two effects (Figure~\ref{fig:rank-matched-policy} in Appendix~\ref{app:regimes}): at \(\gamma\le1\) the group direction is what widens the gap, while most of the loss of knowledge comes from compression; Section~\ref{sec:results-oracle} shows where this loss comes from.

	\subsection{A reference that uses the known equations: erasure, model error and the choice of objective}
	\label{sec:results-oracle}
	The comparisons above use learned models only, so a loss can come from the erasure, from the error of the learned model or from the objective of the controller. To separate them, we add a reference controller that plans with the simulator's \emph{known equations}, Eq.~\eqref{eq:simulator}, with the group or with the group replaced by its mean. It is not an exact oracle, for three reasons that we state once: it plans with the noise-free trajectory of the equations, clipping included, which is not the expected next state because \(\E[\clip(a+\varepsilon)]\neq\clip(a)\); it solves its planning problem by a grid search (\(11^3\) plans, then a local \(5^3\) grid of step \(0.05\)), not exactly; and the budgets are matched approximately. The learned arms solve their quadratic plans exactly. All arms observe the full state and plan three steps ahead with the same objective. Each arm's control weight \(\rho\) is chosen by bisection on two validation cohorts so that its mean absolute help \(\mathcal U_1\) equals that of the reference at \(\rho=1\); on the evaluation cohort, the realized help is within 0.5\% of the reference for every arm except at \(\gamma=2\) with the mean-knowledge objective (within 2.2\% and 4.2\% for the reference without the group and the random removal). We also report every arm at \(\rho=1\). These approximations matter little for the comparisons (Appendix~\ref{app:oracle}): with the objective of the paper, a finer search changes the reference's final knowledge by \(-0.0002\), and planning with the expected noise by \(-0.002\). With the mean-knowledge objective, which is nearly flat in the amount of help, a finer search gives 8\% less help and \(0.054\) less knowledge, so the reference's own level depends on the search; the comparisons at equal budget do not, because every arm is matched to the reference's budget. We use two objectives, mean knowledge (\(\beta_{\mathrm{ind}}=0\)) and the objective of the paper (\(\beta_{\mathrm{ind}}=8\)), three group strengths and 20 replicates. Table~\ref{tab:oracle} gives \(\gamma=1\); Appendix~\ref{app:oracle} gives every setting, with intervals.

	\begin{table}[t]
		\centering\small
		\caption{Controllers against the reference of the simulator (\(\gamma=1\), 20 replicates). Approximately equal budget: each arm's control weight \(\rho\) is chosen on validation cohorts so that its mean absolute help equals that of the reference at \(\rho=1\); final mean knowledge, overall and in each group, and absolute knowledge gap on the evaluation cohort. Last columns: the same arms at \(\rho=1\), with the help \(\mathcal U_1\) they give.}
		\label{tab:oracle}
		\resizebox{\linewidth}{!}{%
			\input{tab_oracle.tex}}
	\end{table}

	\paragraph{With the mean-knowledge objective, erasure costs nothing at equal budget.} At \(\gamma=0.5\) and \(1\), the six controllers reach the same final knowledge to within \(0.001\) once they spend the same budget: the best plan gives nearly the same help to every student, and every model finds it. At \(\rho=1\), in contrast, the learned models give less help than the reference (mean \(|u|\) of \(0.20\) for the original model, \(0.17\) after erasure and \(0.12\) after a random removal, against \(0.30\)), because they underestimate the effect of help, and they end \(0.16\) to \(0.40\) lower in final knowledge. In this case the loss of knowledge of an erased controller is a budget effect, not a loss of information. At \(\gamma=2\) this no longer holds: every learned model ends \(0.13\) below the reference, a model error, and erasure adds nothing to it.

	\paragraph{With the objective of the paper, erasure has a real cost.} At equal budget, the learned original model is close to the reference (\(0.002\), \(0.009\) and \(0.057\) below it at \(\gamma=0.5\), 1 and 2), but erasure lowers final knowledge by \(0.020\), \(0.050\) and \(0.115\) (intervals above zero), and it widens the knowledge gap (from \(0.29\) to \(0.42\) at \(\gamma=1\)). A random removal of the same rank lowers knowledge even more at \(\gamma\ge1\) (\(0.12\) and \(0.26\), with wide intervals over the random draws) but widens the knowledge gap less. Matching the budget removes between a third and 84\% of the loss of the erased controller measured at \(\rho=1\): part of the loss reported in Section~\ref{sec:results-regimes} comes from the smaller help that a compressed model gives.

	\paragraph{The objective decides who is helped, even with the reference.} With the mean-knowledge objective, the reference controller leaves group~1 behind by \(0.51\) at \(\gamma=1\), and at \(\gamma=2\) it stops helping group~1 altogether (its final knowledge is \(0.00\)): maximizing the mean sacrifices the group that learns less from each unit of help. With the objective of the paper, the individual disparity penalty sends help to group~1, which drifts in disparity, and the reference overshoots: group~1 ends \(0.20\) ahead in knowledge. Knowing the group does not help either objective protect the group behind (without the group, the reference leaves a knowledge gap of \(0.06\) under the objective of the paper). Neither objective aims at the knowledge gap; this is why the next controller does.

	So, in prediction, the group direction explains most of the cost of erasure (Section~\ref{sec:results-rank-matched}); in control, a large part of the loss of knowledge comes from elsewhere: from the smaller help that a compressed model gives at a fixed control weight, from the extra compression, and from the objective itself.

	\subsection{Aiming at knowledge, with floors}
	\label{sec:results-pareto}
	The floor-constrained controller of Section~\ref{sec:pareto-fair} narrows the predicted \emph{disparity} gap. Because knowledge is the primary outcome, we also run it with the knowledge gap as its target: the predicted gap in the controller's linear program and constraint is computed from predicted knowledge instead of the disparity state, and nothing else changes. Table~\ref{tab:knowledge-controller} in Appendix~\ref{app:block-simulator} compares, on the same group-aware model, cohorts, noise and budget, the two targets with and without the floors, and a targeted top-up for the group behind in knowledge, with the mean absolute control \(\mathcal U_1\) of the knowledge-targeting controller. So every arm has the same model and the same access to the group; only the objective, the floors and the allocation change.

	Aimed at the disparity gap, the controller with floors does what it was built for, but it \emph{widens the knowledge gap} by \(0.28\), \(0.21\) and \(0.09\) at \(\gamma=0.5\), 1 and 2: it raises the knowledge of group~1, which is behind in disparity but ahead in knowledge, and leaves group~0 unchanged. Without floors it also lowers group~0 by \(0.39\) to \(0.55\). Aimed at the knowledge gap, the controller with floors narrows it by \(0.10\), \(0.18\) and \(0.15\), by raising group~0 (by \(0.11\) to \(0.19\)) while group~1 also gains slightly, with \(9\)--\(14\%\) more absolute control; no group loses (largest floor violation \(1.3\times10^{-9}\)). Without floors, the same target narrows the knowledge gap as much at the reference budget, but by lowering group~1 by \(0.05\) to \(0.09\): this is levelling down. At the same budget, the targeted top-up does as well as the controller with floors at \(\gamma\le1\) (difference in the absolute knowledge gap at most \(0.003\)) and better at \(\gamma=2\) (\(-0.24\) against \(-0.15\)).

	Three conclusions follow. The choice of target matters more than the controller: narrowing the synthetic disparity gap can widen the knowledge gap. Floors are what separates narrowing a gap from levelling down. And, as with the disparity target, the gain comes from directing extra help to the group behind, which a simple top-up also does; the floor-constrained controller is a reference design for auditing, not a method shown to be better.

	\paragraph{Which budget.} The comparisons of Table~\ref{tab:knowledge-controller} hold the absolute help \(\mathcal U_1\) fixed, the budget of Proposition~\ref{prop:no-levelling}. The top-up then uses 1--8\% less quadratic control \(\mathcal U_2\) than the controller with floors (for example 1.14 against 1.18 times the reference at \(\gamma=1\)), so the comparison does not favour the top-up on quadratic cost either. Appendix~\ref{app:disparity-controller} reports the controller aimed at the disparity gap over a range of budgets and against a top-up matched on both \(\mathcal U_1\) and \(\mathcal U_2\) (Table~\ref{tab:equal-budget}), and in three scenarios not used to design the method.

	\section{An audit protocol and practical recommendations}
	\label{sec:general-diagnostic-protocol}
	The results above show that a correct construction does not ensure accurate predictions, and that a smaller predicted gap does not ensure better outcomes. The audit therefore asks three separate questions, one per level (Figure~\ref{fig:validation-workflow} in Appendix~\ref{app:margins}), and Table~\ref{tab:protocol} lists for each level the evidence that must be reported, the warning signs, and the methods to compare with. A \emph{diagnostic flag} is such a warning sign: a measurable condition which, when it is met, tells the auditor that the claim made at that level is not supported by the evidence. A flag does not by itself reject a model; it says where to look. A negative result is useful when it shows which claim fails; a small predicted gap must not be counted as a success.
	\begin{enumerate}
		\item \textbf{Structure.} Write down the state, the change of coordinates, the constraints and the change made to the model. Then test the identities that follow from them.
		\item \textbf{Prediction.} With the same information and the same horizon, compare the predictions with observed outcomes that the model has not seen. Report the group gap together with the error in each group. Compare with simple predictors trained on the same data, and compute intervals at the level of the real sampling unit (for example, the school).
		\item \textbf{Policy.} Compare what really happens to students under policies built on the model. Use paired comparisons, report results for all students and for each group, and compare with controllers that target the gap directly.
	\end{enumerate}

	\begin{table}[t]
		\centering\small
		\caption{The three levels of the audit. For each level: the question it answers, what must be reported, the diagnostic flag (the warning sign that the claim is not supported), and the methods to compare with. The symbols are defined in Section~\ref{sec:method} and recalled below the table. A flag tests against zero; deciding whether an error matters in practice needs the margins of Table~\ref{tab:margins}.}
		\label{tab:protocol}
		\begin{tabularx}{\linewidth}{@{}>{\RaggedRight\arraybackslash}p{2.1cm}*{3}{>{\RaggedRight\arraybackslash}X}@{}}
			\toprule
			Level and question & What must be reported & Diagnostic flag (warning sign) & Methods to compare with \\
			\midrule
			1. Structure: \emph{is the change built as specified?} & Duality, idempotency and metric-duality errors; seed residual \(\epsilon_q\); cross-block residuals \(r_1,r_2\); leakage \(\|R^\top B_P\|\), \(\|R^\top a_P\|\); requested, candidate, applied and numerical ranks; singular-value cutoff; the space (centered or augmented) where they are measured & an identity that should hold is broken beyond numerical error, or the removed direction comes back through the action or constant term & the original model, fitted on the same data \\
			\addlinespace
			2. Prediction: \emph{does the model still predict unseen students?} & Errors at several horizons, for each variable and each group; predicted and observed gaps, with intervals at the level of the sampling unit; paired differences in error & the interval of the paired gap error excludes zero, or the model predicts worse than persistence & persistence; a linear model of the state; a nonlinear model trained on the same data; a recalibration for each group \\
			\addlinespace
			3. Policy: \emph{does a tutor using it help both groups?} & Paired effects on the gap, on all students and on each group, with the worst-group change; the control budget; the uncertainty target and the calibration assumptions; the settings where the result holds & the gap, or the outcome of one group, gets worse than with the original controller & the original controller; a controller that penalizes or limits the gap; changing only the predictions afterwards \\
			\bottomrule
		\end{tabularx}
		\par\smallskip\raggedright\footnotesize
		\emph{Symbols.} \(\epsilon_q\): the part of the fitted group contrast that is left after projection (zero if the group direction is fully removed). \(r_1=\|PKR\|_{\Gamma,F}\) and \(r_2=\|RKP\|_{\Gamma,F}\) (Eq.~\eqref{eq:diagnostics}): the \emph{cross-block residuals}, which measure how much the fitted dynamics move the removed part into the kept part (\(r_1\)) and the kept part into the removed part (\(r_2\)); both are zero exactly when the two parts do not mix. Leakage: the part of the action and constant terms that falls back into the removed directions. Persistence: the prediction ``the next state equals the current state''.
	\end{table}

	\paragraph{When does an error matter?} A flag of Table~\ref{tab:protocol} says that an error can be detected, not that it is large enough to matter. For this, the auditor fixes margins before looking at the results: an absolute and a relative tolerance on the gap, \(\epsilon_\Delta=\max\{\epsilon_{\mathrm{abs}},\epsilon_{\mathrm{rel}}|\delta_{\mathrm{ref}}|\}\), and a tolerance on the increase of the error; a result is then acceptable, an error that matters, or inconclusive (Appendix~\ref{app:margins}). Applied after the fact, with \(\epsilon_{\mathrm{abs}}=0.01\) and \(\epsilon_{\mathrm{rel}}=10\%\), to the 18 held-out splits with a clear observed gap, the error on the gap matters for the projected model in 14 splits and is inconclusive in 4, while the per-group recalibration is acceptable in 15 and inconclusive in 3. These decisions illustrate the rule; they are not an acceptance test (Section~\ref{sec:protocol}).

	\paragraph{Recommendations.} Our results suggest the following practice for a fairness change to a learned model of learning.
	\begin{enumerate}[leftmargin=*,itemsep=2pt,topsep=3pt]
		\item \emph{Estimate the cost before erasing, in three stages.} Fit the predictor and a group model with the same information on training units; announce the cost of the correction and run the diagnostic on validation units (Section~\ref{sec:price-estimation}); test on independent units. A large built cost has two possible sources (Proposition~\ref{prop:built}): the group signal is the predictive signal, and erasure will hide a real gap; or the group score misses part of the group signal, and the minimum is lower. Only the first says that the minimum is large. A more flexible group model with the same information is a sensitivity check, not a test: a lower cost shows that the first model was too weak, but an unchanged cost does not prove that the minimum is reached, because both models may miss the same signal.
		\item \emph{Report the cost with its width, the decision and the transfer.} Give together the announced cost and the width of its interval, the decision of the diagnostic, the measured cost, and whether the gap reduction transfers (Table~\ref{tab:next-item-audit-main}). A measured cost inside a wide interval is a weak check, and a cost that was announced correctly does not show that the correction reaches its parity target.
		\item \emph{Treat the diagnostic as a screen.} When it abstains, report the measured cost on test units, not the formula, and do not read the abstention as evidence that there is no gap. Use a flexible group model: by Proposition~\ref{prop:built}, for the best predictor, a group model that misses part of the group signal gives a cost above the minimum, even when it is calibrated.
		\item \emph{Remove the contrast from the prediction, not from one representation}, if it must be removed. Erasing a hidden state leaves the group signal in the other inputs of the prediction (Section~\ref{sec:results-next-item}).
		\item \emph{Separate the prediction from the decision.} Keep a model that predicts each group well, for example with a correction for each group, and put the fairness requirement into the decision rule, with floors on each group's outcome.
		\item \emph{Aim at the outcome that matters and state the budget.} Narrowing a synthetic disparity gap can widen the knowledge gap; compare any fair controller with a simple top-up for the group behind at the same budget, and say which budget (\(\mathcal U_1\) or \(\mathcal U_2\)) is held fixed.
		\item \emph{Compare with the best available model.} In simulation, a reference that uses the known equations separates the effect of erasure from model error and from the choice of objective (Section~\ref{sec:results-oracle}).
	\end{enumerate}

	\section{Discussion}
	\label{sec:discussion}
	This study asks what can and cannot be concluded from one change to a learned model; it does not propose a new definition of fairness. One pattern holds across the simulator and the three real data sets: a correct construction does not ensure accurate forecasts (Appendix~\ref{app:structural} documents the identities for all 650 simulator fits and every real-data fit), and a smaller predicted gap does not ensure better outcomes.

	\paragraph{What the price explains, and when to trust it.}
	The loss of accuracy is not a general cost of compression: two reductions of the same rank that do not use the group cost nothing, and random directions cost much less. It comes from the group direction, because in these data the prediction is strongly aligned with the group signal, and the price \(p_0p_1\Delta^2/R^2_{\grp}=\operatorname{Var}(f^*)\operatorname{corr}(f^*,q)^2\) says so before any intervention. What an audit can estimate is the cost of the correction it can build. For the best predictor, this cost is at least the minimum (Proposition~\ref{prop:built}). For a learned predictor it is not: its quadratic part is only bounded below by \(p_0p_1\Delta(f)^2/R^2_{\grp}\), which uses the gap of the learned predictor, and the cross term can make the total cost lower or higher; what remains true is that the corrected predictor is worse than the best one by at least the minimum (Section~\ref{sec:price-estimation}). In the controlled study, the cost of the fitted correction was estimated accurately, it was conservative as an estimate of the minimum when the group model missed part of the group signal, and the diagnostic abstained when there was nothing to remove; a change of the group signal between training and use was caught only on independent test units. On Eedi, with the network selected on separate classes, the cost announced on validation classes contained the cost measured on new classes in all five splits, but the intervals were one to two times as wide as the cost, and the gap reduction transferred in four splits of five; on new ASSISTments schools the audit abstained, rightly, because a correction built from that group model would have damaged the predictions. The diagnostic is a screen with conventional thresholds, not a test with known error rates, and its intervals are approximate: in the controlled study, with 1,000 bootstrap draws, the nominal 95\% interval contained the realized cost in 90 to 100\% of the accepted replicates (both learners and both sample sizes; 88 to 100\% with 60 draws).

	\paragraph{Prediction and decision are different tasks.}
	Equal mean predictions are a contestable goal for a tutor (Section~1): a model that predicts equal means for groups that do not perform equally is wrong for at least one of them. Our results support separating the two tasks. A correction for each group keeps the accuracy and predicts the observed gap; erasure in the representation of a knowledge-tracing network does not even remove the gap from the prediction, because the prediction also uses the skill of the next item; and removal at the output, which does remove it, costs about what the audit announced for it. On Eedi, removing the contrast from the selected network costs \(0.021\) in Brier score and \(0.077\) in AUC when it is done at the output with the three-stage audit, and \(0.022\) and \(0.096\) when it is imposed during training.

	\paragraph{What the simulator shows, and what it does not.}
	In prediction, the group direction explains most of the cost. In control, a large part of the loss of knowledge comes from elsewhere: against a reference controller that uses the simulator's known equations, at equal help and with a mean-knowledge objective, erasure costs no knowledge at all, because the main fault of a compressed model is to give less help at a fixed control weight; with the objective of the paper, erasure still costs \(0.02\) to \(0.11\), and a random compression of the same rank costs as much or more. The objective matters most: even the reference leaves group~1 far behind when it maximizes mean knowledge, and overshoots when it penalizes individual disparity. Aimed at knowledge with floors, a controller narrows the knowledge gap without levelling down, but only with extra help, which a simple top-up for the group behind matches at the same absolute budget; aimed at the synthetic disparity gap, the same controller widens the knowledge gap. These results hold across ranks, depths, group strengths and three scenarios not used to design the method, but they come from one simulator family that we built.

	\paragraph{What the real-data results can and cannot say.}
	These are prediction comparisons, not estimates of tutoring effects. ASSISTments uses school-performance groups defined after the fact, so it tests the mechanism; Eedi describes the pupils whose status is recorded (reweighting to all pupils does not change the conclusion, under an assumption that cannot be checked); OULAD does not record when a score becomes known (delays of up to four weeks do not change the conclusion). Rolling and cumulative proxies change slowly, which inflates error ratios: on the OULAD weeks with a new score the projected model's error is 1.5 to 1.7 times the original, not seven to eleven, and on next-item correctness its Brier score is 7 to 11\% higher; but in every case it still hides the gap, and its next-item AUC is close to chance.

	\paragraph{Limits and next steps.}
	Several analysis choices were made after seeing results (Section~\ref{sec:protocol}). Repeated splits describe variation in the analysis, not independent samples, and the population-transfer terms of Appendix~\ref{sec:theory} are not estimated, so the operator bounds do not certify performance with real students. Proposition~\ref{prop:no-levelling} holds for one step and one kind of action. The exploratory analysis of 30 randomized experiments (Appendix~\ref{app:rct}) tests a different decision, choosing between two existing conditions, and finds no detectable narrowing of a completion gap (\(-0.5\) points, interval \(-1.5\) to \(+0.7\)); it does not test the projection, a sequential controller or extra help. Two questions stay open: a method that limits the prediction error of each group separately, and a model-based controller that does better than a simple top-up at the same budget. No experiment uses a language model or a tutor deployed with real students. A smaller predicted gap is a signal to investigate, not evidence of fairness.

	\section{Conclusion}
	Erasing a group signal from a learned model of learning has a price, and the price can be computed before any intervention: the squared gap that the best model predicts, divided by how well the state explains the group, or equivalently the variance of the best prediction times its squared correlation with the group signal. What can be announced before an intervention is the cost of the correction built from a group model; for the best predictor it is at least this minimum; for a learned predictor it can be lower or higher, because it starts from the gap of the learned predictor and adds a cross term. A three-stage audit announces this cost on validation units, abstains when the data do not support an estimate, and checks on independent units both that the gap reduction transfers and that the measured cost agrees with the announced one. Where the prediction is strongly aligned with the group signal, the price is high and erasure hides the gap without closing it: on three real data sets the erased model predicts almost no gap where one is observed, the loss comes from the group direction, and on next-item correctness no way of removing the contrast from a selected knowledge-tracing network is free. For decisions, the cost of erasure depends on the help budget and on the objective: at equal help with a mean-knowledge objective it costs nothing, and the knowledge gap is decided by the objective; a controller aimed at knowledge with floors on each group narrows the gap without levelling down, but only with extra help, which a simple top-up matches. The practical lesson is to compute the price before erasing, to keep the fairness requirement in the decision rather than in the prediction, and to judge any fairness change at three separate levels: what the model suppresses, how well it predicts, and what a policy using it does, with the resources it needs.

	\section*{Broader impact statement}
	This work studies fairness changes to learning systems that make decisions about students step by step. Its main message is a warning. A model that hides a group gap can look fair while it predicts worse and changes nothing for students. And a controller that closes a gap can do it by making one group worse off. The floor-constrained controller, which we report as a reference design, avoids a final group-mean knowledge loss in the tested simulator replicates, but this is not a guarantee for individual students or deployment, and it uses group labels when it decides. Whether this is legal and acceptable depends on the educational context, not on our method. The real-data studies use public, licensed data sets. On ASSISTments, the groups come from school performance and have no demographic meaning. On Eedi, the group is the pupils' recorded financial disadvantage, and on OULAD it is area deprivation, disability or gender. We use these attributes only to measure gaps in predictions, never to make decisions about students. In the check on randomized experiments, the groups are levels of prior achievement, and the policies are only evaluated on past data. No policy of ours was used with real students, and no experiment involves a deployed tutor.

	\paragraph{Who sets the group label, and what if it is wrong?}
	Group-aware controllers, and the recalibration for each group, need a group label when they decide. Someone must define it: the institution, the teacher, the student, or a public register, and this definition decides who counts as ``behind''. Labels can be missing, outdated or wrong (on Eedi, the attribute is recorded for 29\% of pupils), and a wrong label sends extra help to the wrong students. A deployment should therefore document where the label comes from, how often it is wrong, and how students or families can correct it, and it should report results for students without a label.

	\paragraph{Legal context.}
	Using a group attribute to decide is not only a technical choice. In the European Union, data about health, including disability, are a special category under the General Data Protection Regulation, and AI systems that evaluate learning outcomes or steer the learning process are classified as high-risk under the AI Act, with duties of risk management, data governance, human oversight and documentation. In other places, using a protected attribute in individual decisions can be restricted, while some laws require targeted support for disadvantaged students. A method that needs the label may thus be required in one context and not allowed in another. Our results do not settle this question. They show the cost of each option: removing the group from the model hides real gaps, and using it to target help needs the label and a decision about who receives the extra resources.

	\section*{Data and code availability}
	OULAD is available from the UCI Machine Learning Repository under a CC BY 4.0 licence \citep{kuzilek2017open}. The Eedi data are available from the NeurIPS 2020 Education Challenge page under a CC BY-NC-ND 4.0 licence \citep{wang2020diagnostic}; we do not share them or any file derived from them. The ASSISTments 2012--2013 data can be obtained from the ASSISTments data portal under its terms of use; we do not share them either. The randomized-experiment data \citep{sales2023auxiliary} are available from the Open Science Framework (project \texttt{k8ph9}) under the ASSISTments terms of use; we do not share them. The reproduction package is published, without the data, as release \texttt{v1.6} of \url{https://github.com/SostheneHB/Gen-AI-Tutoring}, in the folder \texttt{reproduction/}. It contains the scripts that recompute every number, table and figure: feature construction, the pipelines for the three real data sets, the simulator and the synthetic experiments, the regime study and the comparators, the structural test (Appendix~\ref{app:structural}), the sample flow (Table~\ref{tab:flow}), the floor-constrained MPC and its budget frontier (Figure~\ref{fig:frontier}), the rank-matched controls (Section~\ref{sec:results-rank-matched}), the equal-budget baselines, the solver certification, the scenarios not used to design the method, and the 50-replicate study. It also contains a locked environment file and the commands to run, with the seeds (root seed 20260914). Numerical outputs are not distributed, because some are derived from data that may not be redistributed; the scripts recompute them. The release also contains the scripts for Proposition~\ref{prop:general-price}, the dynamics-only correction, the next-item study with deep knowledge tracing and the 50-replicate simulator study, and, in \texttt{reproduction/randomized\_experiments/}, the analysis of the randomized experiments with its analysis plan and the code as it was when the plan was written. Release \texttt{v1.6}, used for this version of the paper, is archived on Zenodo, \href{https://doi.org/10.5281/zenodo.24147314}{doi:10.5281/zenodo.24147314}; it adds to release \texttt{v1.5} (\href{https://doi.org/10.5281/zenodo.23362913}{doi:10.5281/zenodo.23362913}) the controlled study with 1,000 bootstrap draws and the next-item audit with four sets of units. Release \texttt{v1.5} adds to release \texttt{v1.4} (\href{https://doi.org/10.5281/zenodo.23359853}{doi:10.5281/zenodo.23359853}) the audit with explicit targets, its three separate intervals and the bootstrap stability check. Release \texttt{v1.4} contains the scripts of the controlled study, the three-stage audit, the next-item study with the selected knowledge-tracing network, the reference controller that uses the known equations, and the data checks. We tested the package in a new environment built from its environment file (Python 3.12, with the pinned package versions): run from the package alone, a real-data script reproduces our numbers to within \(10^{-11}\) and a simulator script to within \(3\times10^{-8}\).

	\bibliography{Gen,journal_references}
	\bibliographystyle{plainnat}
	
	\clearpage

	\appendix

	\section{Theory, construction and related work}
	\label{app:block-theory}
	\subsection{Extended related work}
	\label{app:related}
	Each paragraph below states what existing work provides, what it leaves open for our setting, and what this paper adds; Table~\ref{tab:positioning} summarizes the comparison.

	\paragraph{Fairness over time.}
	A fairness rule can look good at one moment and still cause harm later, once the system reacts to it \citep{liu2018delayed}. Simulation studies show the same thing for many kinds of feedback loops \citep{damour2020fairness}. \citet{jabbari2017fairness} study fairness over time through long-term reward, and \citet{alamdari2024remembering} define fairness properties that depend on the whole history. \citet{wen2021algorithms} model the feedback effects of decisions as Markov decision processes, define fairness properties for this setting and give algorithms that learn fair policies. \citet{rateike2024designing} design group-fair policies for a known Markov chain, so that the system reaches a chosen fair state in the long run. In education, \citet{tschiatschek2022equity} show with Bayesian knowledge tracing that predictions that are equally accurate for two groups can still produce inequitable tutoring policies, and a recent audit of deep knowledge-tracing models finds a socioeconomic bias on the Eedi data that reweighting and adversarial debiasing do not remove when accuracy is kept \citep{minh2026accuracy}. \emph{What remains open:} these works either assume a known model of the process or evaluate one level only, prediction or policy. \emph{What we add:} a model learned from data over a finite horizon, an intervention on that model, and a check at three levels; our results show the same separation between prediction and policy.

	\paragraph{Fair representations by projection.}
	Learning fair representations \citep{zemel2013learning}, removing bias directions from embeddings \citep{bolukbasi2016man} and repeated null-space projection \citep{ravfogel2020null} are the ideas behind removing a group contrast. LEACE erases a linear concept in closed form with the smallest change to the representation under its own criteria \citep{belrose2023leace}. Trade-offs proved for classification \citep{zhao2019inherent} motivate caution, but they are proved for one prediction. \emph{What remains open:} all these methods act on a static representation, which is computed once and then used. In a model of dynamics, the representation is pushed forward at every step by the model's update rule, so removing a direction once is not enough. \emph{What we add:} our construction removes a subspace built from the group contrast \emph{and} from the update rule of the model together with its adjoint, so that the dynamics regenerate as little as possible of the removed direction; the residuals \(r_1,r_2\) measure what remains. Section~\ref{sec:results-rank-matched} compares it with LEACE and with rank-matched controls; at rank one, with the sample covariance as metric, our projection and LEACE followed by refitting give the same model.

	\paragraph{Fair regression and fairness through unawareness.}
	Dropping the group from the inputs, ``fairness through unawareness'', does not remove its influence when other features carry it \citep{dwork2012fairness,kusner2017counterfactual}. Fair regression constrains the predictions instead. \citet{calders2013controlling} fit linear regressions under constraints on group mean differences, and \citet{chzhen2020fair} and \citet{legouic2020projection} characterize the best regressor under full demographic parity, through Wasserstein barycentres of the group-wise prediction distributions, for predictors that may use the group. \citet{wei2023mean} solve mean-parity regression in closed form for a predictor that uses the attribute, and \citet{petrulionyte2026directly} train predictors of the non-sensitive features under a mean-parity penalty. \emph{What remains open:} the exact minimum cost of removing the mean group gap from a predictor that uses only the state, how this cost depends on the data, and whether it can be used to diagnose a learned model before an intervention. \emph{What we add:} for a binary group and an exact constraint, Propositions~\ref{prop:erasure-price} and~\ref{prop:general-price} give this cost in closed form, in two equivalent forms (the squared predicted gap divided by how well the state explains the group, and the variance of the best prediction times its squared correlation with the group signal), together with a bound and its horizon profile in a learned dynamic model.

	\paragraph{Fairness in models of students.}
	\citet{doroudi2019fairer} show that knowledge-tracing models used for mastery learning can be inequitable for students who learn at a different rate from the one the model assumes, and that more individualized models are fairer but still not fully equitable. This is the closest work to ours in education: in both cases, the model of learning decides who receives practice and help. \emph{What we add:} we study the opposite remedy, removing group information from the model of learning, and show what it costs.

	\paragraph{Levelling down.}
	Fairness methods can narrow a gap by making the better group worse off. This is called ``levelling down'' \citep{zietlow2022leveling,maheshwari2023fair}. Floors on the outcome of each group are an alternative \citep{mittelstadt2023unfairness,gultchin2023pragmatic}. \emph{What remains open:} floors say what must not happen, but not whether a gap can still be narrowed once they are imposed, nor at what cost; this question has not been answered for a controlled model of learning. \emph{What we add:} Proposition~\ref{prop:no-levelling} answers it for one step. When an action has the same positive effect on both groups and the reference actions are not negative, the gap cannot be narrowed at the same absolute-control budget while both floors hold; with extra budget, a reduction becomes possible when the floors leave slack. Our floor-constrained MPC applies these floors to the predicted mean knowledge of each group at the next step. The result is not an impossibility result when the groups respond differently, when there are several kinds of action, or for every policy over a long horizon.

	\paragraph{Constrained control and Koopman models.}
	Constrained Markov decision processes \citep{altman1999constrained} put a fairness or safety rule into the control problem, not into the model, and \citet{korda2018mpc} use linear Koopman predictors inside model predictive control (MPC). Residual-based EDMD \citep{colbrook2024rigorous} controls errors in the spectrum of Koopman approximations. \emph{What remains open:} these tools have not been used to compare a change to the model with a fairness rule in the controller. \emph{What we add:} we compare our model change with MPC controllers that penalize or limit the predicted gap, on the same data and noise. Our residuals \(r_1,r_2\) measure how much the reduced dynamics still mix the kept and removed parts; they play a role similar to the residuals of \citet{colbrook2024rigorous}, but they do not give the same spectral guarantees.

	\paragraph{Prediction coverage.}
	Split-conformal prediction gives guarantees for finite samples, on average, when the data are exchangeable \citep{lei2018distribution,angelopoulos2023conformal}, and its behaviour when they are not is studied by \citet{barber2023beyond}. \emph{What remains open:} coverage of a group gap for a new cohort under a fixed tutoring policy. \emph{What we add:} a narrow use of conformal prediction to check the calibration of the next cohort's gap under one policy, not as evidence of fairness.

	\paragraph{What is standard and what is new.}
	Proposition~\ref{prop:optimality}, Theorem~\ref{thm:horizon}, Proposition~\ref{prop:amplification-regimes} and the strong-limit argument (Appendix~\ref{sec:theory}) use standard mathematical tools; we include them to make the assumptions clear, not as new general results on operators. The new elements are the three-level audit, the consistent removal for controlled EDMD models, the price of erasure (Propositions~\ref{prop:erasure-price} and~\ref{prop:general-price}; the structure is known from mean-parity regression, \citealp{calders2013controlling,wei2023mean}) and the conditional feasibility result with its floor-constrained controller (Proposition~\ref{prop:no-levelling}). The trajectory score (Appendix~\ref{app:trs}) is an extra result; it is not independent evidence for the main conclusions.

	\begin{table}[t]
		\centering\small
		\caption{How this work relates to methods that remove group information or put fairness into decisions.}
		\label{tab:positioning}
		\begin{tabularx}{\linewidth}{@{}>{\RaggedRight\arraybackslash}p{3.0cm}>{\RaggedRight\arraybackslash}X>{\RaggedRight\arraybackslash}X>{\RaggedRight\arraybackslash}p{1.9cm}@{}}
			\toprule
			Method & What it changes & What it guarantees & Dynamics \\
			\midrule
			Fair representations \citep{zemel2013learning} & a learned representation, trained with a parity term & no formal guarantee for later decisions & no \\
			\addlinespace
			Null-space projection \citep{ravfogel2020null} & removes, again and again, the directions a linear classifier uses for the group & no formal guarantee; stops when a linear classifier fails & no \\
			\addlinespace
			LEACE \citep{belrose2023leace} & erases a linear concept in closed form, with the smallest change & no linear predictor can recover the concept & no \\
			\addlinespace
			Fair regression \citep{calders2013controlling,chzhen2020fair} & constrains the predictions to equal group means or equal distributions & the best predictor under the constraint & no \\
			\addlinespace
			Constrained MDPs \citep{altman1999constrained} & adds the fairness rule to the decision problem & the constraint holds for the model used & yes (model given) \\
			\addlinespace
			Fair reinforcement learning \citep{jabbari2017fairness,wen2021algorithms} & defines fairness over long-term reward or over decision processes, and learns policies & guarantees for their own definitions & yes \\
			\addlinespace
			This paper & removes a group subspace from a learned controlled Koopman model; adds floors to the controller; audits both & structural identities; approximation and horizon bounds; minimum price of erasure for any predictor of the state; one-step feasibility of floors & yes (learned model) \\
			\bottomrule
		\end{tabularx}
	\end{table}

	\subsection{Details of the construction}
	\label{app:construction}
	\paragraph{The ridge of the metric.} The relative ridge \(10^{-3}\) is a numerical default, not a value selected by cross-validation. Its conditioning effect can be stated directly: for nonzero positive-semidefinite \(\Gram\) and \(r=\eta/\lambda_{\max}(\Gram)>0\),
	\[
	\operatorname{cond}_2(\Gamma)=\frac{\lambda_{\max}(\Gram)+\eta}{\lambda_{\min}(\Gram)+\eta}\leq1+r^{-1}.
	\]
	the default bounds the condition number by \(1001\). Because the ridge also changes the adjoint, projection geometry and selected subspace, it may affect the substantive results; Section~\ref{sec:results-ridge} therefore evaluates sensitivity to this choice.

	\paragraph{Augmented form.} With \(\bar z=(z^\top,1)^\top\), the extended objects are
	\begin{equation}
		\bar A=\begin{pmatrix}A&a\\0&1\end{pmatrix},\quad
		\bar\Gamma=\diag(\Gamma,1),\quad
		\bar P=\diag(P,1).
	\end{equation}
	The augmented projection \(\bar A_P=\bar P^\top\bar A\bar P^\top\) leaves the constant coordinate unchanged. It does not, however, guarantee positivity: an orthogonal compression need not remain the Koopman operator of a Markov process. We therefore interpret Eq.~\eqref{eq:affine-projected} as a projected surrogate, not automatically as a valid model of how a real system evolves.
	
	\begin{remark}[Decoding and latent projection]
		We keep the affine decoder \(C\) and its constant \(c\) fixed, so that forecasts under different policies are always in the same physical units. A projected forecast \(z_P\) satisfies \(P^\top z_P=z_P\) in the uncentered affine setting, so \(Cz_P+c=(CP^\top)z_P+c\). Projecting a second time before decoding therefore changes nothing on this range, although it is not wrong. Projecting on the left of the decoder, on the other hand, usually does not even have the right dimensions.
	\end{remark}

	\paragraph{Observed-feedback use of the compressed model.}
	\label{sec:inference-architecture}
	The distinction between observations and model-generated states determines how the compressed model is used. Algorithm~\ref{alg:tutoring-inference} (Appendix~\ref{sec:theory}) starts each step from the newly observed state rather than recursively feeding the previous forecast back into the model. Setting \(u_t\equiv0\) gives the observational replay used for ASSISTments (Section~\ref{sec:results-realdata}); choosing an MPC policy yields the simulator controller described in Section~\ref{sec:protocol}.

	The algorithm thus keeps action selection, model prediction and observation of the next state distinct. This separation makes forecast error measurable at every step and prevents model-generated states from being confused with observed feedback. It also makes the role of consistent projection explicit: leaving \(B\) or \(a\) unprojected could reintroduce the removed direction, as flagged by the structural audit in Table~\ref{tab:protocol}. The algorithm is the numerical procedure evaluated here.

	\subsection{Properties of the compressed operator}
	\label{sec:theory}
	\begin{algorithm}[t]
		\caption{Prediction and control with a reduced model, using observed feedback}
		\label{alg:tutoring-inference}
		\small
		\begin{algorithmic}[1]
			\Require Fitted \(A,B,a\), coefficient projector \(P\), encoder \(\Psi\), decoder \(x=Cz+c\)
			\Require Numerical policy \(\pi\) (possibly \(\pi\equiv0\)), horizon \(T\), first observed state \(x_0\)
			\State \(A_P\gets P^\top AP^\top\), \(B_P\gets P^\top B\), \(a_P\gets P^\top a\) \Comment{dual projector on all terms}
			\For{\(t=0,\ldots,T-1\)}
			\State \(z_t\gets\Psi(x_t)\) \Comment{encode the current observation}
			\State \(u_t\gets\pi(z_t)\)
			\State \(\widehat z_{t+1\mid t}\gets A_Pz_t+B_Pu_t+a_P\), \quad \(\widehat x_{t+1\mid t}\gets C\widehat z_{t+1\mid t}+c\)
			\State apply \(u_t\) and observe the next state \(x_{t+1}\); record \(\widehat x_{t+1\mid t}-x_{t+1}\)
			\EndFor
		\end{algorithmic}
	\end{algorithm}

	The preceding construction defines a finite-dimensional model transformation. The results below clarify what this transformation guarantees mathematically and where those guarantees stop. We distinguish approximation properties of a fixed projector, finite-horizon bounds for an operator pairing, and claims about population outcomes. The first two can support an audit; neither alone establishes that a representation or a policy is fairer.
	
	\begin{proposition}[Fixed-projector approximation]
		\label{prop:optimality}
		Let \(K,P\in\mathbb R^{N\times N}\), let \(\Gamma\) be symmetric positive definite, and suppose that \(P^2=P\) and \(P^{*\Gamma}=P\). Then \(F=PKP\) is the unique minimizer of \(\|K-L\|_{\Gamma,F}\) over all \(L=PLP\). More precisely, every such \(L\) satisfies the Pythagorean identity
		\begin{equation}
			\|K-L\|_{\Gamma,F}^2
			=\|K-F\|_{\Gamma,F}^2+\|F-L\|_{\Gamma,F}^2.
		\end{equation}
		For \(R=I-P\),
		\begin{equation}
			K-F=RKR+PKR+RKP.
			\label{eq:blocks}
		\end{equation}
		These three blocks are pairwise orthogonal in the weighted Frobenius inner product, so
		\begin{equation}
			\|K-F\|_{\Gamma,F}^2
			=\|RKR\|_{\Gamma,F}^2
			+\|PKR\|_{\Gamma,F}^2
			+\|RKP\|_{\Gamma,F}^2.
		\end{equation}
		So, if a functional \(U\) is \(L_U\)-Lipschitz in this matrix norm,
		\begin{equation}
			|U(K)-U(F)|
			\leq L_U\bigl(\|RKR\|_{\Gamma,F}^2+r_1^2+r_2^2\bigr)^{1/2}.
		\end{equation}
	\end{proposition}
	The proof is in Appendix~\ref{app:proofs}. For a fixed admissible projector, \(F=PKP\) is the closest operator to \(K\) among matrices with the prescribed retained subspace. The result is an approximation statement, not an outcome guarantee: it neither identifies a functional \(U\) with educational benefit nor determines its Lipschitz constant in an application, and it does not specify how \(P\) should be selected.
	
	\begin{theorem}[Finite-horizon pairing bound]
		\label{thm:horizon}
		Let \(K\) be bounded on a real Hilbert space, \(P\) an orthogonal projector, \(R=I-P\), \(F=PKP\), and \(q,f\in\cH\). For integer \(T\geq1\), put
		\begin{equation}
			c_T(K,F)=\sum_{j=0}^{T-1}\|K\|^{T-1-j}\|F\|^j.
		\end{equation}
		All operator norms below use the same Hilbert norm. Throughout this theorem, the superscript \(T\) denotes the \(T\)-fold operator power; matrix transpose is written \(\top\). Then
		\begin{equation}
			|\langle q,(K^T-F^T)f\rangle|
			\leq\|q\|\|f\|c_T(K,F)\|K-F\|
			\leq\|q\|\|f\|c_T(K,F)
			\bigl(\|RKR\|+\|PKR\|+\|RKP\|\bigr).
			\label{eq:horizon-error}
		\end{equation}
		Consequently,
		\begin{equation}
			\begin{aligned}
				|\langle q,K^T f\rangle|
				&\leq\|q\|\|f\|c_T(K,F)
				\bigl(\|RKR\|+\|PKR\|+\|RKP\|\bigr)\\
				&\quad+\|Pq\|\|F\|^T\|Pf\|.
			\end{aligned}
			\label{eq:horizon}
		\end{equation}
		Write \(\mathcal B_T(q,f;K,P)\) for the right-hand side.
	\end{theorem}
	The proof is in Appendix~\ref{app:proofs}. The bound separates two sources of a nonzero pairing after \(T\) steps. The first accumulates the discrepancy between the original and compressed operators; the second is the contribution retained within the projected subspace. The latter need not vanish, because the projection removes only part of the outcome direction. In particular, \(F^Tf=F^TPf\), so only the retained component \(Pf\) contributes to this term. If both operators are contractions, \(c_T\leq T\); otherwise, the amplification factor must be evaluated rather than assumed negligible.
	
	\begin{proposition}[Amplification regimes]
		\label{prop:amplification-regimes}
		Let \(\kappa=\|K\|\), \(\varphi=\|F\|\), and \(\rho=\max(\kappa,\varphi)\) for bounded operators on the same normed space. For every integer \(T\geq1\),
		\begin{equation}
			c_T(K,F)\leq T\rho^{T-1},\qquad
			c_T(K,F)=\begin{cases}
				(\kappa^T-\varphi^T)/(\kappa-\varphi),&\kappa\ne\varphi,\\
				T\kappa^{T-1},&\kappa=\varphi.
			\end{cases}
			\label{eq:amplification-regimes}
		\end{equation}
		Zero to the zeroth power is read as one in this finite sum. If \(0\leq\rho<1\), then \(c_T\leq\min\{T\rho^{T-1},(1-\rho)^{-1}\}\). If \(\rho=1\), then \(c_T=T\) when \(\kappa=\varphi=1\), and \(c_T=(1-m^T)/(1-m)\leq(1-m)^{-1}\) when \(m=\min(\kappa,\varphi)<1\). If \(\rho>1\), equal norms give exactly \(T\rho^{T-1}\); unequal fixed norms give \(c_T=\Theta(\rho^T)\) as \(T\to\infty\).
	\end{proposition}
	The proof is in Appendix~\ref{app:proofs}. These regimes show why the condition \(\varphi<1\leq\kappa\) does not, by itself, imply an exploding bound: when \(\kappa=1\), the sum remains bounded. Exponential growth of order \(\kappa^T\) occurs only when \(\kappa>1\) and the reduced norm \(\varphi<\kappa\) remains fixed.
	
	The replicate-0 centered block illustrates the difference between the exact amplification factor and its general upper bound. Here \(\kappa=1.02760679\) and \(\varphi=0.87807283\); at \(T=25\), the exact sum is \(c_{25}=12.951824\), compared with \(25\kappa^{24}=48.060377\). The simpler expression remains valid, but substantially overestimates amplification in this case.
	
	\subsubsection{Why a numerical bound is not a promise about real use}
	The finite-model bound is not yet a bound on the population quantity in Eq.~\eqref{eq:physical-identity}. To make that connection, suppose the true closed-loop operator \(\cK^\pi\) and fitted operator \(K\) are bounded on the same \(L^2(\mu)\), with \(q_0,q,f_D\) square-integrable there, and let \(\nu_0\) denote the initial law of the full state. Decomposing the difference into its sources yields
	\begin{equation}
		|\Delta_D|\leq\mathcal B_T(q,f_D;K,P)
		+E_{\mathrm{model}}+E_{\mathrm{contrast}}+E_{\mathrm{law}},
		\label{eq:transfer}
	\end{equation}
	The exact formulas are given in Appendix~\ref{app:transfer}. The additional terms account respectively for operator-estimation error, error in the estimated initial contrast, and the change from the reference law to the target law. Since the experiments do not estimate these quantities, Eq.~\eqref{eq:transfer} supplies no guarantee for deployment; it identifies the extra evidence such a guarantee would require.
	
	\subsubsection{Convergence needs two separate limits, not one}
	Convergence introduces a further distinction. Let \(E_N\) project onto a sequence of feature spaces whose union is dense. The corresponding Galerkin operators \(\mathcal A_N=E_N\cK E_N\) converge strongly to bounded \(\cK\). Separately, for a fixed \(N\), a positive-definite population Gram matrix and almost-sure convergence of the empirical matrices imply operator-norm convergence of the unregularized EDMD fit to \(\mathcal A_N\). These are different limits—more data at fixed features versus more features—and neither should be conflated with the other \citep{korda2018convergence}.
	
	Data-dependent projection raises a third question: whether the compressed fits converge as both the operator and projector change with \(N\). Embed the finite-dimensional approximations in one Hilbert space and suppose \(K_N\to K\) and \(P_N\to P\) strongly. Then \(P\) is an orthogonal projector and \(P_NK_NP_N\to PKP\) strongly (Appendix~\ref{app:convergence}). This conditional statement does not yet establish convergence for the specific low-rank, data-selected projectors used in our experiments.
	\begin{equation}
		P_N K_N P_N \longrightarrow P K P \quad\text{strongly}.
		\label{eq:compression-limit}
	\end{equation}

	\subsubsection{The exact closure}
	The idealized construction starts from the group contrast \(q\) and seeks the smallest closed reducing subspace containing it. For a bounded operator on a chosen function space, this subspace is
	\begin{equation}
		\Bred=\overline{\Span\{Wq: W\text{ is a finite word in }\cK,\cK^*\}},
		\label{eq:reducing}
	\end{equation}
	where the empty word is included, ensuring \(q\in\Bred\). This closed cyclic subspace is invariant under both \(\cK\) and \(\cK^*\), and is contained in every other closed reducing subspace that contains \(q\). In finite dimensions, the analogous construction repeatedly applies \(K\) and its \(\Gamma\)-weighted adjoint \(K^{*\Gamma}\) to the contrast coefficient vector. Exact invariance under both operators eliminates the off-diagonal blocks between the retained and removed subspaces. The bar denotes the closure, which matters in infinite dimension: for the unilateral shift on \(\ell^2\) and \(q=e_0\), the finite words span only the finitely supported sequences, a dense subspace that is not closed.
	
	In the simulator, the exact closure of the fitted matrix fills the entire \emph{centered} dictionary, leaving a centered complement of dimension zero. In the augmented affine space, however, the projection retains the constant direction, so the complement has dimension one. These conclusions are compatible because they concern different spaces. More broadly, the results distinguish a useful rank-limited surrogate from an exact low-dimensional reducing model: the former need not approximate the latter closely.

	\subsection{Proofs of Appendix~\ref{sec:theory}}
	\label{app:proofs}
	\paragraph{Proof of Proposition~\ref{prop:optimality}.}
	\begin{proof}
		Set \(\widetilde K=\Gamma^{1/2}K\Gamma^{-1/2}\), and define \(\widetilde P,\widetilde L\) similarly. The assumptions on \(P\) imply that \(\widetilde P\) is an ordinary Euclidean orthogonal projector. Hence \(\widetilde{\mathcal Q}:X\mapsto\widetilde P X\widetilde P\) is self-adjoint and idempotent for the Frobenius inner product, so it is the orthogonal projector onto \(\{X:X=\widetilde P X\widetilde P\}\). The Frobenius Pythagorean theorem gives
		\[
		\|\widetilde K-\widetilde L\|_F^2
		=\|\widetilde K-\widetilde P\widetilde K\widetilde P\|_F^2
		+\|\widetilde P\widetilde K\widetilde P-\widetilde L\|_F^2,
		\]
		which becomes the stated weighted identity after undoing the similarity transform. It also proves uniqueness. Expanding \(K=(P+R)K(P+R)\) gives Eq.~\eqref{eq:blocks}. In whitened coordinates, the three displayed blocks occupy mutually orthogonal row--column blocks relative to \(\ran(\widetilde P)\oplus\ran(I-\widetilde P)\); their Frobenius inner products therefore vanish. This proves the squared-norm identity. The Lipschitz bound follows immediately.
	\end{proof}
	
	\paragraph{Proof of Theorem~\ref{thm:horizon}.}
	\begin{proof}
		A standard telescoping identity for non-commuting operators gives
		\[
		K^T-F^T=\sum_{j=0}^{T-1}K^{T-1-j}(K-F)F^j.
		\]
		Taking norms, using Eq.~\eqref{eq:blocks} and applying Cauchy--Schwarz gives Eq.~\eqref{eq:horizon-error}. Since \(F=PF=FP\), we have \(\ran(F^T)\subseteq\ran(P)\) for \(T\geq1\). Orthogonality of \(P\) therefore gives \(\langle q,F^Tf\rangle=\langle Pq,F^Tf\rangle\). Moreover, \(F=FP\) gives \(F^Tf=F^TPf\). Hence
		\[
		|\langle Pq,F^Tf\rangle|
		=|\langle Pq,F^TPf\rangle|
		\leq\|Pq\|\|F\|^T\|Pf\|.
		\]
		Adding this estimate to Eq.~\eqref{eq:horizon-error} proves Eq.~\eqref{eq:horizon}.
	\end{proof}
	
	\paragraph{Proof of Proposition~\ref{prop:amplification-regimes}.}
	\begin{proof}
		Each of the \(T\) terms in the sum is at most \(\rho^{T-1}\), which proves the first inequality. If \(\kappa\ne\varphi\), multiplying the defining sum by \(\kappa-\varphi\) telescopes to \(\kappa^T-\varphi^T\); if \(\kappa=\varphi\), all \(T\) terms equal \(\kappa^{T-1}\). For \(\rho<1\), the \(j\)-th term is at most \(\rho^j\), hence \(c_T\leq\sum_{j=0}^{T-1}\rho^j\leq(1-\rho)^{-1}\). If \(\rho=1\) and \(m=\min(\kappa,\varphi)<1\), the closed form reduces to \((1-m^T)/(1-m)\); the equal case gives \(T\). Finally, if \(\rho>1\) and \(m<\rho\), then
		\[
		c_T=\frac{\rho^T-m^T}{\rho-m}
		=\frac{\rho^T}{\rho-m}\left[1-\left(\frac m\rho\right)^T\right],
		\]
		which is \(\Theta(\rho^T)\). This proves every stated regime.
	\end{proof}

	\subsection{Function-space assumptions}
	\label{app:function-space}
	Our operator statements are made on a real Hilbert space \(\cH=L^2(\mu;\R)\), with the reference measure specified wherever it matters. For a transition kernel \(\mathsf P\), the domination condition \(\mu\mathsf P\leq C_\mu\mu\) implies \(\|\cK\|\leq\sqrt{C_\mu}\) by conditional Jensen. An invariant measure would yield the stronger contraction property. We do not assume invariance, since exploratory and controlled trajectories need not be sampled from an equilibrium distribution.
	
	The distinction between a fixed transition rule and a stationary state law is important here. For example, augment a finite-horizon process with a discrete clock and make the final step absorbing. If \(\mu_t\) is the state law at step \(t\), then \(\mu_*=(T+1)^{-1}\sum_{t=0}^T\mu_t\) satisfies \(\mu_*\widetilde{\mathsf P}\leq2\mu_*\), so the associated operator is bounded by \(\sqrt 2\), even though the process need not be stationary. Since some coordinates may also be unbounded under Gaussian noise, we rely only on the square-integrability and fourth-moment conditions required by a quadratic dictionary.

	\subsection{Further theoretical results}
	\label{app:further-theory}
	\subsubsection{Explicit population-transfer terms}
	\label{app:transfer}
	Under the common-space and integrability assumptions stated before Eq.~\eqref{eq:transfer}, define
	\begin{align}
		E_{\mathrm{model}}&=\|q\|\|f_D\|\|\cK^\pi-K\|
		\sum_{j=0}^{T-1}\|\cK^\pi\|^{T-1-j}\|K\|^j,\\
		E_{\mathrm{contrast}}&=\|q_0-q\|\|\cK^\pi\|^T\|f_D\|,\\
		E_{\mathrm{law}}&=\left|\langle q_0,(\cK^\pi)^Tf_D\rangle_{\nu_0}
		-\langle q_0,(\cK^\pi)^Tf_D\rangle_\mu\right|.
	\end{align}
	All unmarked norms use \(\mu\). The finite diagnostics on coefficients alone do not give a value to any of these three terms.
	
	To prove Eq.~\eqref{eq:transfer}, write \(L=\cK^\pi\). Successively adding and subtracting the same quantities gives
	\begin{align*}
		|\Delta_D|
		&\leq E_{\mathrm{law}}+
		\bigl|\langle q_0,L^Tf_D\rangle_\mu\bigr|\\
		&\leq E_{\mathrm{law}}+E_{\mathrm{contrast}}
		+\bigl|\langle q,L^Tf_D\rangle_\mu\bigr|\\
		&\leq E_{\mathrm{law}}+E_{\mathrm{contrast}}
		+\bigl|\langle q,K^Tf_D\rangle_\mu\bigr|
		+\|q\|\,\|(L^T-K^T)f_D\|.
	\end{align*}
	The noncommutative telescoping identity
	\[
	L^T-K^T=\sum_{j=0}^{T-1}L^{T-1-j}(L-K)K^j
	\]
	therefore yields
	\[
	\|q\|\,\|(L^T-K^T)f_D\|
	\leq E_{\mathrm{model}}.
	\]
	Finally, Theorem~\ref{thm:horizon} bounds
	\(\bigl|\langle q,K^Tf_D\rangle_\mu\bigr|\) by
	\(\mathcal B_T(q,f_D;K,P)\), which proves Eq.~\eqref{eq:transfer}.
	
	\FloatBarrier

	\subsubsection{Strong convergence of compressed operators}
	\label{app:convergence}
	After the embeddings described in Appendix~\ref{sec:theory}, let all operators act
	on one Hilbert space. Let \(P_N\) be orthogonal projectors with \(P_N\to P\)
	strongly, and let the bounded operators \(K_N\to K\) strongly. The strong limit
	\(P\) is an orthogonal projector: self-adjointness follows by passing to the limit
	in \(\langle P_Nf,g\rangle=\langle f,P_Ng\rangle\), and idempotence follows from
	\[
	\|P_N^2f-P^2f\|
	\leq\|P_N(P_N-P)f\|+\|(P_N-P)Pf\|\longrightarrow0.
	\]
	Because \(P_N^2f=P_Nf\to Pf\), this also gives \(P^2f=Pf\).
	Strong convergence \(K_N\to K\) also implies
	\(M_*:=\sup_N\|K_N\|<\infty\), by the uniform boundedness principle. For a fixed
	\(f\),
	\begin{align*}
		(P_N K_N P_N - P K P)f
		={}&P_N K_N (P_N - P)f\\
		&+P_N(K_N-K)P f+(P_N-P)KP f.
	\end{align*}
	The three terms are bounded, respectively, by
	\(M_*\|(P_N-P)f\|\), \(\|(K_N-K)P f\|\), and
	\(\|(P_N-P)KP f\|\). All three go to zero under the separate
	strong-convergence assumptions. Hence \(P_NK_NP_Nf\to PKPf\) for every fixed
	\(f\), which proves Eq.~\eqref{eq:compression-limit}.
	\FloatBarrier

	\subsubsection{Marginal cohort coverage}
	\label{app:coverage}
	Condition on the independent training data, so that the fitted model, prediction rule, and policy are fixed. Let the \(n\) calibration scores and the new-cohort score be exchangeable. For each of these \(n+1\) scores, define its lower rank as one plus the number of strictly smaller scores among all \(n+1\) scores. For every realized multiset of scores, at least \(\ell_\alpha\) indices have lower rank at most \(\ell_\alpha\). Conditional exchangeability therefore gives
	\[
	\Pr\{\operatorname{rank}_{\mathrm{lower}}(s_{n+1})\leq\ell_\alpha
	\mid\text{training data}\}
	\geq\frac{\ell_\alpha}{n+1}.
	\]
	When \(\ell_\alpha\leq n\), the event on the left is equivalent to
	\(s_{n+1}\leq s_{(\ell_\alpha)}\), where the order statistic uses only the
	calibration scores; using a closed interval includes all ties. Since
	\(s_{n+1}=|\delta_{n+1}-\widehat\delta_{n+1}|\), this event is exactly
	\(\delta_{n+1}\in C_{n+1}\). Thus the conditional, and hence unconditional,
	coverage is at least \(\ell_\alpha/(n+1)\geq1-\alpha\). If
	\(\ell_\alpha=n+1\), the prescribed interval is the whole real line and the
	coverage probability is one.
	
	\FloatBarrier

	\subsection{Structural diagnostics of the fitted models}
	\label{app:structural}
	Table~\ref{tab:structural} gives the level-1 diagnostics for the simulator and ASSISTments fits. The simulator family contains every fit of the 50-replicate study: the 13 combinations of group strength, rank and depth, each with seeds 0--49 (650 fits; the horizon and \(\beta_{\mathrm{ind}}\) do not change the fit). In an earlier version only seeds 0--9 had been tested; seeds 10--49 now pass the same test. The table covers the simulator and ASSISTments; the Eedi, OULAD and group-aware fits are reported below it. For each quantity it gives the largest value over the fits of the family; for \(r_1\) and \(r_2\), it gives the minimum and maximum across those fits. These ranges describe the observed spread, not confidence intervals or acceptable margins. The duality error is zero by construction, because \(K\) is defined as \(A^\top\) in Eq.~\eqref{eq:controlled-fit}; what the test checks is that the same matrix is used to build the projector and to simulate. The metric-duality error \(\max|\Gamma^{-1}P^\top\Gamma-P|\) checks that \(P\) is self-adjoint for \(\Gamma\). The leakage norms use the value-space complement \(R^\top=I-P^\top\) and check that the removed direction does not come back through the control or the constant term; the ASSISTments model has no control, so \(\|R^\top B_P\|\) does not apply. All identities hold to \(4\times10^{-12}\) or better. The cross-block residuals are far from zero (\(r_1\) from \(0.05\) to \(0.57\), \(r_2\) from \(0.09\) to \(0.58\)): the selected spaces are only approximately reducing, as Section~\ref{sec:method} explains. In the simulator, the numerical rank of \(\Gram\) is 10 (full); on ASSISTments it is 13, because \(p^2=p\). The singular-value cutoff is \(10^{-8}\) (relative) everywhere. The script \path{structural_diagnostics.py} recomputes the table and stops with an error if an identity fails beyond \(10^{-9}\).
	
	\begin{table}[!ht]
		\centering\small
		\caption{Level-1 diagnostics for all families of fits: largest errors over the fits of each family, minimum--maximum ranges of \(r_1,r_2\) (in the \(\Gamma\)-weighted Frobenius norm), and requested, candidate, applied and numerical ranks. These ranges are observed values across fits, not confidence intervals or acceptance margins. Columns: simulator; ASSISTments school-disjoint splits, other-student-label variant, student splits, 5{,}000-student subsamples and full-corpus label. The simulator family contains the 650 fits of the 50-replicate study (13 combinations of \(\gamma\), rank and depth, 50 replicates each). The other-student-label fits use the same training schools as the school-disjoint fits, so their numbers are identical.}
		\label{tab:structural}
		\resizebox{\linewidth}{!}{%
			\input{tab_structural.tex}}
	\end{table}

	\paragraph{Eedi, OULAD and the group-aware model.}
	We computed the same diagnostics for the five Eedi splits and for the fifteen OULAD fits (three attributes, five splits each). All identities hold to \(2\times10^{-12}\) or better, and the leakage through the constant term is at most \(3\times10^{-15}\). The cross-block residuals are again far from zero (Eedi: \(r_1\) from \(0.54\) to \(0.67\), \(r_2\) from \(0.25\) to \(0.34\); OULAD: \(r_1\) and \(r_2\) from \(0.23\) to \(0.33\)). The group-aware model has no projection, so its level-1 check is different: the least-squares normal equations hold to a relative residual of \(7\times10^{-14}\), and in the thirty runs of the floor-constrained MPC (\(\gamma\in\{0.5,1,2\}\), ten replicates, \(c=2\)) the knowledge floors are met to \(1.1\times10^{-9}\) and the budget exactly.

	\section{Data and real-data results}
	\label{app:block-data}
	\subsection{Data sets in detail}
	\label{app:data}
	\subsubsection{ASSISTments 2012--2013}
	\paragraph{Data.}
	For each included student, prediction uses the first 26 states \(x_0,\ldots,x_{25}\) (\(T=25\) transitions). Each student belongs to one school, and each school belongs to one group.
	
	\paragraph{Design.} Each student's rows are taken in timestamp order. The state is a four-dimensional vector of performance proxies, lifted by a fixed quadratic dictionary. We use five school-disjoint 60/20/20 splits, compute every held-out interval with a school-cluster bootstrap, and compare with persistence, an affine model, gradient boosting and a recalibration for each group. Appendix~\ref{app:assist-details} gives the details.

	\paragraph{Group variable and order of estimation.}
	The data set has no validated sensitive attribute for students. We therefore use the general label of Section~\ref{sec:problem} with \(\grp=\grp_{\mathrm{school}}\), a \emph{group of schools defined by their own performance}. A school is in group 1 if its average success is above the median of the training schools. We compute the school averages on a \emph{label window}: the steps after \(t=25\), which never overlap the steps we evaluate. The median uses training schools only. A school that the model has not seen is labelled by comparing its own later average with this training median. This label is \emph{retrospective}: it uses later steps of the same students whose earlier steps we predict. It would not be available when a truly new school starts, and neither would the recalibration for each group. So we read the main analysis as a retrospective grouping of schools, not as a prediction for the future. We also run a variant where \emph{other students} give the label: in every validation and test school, a random half of the students is used only to compute the school's label, and these students are removed from the evaluation. This separates the students who give the label from the students we evaluate, but it does not show that the label would be known earlier in calendar time. The label comes from performance; it is not a social or demographic attribute. After the split, and with training students only, we fit the skill averages in \(r_t\), the scaling of the features, the EDMD operator and the decoder, \(\Gamma\), the contrast, the projector and every baseline. Everything is then frozen before evaluation.
	
	\subsubsection{Eedi}
	The ASSISTments groups are built from school performance. To test the same claims with a sensitive attribute that is defined outside the data, we use the Eedi data of the NeurIPS 2020 Education Challenge \citep{wang2020diagnostic}. They contain 15.9 million answers of 118{,}971 pupils to multiple-choice questions in mathematics, from September 2018 to May 2020, with the time of each answer and the class in which it was given. The group is \textbf{PremiumPupil}: whether the pupil can receive free school meals or pupil premium because the family has little money. This attribute is recorded for 34{,}386 pupils (29\%), and 8{,}871 of them (25.8\%) are disadvantaged. We analyse only these pupils (6.0 million answers), who may not represent all Eedi pupils. The state is built as for ASSISTments, in time order. \(k_t\) is the share of correct answers in the last ten, \(p_t=\mathbf 1\{k_t>0.5\}\), and \(r_t\) is the answer minus the training mean of the question's most specific subject. Eedi has no affect detectors, so \(e_t\) is the confidence that the pupil reports, rescaled to \([-0.5,0.5]\). Pupils report it for only 11\% of the answers, so we carry forward the last reported value. The sampling unit is the class. Pupils often belong to several classes (median 2), so each pupil is assigned to the class where the pupil answered most often (2{,}369 classes). The five splits are disjoint in these classes, with about 20{,}600 training, 6{,}900 validation and 6{,}900 test pupils, and about 480 test classes. A pupil's other classes can still appear on both sides. So a \emph{strict} variant also removes every validation and test pupil who answered in any class of a training pupil; 1{,}928 to 2{,}870 test pupils remain. Everything else is unchanged: the predictors, the projector, the recalibration, the class-level bootstrap and the level-2 rule. The label is the pupil's own attribute and is known before any prediction, so this study is not a retrospective grouping. It is still a test of prediction: Eedi does not record any tutor action whose alternative we could observe.
	
	\subsubsection{OULAD}
	Both studies above are about mathematics practice at school. To see whether the picture also holds in a very different setting, we use the Open University Learning Analytics Dataset (OULAD) \citep{kuzilek2017open}. It describes 32{,}593 registrations of 28{,}785 students in 22 presentations of seven distance-learning university courses in the United Kingdom. It contains declared demographic attributes, the daily number of clicks in the online learning environment, and assessment scores. We keep the first registration of each student and the students whose deprivation band is known (27{,}814 students). The state is weekly, over the first 26 weeks (\(x_0,\ldots,x_{25}\)). \(k_t\) is the mean score (divided by 100) of the assessments submitted so far, and 0.5 before the first one. \(e_t\) is the logarithm of the week's clicks, scaled to \([-0.5,0.5]\); it measures engagement directly. \(p_t=\mathbf 1\{k_t\geq0.4\}\), where 0.4 is the Open University pass mark. \(r_t\) is the week's activity (at least one click) minus the training mean of the same course presentation and week. The main attribute is \textbf{deprivation}: a student is in group 1 if the home area is among the 30\% most deprived areas of the Index of Multiple Deprivation (33.5\% of students). Like the Eedi attribute, it is socioeconomic, but it describes the home area, not the student's own income. We also report \textbf{disability} (declared by 9.8\% of students) and \textbf{gender} (48.7\% women). The five splits assign students at random (60/20/20) within the 22 course presentations, and the intervals resample presentations, because the students of one presentation share its content, calendar and assessments. So the main splits test new students in presentations that the model has seen. Section~\ref{sec:results-oulad} also reports splits that hold out whole presentations, and a split that trains on the 2013 presentations and tests on the 2014 ones. Everything else is unchanged. Like the two other studies, this one tests prediction, not the effect of an intervention.

	\subsection{Details of the ASSISTments study}
	\label{app:assist-details}
	ASSISTments is an online platform for learning mathematics. Its 2012--2013 data set has one row for each attempt at a problem, with a timestamp, a school identifier and the output of affect detectors \citep{pardos2013affective,pardos2014affective}. Of its 6{,}123{,}270 rows, 2{,}711{,}753 rows from 29{,}006 students have a skill and a school. If we keep students with at least 20 rows, we have 2{,}622{,}847 interactions from 18{,}366 students in 243 schools. There are 16{,}668 eligible students with at least 26 rows, in 216 schools. Of these, 16{,}661 students in 211 schools enter the analysis (Table~\ref{tab:flow}). The seven others are the only students of five very small schools (with 2, 1, 2, 1 and 1 students). These schools have no step after \(t=25\), so they have no label window and cannot receive a group label (see below). We remove them. The rule uses only the number of rows, never an outcome. Every split of Table~\ref{tab:splits} partitions these 16{,}661 students and 211 schools. \begin{table}[htbp]
		\centering\small
		\caption{Sample flow of the ASSISTments study. Every rule uses row counts only. The script \texttt{sample\_flow.py} recomputes each line from the licensed file.}
		\label{tab:flow}
		\input{tab_flow.tex}
	\end{table}

	\paragraph{Temporal order.}
	A transition \((x_t,x_{t+1})\) must connect two interactions that really follow each other. So we rebuilt all features from the raw data, with the rows of each student sorted by the time of the attempt, \texttt{start\_time}, and by the log identifier when two times are equal. Repeated attempts at the same problem (127{,}170 rows) stay as separate steps. A feature file sorted by problem identifier is also common for this data set; we rebuild the features in that order as well, so we can measure the difference precisely. Of the 416{,}700 transitions analysed in problem order, only 9.6\% follow each other in time, and only 39.5\% of the rows in the first 26 steps are the same as in time order. In other words, an analysis in problem order mostly looks at pairs that are not consecutive. All results below use time order; the problem order is kept only as a sensitivity check.
	
	\paragraph{State and dictionary.}
	The state of a student is \(x_t=(k_t,e_t,p_t,r_t)\):
	\begin{itemize}
		\item \(k_t\): the share of correct answers in the last ten attempts. It is a \emph{rolling-correctness proxy} of performance, not a measure of hidden knowledge; we keep the letter \(k\) only to match the simulator.
		\item \(e_t\): the confidence of the \texttt{CONCENTRATING} detector minus that of the \texttt{BORED} detector.
		\item \(p_t=\mathbf 1\{k_t>0.5\}\).
		\item \(r_t\): the answer (correct or not) minus the average success on the same skill among training students, kept inside \([-1,1]\).
	\end{itemize}
	The dictionary is \(\Psi_{\mathrm{A}}(x)=(k,e,p,r,k^2,e^2,p^2,r^2,ke,kp,kr,ep,er,pr)^\top\). We write \(r_t\), not \(d_t\), because this skill residual has nothing to do with the synthetic disparity \(d_t\) of the simulator. Since \(p^2=p\), the Gram matrix has numerical rank 13, and the least-squares fits use the minimum-norm solution. Multi-step forecasts iterate in the lifted space, \(\widehat z_{t+1}=A\widehat z_t+a\), and are \emph{not} encoded again from the decoded state, so the forecast can leave the set where \(z_{p^2}=z_p\) and \(p\in\{0,1\}\). In practice the two coordinates stay equal (exactly for the original model, within \(5\times10^{-15}\) after projection), because the minimum-norm fit treats the two identical columns symmetrically. Decoded values are not clipped: over 50 free steps, the original model keeps \(k\) in \([0.11,0.98]\) and \(p\) in \([-0.14,1.03]\), and the projected model keeps \(k\) in \([0.68,0.73]\). The linear, gradient-boosting and persistence baselines work in the physical state and treat \(p\) as a continuous number. As a check, we removed \(p^2\) from the dictionary (13 features): the errors of the original model at \(h=1,5,10,25\) do not change (\(0.0074\), \(0.0238\), \(0.0406\), \(0.0459\) in the school-disjoint split we checked), and those of the projected model change by less than \(0.0005\). We use rank 4, depth 3 and \(\Gamma=\Gram+10^{-3}\lambda_{\max}(\Gram)I\). The data contain no action that we can change and whose alternative outcome we can observe. So the model has no control input, \(z^+=Az+a\), and this study tests prediction, not the effect of a policy.
	
	\paragraph{Splits.}
	The \emph{main analysis} uses five splits in which training, validation and test schools are different (60/20/20). Schools are shuffled at random and assigned to training, validation and test until these hold 60\%, 20\% and 20\% of the students, so no school appears in two parts. Table~\ref{tab:splits} gives the number of students and schools in each part. The five splits reuse the same 211 schools in different ways, so they repeat the whole analysis; they are not independent samples. \emph{Secondary analyses} split students at random, so the same schools appear in training and test: five re-fits with different seeds on all students, and three re-fits on separate samples of 5{,}000 students. \emph{Sensitivity analyses} repeat everything with a group label built on all data, including validation and test outcomes, and with the problem-identifier order.
	
	\begin{table}[t]
		\centering\small
		\caption{The five splits of the main analysis, with different schools in each part. Schools are given as total (group 0 / group 1). The last three columns report the share of students in group 1 in training and test (\(p_1\)) and the training median of school averages used as the cutoff for group assignment. This median is across schools, not students; because school sizes differ, the group-1 student share need not be \(0.5\). No school appears in two parts.}
		\label{tab:splits}
		\resizebox{\linewidth}{!}{%
		\begin{tabular}{@{}lrrrrrrrrr@{}}
				\toprule
			& \multicolumn{2}{c}{Training} & \multicolumn{2}{c}{Validation} & \multicolumn{2}{c}{Test} & \multicolumn{3}{c}{Group assignment} \\
			\cmidrule(lr){2-3}\cmidrule(lr){4-5}\cmidrule(lr){6-7}\cmidrule(lr){8-10}
			Split & Students & Schools & Students & Schools & Students & Schools & \(p_1\) train & \(p_1\) test & Cutoff \\
				\midrule
			1 & 9{,}953 & 120 (60/60) & 3{,}340 & 41 (13/28) & 3{,}368 & 50 (22/28) & 0.58 & 0.81 & 0.671 \\
			2 & 9{,}788 & 125 (63/62) & 3{,}539 & 37 (17/20) & 3{,}334 & 49 (17/32) & 0.74 & 0.60 & 0.671 \\
			3 & 9{,}890 & 142 (71/71) & 3{,}084 & 36 (19/17) & 3{,}687 & 33 (17/16) & 0.52 & 0.61 & 0.689 \\
			4 & 9{,}853 & 113 (57/56) & 3{,}002 & 55 (31/24) & 3{,}806 & 43 (31/12) & 0.53 & 0.20 & 0.696 \\
			5 & 9{,}685 & 138 (69/69) & 3{,}637 & 31 (14/17) & 3{,}339 & 42 (23/19) & 0.62 & 0.51 & 0.688 \\
				\bottomrule
		\end{tabular}}
	\end{table}
	The training cutoff is the median of school-level average success, rather than an individual-student threshold. Because school sizes differ, this school-level median does not create equally sized student groups: the group-1 share ranges from \(0.52\) to \(0.74\) in training and from \(0.20\) to \(0.81\) in test. Reporting these shares makes the composition of each split visible and helps interpret the uncertainty of group comparisons.
	
	\paragraph{Uncertainty.}
	Students in the same school share the group label and much of their variation in performance. So every result on test data has a 95\% interval from a school-level bootstrap: we draw test schools with replacement (2{,}000 draws) and keep all their students \citep{field2007bootstrapping,cameron2015practitioner}. We compute the difference between two predictors for each student and then bootstrap it, so the comparison is paired. Over the five main splits, we report the mean, a Student-\(t\) interval with 4 degrees of freedom, and the range. The splits share schools, so this interval shows how much the analysis varies between splits; it is not a confidence interval for the population. We use balanced cohorts of 150 test students (75 per group) only for the trajectory reorientation score, which is defined per cohort; the smaller test group gives 8 to 10 cohorts per split.
	
	\paragraph{Predictors and group-specific recalibration.}
	All predictors are trained on one-step pairs from the training data, and for longer horizons they are applied again and again to their own output. \emph{Simple baselines}: persistence \(\widehat x_{t+h}=x_t\) (``tomorrow looks like today''); a linear least-squares model of the state; gradient-boosted regression trees on the state (one model per variable, 200 iterations, learning rate 0.1); and the original EDMD model. \emph{Group-specific comparisons}: the projected EDMD model, \(A_P=P^\top AP^\top\); partial projections \(P_\alpha=I-\alpha R\) with \(\alpha\in\{0.25,0.5,0.75\}\); and group-specific recalibration of the original EDMD \citep{pleiss2017fairness,hebert2018multicalibration}. For this recalibration baseline, we fit a separate straight-line correction for each group on its training predictions and apply it to the predicted \(k\). It uses the group label at prediction time and tests whether predictions are calibrated within each group; it is not designed to force the group gap to zero. We evaluate horizons \(h\in\{1,5,10,25\}\) from every observed starting point \(x_t\), \(t\leq25-h\), for each variable and each group. We also evaluate the one-step prediction of the final gap in \(k_{25}\), and the long free runs from \(x_0\). The starting points are not the same for every horizon: \(h=25\) uses only \(t=0\), while shorter horizons also start from later states. At short horizons, the windows of rolling correctness also overlap. This information was observed, so it is not a leak of future outcomes, but it makes the target different from hidden learning or from the correctness on a new item. Section~\ref{sec:results-next-item} adds a next-item target, and Section~\ref{sec:results-rank-matched} compares horizons on a common set of starting points; the error curves of this appendix still use different starting points for different horizons.

	\subsection{Further real-data results}
	\label{app:more-real}
	\subsubsection{ASSISTments: predictions for new schools}
	\label{sec:results-assist}
	We now assess whether the simulator findings have a predictive analogue in observational data. All ASSISTments analyses use time-ordered interactions and the five school-disjoint splits in Table~\ref{tab:splits}, so test schools are excluded from model fitting. These data permit an out-of-sample test of performance prediction, not an estimate of intervention effects. We use ASSISTments as a stress test of contrast suppression: does removing the group contrast also suppress the gap predicted for schools held out from training? Under the level-2 rule in Table~\ref{tab:protocol}, a predictor is flagged if its school-level interval for gap error excludes zero or if its error exceeds persistence at the horizon of interest.
	
\begin{figure}[t]
		\centering
		\includegraphics[width=\textwidth]{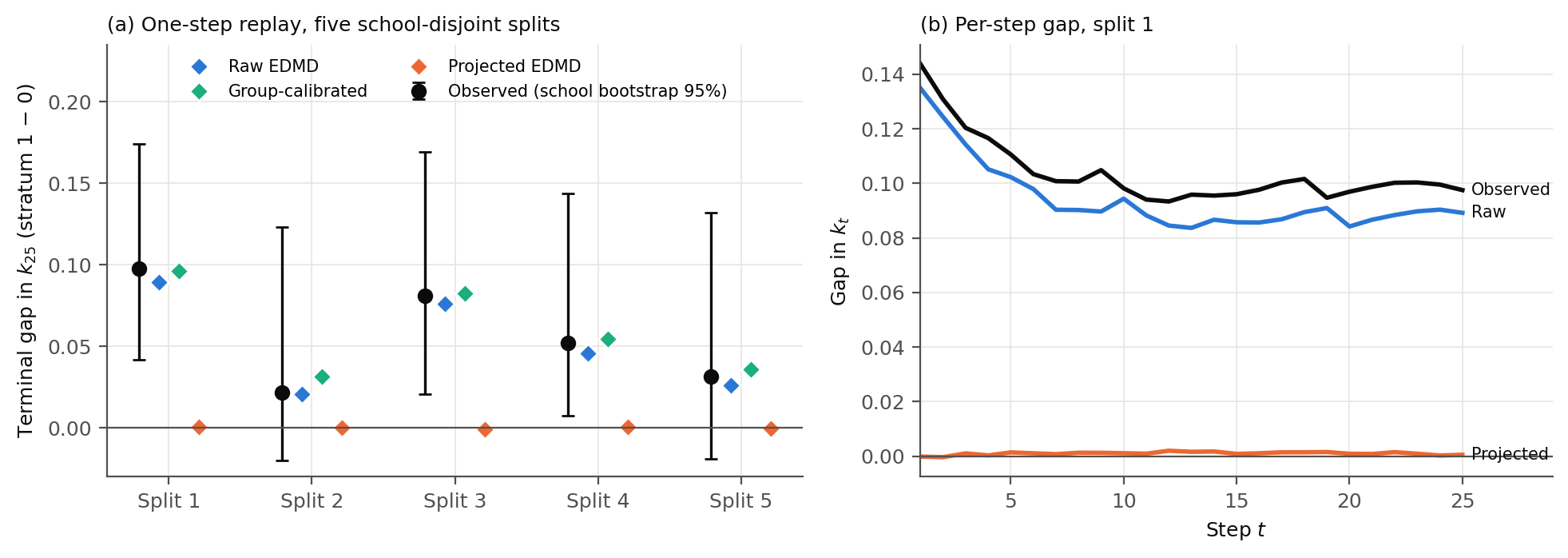}
		\caption{Replay of the observed data on ASSISTments schools never seen in training. Panel (a): observed final gap with its 95\% school-level bootstrap interval, and the one-step predictions of the original, recalibrated and projected EDMD, for each split. Panel (b): the gap at every step in split 1. This tests prediction, not the effect of an intervention.}
		\label{fig:real-replay}
	\end{figure}

	\paragraph{Observed and predicted gaps.}
	We replay the observed sequences using Algorithm~\ref{alg:tutoring-inference} with \(u_t\equiv0\): the model forecasts the next state from each newly observed state. Because every step starts from what really happened, this replay tests the model under the policy that actually generated the data, whatever help the students received, not under an imagined policy. Table~\ref{tab:real-primary} and Figure~\ref{fig:real-replay} compare the observed final gap in \(k_{25}\) with one-step forecasts of that gap. Averaged over the five splits, the observed gap is \(+0.057\) (split-level interval \([0.017,0.097]\); range \(0.022\)--\(0.098\)), and its school-level interval excludes zero in three splits. The original model predicts \(+0.052\) (\([0.014,0.089]\)), modestly underestimating the gap; its gap error is distinguishable from zero in two splits. The projected model instead predicts \(-0.0001\) (\([-0.0010,0.0009]\)), with a gap error distinguishable from zero in the same three splits where the observed gap is. Group-specific recalibration predicts \(+0.060\), and its gap error is compatible with zero in all five splits.

	The contrast is therefore substantive: the projection brings the predicted gap near zero even though a positive gap is observed in held-out schools, while both the original and recalibrated models approximately recover that gap. Under the level-2 rule, projection is flagged in every split: its gap-error interval excludes zero in three and its prediction error exceeds persistence at both one and five steps in all five. The original model is flagged for gap error in two splits, underestimating the gap by about \(0.005\); recalibration triggers neither flag. These zero-based flags are descriptive and should not be read as a prespecified equivalence or non-inferiority assessment.
	
	\paragraph{Labels computed from other students.} 	The held-out-student label variant provides a further check on the retrospective school grouping (Table~\ref{tab:real-sensitivity}). Although only 1{,}655--1{,}894 students per split remain for evaluation, the pattern is similar: the observed gap is \(+0.057\) (\([0.032,0.082]\)); the original and recalibrated models predict \(+0.052\) and \(+0.060\), while projection predicts \(+0.0004\). The projected model's one-step error is 7.6--8.2 times larger. Recalibration's gap error remains compatible with zero in every split, whereas projection's error is significantly negative in three. Thus, the central result does not depend on using evaluated students' future outcomes to define their school's label, although the grouping itself remains retrospective.
	
	\begin{table}[t]
		\centering\small
		\caption{Main ASSISTments results: five splits with new test schools, time order. Observed: final gap in \(k_{25}\) on test schools (group 1 minus group 0), with a 95\% school-level bootstrap interval. Raw, recal., proj.: one-step predictions of this gap by the original EDMD, the EDMD recalibrated for each group, and the projected EDMD. MSE ratio: one-step error on \(k\) of the projected model divided by that of the original model. \(\widehat\tau_\Delta\): long free run from \(x_0\), final gap of the projected model minus that of the original model, with a school-level interval. TRS: mean trajectory reorientation score.}
		\label{tab:real-primary}
		\resizebox{\linewidth}{!}{%
			\begin{tabular}{@{}lrrrrrrr@{}}
				\toprule
				Split & Observed [95\% CI] & Raw & Recal. & Proj. & MSE ratio & \(\widehat\tau_\Delta\) [95\% CI] & TRS\\
				\midrule
				1 & \(0.098\) \([0.042,0.174]\) & \(0.089\) & \(0.096\) & \(0.0005\) & 7.4 & \(-0.0078\) \([-0.0124,-0.0033]\) & 1.50\\
				2 & \(0.022\) \([-0.020,0.123]\) & \(0.021\) & \(0.032\) & \(0.0002\) & 7.7 & \(+0.0022\) \([-0.0052,+0.0068]\) & 1.27\\
				3 & \(0.081\) \([0.021,0.169]\) & \(0.076\) & \(0.082\) & \(-0.0011\) & 8.1 & \(-0.0011\) \([-0.0081,+0.0056]\) & 1.17\\
				4 & \(0.052\) \([0.007,0.144]\) & \(0.046\) & \(0.055\) & \(0.0007\) & 7.8 & \(-0.0044\) \([-0.0106,+0.0011]\) & 1.31\\
				5 & \(0.031\) \([-0.019,0.132]\) & \(0.026\) & \(0.036\) & \(-0.0006\) & 8.0 & \(-0.0060\) \([-0.0088,-0.0029]\) & 1.29\\
				\midrule
				Mean [\(t_4\) interval] & \(0.057\) \([0.017,0.097]\) & \(0.052\) & \(0.060\) & \(-0.0001\) & 7.8 & \(-0.0034\) \([-0.0084,+0.0015]\) & 1.31\\
				\bottomrule
		\end{tabular}}
	\end{table}
	
\begin{figure}[t]
		\centering
		\includegraphics[width=\textwidth]{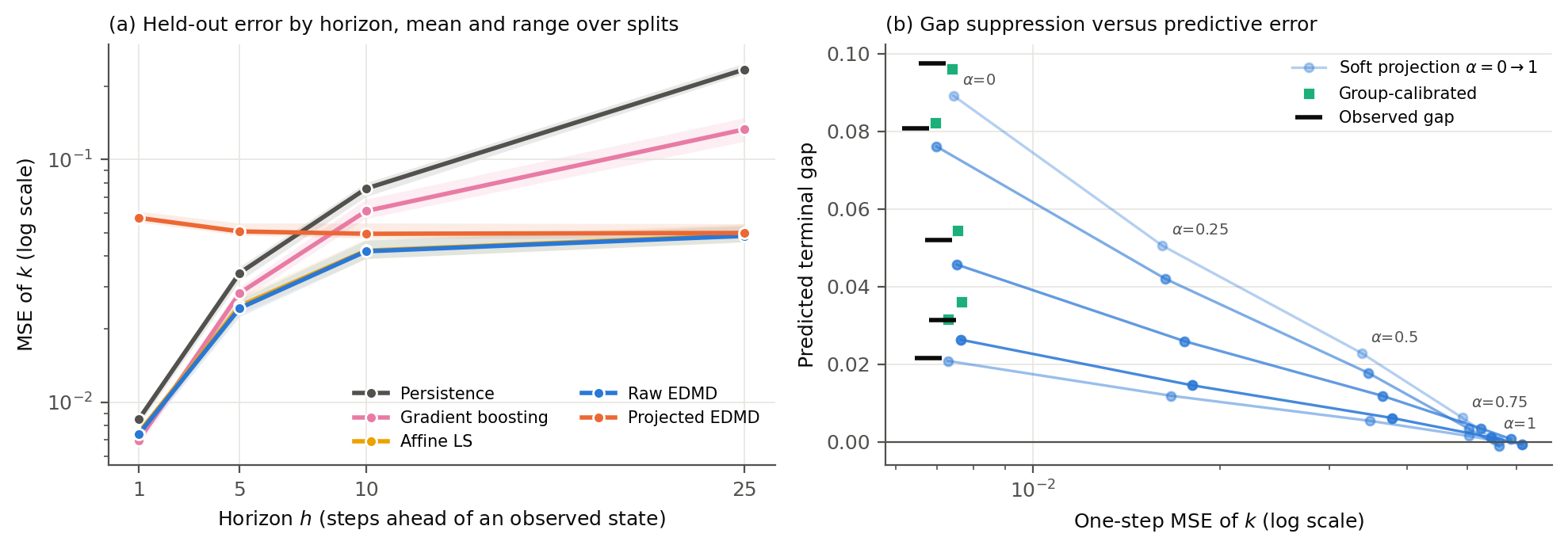}
		\caption{Panel (a): error on \(k\) for new schools, by horizon (log scale); lines are means over the five splits and bands are ranges. Panel (b): one-step error against predicted final gap for partial projections \(P_\alpha=I-\alpha R\), \(\alpha=0,0.25,0.5,0.75,1\) (one line per split, labels for split 1), with the recalibrated model (squares) and the observed gap of each split (bars).}
		\label{fig:real-baselines}
	\end{figure}

	\paragraph{Comparison with simple predictors.}
	Is the original EDMD a good model in the first place? Table~\ref{tab:baselines} and Figure~\ref{fig:real-baselines}(a) compare the error on \(k\) for new schools. Table~\ref{tab:paired-differences} gives, for every split, the paired difference with the original EDMD and its school-level interval. One step ahead, gradient boosting is the best (\(0.0070\)), then the original EDMD (\(0.0074\)), the linear model (\(0.0075\)) and persistence (\(0.0085\)). For each of these differences, the paired school-level interval excludes zero in at least four of the five splits. Five steps ahead and more, the original EDMD and the linear model are the best, and they are almost equal (\(0.0485\) and \(0.0488\) at \(h=25\)), while gradient boosting and persistence get much worse (\(0.133\) and \(0.236\)). So the quadratic features add little to a linear model of the state. The original EDMD is a good model, but not the best one, and this makes it a fair reference for judging the projection.
	
	The projected model has a one-step error of \(0.0575\), which is 7.4 to 8.1 times the error of the original EDMD. It is even worse than simply repeating the last value (persistence) one and five steps ahead. Section~\ref{sec:results-rank-matched} shows that this increase comes mainly from removing the group direction: random directions of the same rank cost much less. The extra error is similar in the two groups (\(0.0615\) against \(0.0077\) in group 0, \(0.0567\) against \(0.0073\) in group 1) and appears in every variable (\(p_t\): \(0.165\) against \(0.056\); \(e_t\) and \(r_t\): 5\% and 7\% higher). At \(h=25\), all EDMD versions reach about the same error (close to \(0.049\)), because they all settle towards a fixed point; even there, the paired difference with the original EDMD stays positive in all five splits.

\begin{figure}[htbp]
		\centering
		\includegraphics[width=\textwidth]{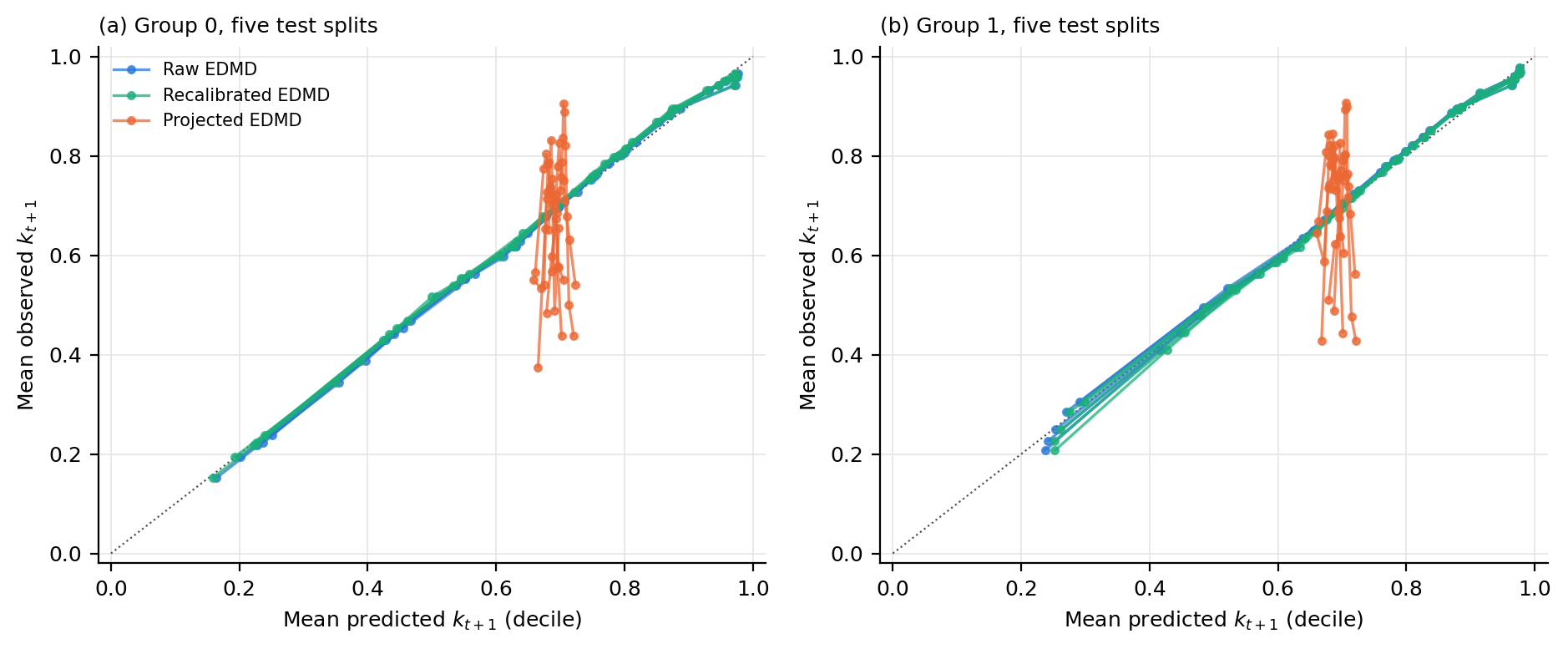}
		\caption{Reliability of the one-step prediction of \(k\) on test schools: mean observed against mean predicted value, by decile of the prediction, one line per split. The dotted line is perfect calibration.}
		\label{fig:real-calibration}
	\end{figure}

	\paragraph{Alternatives that keep the prediction accurate.}
	Could a softer projection find a middle way? Figure~\ref{fig:real-baselines}(b) shows the trade-off. Already at \(\alpha=0.25\), the one-step error is 2.3 times that of the original EDMD, while the predicted gap only falls from \(0.052\) to \(0.029\). No value of \(\alpha\) keeps the error low and removes the gap at the same time. The recalibration for each group behaves very differently. It has the same error as the original EDMD (paired one-step difference below \(10^{-4}\); at \(h=25\) it is even slightly better, with the interval below zero in three splits), and it predicts the observed gap without visible bias. It does not make the predicted gap small, because the gap is present in the held-out data. Is it also well calibrated in each group? Table~\ref{tab:calibration} and Figure~\ref{fig:real-calibration} check this on the test schools, one step ahead, for each group and each split. Its calibration-in-the-large (mean observed minus mean predicted) is compatible with zero in all ten group-split cases. Its calibration slope is within \(0.04\) of 1, and the interval of the slope contains 1 in eight of ten cases (the two others are \(0.984\) and \(1.040\)). The original model is already close to calibrated, but it under-predicts group~1 slightly in two splits; the group correction removes this. The projected model, in contrast, predicts almost the same value for every student, and its slopes are unstable (from \(-2.2\) to \(5.1\)). We therefore describe the recalibration as a \emph{group-specific linear correction that reproduces the observed gap and is approximately calibrated in both groups}, not as a guarantee of calibration. Remember also that it needs the group label when it predicts. Which goal is the right one is an ethical choice. Our check does not make this choice, but it makes the consequences of each option visible.

	\subsubsection{Eedi: predictions for new classes}
	\label{sec:results-eedi}
\begin{figure}[htbp]
		\centering
		\includegraphics[width=\textwidth]{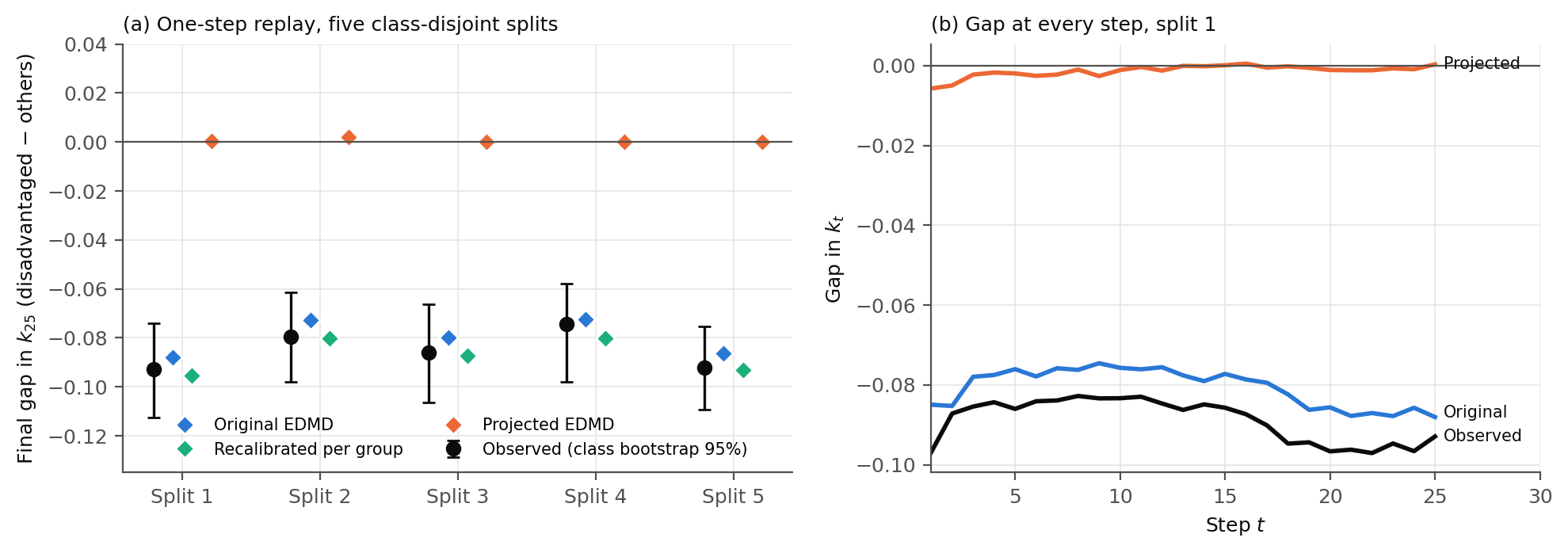}
		\caption{Eedi study (disadvantaged minus other pupils), on classes never seen in training. Panel (a): observed final gap with its 95\% class-level bootstrap interval, and the one-step predictions of the original, recalibrated and projected EDMD, for each split. Panel (b): the gap at every step in split 1. The projected model predicts no gap at any step.}
		\label{fig:eedi-replay}
	\end{figure}

\begin{figure}[htbp]
		\centering
		\includegraphics[width=\textwidth]{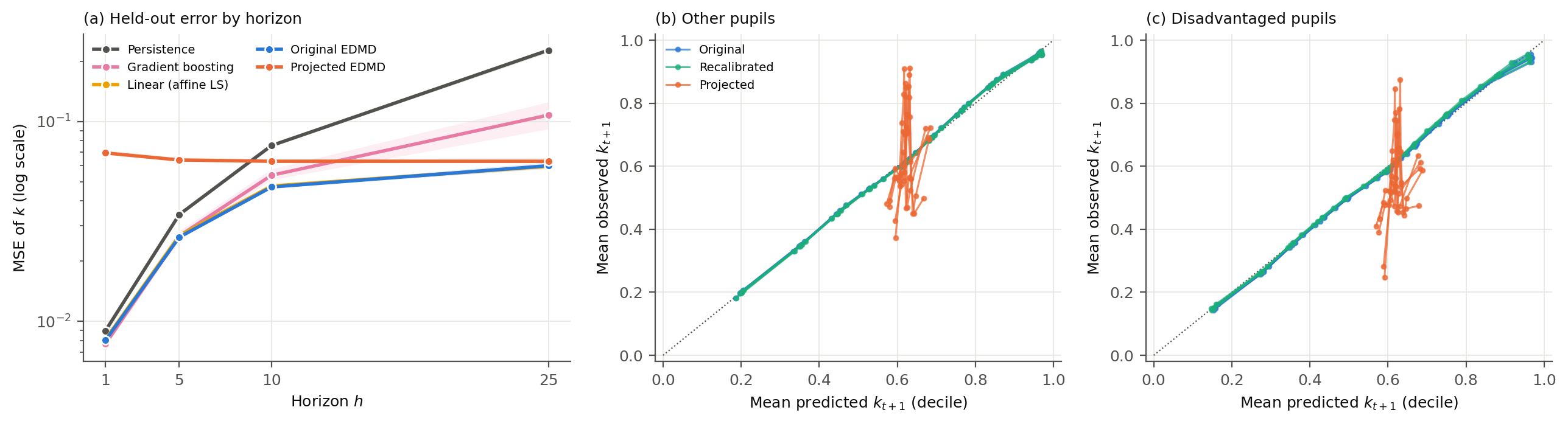}
		\caption{Eedi study. Panel (a): error on \(k\) by horizon, mean over the five splits (bands: range). Panels (b, c): reliability of the one-step prediction of \(k\) for the other and the disadvantaged pupils, by decile of the prediction, one line per split; the dotted line is perfect calibration.}
		\label{fig:eedi-baselines}
	\end{figure}

	Table~\ref{tab:eedi} and Figures~\ref{fig:eedi-replay} and~\ref{fig:eedi-baselines} report predictions for Eedi classes excluded from training. This study tests whether the ASSISTments pattern persists when group membership is a pupil-level socioeconomic attribute recorded independently of performance. By step 25, disadvantaged pupils' recent correctness is about 8.5 percentage points lower than that of other pupils: the mean gap is \(-0.085\), ranging from \(-0.074\) to \(-0.093\) across splits, and every class-level interval excludes zero. The original EDMD predicts \(-0.080\), with a modest calibration-in-the-large bias of about \(-0.006\); its gap-error interval excludes zero in four splits. Group-specific recalibration predicts \(-0.087\), with gap error compatible with zero in four of five splits and calibration slopes between \(0.988\) and \(1.010\).

	\paragraph{A selected sample.} The attribute is recorded for 29\% of pupils, and these pupils differ from the others: they answer more (median 116 answers against 81) and less accurately (mean accuracy 0.58 against 0.68), with standardized differences of about 0.5 on both. The Eedi results therefore describe the pupils whose status is recorded, not all Eedi pupils.

	Projection again removes the gap from the forecast: its predicted value is \(+0.0005\), despite the observed negative gap. Its one-step error is 8.4--9.0 times that of the original model; it is worse than persistence at both one and five steps, and its gap-error interval excludes zero in all five splits. The baseline ranking otherwise resembles ASSISTments: gradient boosting is best one step ahead (\(0.0077\) versus \(0.0080\) for EDMD), while original EDMD and the linear model perform best from five steps onward. Unlike the ASSISTments free runs, the projected model's long-run gap effect is distinguishable from zero in all five splits (\(\widehat\tau_\Delta=+0.007\)). A stricter class-overlap variant yields observed, original, recalibrated and projected gaps of \(-0.080\), \(-0.076\), \(-0.083\) and \(+0.0016\), respectively, with an error ratio of 7.7--8.0. This supports the same conclusion on the smaller selected sample, but does not establish representativeness for pupils whose attribute is unrecorded. Taken together, the Eedi results show that the predictive failure is not confined to retrospectively defined school-performance groups.
	
	\begin{table}[htbp]
		\centering\small
		\caption{Eedi study (group: PremiumPupil, financially disadvantaged pupils minus the others), five class-disjoint splits, time order. Observed: final gap in \(k_{25}\) on test classes with a 95\% class-level bootstrap interval. Raw, recal., proj.: one-step predictions of this gap by the original EDMD, the EDMD recalibrated for each group, and the projected EDMD. MSE ratio: one-step error on \(k\) of the projected model over that of the original model. \(\widehat\tau_\Delta\): long free run from \(x_0\), projected minus original final gap. Last column: whether the class-level interval of the gap error excludes zero, for raw/recal./proj. Strict: mean over five splits without any test pupil who answered in a training class.}
		\label{tab:eedi}
		\resizebox{\linewidth}{!}{%
			\input{tab_eedi.tex}}
	\end{table}

	\subsubsection{OULAD: predictions in distance higher education}
	\label{sec:results-oulad}
\begin{figure}[htbp]
		\centering
		\includegraphics[width=\textwidth]{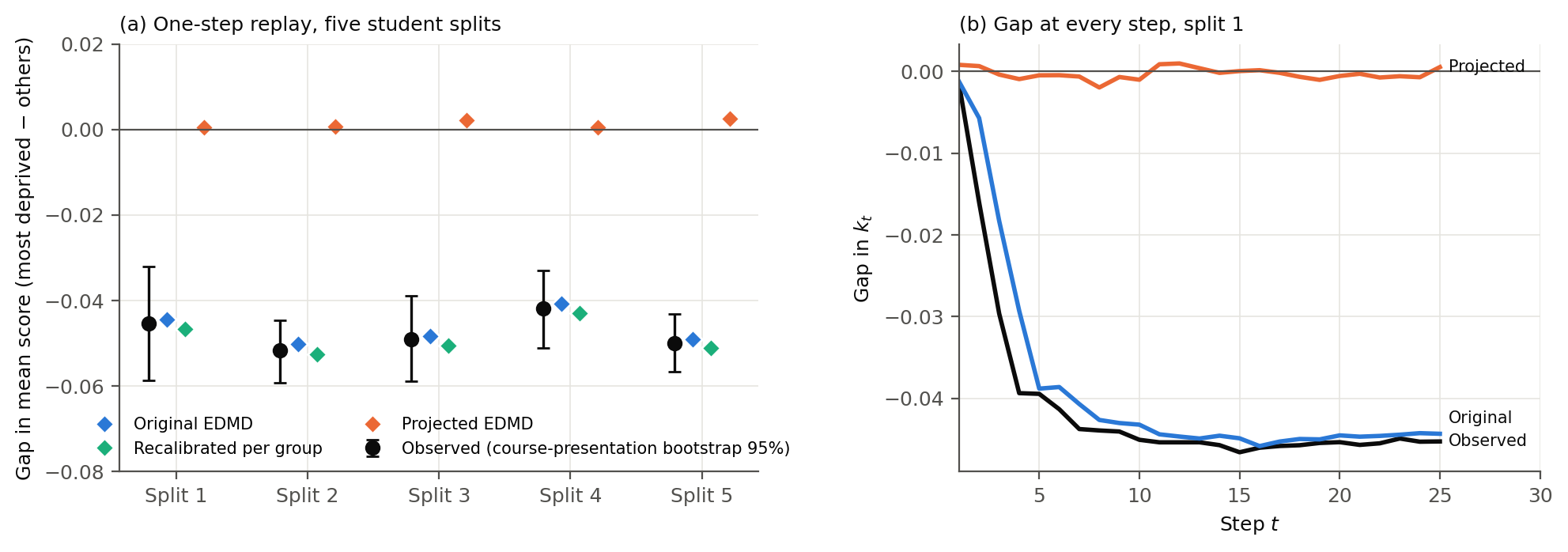}
		\caption{OULAD study, deprivation. Panel (a): observed gap in mean score after 25 weeks (most deprived minus other students) with its 95\% bootstrap interval over course presentations, and the one-step predictions of the three models, for each split. Panel (b): the gap at every week in split 1.}
		\label{fig:oulad-replay}
	\end{figure}

\begin{figure}[htbp]
		\centering
		\includegraphics[width=\textwidth]{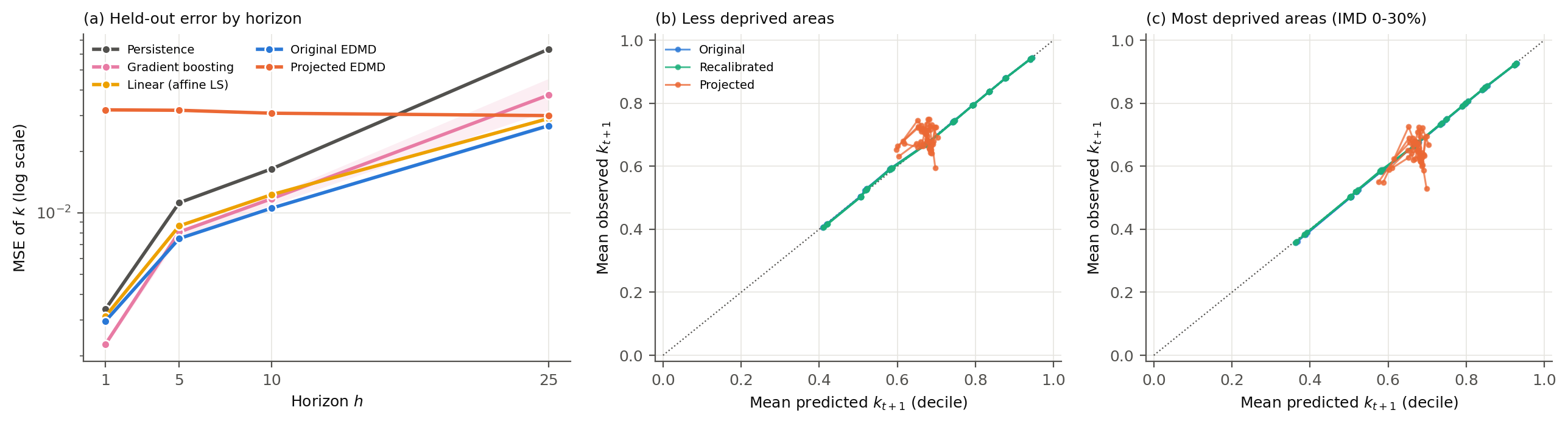}
		\caption{OULAD study, deprivation. Panel (a): error on \(k\) (mean score) by horizon, mean over the five splits (bands: range). Panels (b, c): reliability of the one-step prediction of \(k\) for students from less and most deprived areas, by decile of the prediction, one line per split.}
		\label{fig:oulad-baselines}
	\end{figure}

	Table~\ref{tab:oulad} and Figures~\ref{fig:oulad-replay} and~\ref{fig:oulad-baselines} extend the evaluation to distance higher education. After 25 weeks, students from the most deprived areas have a mean assessment score 4.8 percentage points lower on the normalized 0--1 scale (gap \(-0.048\)); every split's interval excludes zero. The original and recalibrated models predict gaps of \(-0.047\) and \(-0.049\), whereas projection predicts \(+0.0013\). Its one-step error is 10.6--11.0 times the original model's and exceeds persistence in every split. The level-2 diagnostic therefore flags the projected model in all five splits.
	\paragraph{Timing and missing data.} A score enters the state in the week it was submitted; OULAD does not record when the mark became known, so the state can contain a score a few weeks before the system would have it. Before a student's first score, the score proxy is set to 0.5 (30\% of student-weeks; 21\% of students have no score by week 25), and it is carried forward between assessments. Students who withdraw (28\% before week 26) stay in the data with zero activity after leaving, and 46\% of student-weeks have no activity. Only 0.1\% of submitted assessments lack a score; these are dropped.

	The other OULAD attributes help interpret this result. For disability, the smaller observed gap (\(-0.025\)) is approximately reproduced by the original model (\(-0.024\)) but removed by projection (\(-0.0001\)); the error ratio is 8.4--8.9. Gender provides a useful contrast: its observed gap is small (\(-0.006\)) and compatible with zero in every split, yet projection still increases one-step error by a factor of 7.5--9.3. Here the loss does come from compression: removing the gender direction alone costs little, and the extra error comes from the three other removed directions (Section~\ref{sec:results-rank-matched}). The baseline pattern is consistent with the other studies except that gradient boosting is clearly best one step ahead (\(0.0023\) versus \(0.0030\) for EDMD); original EDMD performs best from five steps onward.
	
	The OULAD results also illustrate why statistical detection and practical importance must be distinguished. With large samples, a gap error of \(0.001\)—about 2\% of the deprivation gap—can exclude zero and trigger the original diagnostic flag. A post hoc sensitivity analysis using 10\% of the observed gap as a tolerance still flags projection in all 18 splits with a clearly nonzero observed gap: three ASSISTments splits, five Eedi splits, and five each for OULAD deprivation and disability. Under the same analysis, the original and recalibrated models are each flagged in one split. These counts do not constitute an equivalence assessment with prespecified margins. Moreover, a purely relative tolerance becomes unstable when the reference gap is close to zero, as the gender results demonstrate. A future assessment should instead prespecify the combined absolute and relative margin in Eq.~\eqref{eq:practical-gap-margin} and allow inconclusive outcomes.
	
	\begin{table}[htbp]
		\centering\small
		\caption{OULAD study, five student splits, weekly states over the first 26 weeks; intervals from a bootstrap over the 22 course presentations. Top: deprivation (students from the three most deprived IMD deciles minus the others), per split. Bottom: means over splits for deprivation, disability and gender; in the Observed column, the number of splits whose interval excludes zero. Columns as in Table~\ref{tab:eedi}.}
		\label{tab:oulad}
		\resizebox{\linewidth}{!}{%
			\input{tab_oulad.tex}}
	\end{table}

	\paragraph{Unseen presentations and a later year.}
	In the analysis above, the test students come from the same 22 presentations as the training students. We therefore repeated the OULAD study with two stricter splits: five splits in which whole presentations are held out, and one split that trains on the 2013 presentations and tests on the 2014 ones. The pattern does not change. With held-out presentations, the deprivation gap is \(-0.048\) observed, \(-0.047\) for the original model, \(-0.049\) for the recalibration and \(-0.001\) for the projection, whose one-step error is 11.1 times larger; for disability, the corresponding values are \(-0.032\), \(-0.031\), \(-0.033\) and \(-0.005\) (error ratio 9.1). In the time split, the projected model predicts \(-0.004\) for an observed deprivation gap of \(-0.046\) (error ratio 12.4), and the bootstrap interval of its gap error over presentations is \([0.035,0.051]\); for disability it is \([0.013,0.029]\). Gender again shows no clear gap (\(+0.001\) with held-out presentations).

	\paragraph{OULAD: when does a score become known?} OULAD records when an assessment was submitted, not when its mark became known, so the state can contain a score a few weeks before a system would have it. We therefore delayed every score by one, two or four weeks: since \(k_t\) is the running mean of the scores submitted up to week \(t\), the delayed proxy is \(k_{t-\ell}\) (0.5 before), and the pass indicator is recomputed. Table~\ref{tab:oulad-delay} gives the result, with the error ratio over all transitions and separately over the transitions in which a new score becomes known. The conclusion does not depend on the delay: the projected model predicts almost no gap for every attribute and delay, while the original and group-aware models follow the observed gap. On the weeks with a new score, the projected model's error is 1.2 to 1.7 times the original, the size of the ratio on the next-item target; the large overall ratios come from the weeks without a new score, where the original model only has to carry the score forward.

	\begin{table}[htbp]
		\centering\small
		\caption{OULAD, sensitivity to the delay with which a score becomes known (mean over five student splits). Observed: final gap in \(k_{25}\); original, group-aware, projected: one-step predicted gaps. Error ratio: one-step error on \(k\), projected over original (range over splits), over all transitions and over the transitions with a new score (share in the last column).}
		\label{tab:oulad-delay}
		\resizebox{\linewidth}{!}{%
		\begin{tabular}{@{}llrrrrrrr@{}}
\toprule
Attribute & Delay & Observed & Original & Group-aware & Projected & Ratio, all & Ratio, new score & New-score share\\
\midrule
Deprivation & 0 wk & \(-0.048\) & \(-0.047\) & \(-0.048\) & \(+0.0013\) & 10.6--11.0 & 1.68--1.74 & 0.149\\
 & 1 wk & \(-0.048\) & \(-0.046\) & \(-0.047\) & \(+0.0011\) & 10.4--10.7 & 1.71--1.79 & 0.148\\
 & 2 wk & \(-0.047\) & \(-0.046\) & \(-0.048\) & \(+0.0007\) & 9.8--10.1 & 1.65--1.69 & 0.142\\
 & 4 wk & \(-0.047\) & \(-0.046\) & \(-0.048\) & \(+0.0006\) & 9.3--9.8 & 1.62--1.67 & 0.128\\
\addlinespace
Disability & 0 wk & \(-0.025\) & \(-0.024\) & \(-0.025\) & \(-0.0001\) & 8.4--8.9 & 1.52--1.57 & 0.149\\
 & 1 wk & \(-0.025\) & \(-0.025\) & \(-0.025\) & \(-0.0004\) & 8.1--8.7 & 1.57--1.62 & 0.148\\
 & 2 wk & \(-0.025\) & \(-0.025\) & \(-0.026\) & \(-0.0007\) & 7.4--8.1 & 1.42--1.48 & 0.142\\
 & 4 wk & \(-0.025\) & \(-0.025\) & \(-0.026\) & \(-0.0004\) & 7.1--7.8 & 1.41--1.48 & 0.128\\
\addlinespace
Gender & 0 wk & \(-0.0062\) & \(-0.0067\) & \(-0.0043\) & \(-0.0040\) & 7.5--9.3 & 1.46--1.58 & 0.149\\
 & 1 wk & \(-0.0063\) & \(-0.0069\) & \(-0.0041\) & \(-0.0051\) & 6.3--8.2 & 1.47--1.61 & 0.148\\
 & 2 wk & \(-0.0067\) & \(-0.0072\) & \(-0.0057\) & \(-0.0058\) & 4.7--6.4 & 1.30--1.37 & 0.142\\
 & 4 wk & \(-0.0086\) & \(-0.0088\) & \(-0.0078\) & \(-0.0071\) & 3.3--4.5 & 1.19--1.25 & 0.128\\
\bottomrule
\end{tabular}}
	\end{table}

	\paragraph{Eedi: the target population.} The attribute is recorded for 29\% of Eedi pupils, so the main analysis describes the recorded pupils. To see how much this matters, we modelled the probability of being recorded from what is known for every pupil (number of answers, accuracy, gender and year of birth, with indicators of missing values; logistic regression), and weighted each recorded pupil by the inverse of this probability, trimmed at the 99th percentile. The weights remove the imbalance on answers and accuracy (standardized differences from \(0.32\) and \(-0.38\) to below \(0.002\); effective sample size 26{,}500 of 34{,}386). With these weights, the observed gap on the test classes is \(-0.081\) (unweighted \(-0.085\)), the original model predicts \(-0.076\), the group-aware model \(-0.084\) and the projected model \(0.0000\), with an error ratio of 9.2 (8.6--9.8 over splits); fitting the models with the same weights changes the gaps by at most \(0.001\) and the error ratio to 8.9. Under the assumption that the status is missing at random given these observables, the conclusion therefore also describes all Eedi pupils; the assumption itself cannot be checked with these data.

	\paragraph{Long free runs and other designs.} Started from the observed \(x_0\) and run on their own for 25 steps, both models lose most of the gap, so the long-run effect of the projection is small (\(-0.0034\), interval \([-0.0084,+0.0015]\)) and not established on new schools. The main replay results hold in every other design we tried (student splits, subsamples, other labels, problem order); Appendix~\ref{app:assist-more} gives these analyses.

	\begin{table}[t]
		\centering\small
		\caption{Error (MSE) on \(k\) for new schools, by horizon \(h\), mean over the five splits (paired differences with school-level intervals for every split are in Table~\ref{tab:paired-differences}). Last column: predicted final gap one step ahead, mean over splits; the observed gap is \(0.057\).}
		\label{tab:baselines}
		\begin{tabular}{@{}lrrrrr@{}}
			\toprule
			Predictor & \(h=1\) & \(h=5\) & \(h=10\) & \(h=25\) & Predicted gap\\
			\midrule
			Persistence & 0.0085 & 0.0341 & 0.0759 & 0.2356 & \\
			Affine least squares & 0.0075 & 0.0249 & 0.0421 & 0.0488 & \\
			Gradient boosting & 0.0070 & 0.0282 & 0.0614 & 0.1334 & \\
			Raw EDMD & 0.0074 & 0.0244 & 0.0419 & 0.0485 & 0.052\\
			Recalibrated EDMD & 0.0074 & 0.0244 & 0.0418 & 0.0483 & 0.060\\
			Soft projection, \(\alpha=0.25\) & 0.0169 & 0.0480 & 0.0494 & 0.0499 & 0.029\\
			Soft projection, \(\alpha=0.5\) & 0.0355 & 0.0505 & 0.0494 & 0.0499 & 0.013\\
			Soft projection, \(\alpha=0.75\) & 0.0514 & 0.0506 & 0.0495 & 0.0499 & 0.003\\
			Projected EDMD & 0.0575 & 0.0506 & 0.0495 & 0.0499 & \(-0.0001\)\\
			\bottomrule
		\end{tabular}
	\end{table}

	\begin{table}[htbp]
		\centering\small
		\caption{Nonlinear erasure on held-out students (Proposition~\ref{prop:general-price}), one-step knowledge, mean over five splits. Price at \(\tau=0\): estimated on the training students, with the share of the group's variance explained in brackets. Achieved \(\tau\): held-out gap of \(f_0\) divided by held-out gap of \(f^*\) (negative: the gap is reversed). Measured: extra MSE of \(f_0\) on held-out students. Frontier at \(\tau\): \((1-\tau)^2\) times the nonlinear or the linear price, split by split. ASSISTments is excluded from the comparison with the frontier: its held-out gap is close to zero in two of the five splits, so \(\tau\) is not reliable (\(^\dagger\)); its row is shown for completeness.}
		\label{tab:general-price}
		\resizebox{\linewidth}{!}{%
		\begin{tabular}{@{}lrrrrrr@{}}
			\toprule
			& \multicolumn{2}{c}{Price at \(\tau=0\)} & & & \multicolumn{2}{c}{Frontier at achieved \(\tau\)}\\
			\cmidrule(lr){2-3}\cmidrule(l){6-7}
			Setting & Linear (\(R^2\)) & Nonlinear (\(R^2_{\grp}\)) & Achieved \(\tau\) & Measured & Nonlinear & Linear\\
			\midrule
			ASSISTments (school) & 0.0438 (0.030) & 0.0122 (0.120) & \(+0.89^\dagger\) & 0.0108 & 0.0071 & 0.0203\\
			Eedi (PremiumPupil) & 0.0592 (0.017) & 0.0412 (0.025) & \(+0.18\) & 0.0391 & 0.0276 & 0.0397\\
			OULAD (deprivation) & 0.0245 (0.016) & 0.0184 (0.021) & \(+0.12\) & 0.0176 & 0.0142 & 0.0188\\
			OULAD (disability) & 0.0185 (0.004) & 0.0090 (0.008) & \(+0.40\) & 0.0081 & 0.0035 & 0.0070\\
			OULAD (gender) & 0.0023 (0.060) & 0.0009 (0.158) & \(-0.14\) & 0.0009 & 0.0011 & 0.0030\\
			\bottomrule
		\end{tabular}}
	\end{table}

	\subsubsection{Correcting the dynamics only: details}
	Proposition~\ref{prop:general-price} says that erasing the group from the state is expensive because the state carries the gap. A different correction leaves the state alone and removes only the effect of the group on the dynamics. We fit a group-aware model \(z_{t+1}=Az_t+a+(\grp-p_1)(A_{\grp}z_t+a_{\grp})\) on the training students and use it with the group term set to zero, \(z_{t+1}=Az_t+a\). Two students in the same state then receive the same forecast (equal treatment given the state), while the differences between their states are kept. Table~\ref{tab:dynamics-only} compares this model with the raw, projected and group-aware models, on the splits of Section~\ref{sec:results-realdata}.

	\begin{table}[t]
		\centering\small
		\caption{Correcting the dynamics only (held-out students, mean over five splits). One-step error: MSE of knowledge. Free-running gap: knowledge gap forecast 25 steps ahead from the first state of each student. Same-state difference: mean absolute difference between the one-step knowledge forecasts of the group-aware model for \(\grp=1\) and \(\grp=0\) at the same state (zero by construction for the other models).}
		\label{tab:dynamics-only}
		\resizebox{\linewidth}{!}{%
		\begin{tabular}{@{}lrrrrrrrrr@{}}
			\toprule
			& \multicolumn{3}{c}{One-step error} & & \multicolumn{4}{c}{Free-running gap} & \\
			\cmidrule(lr){2-4}\cmidrule(lr){6-9}
			Setting & Raw & Projected & Dyn.-only & Observed gap & Raw & Projected & Group-aware & Dyn.-only & Same-state diff.\\
			\midrule
			ASSISTments (school) & 0.0074 & 0.0575 & 0.0074 & \(+0.057\) & \(+0.003\) & \(0.000\) & \(+0.067\) & \(+0.003\) & 0.0090\\
			Eedi (PremiumPupil) & 0.0080 & 0.0695 & 0.0080 & \(-0.085\) & \(-0.007\) & \(0.000\) & \(-0.079\) & \(-0.007\) & 0.0077\\
			OULAD (deprivation) & 0.0030 & 0.0319 & 0.0030 & \(-0.048\) & \(-0.003\) & \(0.000\) & \(-0.051\) & \(-0.003\) & 0.0028\\
			OULAD (disability) & 0.0030 & 0.0257 & 0.0030 & \(-0.025\) & \(-0.002\) & \(0.000\) & \(-0.035\) & \(-0.002\) & 0.0025\\
			OULAD (gender) & 0.0030 & 0.0242 & 0.0030 & \(-0.006\) & \(-0.011\) & \(-0.001\) & \(-0.005\) & \(-0.011\) & 0.0038\\
			\bottomrule
		\end{tabular}}
	\end{table}

	The correction keeps accuracy: its one-step error equals that of the raw model in every setting, while the projection multiplies it by seven to eleven (averaged over all starting points). It also treats equal states equally, whereas the forecasts of the group-aware model for the same state differ by \(0.003\) to \(0.009\) per step. But when it forecasts from the start of the 25-step window, it predicts almost none of the gap: \(+0.003\) against an observed \(+0.057\) on ASSISTments, and \(-0.007\) against \(-0.085\) on Eedi. The group-aware model forecasts these gaps well (\(+0.067\) and \(-0.079\)). In the four settings with a clear gap, 91--95\% of the gap that the group-aware model forecasts over 25 steps comes from its group-specific dynamics; the starting state carries the rest. (For gender the gap is too small to split.)

	On these data, the corrected model behaves almost exactly like the raw EDMD model. This is expected: the raw model never sees the group, so in it the group already acts only through the state. The correction therefore does not give a new way out. It makes explicit a choice between two kinds of blindness. Erasing the group from the state hides the gap that exists now and costs accuracy (Proposition~\ref{prop:general-price}); removing the group from the dynamics keeps accuracy but hides how the groups drift apart from similar states. For a decision about one student, treating equal states equally may be the goal. For auditing a tutor, or planning help for a group, a model that cannot forecast the divergence hides it.

	\subsubsection{Further ASSISTments analyses}
	\label{app:assist-more}
	\begin{sidewaystable}
		\centering\small
		\caption{Paired differences on test schools, for every split. Top: MSE of \(k\) of each predictor minus that of the original EDMD (\(\times10^3\)), at horizons 1 and 25; positive means worse than the original EDMD. Bottom: predicted minus observed final gap (\(\times10^3\)). Brackets: 95\% school-level bootstrap intervals; last column: mean and range over the five splits.}
		\label{tab:paired-differences}
		\resizebox{\textheight}{!}{%
			\input{tab_paired_differences.tex}}
	\end{sidewaystable}
	\begin{table}[htbp]
		\centering\small
		\caption{Calibration of the one-step prediction of \(k\) on test schools, for each group and each split. CITL: calibration-in-the-large, mean observed minus mean predicted (\(\times10^3\)); slope: regression of observed on predicted. Brackets: 95\% school-level bootstrap intervals.}
		\label{tab:calibration}
		\resizebox{\linewidth}{!}{%
			\input{tab_calibration.tex}}
	\end{table}
	\paragraph{ASSISTments: long free runs and the ridge value}
	\label{sec:results-real-ridge}
	What happens if we start from the observed \(x_0\) and let both models run on their own for 25 steps? Both lose most of the observed gap. The final gap of the original model is never larger than \(0.008\) in size, because the model settles towards its fixed point (spectral radius 0.89; 0.59 to 0.64 after projection). So the difference \(\widehat\tau_\Delta\) between the projected and original models is almost exactly minus the gap of the original model. On average it is \(-0.0034\) (\([-0.0084,+0.0015]\), range \(-0.0078\) to \(+0.0022\)); its school-level interval excludes zero in only two splits, and 36 of the 48 balanced cohorts give a negative effect. The effect on the average predicted \(k_{25}\) is \(-0.0011\) (\([-0.0039,+0.0017]\)). Changing the ridge over \(\eta/\lambda_{\max}(\Gram)\in\{10^{-6},\ldots,10^{-1}\}\) changes the residuals (\(r_1\) goes from \(0.18\) to \(0.94\) over settings and splits) but does not change \(\widehat\tau_\Delta\) at the fourth decimal, because the projected long-run gap is zero for every metric. So, unlike the replay result, this long-run effect is not established on new schools.
	
	\paragraph{ASSISTments: other designs and sensitivity checks}
	\label{sec:results-closed-loop}
	Table~\ref{tab:real-sensitivity} repeats the main numbers with the other designs. Three results hold everywhere: the original model reproduces most of the observed gap, the projected model predicts a gap within \(0.001\) of zero, and the projected one-step error is 7.3 to 8.1 times the original one. Three other numbers are not stable, and each one teaches us something. First, the observed gap is larger with the group label built on all data (\(+0.088\) against \(+0.062\) with labels from training data only, on the same kind of split). This confirms the concern that a label built with test outcomes makes the gap look larger. Second, the observed gap also depends on the order of the steps (\(+0.084\) in problem order against \(+0.057\) in time order, with new test schools), because the two orders select different rows as the first 26 steps. Third, the long-run effect is clear in all five time-ordered student splits, where the same schools appear in training and test, but only in two of the five splits with new schools, whatever the order. Finally, for the student split with the full-corpus label, the school-level interval of the observed gap, \([0.056,0.120]\), is about twice as wide as the student-level interval \([0.068,0.099]\), which ignores the clustering of students in schools.
	
	\begin{table}[t]
		\centering\small
		\caption{Other designs and sensitivity checks. Values are means over the listed re-fits; ``CI \(\not\ni0\)'' counts the re-fits whose school-level interval for \(\widehat\tau_\Delta\) does not contain zero. In student splits, the same schools appear in training and test.}
		\label{tab:real-sensitivity}
		\resizebox{\linewidth}{!}{%
			\begin{tabular}{@{}llrrrrrrr@{}}
				\toprule
				Order & Design (re-fits) & Observed & Raw & Proj. & MSE ratio & \(\widehat\tau_\Delta\) & CI \(\not\ni0\) & TRS\\
				\midrule
				Time & School-disjoint, retrospective labels (5) & 0.057 & 0.052 & \(-0.0001\) & 7.8 & \(-0.0034\) & 2/5 & 1.31\\
				Time & Student split, training-student labels (5, seeded) & 0.062 & 0.056 & \(-0.0004\) & 7.9 & \(-0.0044\) & 5/5 & 1.35\\
				Time & School-disjoint, labels from other students (5) & 0.057 & 0.052 & \(0.0004\) & 7.8 & \(-0.0033\) & 2/5 & 1.31\\
				Time & Disjoint 5{,}000-student subsamples (3) & 0.056 & 0.050 & \(-0.0005\) & 7.9 & \(-0.0033\) & 2/3 & 1.23\\
				Time & Student split, full-corpus label (1) & 0.088 & 0.077 & \(-0.0005\) & 7.6 & \(-0.0067\) & 1/1 & 1.27\\
				\midrule
				Problem id & School-disjoint, retrospective labels (5) & 0.084 & 0.075 & \(0.0002\) & 7.8 & \(-0.0043\) & 2/5 & 1.35\\
				Problem id & Student split, training-student labels (5, seeded) & 0.067 & 0.059 & \(0.0003\) & 7.7 & \(-0.0029\) & 3/5 & 1.46\\
				Problem id & Student split, full-corpus label (1) & 0.084 & 0.075 & \(-0.0007\) & 7.6 & \(-0.0035\) & 1/1 & 1.47\\
				\bottomrule
		\end{tabular}}
	\end{table}

	\subsection{Trajectory reorientation score}
	\label{app:trs}
	\subsubsection{Definition}
	\label{sec:trs}
	A summary at group level can hide large changes for individual students. To look at individuals, we adapt the Individual Bias measure of \citet{corriera2026individual}. We call it a \emph{reorientation} score because it measures a change of direction in a geometric sense, which is not, on its own, an injustice to a student.
	
	\begin{definition}[Trajectory reorientation score]
		\label{def:trs}
		Fix a cohort \(\{(x_{i, 0}, \grp_i)\}_{i=1}^n\). For a policy or surrogate \(\pi\) with matched noise, let \(\tau_i^\pi\in\R^{m}\) be a stacked trajectory of student \(i\) (controls \(\tau_{u,i}^\pi\in\R^{Td_u}\) or predicted states \(\tau_{x,i}^\pi=(x_{i,1}^\pi,\ldots,x_{i,T}^\pi)\in\R^{Td_x}\)), let \(\mu^\pi=n^{-1}\sum_i\tau_i^\pi\), and put \(g_i^\pi=\tau_i^\pi-\mu^\pi\). For two vectors define the extended cosine distance
		\[
		d_{\cos}^0(a,b)=
		\begin{cases}
			0,&a=0\text{ and }b=0,\\
			1,&\text{exactly one of }a,b\text{ is zero},\\
			1-\dfrac{\langle a,b\rangle}{\|a\|_2\|b\|_2},&a\neq0\text{ and }b\neq0.
		\end{cases}
		\]
		The node-level score is \(\mathrm{TRS}_i^{\pi_0\to\pi_1}=d_{\cos}^0(g_i^{\pi_0},g_i^{\pi_1})\in[0,2]\). We report its cohort mean \(\overline{\mathrm{TRS}}\) and its standard deviation with divisor \(n\), \(\mathrm{TRS}_G\in[0,1]\) (Popoviciu's inequality).
	\end{definition}
	
	Two centered trajectories that are both zero have distance 0, and a zero and a nonzero trajectory have distance 1; every summary says how many such pairs there are. For nonzero pairs, \(\mathrm{TRS}_i=1-\cos\theta_i\), but \(\arccos(1-\overline{\mathrm{TRS}})\) is not an average angle.
	
	\begin{proposition}[Geometric bound]
		\label{prop:tib-bound}
		Let \(a,b\) be nonzero centered trajectory vectors with \(\|a\|_2\geq m\) and \(\|b\|_2\geq m\), where \(m>0\). Then
		\begin{equation}
			d_{\cos}^0(a,b)\leq \frac{\|a-b\|_2^2}{2m^2},
		\end{equation}
		and the constant \(1/2\) is sharp.
	\end{proposition}
	\begin{proof}
		For nonzero \(a,b\), polarization gives
		\(d_{\cos}^0(a,b)=\bigl(\|a-b\|_2^2-(\|a\|_2-\|b\|_2)^2\bigr)/(2\|a\|_2\|b\|_2)\).
		Dropping the nonnegative squared term and using \(\|a\|_2\|b\|_2\geq m^2\) proves the inequality; equality holds for \(a=-b\) with \(\|a\|_2=m\).
	\end{proof}
	The bound explains why the score is unstable when a norm is small: the same small change \(a-b\) moves the score by an amount of order \(\|a-b\|^2/m^2\). The score also ignores size and depends on how we center the data. For these reasons, we always report it together with four checks on size: (i) the distributions of \(\|g_i^{\pi_0}\|\) and \(\|g_i^{\pi_1}\|\) and their ratio; (ii) the normalized Euclidean distance \(\|g_i^{\pi_1}-g_i^{\pi_0}\|/(\|g_i^{\pi_0}\|+\|g_i^{\pi_1}\|)\in[0,1]\), set to \(0\) when both centered trajectories are zero; (iii) Spearman rank correlations across students between the two terminal predictions and between the two path means; and (iv) the score recomputed with common centering (both families centered at \(\mu^{\pi_0}\)), with no centering, and after standardizing every coordinate by its standard deviation under \(\pi_0\). Note that a state trajectory does not live in the coefficient space where \(R\) acts, so a product \(R\tau_i\) has no meaning. To study the projection, we encode the trajectory again as \(z_i^\pi=\operatorname{vec}(\Psi(x_{i,1}^\pi),\ldots,\Psi(x_{i,T}^\pi))\in\R^{TN}\), center it within the cohort, and report the relative residual \(\rho_i^\pi=\|\mathcal R_T(z_i^\pi-\bar z^\pi)\|/\|z_i^\pi-\bar z^\pi\|\) with \(\mathcal R_T=I_T\otimes R^\top\) and norm \(I_T\otimes\Gamma^{-1}\), set to \(0\) when the centered vector is zero. The code uses exactly these conventions, and every summary reports how many zero cases occur (none on ASSISTments).
	
	In an illustrative simulator run (seed 42, outside the ten replicates of the main study), projection changes the gap by \(-0.133\), and the mean score on the control trajectories is \(0.335\). Both the group summary and the individual summary are unfavorable here; they simply describe two different sides of the same change.
	
	\subsubsection{Results on ASSISTments}
	\label{sec:results-tib-real}
\begin{figure}[!htbp]
		\centering
		\includegraphics[width=\textwidth]{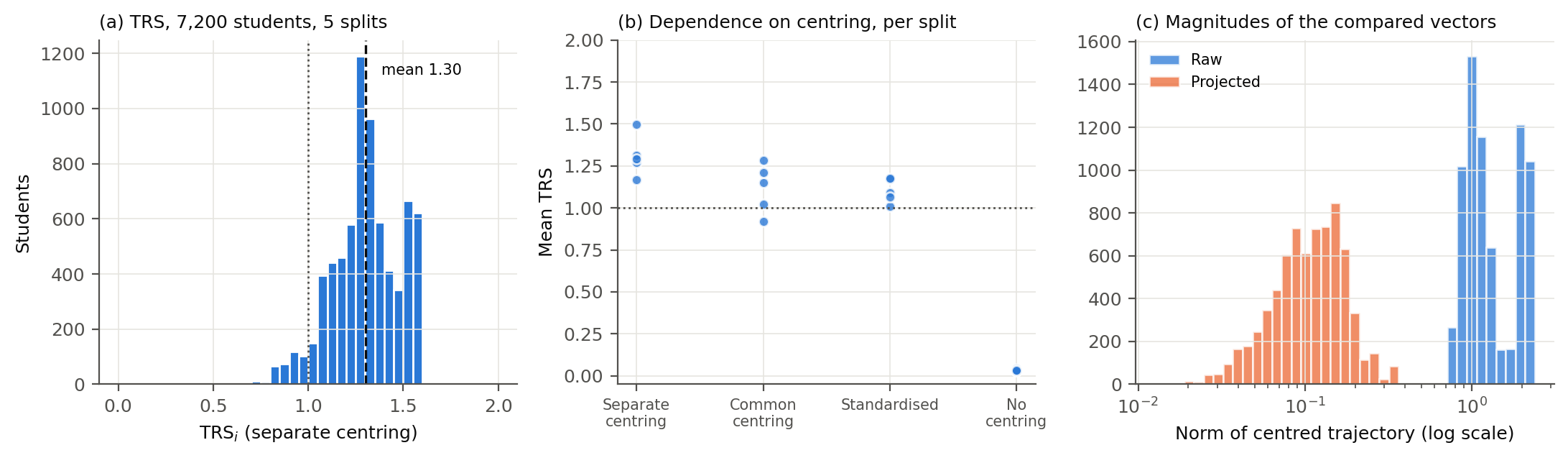}
		\caption{Change of direction between the long-run trajectories of the original and projected models, for 7{,}200 test students in five splits. Panel (a): distribution of \(\mathrm{TRS}_i\); the dotted line marks orthogonal directions. Panel (b): mean score per split with four ways of centering. Panel (c): size of the centered trajectories that the score compares (log scale).}
		\label{fig:real-trs}
	\end{figure}

	For the 7{,}200 students of the 48 cohorts, the mean score between the original and projected long-run trajectories is \(1.30\) (median \(1.30\), \(\mathrm{TRS}_G=0.175\)). For 94.6\% of the students the score is above 1, and there are no zero--zero or zero--nonzero pairs (Figure~\ref{fig:real-trs}(a)). At first sight, this suggests that the projection turns individual predictions around. The checks on size tell a different story. After projection, the centered trajectories are about twelve times shorter (median ratio of norms \(0.067\) to \(0.094\) over splits, Figure~\ref{fig:real-trs}(c)), and the normalized distance is \(0.95\). In other words, the projection almost removes the differences between students, and the score only measures the direction of what is left, which is very small. The value also depends on the centering (Figure~\ref{fig:real-trs}(b)): \(0.92\) to \(1.28\) with common centering, \(1.01\) to \(1.18\) after standardization, and only \(0.03\) without centering, because then both sets of trajectories are dominated by the same average path. The rank correlation between the final predictions of the two models goes from \(-0.40\) to \(0.83\) over splits, so the order of students is not reversed in a stable way. The projection residual is \(0.54\) for the original trajectories and \(0.18\) for the projected ones (\(0.10\) to \(0.25\) over splits). Our conclusion: the projection removes most of the differences between individual students in the model. A mean score above 1 should not be read as a stable reversal of predictions, and it is not a measure of unfairness to individuals.
	
	\FloatBarrier

	\FloatBarrier

	\section{The price, its estimate and next-item prediction}
	\label{app:block-prediction}
	\subsection{The controlled study in detail}
	\label{app:controlled}
	\paragraph{Design.} In every replicate we draw independent training, validation and test sets of \(n\) units each. In the Gaussian cells, \(\grp\sim\mathrm{Bernoulli}(0.4)\), \(z=\varepsilon_z+\grp\,a\,u\) with \(\varepsilon_z\sim\mathcal N(0,I_6)\), and \(y=\beta^\top z+\varepsilon\) with \(\varepsilon\sim\mathcal N(0,1)\), \(\beta=e_1\) and \(u=\cos\theta\,e_1+\sin\theta\,e_2\). The true group posterior is \(q(z)=\sigma(a\,u^\top z-a^2/2+\log(p_1/p_0))\), the true predicted gap is \(\Delta^*=a\cos\theta\), and \(R^2_{\grp}=\operatorname{Var}(q)/(p_0p_1)\) is computed from two million draws. In the shift cells, the training and validation sets use one angle and the test set another. In the misspecified cells, \(z=(X_1,X_2)\) is uniform on \(\{-1,1\}^2\), \(q=0.5+0.2X_1+0.2X_1X_2\) and \(y=X_1+\varepsilon\); the group model is a logistic regression on \((X_1,X_2)\), which is calibrated but misses the interaction (its population limit is \(0.5+0.2X_1\)), or on \((X_1,X_2,X_1X_2)\), which is correct. The learners for \(f\) are least squares with an intercept, refitted in each bootstrap draw, and gradient boosting (100 trees, learning rate 0.1), kept fixed; the group model is an unpenalized logistic regression, refitted in each draw. Each bootstrap draw (1,000 per replicate in Tables~\ref{tab:controlled} and~\ref{tab:controlled-robust}; 60 in a first run on the same data, Tables~\ref{tab:controlled-full-ls} and~\ref{tab:controlled-full-gb}) resamples the training set (multinomial weights, with the refits above) and the validation set; this gives the \emph{audit} interval, used on the real data. A second, \emph{conditional} interval resamples the validation set only, with the fitted \(f\) and \(s\) fixed (400 draws). The announced quadratic part is \(\widehat c_s^2\widehat{\operatorname{Var}}(s)\) on the validation set, and the announced total adds \(2\widehat\E_{\mathrm{val}}[(f-y)d]\).

	\paragraph{The target.} For each replicate we keep the fitted \(f\), \(s\), \(\widehat c_s\) and centre, draw 200{,}000 new units from the validation law (and from the test law, which differs only in the shift cells), and compute the realized quadratic part \(\E[d^2]\) and the realized total change of the squared error. This is the cost of the correction that was actually fitted, the target named in Section~\ref{sec:price-estimation}. We report how often each interval contains it, and how often the audit interval contains the cost of the population built correction. The scripts are \path{price_reliability_synthetic_v3.py}, \path{price_reliability_synthetic_v4.py} (1,000 draws, at the end of this appendix) and, for the first version, \path{price_reliability_synthetic_v2.py}; each cell has 100 replicates with least squares and 50 with gradient boosting.

	\paragraph{All cells.} Tables~\ref{tab:controlled-full-ls} and~\ref{tab:controlled-full-gb} report, for each cell and sample size: the median announced and realized costs (quadratic part and total); the shares of replicates accepted on validation units and whose gap reduction transfers to test units (T1--T2); and the coverage of the realized total by the audit interval (over all replicates and among accepted ones), by the conditional interval, and the coverage of the population built cost among accepted replicates.

	\begin{table}[htbp]
		\centering\footnotesize
		\caption{Controlled study, every cell, least squares (refitted in each bootstrap draw; first run, 60 draws per replicate). Announced and realized: medians over replicates; realized under the validation law. Acc., transf.: shares of replicates accepted on validation units and whose gap reduction transfers to test units (T1--T2, Definition~\ref{def:interpretable}). Coverage: of the nominal 95\% interval, over 100 replicates. Coverage of the realized total: by the audit interval (all replicates; accepted ones) and by the conditional interval. Last column: coverage of the population built cost by the audit interval of the quadratic part, among accepted replicates.}
		\label{tab:controlled-full-ls}
		\resizebox{\linewidth}{!}{%
			\input{tab_controlled_full_ls.tex}}
	\end{table}

	\begin{table}[htbp]
		\centering\footnotesize
		\caption{Controlled study, every cell, gradient boosting (kept fixed in the bootstrap; first run, 60 draws per replicate). Columns as in Table~\ref{tab:controlled-full-ls}; shares over 50 replicates (an observed coverage of 0.88 has a binomial interval of 0.76 to 0.94).}
		\label{tab:controlled-full-gb}
		\resizebox{\linewidth}{!}{%
			\input{tab_controlled_full_gb.tex}}
	\end{table}

	\paragraph{Stability of the bootstrap.} In four cells (\(n=2{,}000\), both learners, 30 replicates each) we drew 400 bootstrap samples and recomputed the audit interval from the first 40, 60 and 200 of them (script \path{price_reliability_synthetic_v3.py --stability}). Table~\ref{tab:controlled-stability} gives the coverage of the realized total cost and the median of the largest change of an end-point, relative to the interval with 400 draws, as a share of its width. Except for the weak aligned signal (\(a=0.2\)), where the interval is very wide and the diagnostic rarely accepts, 40 draws move the end-points by about a tenth of the width and 200 draws by 3--4\%; for the weak signal, 200 draws still move them by 13--16\%. The next-item audit uses 200 draws.

	\begin{table}[htbp]
		\centering\small
		\caption{Stability of the audit interval with the number of bootstrap draws \(B\) (\(n=2{,}000\), 30 replicates per cell). Left: coverage of the realized total cost of the fitted correction. Right: median over replicates of the largest change of an end-point with respect to \(B=400\), as a share of the width at \(B=400\).}
		\label{tab:controlled-stability}
		\input{tab_controlled_stability.tex}
	\end{table}

	\paragraph{More draws, a second interval and other thresholds.} To check that the coverage does not depend on a small number of draws or on our choices, we ran the study again with 1,000 draws per replicate (script \path{price_reliability_synthetic_v4.py}). The data of every replicate and its first 60 draws are those of the study above, so the columns \(B=60\) of Table~\ref{tab:controlled-robust} repeat the first run (Table~\ref{tab:controlled-full-ls}), and the percentile columns with 1,000 draws repeat Table~\ref{tab:controlled}. From the same draws we also computed a second interval, the basic bootstrap interval \([2x-b_{97.5},\,2x-b_{2.5}]\), where \(x\) is the announced value and \(b_{2.5}\), \(b_{97.5}\) are percentiles of the draws. With least squares, 1,000 draws raise the lowest coverage from 88 to 91\% (\(n=20{,}000\)) and from 94 to 96\% (\(n=2{,}000\)); the basic interval gives 92 to 100\% and 98 to 100\%. With gradient boosting (50 replicates), 1,000 draws give 90 to 100\% (\(n=2{,}000\)) and 92 to 100\% (\(n=20{,}000\)), and the basic interval gives the same ranges. The coverage is lowest where the interval is narrowest relative to the realized cost: 5 to 6\% of the cost for the aligned signal with \(a=1.5\) and for the misspecified group model (\(n=20{,}000\)), against 13\% or more in the cells covered at 97\% or more.

	Panel~(b) changes the thresholds. In every cell with a gap, the coverage among accepted replicates is the same, to two decimals, for the D3 thresholds 0.25, 0.5, 1 and none; the threshold changes only how often the audit abstains. Without D3, or with a threshold of 1, the audit accepts up to 7\% of the replicates where there is no gap to remove (panel b; up to 12\% with \(n=20{,}000\) or gradient boosting); with 0.5 it accepts none. With 0.25 it also rejects replicates whose cost is well estimated: 44\% for the partly aligned signal with \(a=0.7\) and \(n=2{,}000\), and 62\% with \(a=0.2\) and \(n=20{,}000\). With 1,000 draws the diagnostic is also stricter where the signal is weak: it accepts 5\% of the aligned replicates with \(a=0.2\) and \(n=2{,}000\), against 13\% with 60 draws, because the coefficient of variation and the intervals of D1--D2 are estimated more precisely. For the transfer check, the T2 thresholds 0.2 and 0.3 give the same shares except for the weak partly aligned signal (72 and 90\%), and 0.1 is stricter there (42\%); when the effect of the group signal changes between training and use (utility lost), at most 1\% of the replicates transfer, for every threshold.

	\begin{table}[htbp]
		\centering\footnotesize
		\setlength{\tabcolsep}{4pt}
		\caption{Controlled study, least squares: more bootstrap draws, a second interval and other thresholds (100 replicates per cell, the data of Table~\ref{tab:controlled}). (a) Share of the replicates accepted with the thresholds of Definition~\ref{def:interpretable} whose nominal 95\% audit interval contains the realized total cost of the fitted correction: percentile interval from the first 60 draws (the first run, as in Table~\ref{tab:controlled-full-ls}) and from 1,000 draws, and basic interval from 1,000 draws. Cells without a gap are not shown: the audit abstains there. (b) Share of the replicates accepted on validation units for four thresholds of D3 on the coefficient of variation of the announced price (0.5 is ours; ``none'' keeps D1--D2 only), and share of the replicates accepted with 0.5 whose gap reduction transfers to test units, for three thresholds of T2 on \(|\tau|\) (0.2 is ours). Panel (b) uses 1,000 draws.}
		\label{tab:controlled-robust}
		\input{tab_controlled_robust.tex}
	\end{table}

	\subsection{The first version of the estimation procedure}
	\label{app:v1-procedure}
	We report the first version of the procedure as a comparator. The group model, the price \(p_0p_1\widehat\Delta^2/\widehat R^2\) with \(\widehat R^2=\widehat{\operatorname{Var}}(\widehat q)/(p_0p_1)\), and the coefficient \(\widehat c=\widehat{\operatorname{Cov}}(f,\widehat q)/\widehat{\operatorname{Var}}(\widehat q)\) were all computed on the training units, and the estimate was declared interpretable on the test units if (R1) the test gap of \(f\) was detectable, (R2) \(\widehat q\) was calibrated on the test units (slope in \([0.8,1.2]\) and expected calibration error at most 0.05), (R3) the correction left at most 20\% of the test gap, and (R4) the coefficient of variation of the price over refits was at most 0.5. It has two problems, which the three-stage audit of Section~\ref{sec:price-estimation} corrects.

	First, \(\widehat{\operatorname{Var}}(\widehat q)/(p_0p_1)\) equals \(\operatorname{corr}(\widehat q,\grp)^2\) only for a calibrated group model, and a group model evaluated on its own training data looks more informative than it is. The announced price was therefore too low when the sample was small or the group signal weak: in the controlled study, 0.79 for a true 1.01 (aligned signal, \(a=0.2\), \(n=2{,}000\)), and at \(n=20{,}000\) its interval covered the truth in only 41\% of the replicates that the rule accepted (column v1 of Table~\ref{tab:controlled}; script \path{price_reliability_synthetic_v2.py}). On the Eedi network, the first version announced 0.016 to 0.018, below the measured costs, and the rule accepted one split of five, because the group model was overconfident on new classes (calibration slopes 0.73 to 0.76). Recalibrating the group model on the validation classes by Platt scaling raises the acceptance to three splits of five; on ASSISTments recalibration does not help, because the calibration learned on validation schools does not transfer to the test schools (Table~\ref{tab:next-item-rel}).

	Second, calibration and stability do not make the estimate the minimum price: a calibrated group model that misses part of the group signal passes all four conditions while the minimum is overstated (Proposition~\ref{prop:built} and the example of Section~\ref{sec:price-estimation}). The three-stage audit does not remove this limit; it states it, by announcing the cost of the correction that can be built, which, for the best predictor, is an upper bound of the minimum.

	\subsection{Next-item prediction in detail}
	\label{app:next-item}
	\paragraph{Data and windows.} Each student contributes the first 26 items of the window used everywhere in the paper; the target is the correctness of items \(2\) to \(26\) (partial credit, 0.01\% of the ASSISTments answers, counts as incorrect). The splits are the five school-disjoint (ASSISTments) or class-disjoint (Eedi) splits of the main study, with their training, validation and test units.

	\paragraph{Selection of the network.} The network embeds the skill and the correctness of each item in 32 dimensions, reads the sequence with a GRU, and gives each skill its own output weights, applied to the hidden state after dropout. For each split we train the twelve configurations of the grid with Adam (batch 256, at most 30 epochs) and keep, for each, the epoch with the lowest log-loss on the validation units (patience 3); the configuration with the lowest validation log-loss is used. Table~\ref{tab:next-item-grid} gives the selected configurations.

	\begin{table}[htbp]
		\centering\small
		\caption{Selected DKT configuration in each split, with its number of epochs and validation log-loss, and the range of the validation log-loss over the twelve configurations.}
		\label{tab:next-item-grid}
		\resizebox{\linewidth}{!}{%
			\input{tab_next_item_grid.tex}}
	\end{table}

	\paragraph{The four removals.} The LEACE erasers use the closed form for a binary concept on the training hidden states; the skill-specific erasers are fitted for every next skill with at least 50 training rows in each group, the global eraser being used for the other skills. The output layer is refitted on the erased training states with early stopping on the erased validation states. The mean-parity fine-tuning starts from the selected network, with one third of its learning rate, for at most ten epochs, and adds \(\lambda\) times the squared difference of the group means of the batch predictions to the loss; early stopping uses the validation log-loss plus the same penalty. The removal at the output uses a logistic model of the group with a linear term in the hidden state and one intercept per next skill, fitted by L-BFGS on the training rows.

	\paragraph{The cost of the correction, split by split.} Table~\ref{tab:next-item-rel} compares, in every split, three ways of announcing the cost of the output correction. The first version (Appendix~\ref{app:v1-procedure}) fits the group model and computes the price and the coefficient on the training units. The second recalibrates the group model on the validation units with Platt scaling before doing the same. The third is the three-stage audit of Section~\ref{sec:price-estimation}. For each split the selected network is retrained with the configuration and seed of Table~\ref{tab:next-item-grid}; it reproduces the AUC of Table~\ref{tab:next-item}. The scripts are \path{next_item_v2.py} (first version and Platt) and \path{next_item_v3.py} (three-stage audit).

	Tables~\ref{tab:next-item-audit} (Eedi) and~\ref{tab:next-item-audit-assist} (ASSISTments) give the three intervals of the three-stage audit (Table~\ref{tab:audit-data}) with three sets of units, the validation units selecting the network and announcing the cost (the main audit, with four sets of units, is in Table~\ref{tab:next-item-audit-main}), each from cluster-bootstrap draws:
	\begin{itemize}[leftmargin=*,itemsep=1pt,topsep=2pt]
		\item \emph{announced}: 200 draws that resample the training schools or classes, with the group model refitted (60 L-BFGS iterations from the fitted value), and the validation schools or classes; the test units are not used;
		\item \emph{measured}: 200 draws that resample the test schools or classes, with the fitted correction fixed;
		\item \emph{difference}: in each of the 200 draws of the announced interval, the test schools or classes are resampled too, and the measured cost of that draw's correction minus its announced total is recorded.
	\end{itemize}
	The network is not refitted, so all intervals are conditional on the fitted network. With 40 draws instead of 200 (\path{next_item_v2.py}), the announced intervals on Eedi were similar.

	\paragraph{The training-fitted correction.} Table~\ref{tab:next-item} reports the three-stage output correction. When instead the group model, the coefficient and the centre are all fitted on training classes (the first version), the correction leaves 11\% of the Eedi gap (median), lowers the AUC by \(0.055\) and raises the Brier score by \(0.016\); on ASSISTments it lowers the AUC by \(0.015\) and raises the Brier score by \(0.003\).

	\begin{table}[htbp]
		\centering\footnotesize
		\caption{Three-stage audit of the output correction on Eedi, per split: the three intervals of Table~\ref{tab:audit-data} (nominal 95\%, class bootstrap, 200 draws, conditional on the fitted network). Announced: quadratic part \(\widehat\pi_s\) and total cost on validation classes. Measured: \(\E[d^2]\) and total extra Brier score on test classes (unclipped predictor). Last columns: measured minus announced. The announced intervals are asymmetric because the cost varies as \(1/\widehat\rho_s^2\).}
		\label{tab:next-item-audit}
		\resizebox{\linewidth}{!}{%
			\input{tab_next_item_audit_eedi.tex}}
	\end{table}

	\begin{table}[htbp]
		\centering\footnotesize
		\caption{Three-stage audit of the output correction on ASSISTments, per split; columns as in Table~\ref{tab:next-item-audit}. The audit abstains in every split (Table~\ref{tab:next-item-rel}); the very wide announced intervals show that the cost is not identified on new schools.}
		\label{tab:next-item-audit-assist}
		\resizebox{\linewidth}{!}{%
			\input{tab_next_item_audit_assist.tex}}
	\end{table}

	\begin{table}[tbp]
		\centering\small
		\caption{Cost of the output correction, per split, for the first version of the procedure (group model, price and coefficient on training units), the same after Platt recalibration of the group model on validation units, and the three-stage audit. Slope: calibration slope of the group model on test units. \(\widehat\tau\): remaining share of the test gap. Accepted / transfers: the diagnostic D1--D3 of Definition~\ref{def:interpretable} on validation units, and whether the gap reduction transfers to test units (T1--T2), with the conditions that fail.}
		\label{tab:next-item-rel}
		\resizebox{\linewidth}{!}{%
			\input{tab_next_item_rel.tex}}
	\end{table}

	\paragraph{Three sets of units against four.} In the analysis above, the validation units select the network, stop its training and announce the cost. The main audit of Section~\ref{sec:results-next-item} removes this overlap: we divided the validation schools (ASSISTments) or classes (Eedi) of each split into two halves with about the same number of students (script \path{next_item_v4.py}); the training and test units are those of the three-split analysis. The first half (\emph{selection}: about 3{,}400 pupils on Eedi, 1{,}300 to 1{,}700 students on ASSISTments) selects the network on the grid of Table~\ref{tab:next-item-grid}, stops its training and the refit of the LEACE readouts, and sets the weight of mean-parity fine-tuning. The second half (\emph{announcement}: about 3{,}450 pupils on Eedi, 1{,}700 to 2{,}000 students on ASSISTments) is used only to fit \(\widehat c_s\) and the centre, to announce the cost and to run D1--D3. The intervals use 400 cluster-bootstrap draws. Panel~(a) of Table~\ref{tab:next-item-v4} compares the two versions. On Eedi the selected networks have almost the same AUC (0.715 to 0.723, differences of at most 0.001), the decision is the same in every split, and the split that fails T2 is the same; with four sets of units, the interval of the difference between measured and announced cost contains zero in all five splits (four of five with three sets), and the announced intervals are 1.3 to 2.5 times wider, because half as many pupils announce the cost. On ASSISTments both versions abstain in every split. Panel~(b) applies the same correction to the test units without and with clipping to \([0,1]\) (200 cluster-bootstrap draws, correction fixed); the clipped predictor is the one of Table~\ref{tab:next-item}, the unclipped one is the one of Proposition~\ref{prop:built} and of the audit.

	\begin{table}[htbp]
		\centering\footnotesize
		\setlength{\tabcolsep}{4pt}
		\caption{Next-item prediction: the audit of the output correction with three sets of units (the validation units select the network and announce the cost; Tables~\ref{tab:next-item-audit} and~\ref{tab:next-item-rel}) and with four (a selection half and an announcement half of the validation units; Table~\ref{tab:next-item-audit-main}, 400 cluster-bootstrap draws). (a) AUC of the selected DKT on test units, total cost announced, width of its nominal 95\% interval, cost measured on test units, and decision of Definition~\ref{def:interpretable}: tr., accepted and the gap reduction transfers; acc., accepted but the gap reduction does not transfer; abst., the audit abstains. (b) With four sets of units, the same correction on test units: share of its values outside \([0,1]\), extra Brier score and remaining share \(\widehat\tau\) of DKT's predicted gap, without and with clipping to \([0,1]\), and whether the clipped cost lies in the announced interval (200 cluster-bootstrap draws, correction fixed).}
		\label{tab:next-item-v4}
		\input{tab_next_item_v4.tex}
	\end{table}

	\paragraph{Longer histories.} On Eedi, the students with at least 101 items (about 11{,}500 training and 3{,}800 test students per split, two splits) are predicted on the same 25 target items (items 77 to 101) from windows of 25, 50 and 100 previous items. The network is selected on the validation classes from two configurations (hidden size 64 or 128, learning rate \(3\times10^{-3}\)); the price and the removal at the output are computed as above. Table~\ref{tab:next-item-long} gives the means over the two splits.

	\begin{table}[htbp]
		\centering\small
		\caption{Eedi, longer histories: the same 25 targets predicted from windows of 25, 50 and 100 items (mean over two splits). Removal: the output correction of the first version of the procedure (group model, price and coefficient on training classes; Appendix~\ref{app:v1-procedure}), paired with the selected DKT. Price: announced on training classes; measured: on test classes. Interpretable: number of splits that pass the four conditions R1--R4 of that first version (not the diagnostic of Definition~\ref{def:interpretable}). Predictions are clipped to \([0,1]\) for the AUC only; costs use the unclipped predictor.}
		\label{tab:next-item-long}
		\input{tab_next_item_long.tex}
	\end{table}

	\begin{table}[htbp]
		\centering\small
		\caption{Eedi, longer histories, three-stage audit of the output correction (mean over two splits; network retrained with the selected configuration; 200 bootstrap draws). Announced total: on validation classes; \(\widehat\tau\) and measured total extra Brier score (unclipped predictor; 0.5 to 1.1\% of the values leave \([0,1]\)): on test classes; last columns: number of splits in which the measured cost lies in the announced interval, in which the diagnostic of Definition~\ref{def:interpretable} accepts the estimate, and in which the gap reduction transfers (T1--T2). Script \texttt{next\_item\_v3.py --long}.}
		\label{tab:next-item-long-v2}
		\resizebox{\linewidth}{!}{%
			\input{tab_next_item_long_v2.tex}}
	\end{table}

	\subsubsection{The fixed DKT configuration of the previous version}
	\label{sec:next-item-fixed}
	Before the DKT was selected on validation data (Section~\ref{sec:results-next-item}), we used a fixed configuration. We report these results because they include the group-aware and dynamics-only EDMD readouts. We predict whether the student answers item \(t+1\) correctly, for \(t=0,\dots,24\), on the held-out students of the same splits, for ASSISTments and Eedi, where item-level correctness is recorded (partial credit, 0.01\% of the ASSISTments answers, counts as incorrect). Each EDMD model predicts the next state, and a logistic readout fitted on the training students turns it into a probability. All models, DKT included, see the same 26-item windows, so the comparison is fair within this study; it is not a comparison with published DKT results, which use full histories. As a knowledge-tracing baseline we train deep knowledge tracing (DKT, \citealp{piech2015deep}), a recurrent network over the skill and correctness of past items, which is also given the skill of the next item. Its architecture and training are fixed, without a hyperparameter search: the skill and the correctness of each item are embedded in 32 dimensions, a GRU with 64 hidden units reads the sequence, and each skill has its own output weights; it is trained with Adam (learning rate 0.003, batch 256) for 20 epochs, with fixed seeds and the same settings on both data sets. We then remove the mean group contrast from DKT in two ways: with the construction of Proposition~\ref{prop:general-price} (first version, Appendix~\ref{app:v1-procedure}), \(f_0=f-c\,(q-p_1)\), where \(z\) is the information that DKT uses for its prediction, its hidden state and the skill of the next item, and \(q\) is a logistic regression of the group on this same \(z\), fitted on the training students; and with LEACE applied to the hidden state, the output layer being refitted.

	\begin{table}[t]
		\centering\small
		\caption{Next-item correctness on held-out students (mean over five splits). AUC: overall and, in brackets, for the worse group. Gap: mean predicted probability of group 1 minus group 0; the observed gap is \(+0.044\) on ASSISTments and \(-0.082\) on Eedi. Brier: mean squared error of the probabilities, the error of Proposition~\ref{prop:general-price}.}
		\label{tab:next-item-fixed}
		\resizebox{\linewidth}{!}{%
		\begin{tabular}{@{}lrrrrrr@{}}
			\toprule
			& \multicolumn{3}{c}{ASSISTments (school)} & \multicolumn{3}{c}{Eedi (PremiumPupil)}\\
			\cmidrule(lr){2-4}\cmidrule(l){5-7}
			Model & AUC & Brier & Gap & AUC & Brier & Gap\\
			\midrule
			EDMD, raw & 0.659 (0.653) & 0.1936 & \(+0.021\) & 0.687 (0.673) & 0.2145 & \(-0.046\)\\
			EDMD, projected & 0.518 (0.509) & 0.2082 & \(+0.001\) & 0.521 (0.505) & 0.2377 & \(-0.001\)\\
			EDMD, group-aware & 0.660 (0.653) & 0.1935 & \(+0.036\) & 0.688 (0.673) & 0.2143 & \(-0.082\)\\
			EDMD, dynamics only & 0.659 (0.653) & 0.1936 & \(+0.021\) & 0.687 (0.673) & 0.2145 & \(-0.046\)\\
			DKT & 0.648 (0.640) & 0.2047 & \(-0.000\) & 0.710 (0.697) & 0.2104 & \(-0.050\)\\
			DKT, LEACE on the hidden state & 0.650 (0.642) & 0.2040 & \(+0.004\) & 0.689 (0.677) & 0.2162 & \(-0.033\)\\
			DKT, mean contrast removed (Prop.~\ref{prop:general-price}, first version) & 0.636 (0.629) & 0.2075 & \(-0.001\) & 0.658 (0.645) & 0.2253 & \(-0.008\)\\
			\bottomrule
		\end{tabular}}
	\end{table}

	Table~\ref{tab:next-item-fixed} gives the results. Paired with the raw EDMD readout, with a school- or class-level bootstrap in each split, the projection lowers the AUC by \(0.14\) on ASSISTments and \(0.17\) on Eedi, with intervals below zero in every split; DKT is better than the EDMD readout on Eedi by \(0.023\), with intervals above zero in every split, while on ASSISTments the difference changes sign across splits (from \(-0.05\) to \(+0.07\)). The projection fails on this target too: it brings the AUC close to chance (0.52 on both data sets), while the raw, group-aware and dynamics-only models keep it (0.66 and 0.69). DKT is the best model on Eedi (AUC 0.71); on the 26-item windows of ASSISTments it is not clearly better or worse than the EDMD readout. As before, the group-blind models predict about half of the observed gap on Eedi (\(-0.046\) and \(-0.050\) against \(-0.082\)), and the group-aware model predicts all of it.

	On Eedi, this construction (first version, Appendix~\ref{app:v1-procedure}, with the group model fitted on training pupils) removes 84\% of the gap that DKT predicts and lowers the AUC from 0.71 to 0.66 (Table~\ref{tab:next-item-fixed}). The estimated formula is compatible with the measured cost, but not identical to it. At the share of the gap left, the frontier predicts an extra Brier score of \(0.014\) (mean over five splits) and the measured increase is \(0.015\); the cross term of Eq.~\eqref{eq:cross-term} is at most \(0.0006\). The class-bootstrap interval of the difference between measured cost and frontier contains zero in three splits; in the other two, the measured cost is higher by about \(0.002\) to \(0.003\). The group model is reasonably calibrated on held-out pupils (calibration slope 0.73 to 1.06, expected calibration error at most 0.027). The removed predictor can leave \([0,1]\) (0.8\% of its values); we clip it only for the AUC and the log-loss, and the costs above use the unclipped predictor, which is the one of the proposition. These are estimates: they show that the formula, evaluated with DKT's prediction and an estimated group model, describes the measured cost well; they do not test the assumptions of the proposition, which concern the true \(f^*\) and \(q\). On ASSISTments, the group model does not transfer to the new schools (expected calibration error 0.23), so this check cannot be interpreted there. LEACE on the hidden state removes only a third of the gap on Eedi: the output layer of DKT uses the skill of the next item, so a group signal removed from the hidden state can come back through the skill-specific weights. Removing a contrast from a representation is not the same as removing it from the prediction.

	\subsection{Margins fixed in advance}
	\label{app:margins}
	The flags of Table~\ref{tab:protocol} test whether an error can be \emph{detected}, that is, whether its interval excludes zero. They do not say whether the error is \emph{large enough to matter}. For this, the auditor fixes \emph{margins} before looking at the test results. A margin is the largest error that the user of the model is ready to accept: \(\epsilon_{\mathrm{abs}}\) is the largest acceptable error on the gap, in the units of the outcome (for example \(0.01\) on a knowledge score between 0 and 1); \(\epsilon_{\mathrm{rel}}\) is the largest acceptable error relative to the size of the gap (for example 10\%); and \(\epsilon_{\mathrm{MSE}}\) is the largest acceptable increase of the mean squared error. With a reference size of the gap, \(\delta_{\mathrm{ref}}\), fixed from training or validation data, the margin on the gap is
	\begin{equation}
		\epsilon_\Delta=\max\{\epsilon_{\mathrm{abs}},\epsilon_{\mathrm{rel}}|\delta_{\mathrm{ref}}|\}.
		\label{eq:practical-gap-margin}
	\end{equation}
	Table~\ref{tab:margins} gives the resulting decisions. They are about the stated target, horizon and sampling unit, not about fairness in general.

	\begin{table}[t]
		\centering\small
		\caption{Decisions with margins fixed in advance. ``Interval'' is the 95\% interval of the paired difference (changed model minus reference model), computed at the level of the sampling unit.}
		\label{tab:margins}
		\begin{tabularx}{\linewidth}{@{}>{\RaggedRight\arraybackslash}p{3.0cm}>{\RaggedRight\arraybackslash}X>{\RaggedRight\arraybackslash}X@{}}
			\toprule
			Decision & Error on the gap & Increase of the MSE \\
			\midrule
			Acceptable & the whole interval lies inside \([-\epsilon_\Delta,\epsilon_\Delta]\) (equivalence) & the upper end of the interval is at most \(\epsilon_{\mathrm{MSE}}\) (non-inferiority) \\
			\addlinespace
			Error that matters & the whole interval lies outside \([-\epsilon_\Delta,\epsilon_\Delta]\) & the lower end of the interval is above \(\epsilon_{\mathrm{MSE}}\) \\
			\addlinespace
			Inconclusive & the interval crosses \(-\epsilon_\Delta\) or \(\epsilon_\Delta\) & the interval contains \(\epsilon_{\mathrm{MSE}}\) \\
			\bottomrule
		\end{tabularx}
	\end{table}

	We did not fix margins before our experiments. After them, we applied the rule once, with \(\epsilon_{\mathrm{abs}}=0.01\), \(\epsilon_{\mathrm{rel}}=10\%\) and \(\epsilon_{\mathrm{MSE}}\) equal to 10\% of the one-step error of the original model in the same split, to the 18 held-out splits with a clear observed gap (an interval that excludes zero: 3 on ASSISTments, 5 on Eedi and 10 on OULAD). For the projected model, the error on the gap matters in 14 splits and is inconclusive in 4, and the increase of the error matters in all 18. For the per-group recalibration, the error on the gap is acceptable in 15 splits and inconclusive in 3, and the increase of the error is acceptable in all 18; the original model is acceptable on the gap in 14 splits and inconclusive in 4. Because the margins were chosen after the results, these decisions illustrate the rule; they are not an acceptance test fixed in advance. The 10\% tolerance in Section~\ref{sec:results-oulad} is likewise a sensitivity analysis done after the results.

	\begin{figure}[htbp]
		\centering
		\includegraphics[width=0.85\linewidth]{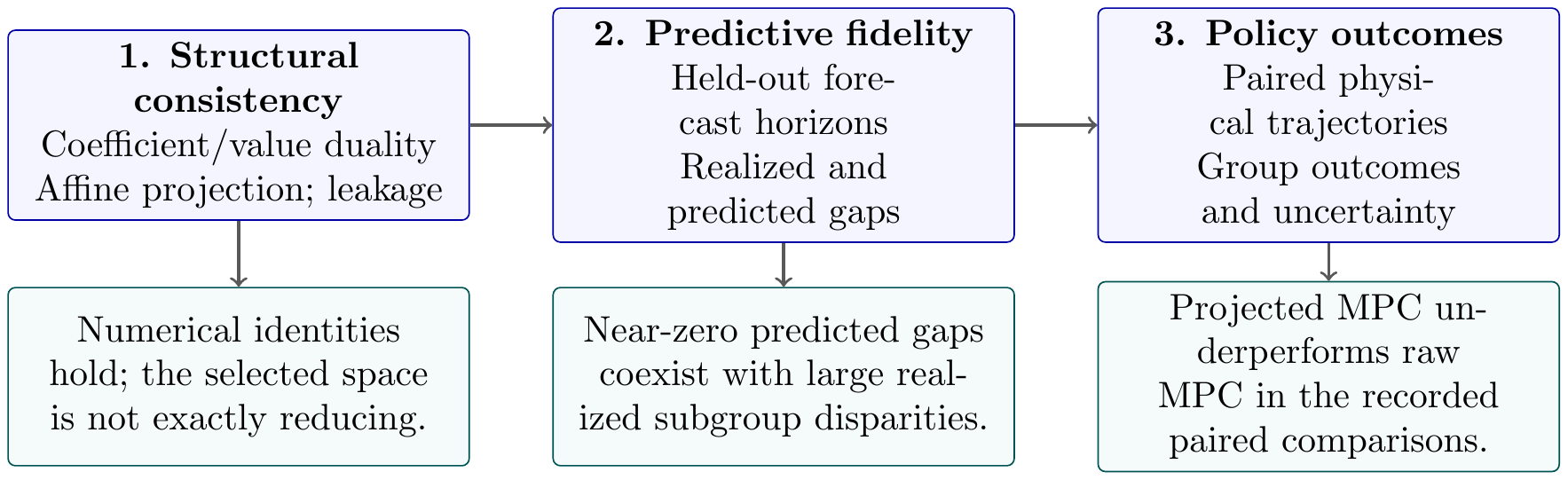}
		\caption{The three levels of the audit, and the evidence we report for each. Passing one level does not mean passing the next one: a model that is built correctly can still predict badly, and can still lead to a worse result for students in simulation.}
		\label{fig:validation-workflow}
	\end{figure}

	\section{The tutoring simulator and the controllers}
	\label{app:block-simulator}
	\subsection{Simulator and controllers in detail}
	\label{app:sim-details}
	\paragraph{Dictionary and training.} The dictionary has ten features:
	\begin{equation}
		\Psi_{\mathrm{raw}}(x)=(k,e,p,d,k^2,e^2,d^2,ke,de,kd)^\top.
	\end{equation}
	For training, we simulate 150 trajectories of 25 transitions with a random control \(u_t\sim\mathrm{Uniform}[-1,1]\) at each step. This gives 3{,}750 transitions, but only 150 truly independent trajectories. We standardize the features with the states at times \(0,\ldots,24\). The regression has 12 columns: ten features, one control and one constant. We fit the dynamics and the decoder by plain least squares, without a ridge. The projection uses rank 4 and a relative singular-value cutoff of \(10^{-8}\). These training trajectories, the scaling and the fitted model remain based on the 25-step window when the regime study changes \(T\); thus \(T=50\) is an evaluation extrapolation beyond the training window, not a refit on 50-step trajectories.

	\paragraph{Replicates.} The walk-through of Appendix~\ref{app:main-setting} uses ten replicates (\(0\) through \(9\)); the regime study and the controller comparisons use 50 (\(0\) through \(49\)). We generate independent random streams from root seed 20260914 using separate keys for replicate, role and cohort. Within each replicate, we compute paired differences before aggregating results, preserving the matched design in the uncertainty summaries.

	\paragraph{Scales of the gap penalties.} The two penalties act at different scales: if \(J_i\) denotes student \(i\)'s knowledge-and-control objective, dividing the cohort objective by \(n\) yields
	\[
	\frac1n\sum_i J_i+\frac{\beta_{\mathrm{gap}}}{n}\widehat\delta_{t+H\mid t}^{\,2}.
	\]
	Thus, numerically equal values of \(\beta_{\mathrm{ind}}\) and \(\beta_{\mathrm{gap}}\) do not imply equally strong penalties: one is applied to individual disparities, the other to a difference between group means. The reported grid should therefore be read as a sensitivity analysis of specified weights, not as a validation-tuned ranking of fairness methods.

	As in standard MPC, only the first planned action is applied before the next observation. Unlike the original and projected MPC, both gap-targeting controllers use group labels at decision time, and neither is tuned to match the original controller's control budget. We therefore report realized \(\mathcal U_1\) and budget sensitivity analyses. As a diagnostic contrast, \emph{predictive post-processing} changes only the final forecast under the original policy; because it does not alter actions, it cannot change student outcomes. For each controller, we record both one-step error and final-gap forecast error.

	\subsubsection{The floor-constrained controller: solver and proof}
	\label{app:floor-controller}
	We solve the second stage exactly through its dual, which has four multipliers: one for the gap, two for the floors and one for the budget. For given multipliers, each student solves the box-constrained quadratic program of Eq.~\eqref{eq:mpc} with one extra linear term and one extra term \(\nu|u_i|\). We solve it exactly by checking the \(5^3\) regions \(\{-1,\text{free}<0,0,\text{free}>0,1\}\) of the three controls. The implementation uses relative numerical margins of \(10^{-3}\) for the floors and the budget in the linear program. Proposition~\ref{prop:no-levelling} is about the exact set \(\mathcal F_c\), not about these numerical margins. Appendix~\ref{app:structural} reports the numerical checks of this controller: the floors are met to \(1.1\times10^{-9}\) and the budget exactly.

	\begin{proof}[Proof of Proposition~\ref{prop:no-levelling}]
		Let \(x_g=\sum_{i:\grp_i=g}(u_i-u^{\mathrm{ref}}_i)\). The floors and \(v_k>0\) give \(x_0\geq0\) and \(x_1\geq0\). (i) Since \(u^{\mathrm{ref}}_i\geq0\), \(\sum_iu_i\leq\sum_i|u_i|\leq c\sum_i|u^{\mathrm{ref}}_i|\leq\sum_iu^{\mathrm{ref}}_i\), so \(x_0+x_1\leq0\). Hence \(x_0=x_1=0\), and by Eq.~\eqref{eq:next-step} the gap changes by \(v_d(x_1/n_1-x_0/n_0)=0\). (ii) Take \(\widehat\delta^{\mathrm{ref}}<0\) (the other case is symmetric) and a student \(j\) of group 1 with \(u^{\mathrm{ref}}_j<1\). Raise only \(u_j\) by \(\varepsilon>0\) with \(\varepsilon\leq1-u^{\mathrm{ref}}_j\), \(\varepsilon\leq(c-1)\sum_iu^{\mathrm{ref}}_i\) and \(\varepsilon<2n_1|\widehat\delta^{\mathrm{ref}}|/v_d\). The control stays in the box; \(x_1=\varepsilon>0\) and \(x_0=0\), so both floors hold; the budget grows by \(\varepsilon\), which is allowed; and the gap moves by \(v_d\varepsilon/n_1\in(0,2|\widehat\delta^{\mathrm{ref}}|)\) towards and possibly past zero, so its size falls. (iii) \(\mathcal F_c\) is a polytope (write \(|u_i|\) with auxiliary variables), \(|\widehat\delta|\) is the maximum of two affine functions, and \(\mathcal F_c\) grows with \(c\).
	\end{proof}
	In the group-aware fits, \(v_k\in[0.13,0.18]\) and \(v_d\in[0.049,0.051]\); the simulator values are \(0.20\) and \(0.05\). The reference controls are not negative, except in 0.03\% of the steps at \(\gamma=2\); we do not apply the proposition to these few steps. Part (i) says that, at equal absolute-control budget, the predicted gap cannot change in one step while both floors on predicted group-mean knowledge hold. Part (ii) gives a strictly positive reduction for a positive increase of the budget, when its slack and nonzero-gap assumptions hold. It does not promise a large reduction for a very small increase. Neither part guarantees outcomes over many steps or protects each individual student. The script is \path{pareto_fair.py}.

	\subsection{The main simulator setting in detail}
	\label{app:main-setting}
	\subsubsection{One replicate}
	\label{sec:results-edmd}
	We begin with one paired simulator cohort to make the comparison concrete. In replicate 0, MPC using the projected model yields more final knowledge than no control, but this comparison does not isolate the effect of projection: the original-model MPC yields still more knowledge and a smaller disparity gap. The relevant contrast is therefore projected MPC against original MPC, with students and noise held fixed. Under that comparison, the projection changes the signed gap by \(-0.135080\) and mean disparity by \(-0.022040\); because the baseline gap is negative, the first change increases its magnitude.
	
	\begin{table}[t]
		\centering\small
		\caption{Paired EDMD--MPC results for replicate 0, with 150 students and 25 turns. All gaps are group 1 minus group 0. The two arms without control are identical by design.}
		\label{tab:edmd}
		\begin{tabular}{@{}lrrrr@{}}
			\toprule
			Policy & \(\delta\) & \(\overline D_T\) & \(\overline k_T\) & \(\overline k_{1,T}-\overline k_{0,T}\)\\
			\midrule
			Zero control, raw & \(-0.46854\) & \(-0.23221\) & \(0.12011\) & \(-0.23514\)\\
			Zero control, projected & \(-0.46854\) & \(-0.23221\) & \(0.12011\) & \(-0.23514\)\\
			MPC, projected & \(-0.45663\) & \(-0.11481\) & \(0.55660\) & \(-0.48414\)\\
			MPC, raw & \(-0.32155\) & \(-0.09277\) & \(0.67082\) & \(+0.19836\)\\
			\bottomrule
		\end{tabular}
	\end{table}
	
	The paired effect is consistent across the ten replicates. The mean change in signed gap is \(-0.136428\) (standard deviation \(0.003699\); 95\% Student-\(t\) interval \([-0.139074,-0.133781]\)), with replicate effects ranging from \(-0.142280\) to \(-0.130742\). Since the reference gap is negative, this negative change moves it farther from zero and therefore widens the disparity gap. Mean knowledge also declines: the average paired change is \(-0.159925\) (standard deviation \(0.038160\); 95\% interval \([-0.187223,-0.132626]\)).
	
\begin{figure}[t]
		\centering
		\resizebox{0.9\textwidth}{!}{\input{journal_edmd_results.tex}}
		\caption{Paired trajectories for replicate 0, from the same data as Table~\ref{tab:edmd}. Panel (a): group gap at each turn. Panel (b): average knowledge. The two arms without control are the same. Compared with the original MPC, the projected MPC ends with a larger gap and less knowledge. The curves show one paired run, not an uncertainty range.}
		\label{fig:edmd}
	\end{figure}
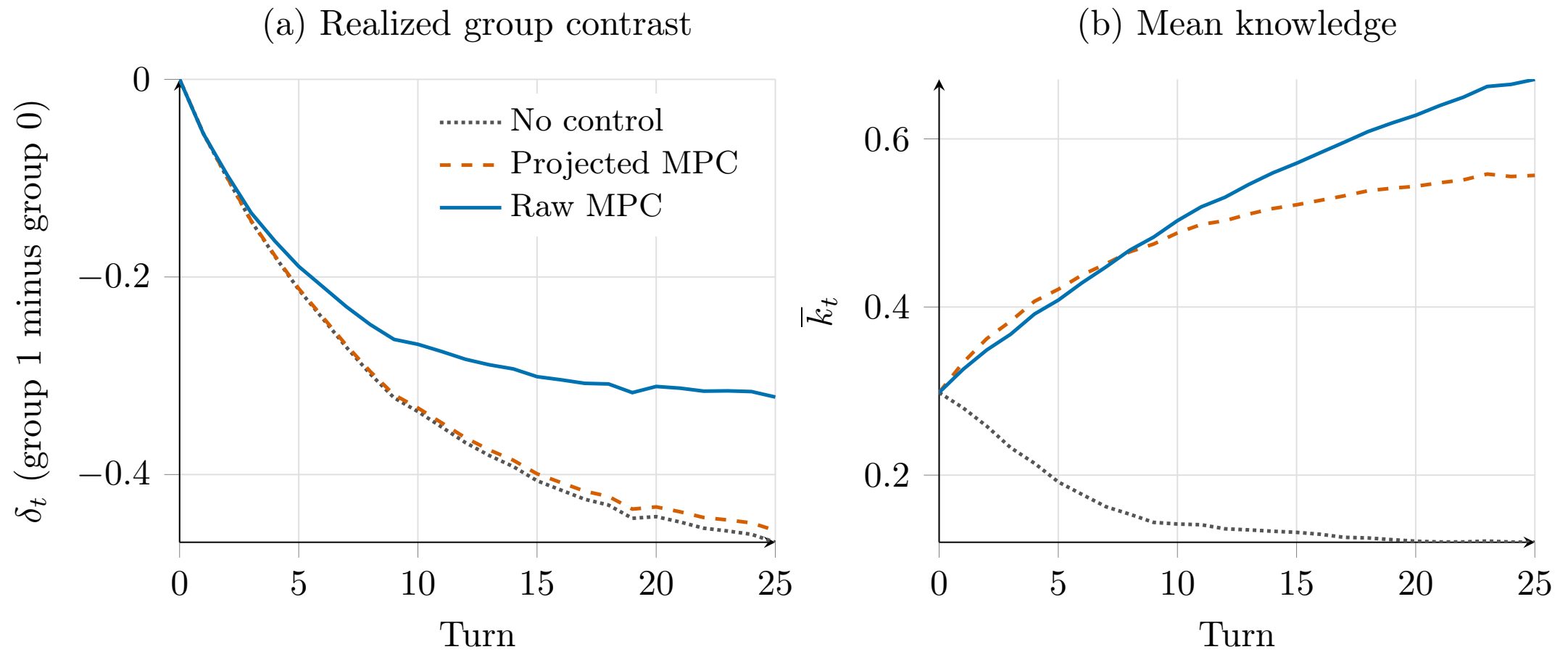

	Figure~\ref{fig:edmd} traces these outcomes over all 25 turns for the same cohort. The two no-control arms coincide by construction, whereas the projected-MPC trajectory ends with both a larger group gap and lower average knowledge than the original-MPC trajectory. The figure therefore shows not just a final difference, but how the paired outcomes diverge under the two controllers.
	
	\subsubsection{Across replicates, the effect is stable but uneven between groups}
	\label{sec:results-replicates}
\begin{figure}[t]
		\centering
		\resizebox{0.8\textwidth}{!}{\input{journal_paired_effects.tex}}
		\caption{Paired effects (projected MPC minus original MPC) in all ten replicates. Panel (a): effect on the gap, \(\tau_\delta\). Panel (b): effect on knowledge, \(\tau_k\). Orange points are single replicates; the blue diamond and bar give the mean and its 95\% Student-\(t\) interval.}
		\label{fig:paired-effects}
	\end{figure}
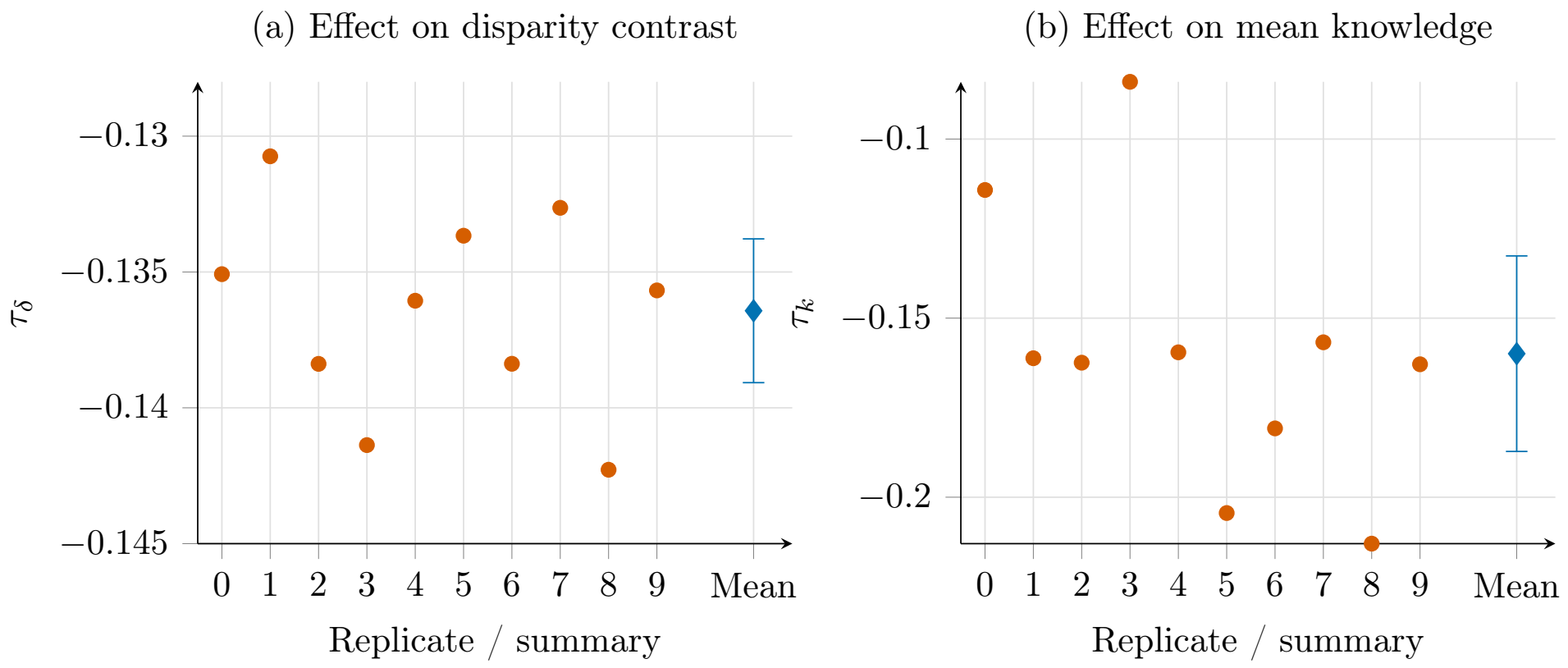

	Figure~\ref{fig:paired-effects} extends the comparison across all ten replicates. Both the signed-gap change \(\tau_\delta\) and the average-knowledge change \(\tau_k\) are negative in every replicate; the blue diamond and interval summarize their means and 95\% Student-\(t\) uncertainty. This consistency establishes the direction of the effect for the chosen simulator, seeds, rank and controller, but not its generality. We examine that question across alternative regimes in Section~\ref{sec:results-regimes}.
	
	The overall mean conceals an important asymmetry. As Table~\ref{tab:paired-summary} shows, relative to the original MPC, the projection increases group~0's knowledge by \(0.184695\) on average but decreases group~1's by \(0.504544\). With equally sized groups, these changes yield the overall loss of \(0.159925\). Reporting group-specific effects is therefore necessary: the aggregate alone would understate the loss experienced by group~1.
	
	\begin{table}[t]
		\centering\small
		\caption{Paired effects over ten replicates. The intervals are for the mean over replicates; they assume that replicate effects are close to independent and close to normal, and they are not corrected for multiple comparisons; a repetition with 50 replicates and a Holm correction is reported in Section~\ref{sec:results-regimes}.}
		\label{tab:paired-summary}
		\input{journal_paired_summary.tex}
	\end{table}
	
	\paragraph{The ridge value.} This result is not sensitive to the tested projection ridge. Varying \(\eta/\lambda_{\max}(\Gram)\) from \(10^{-6}\) to \(10^{-1}\) changes neither the selected ranks nor the direction of the effect: the gap widens in all 60 runs, with mean paired changes between \(-0.147\) and \(-0.136\) (Appendix~\ref{app:ridge}).
	
	\subsubsection{The construction is consistent, but not a contraction}
	\label{sec:results-consistency}
	These policy results follow a numerically consistent construction, but not an exact reduction. For replicate 0, the centered cross-block residuals are \(r_1=0.175484\) and \(r_2=0.172083\), while \(\|K\|_{\Gamma,2}=1.027607\); in the augmented space, the residuals are \(0.186097\) and \(0.172083\). Thus, the retained and removed subspaces remain coupled, and the fitted operator is not a contraction. The centered closure dimensions across depth are \(1,3,7,10,10\), so the full dictionary is reached and the centered complement has dimension zero. In the augmented affine space, retaining the constant coordinate leaves a complement of dimension one. By contrast, construction-level checks are near machine precision: the seed residual is \(8.38\times10^{-14}\), and the idempotency error is \(2.89\times10^{-13}\). The model therefore passes these implementation checks without satisfying exact invariance or contraction.

	\subsection{Solving and certifying the gap-targeting controllers}
	\label{app:solver}
	The final predicted disparity is affine in each student's three planned actions. Introducing a multiplier \(\lambda\) for the gap constraint makes the Lagrangian separable across students, with each subproblem a box-constrained, three-dimensional quadratic program plus the linear term \(\lambda s_i v^\top u_i\). Here \(s_i=1/n_1\) for group~1 and \(-1/n_0\) for group~0, and \(v\) gives the effect of the action sequence on final disparity. We solve each subproblem by enumerating its active sets and find the multiplier by 60 bisection steps over \([-10^4,10^4]\). For the penalty controller, the corresponding equation is \(\lambda=2\beta_{\mathrm{gap}}\widehat\delta(\lambda)\) on the same interval. If the equality-constrained root falls outside the interval, the implementation selects the endpoint with the smaller absolute predicted gap. This occurs only for \(\gamma=2\), in 7.2\% of steps, with residual at most \(0.031\); otherwise the equality is satisfied to numerical precision. A root outside the chosen interval indicates a solver limit, not necessarily physical infeasibility.
	
	To determine whether these residual cases are infeasible or merely beyond the multiplier search interval, stack all planned controls as \(u\in[-1,1]^{nH}\) and write \(\widehat\delta=b_\delta+w_\delta^\top u\). The exact attainable range over the control box is
	\begin{equation}
		[b_\delta-\|w_\delta\|_1,\ b_\delta+\|w_\delta\|_1],
		\label{eq:gap-box-range}
	\end{equation}
	because each control coordinate can independently attain either bound. The equality is feasible exactly when this interval contains zero. We performed this certification for every setting of the regime study, first with ten replicates, and then in the 50-replicate study of Section~\ref{sec:results-regimes}, where the certified solver is used throughout; the counts below are those of the ten-replicate run. At every step, we test the interval of Eq.~\eqref{eq:gap-box-range}; when the equality is feasible and the root is not bracketed, we widen the multiplier interval by factors of ten up to \(10^{12}\); and we record the primal residual \(|\widehat\delta_{t+H\mid t}|\). Every unresolved step is certified infeasible: such steps occur only at \(\gamma=2\), in 18 of 250 steps (7.2\%), where zero lies outside the attainable interval, so no control in the box can close the predicted gap. At every feasible step the equality holds to \(2\times10^{-16}\), no feasible step is left unresolved, and the certified solver changes no reported outcome (largest difference \(2\times10^{-16}\)). Table~\ref{tab:regimes-full} therefore reports certified infeasibility rates.

	\subsection{The controller aimed at the disparity gap}
	\label{app:disparity-controller}
	This appendix reports the controller aimed at the disparity gap, as in Proposition~\ref{prop:no-levelling}, over a range of budgets. Read it with Table~\ref{tab:knowledge-controller} in mind: its gains in disparity come with a wider knowledge gap.

	Can the disparity gap be reduced while maintaining both groups' mean knowledge? To answer this, we compare two strategies using the same cohorts, noise and group-aware model: a reference MPC with a lower control cost, \(\rho\in\{0.6,0.4,0.25,0.15,0.1\}\), which increases teaching effort, and the floor-constrained MPC from Section~\ref{sec:pareto-fair}, with budget factor \(c\in\{1,1.1,1.25,1.5,1.75,2\}\). All effects are paired against the reference setting \((\rho=1,\beta_{\mathrm{ind}}=8)\). Figure~\ref{fig:frontier} shows the trade-offs; Table~\ref{tab:frontier} in Appendix~\ref{app:regimes} reports their intervals.

	The comparison first reveals a binding role for the resource constraint. At \(c=1\), the linear program finds no allowable reduction at any step or replicate for any tested \(\gamma\), consistent with Proposition~\ref{prop:no-levelling}(i) whenever its assumptions hold; the controller therefore retains the reference plan. By contrast, at each tested \(c>1\), a reduction is feasible. Across all settings and replicates, the worst group's change in final simulated knowledge is no less than \(-7\times10^{-10}\). The policy achieves this by leaving group~0's actions unchanged and allocating additional control to group~1, which is behind in the synthetic disparity state. The largest knowledge gains for group~1 are \(0.28\), \(0.21\) and \(0.09\) at \(\gamma=0.5,1,2\), respectively.

	The resulting frontier also shows how the reduction depends on the budget and strategy. At \(\gamma=1\), floor-constrained MPC reduces the gap by \(0.157\) (\([0.154,0.160]\)) using \(1.61\) times the reference control. Increasing teaching alone reduces it by \(0.109\) at \(1.79\) times the budget and requires \(2.2\) times the reference control to reach a reduction of \(0.158\). At \(\gamma=0.5\), the corresponding floor-constrained reduction is \(0.122\) at \(1.61\) times the control, compared with \(0.054\) from teaching more at \(1.78\) times; teaching more requires \(2.8\) times the reference budget to reach \(0.120\). At \(\gamma=2\), the two strategies are closer at the reported budget points: \(0.104\) at \(1.36\) times for floor-constrained MPC, versus \(0.087\) at \(1.37\) times for teaching more. At still higher teaching effort, the latter widens the gap again. These are sampled trade-offs rather than exact equal-budget comparisons.

	The gap-constrained controller can reduce the gap further—in the main setting, by \(0.228\)—but does so by lowering one group's outcome. Simply increasing teaching produces larger gains in overall knowledge because group~0 also benefits. Floor-constrained MPC occupies a different point on this frontier: its first stage finds the largest one-step reduction allowed by the model and group-mean floors, and its second stage targets 95\% of that amount. Its protection remains conditional on the model and the group labels being available to the decision rule. In particular, using the pooled model employed by the other controllers widens the gap by \(0.029\), because it does not represent group-dependent drift; this is a predictive failure, not a failure of the optimization constraints. Moving the floors from the next step to the end of the three-step plan also failed in our experiments, as the controller repeatedly deferred the actions needed to satisfy them. Both variants are documented in \path{pareto_fair.py}.
	
	\paragraph{Equal-budget baselines.}
	The frontier above joins sampled budget points. To compare at exactly the same budget, we add two baselines that use the same group-aware model and reference plan. The \emph{targeted top-up} gives every student of the group behind the reference action plus a constant extra amount \(\kappa\), clipped to the box; the \emph{uniform increase} lowers the control cost \(\rho\) of the reference MPC for every student. In each replicate, \(\kappa\) or \(\rho\) is set by bisection so that the realized mean absolute control \(\mathcal U_1\) equals that of the floor-constrained MPC with \(c=2\); for the top-up, we also match the quadratic cost \(\mathcal U_2\). Table~\ref{tab:equal-budget} gives the results. At equal budget, the targeted top-up narrows the gap as much as the floor-constrained MPC: the difference is \(0.001\) in favour of the floor-constrained MPC at \(\gamma=1\), and \(0.004\) and \(0.007\) in favour of the top-up at \(\gamma=0.5\) and \(\gamma=2\); neither lowers the knowledge of group~0. The uniform increase narrows the gap less (by \(0.043\) to \(0.086\)) but raises the knowledge of both groups. In this simulator, therefore, the gain of the floor-constrained MPC comes from directing additional help to the group behind, not from a more efficient use of a fixed budget. Its value is that it derives this allocation from the model and checks the floors at every step; the simple rule happens to work here because every action has a positive effect for every student and students do not interact.

	\begin{table}[htbp]
		\centering\small
		\caption{Equal-budget baselines (group-aware model, rank 4, depth 3, \(T=25\), \(\beta_{\mathrm{ind}}=8\), ten replicates). Paired changes compared with the reference group-aware MPC, mean with 95\% Student-\(t\) interval: absolute final gap, and final mean knowledge of group 0 and group 1. Ratios: mean absolute (\(\mathcal U_1\)) and quadratic (\(\mathcal U_2\)) control relative to the reference. The extra control of each baseline is set, in each replicate, to match the budget of the floor-constrained MPC.}
		\label{tab:equal-budget}
		\resizebox{\linewidth}{!}{\input{tab_equal_budget.tex}}
	\end{table}

	\paragraph{Scenarios not used to design the method.}
	The simulator was designed together with the methods it tests. To check whether the conclusions depend on its particular form, we ran three scenarios that were not used in any earlier choice, with \(\gamma=1\), \(\beta_{\mathrm{ind}}=8\) and ten replicates (Table~\ref{tab:scenarios}): group~1 responds to help only half as much as group~0; learning is slower (knowledge decays with factor \(0.92\) instead of \(0.85\)) and the noise is doubled; and the effect of help on knowledge has diminishing returns, \(0.40\,u(1-k)\) instead of \(0.20\,u\). In the first and third scenarios, the assumption of Proposition~\ref{prop:no-levelling} that an action has the same constant effect in both groups does not hold. The three conclusions of this section remain. The projected MPC widens the gap in every scenario (by \(0.080\) to \(0.139\)); the gap-constrained MPC narrows it by levelling down group~0, whose knowledge falls by \(0.53\) to \(0.81\); and the floor-constrained MPC narrows it with no group losing (largest floor violation \(6\times10^{-10}\)). The equal-budget top-up again does about as well (within \(0.005\)). A fourth scenario, with a knowledge gap at the start but no group effect on the disparity variable, was uninformative for the disparity gap studied here, which stays near zero in every arm; we do not use it.

	\begin{figure}[htbp]
		\centering
		\includegraphics[width=\textwidth]{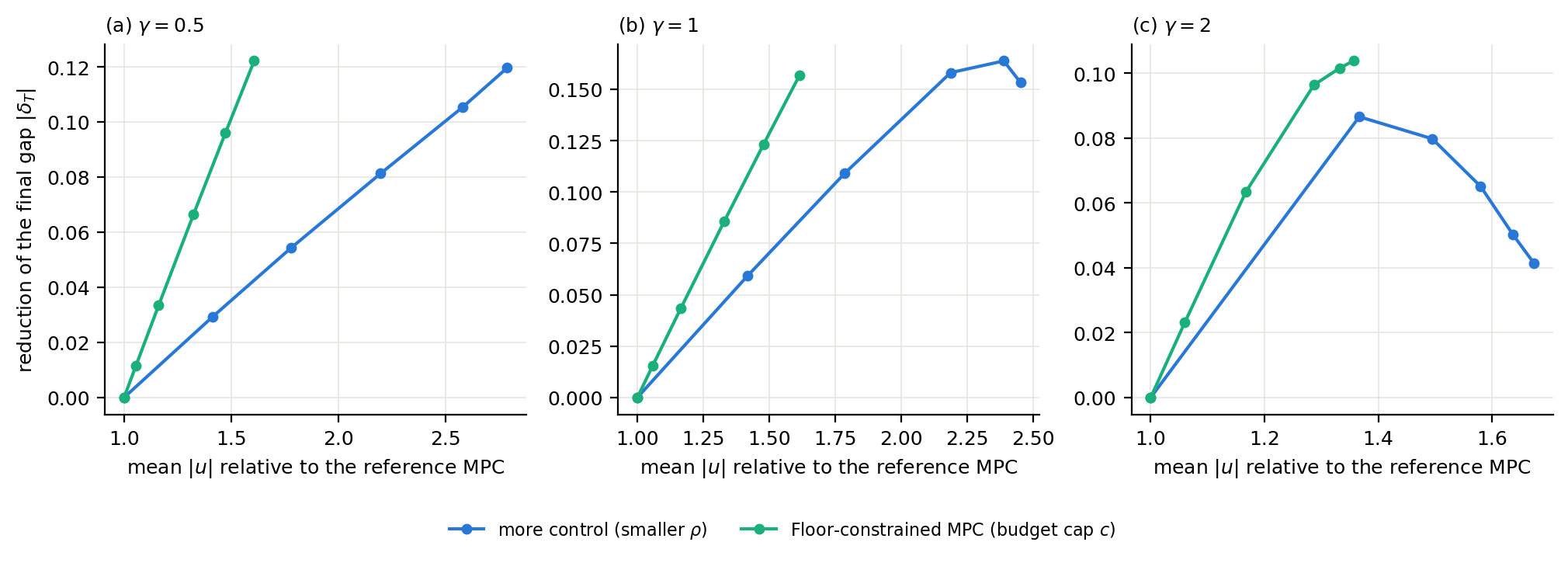}
		\caption{Narrowing the gap without levelling down (rank 4, depth 3, \(T=25\), \(\beta_{\mathrm{ind}}=8\), group-aware surrogate, mean over ten replicates). Reduction of the final gap against the mean control relative to the reference MPC. Blue: the reference MPC with a smaller control cost \(\rho\). Green: the floor-constrained MPC with budget factor \(c\); no group loses knowledge in any replicate. At \(c=1\), the reference plan is kept; the conditional one-step result is Proposition~\ref{prop:no-levelling}. The curves join sampled settings, not settings with exactly the same budget; the intervals of the effects are in Table~\ref{tab:frontier}.}
		\label{fig:frontier}
	\end{figure}
	
	\paragraph{Model error on the same grid.}
	The predictive cost of projection is present across this same grid. For each arm's trajectories, the projected model's one-step error is 5--29 times that of the original model; in the primary setting, the errors are \(0.0165\) and \(0.0021\), respectively. Thus, even settings with no detectable change in the final gap show a substantial loss of predictive accuracy. This distinction suggests that the policy-level effect depends on whether the controller uses the information removed by projection, whereas the predictive loss is more general across the tested settings.

	\subsection{The reference that uses the known equations, in detail}
	\label{app:oracle}
	\paragraph{Controllers.} The reference controller minimizes the objective of Eq.~\eqref{eq:mpc} over three-step plans, with the noise-free next state of Eq.~\eqref{eq:simulator} (clipping kept; this is not the expected next state), for each student separately: a grid of \(11^3\) plans in \([-1,1]^3\), then a finer grid of \(5^3\) plans of step \(0.05\) around the best one; only the first action is applied. The variant without the group uses the group's mean \(p_1=0.5\) in place of each student's group. The learned arms are the MPC of Section~\ref{sec:sim-protocol} on the original EDMD surrogate, on the group-aware surrogate, on the projected surrogate, and on the surrogate with a random rank-4 subspace removed (random directions in whitened coordinates, one draw per replicate). Every arm observes the full physical state at every step.

	\paragraph{Budget and validation.} For each replicate, group strength and objective, the budget is the mean absolute help of the reference controller with \(\rho=1\) on two validation cohorts (stream \texttt{calibration}). For every other arm, \(\log\rho\) is found by ten bisection steps in \([\log0.05,\log20]\) so that its mean absolute help on the same two cohorts equals this budget; the arm is then run once on the evaluation cohort, the same cohort as in the rest of the paper. On the evaluation cohort, the realized help of every arm is within 5\% of the budget (Table~\ref{tab:oracle}). The script is \path{oracle_reference.py}.

	\paragraph{Sensitivity of the reference.} On \(\gamma=1\), for both objectives and 10 replicates at \(\rho=1\) (script \path{oracle_sensitivity.py}), we compared the reference with a finer search (\(21^3\) plans, then a local \(9^3\) grid of step \(0.0125\)) and with planning on the expected cost over 16 draws of the simulator noise (clipping kept, so that the plan uses \(\E[\clip(a+\varepsilon)]\)). With the objective of the paper, final knowledge changes by \(-0.0002\) [\(-0.0005,+0.0001\)] with the finer search and by \(-0.0018\) [\(-0.0023,-0.0014\)] with the expected noise, and the help is unchanged (0.345 to 0.346). With the mean-knowledge objective, the expected noise changes final knowledge by \(-0.0056\) and the help from 0.300 to 0.298, but the finer search gives less help (0.275) and \(0.054\) less knowledge: this objective is nearly flat around its optimum, so the amount of help that the reference chooses depends on the resolution of the search. Because every learned arm is matched to the reference's budget, and at \(\gamma=1\) all arms reach the same knowledge at equal budget under this objective, the comparisons of Section~\ref{sec:results-oracle} do not depend on this choice.

	\paragraph{All settings.} Table~\ref{tab:oracle-diff} gives the paired differences in final mean knowledge, with 95\% Student-\(t\) intervals over the 20 replicates. Model error: true model minus learned model (with the group: group-aware surrogate; without: true model without the group minus original surrogate). Erasure: original surrogate minus projected surrogate. Random compression: original surrogate minus the surrogate with a random rank-4 removal. Group information: true model minus true model without the group. Positive values mean that the first controller leaves more knowledge.

	\begin{table}[htbp]
		\centering\scriptsize
		\caption{True-model reference, every setting: paired differences in final mean knowledge (mean and 95\% interval over 20 replicates), at equal budget and at \(\rho=1\).}
		\label{tab:oracle-diff}
		\resizebox{\linewidth}{!}{%
			\input{tab_oracle_diff.tex}}
	\end{table}

	\begin{table}[htbp]
		\centering\small
		\caption{Controllers aimed at the disparity gap or at the knowledge gap, with and without floors (group-aware model for every arm, budget factor \(c=2\) on the applied control, 50 replicates). Paired changes compared with the reference group-aware MPC: final mean knowledge of each group, signed and absolute knowledge gap, absolute disparity gap; last column: mean absolute and quadratic control relative to the reference. Every 95\% interval excludes zero except where the mean is \(0.000\); the intervals are in the reproduction output \texttt{knowledge\_controller.json}.}
		\label{tab:knowledge-controller}
		\resizebox{\linewidth}{!}{%
		\begin{tabular}{@{}lrrrrrr@{}}
			\toprule
			Arm & \(\Delta\overline k_{T,0}\) & \(\Delta\overline k_{T,1}\) & \(\Delta\delta^k_T\) & \(\Delta|\delta^k_T|\) & \(\Delta|\delta_T|\) & \(\mathcal U_1\) / \(\mathcal U_2\)\\
			\midrule
			\multicolumn{7}{@{}l}{\(\gamma=0.5\): reference \(\overline k_{T,0}=0.613\), \(\overline k_{T,1}=0.721\), knowledge gap \(+0.109\)}\\
			\quad Disparity target, floors & \(-0.000\) & \(+0.277\) & \(+0.277\) & \(+0.277\) & \(-0.121\) & 1.60 / 3.07\\
			\quad Disparity target, no floor & \(-0.392\) & \(+0.265\) & \(+0.657\) & \(+0.657\) & \(-0.154\) & 1.09 / 1.91\\
			\quad Knowledge target, floors & \(+0.107\) & \(+0.008\) & \(-0.100\) & \(-0.099\) & \(+0.022\) & 1.09 / 1.15\\
			\quad Knowledge target, no floor & \(+0.050\) & \(-0.050\) & \(-0.100\) & \(-0.099\) & \(+0.022\) & 1.00 / 0.97\\
			\quad Top-up, same \(\mathcal U_1\) & \(+0.099\) & \(+0.000\) & \(-0.099\) & \(-0.097\) & \(+0.019\) & 1.09 / 1.14\\
			\addlinespace
			\multicolumn{7}{@{}l}{\(\gamma=1\): reference \(\overline k_{T,0}=0.569\), \(\overline k_{T,1}=0.788\), knowledge gap \(+0.219\)}\\
			\quad Disparity target, floors & \(-0.000\) & \(+0.212\) & \(+0.212\) & \(+0.212\) & \(-0.155\) & 1.61 / 3.00\\
			\quad Disparity target, no floor & \(-0.551\) & \(+0.212\) & \(+0.763\) & \(+0.763\) & \(-0.308\) & 1.71 / 3.10\\
			\quad Knowledge target, floors & \(+0.192\) & \(+0.014\) & \(-0.177\) & \(-0.177\) & \(+0.040\) & 1.14 / 1.18\\
			\quad Knowledge target, no floor & \(+0.093\) & \(-0.090\) & \(-0.183\) & \(-0.183\) & \(+0.041\) & 1.00 / 0.94\\
			\quad Top-up, same \(\mathcal U_1\) & \(+0.180\) & \(+0.000\) & \(-0.180\) & \(-0.180\) & \(+0.035\) & 1.14 / 1.14\\
			\addlinespace
			\multicolumn{7}{@{}l}{\(\gamma=2\): reference \(\overline k_{T,0}=0.492\), \(\overline k_{T,1}=0.910\), knowledge gap \(+0.418\)}\\
			\quad Disparity target, floors & \(-0.000\) & \(+0.089\) & \(+0.089\) & \(+0.089\) & \(-0.103\) & 1.35 / 1.90\\
			\quad Disparity target, no floor & \(-0.492\) & \(+0.090\) & \(+0.582\) & \(+0.582\) & \(-0.543\) & 2.06 / 2.95\\
			\quad Knowledge target, floors & \(+0.173\) & \(+0.025\) & \(-0.148\) & \(-0.148\) & \(+0.033\) & 1.13 / 1.15\\
			\quad Knowledge target, no floor & \(+0.093\) & \(-0.079\) & \(-0.172\) & \(-0.172\) & \(+0.044\) & 1.02 / 0.96\\
			\quad Top-up, same \(\mathcal U_1\) & \(+0.237\) & \(+0.000\) & \(-0.237\) & \(-0.237\) & \(+0.046\) & 1.13 / 1.07\\
			\bottomrule
		\end{tabular}}
	\end{table}

	\subsection{The horizon \texorpdfstring{\(T=50\)}{T=50}}
	\label{app:t50}
\begin{figure}[htbp]
		\centering
		\includegraphics[width=\textwidth]{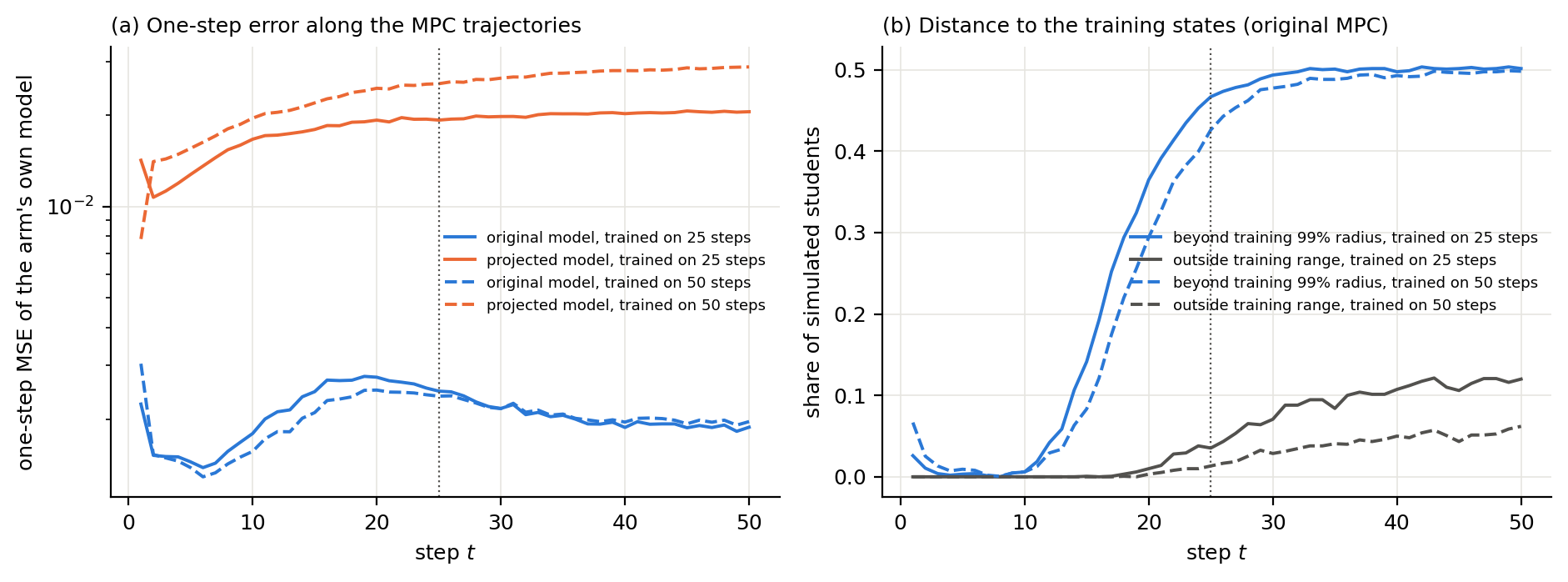}
		\caption{Evaluation at \(T=50\) (\(\gamma=1\), \(\beta_{\mathrm{ind}}=8\), mean over ten replicates). Solid lines: model trained on 25-step trajectories, as in the main study; dashed lines: trained on 50-step trajectories. (a) One-step error of each arm's own model along its MPC trajectory. (b) Share of simulated students beyond the 99th percentile of the training Mahalanobis distance, and outside the training range. The dotted line marks the end of the training window.}
		\label{fig:horizon50}
	\end{figure}

	We checked in the code that a longer evaluation horizon changes only the evaluation cohort: the model, the contrast, the metric and the projector are fitted once, on the 25-step exploratory trajectories, and are not estimated again. So \(T=50\) is an extrapolation in time. Figure~\ref{fig:horizon50} shows what happens after step 25. The one-step error of the original model along its own MPC trajectories does not grow (\(0.0021\) on steps 1--25, \(0.0020\) on steps 26--50), while that of the projected model grows from \(0.0166\) to \(0.0202\). The states do move away from the training data: on steps 26--50, half of the simulated students (50\%) are further from the training states than 99\% of the training states themselves (Mahalanobis distance in the lifted space), and up to 12\% have a coordinate outside the training range. As a sensitivity check, we trained the model on 50-step exploratory trajectories. The projection still widens the gap at \(T=50\), by \(0.174\) (\([0.170,0.178]\)) instead of \(0.155\) (\([0.151,0.159]\)), and still lowers average knowledge (\(-0.088\) instead of \(-0.204\)). The \(T=50\) result therefore does not depend on the extrapolation.
	
	\FloatBarrier

	\subsection{All settings of the regime study}
	\label{app:regimes}
	\begin{figure}[htbp]
		\centering
		\includegraphics[width=\textwidth]{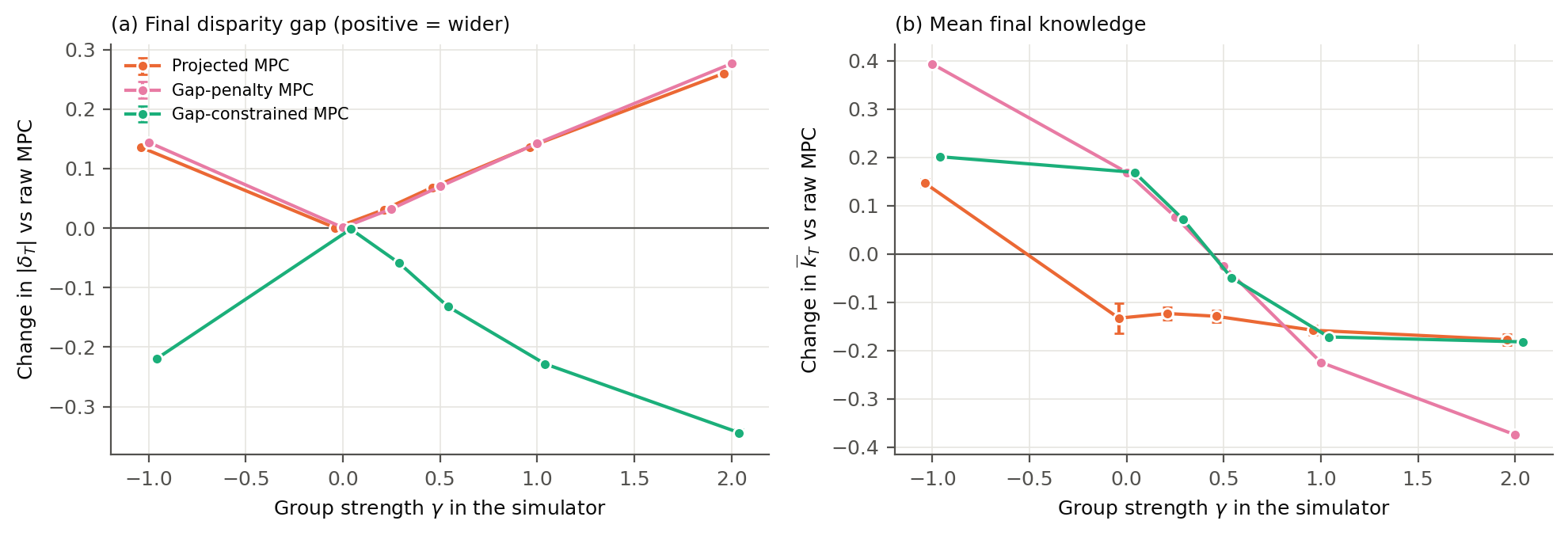}
		\caption{Effect of the group strength \(\gamma\) (rank 4, depth 3, \(T=25\), \(\beta_{\mathrm{ind}}=8\)). Mean paired change compared with the original MPC over 50 replicates, with 95\% Student-\(t\) intervals: (a) size of the final gap; (b) average final knowledge.}
		\label{fig:regimes}
	\end{figure}
\begin{figure}[htbp]
		\centering
		\includegraphics[width=\linewidth]{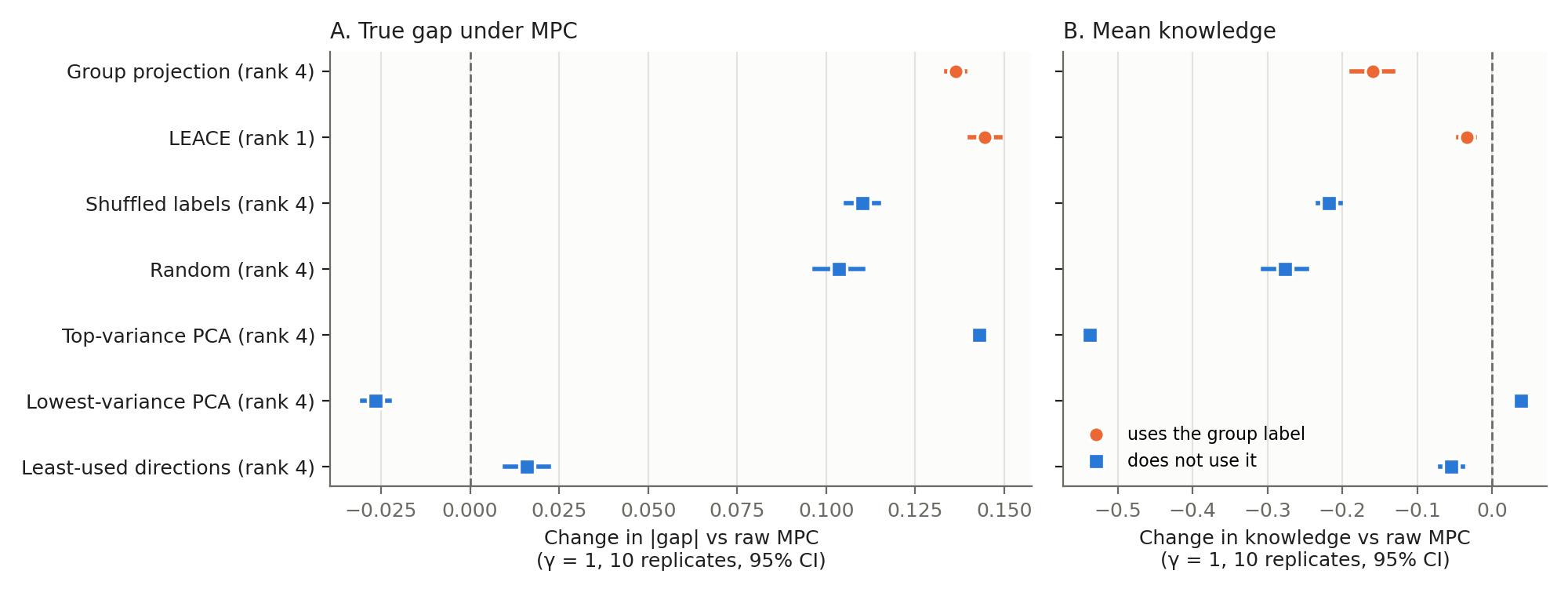}
		\caption{Rank-matched controls in the simulator (\(\gamma=1\), 10 replicates, 95\% intervals): change of the absolute disparity gap (A) and of mean knowledge (B) under MPC with each compressed model, compared with the MPC on the raw model.}
		\label{fig:rank-matched-policy}
	\end{figure}

	\paragraph{Rank-matched controls in the simulator.} Figure~\ref{fig:rank-matched-policy} shows the MPC outcomes for group strength \(\gamma=1\) (10 replicates; the raw controller ends with a gap of \(0.317\) and a mean knowledge of \(0.675\)). The picture is more mixed than for prediction, and we report it as it is.
\begin{itemize}[leftmargin=*,itemsep=1pt]
	\item \emph{The wider gap comes from the group direction.} LEACE alone (one direction) widens the gap by \(0.145\) \([0.140,0.149]\), as much as the rank-4 group projection (\(0.136\)); one random direction widens it by only \(0.025\).
	\item \emph{The loss of knowledge comes mostly from compression.} Random rank-4 directions lose more knowledge (\(-0.277\)) than the group projection (\(-0.160\)); LEACE loses only \(-0.034\). With equal groups (\(\gamma=0\)), the group projection still loses \(0.17\) of knowledge, close to random directions (\(0.14\)).
	\item \emph{Compression does not have to hurt.} Lowest-variance PCA slightly narrows the gap (\(-0.026\)) and raises knowledge (\(+0.038\)).
\end{itemize}

A likely reason for the first point is that the raw controller sees the predicted disparity of each group and gives more help to the group behind, while a model without the group direction cannot see this difference; we have not tested this explanation directly. At \(\gamma=2\) the same pattern holds, except that random rank-4 directions widen the gap as much as the group projection.

	\paragraph{Budget and penalty weight.} To see whether the budget explains the result, we raised the control cost \(\rho\) from 1 to 5 (Table~\ref{tab:budget}). At \(\gamma=0.5\), the constrained controller already uses the same budget as the original MPC (\(|u|=0.233\) against \(0.236\)) and still reduces the gap by \(0.132\). A larger \(\rho\) lowers its budget further (to \(0.204\)), and the gap still falls by \(0.130\), but knowledge falls more. At \(\gamma=1\) and \(\gamma=2\), the constraint is binding: raising \(\rho\) hardly changes the control it needs (\(|u|\) about \(0.39\) against \(0.29\), and \(0.69\) against \(0.39\)). So when the group effect is strong, closing the predicted gap needs more control than the original MPC uses. At equal budget, we can confirm the result only for the weaker group effect. In the regime study, \(\beta_{\mathrm{gap}}\) had the same value as \(\beta_{\mathrm{ind}}\). To see whether a larger penalty works, we ran a grid \(\beta_{\mathrm{gap}}\in\{1,8,64,512,4096,32768\}\) for \(\gamma\in\{0.5,1,2\}\) (Table~\ref{tab:beta-gap} in Appendix~\ref{app:regimes}). Up to \(\beta_{\mathrm{gap}}=512\), the penalty widens the gap, as in the regime study. From \(\beta_{\mathrm{gap}}=4096\) on, it narrows the gap (by \(0.068\), \(0.137\) and \(0.218\) for \(\gamma=0.5\), 1 and 2). At \(32768\) it comes close to the constraint (\(-0.121\), \(-0.212\), \(-0.321\), against \(-0.132\), \(-0.228\), \(-0.344\)). The weight must be so large because the predicted gap is small. The constrained controller needs multipliers with a median \(|\lambda|\) of 789, 1444 and 2353, and a penalty acts like the constraint only when its own multiplier \(2\beta_{\mathrm{gap}}\widehat\delta_{t+H\mid t}\) reaches this size. So the gap penalty does not fail in principle; it fails at the weights of the regime study. The grid also allows a comparison at equal budget. At \(\gamma=0.5\), every penalty uses a little less control than the original MPC (\(|u|=0.229\) against \(0.236\)). At \(\gamma=1\), the penalty with \(\beta_{\mathrm{gap}}=4096\) uses the same budget (\(0.291\) against \(0.287\)). Both still narrow the gap. At \(\gamma=2\), narrowing the gap needs more control (\(0.522\) against \(0.390\)). In every case where the gap narrows, the knowledge of group~0 falls, by \(0.20\) to \(0.56\), as with the constraint.

	Table~\ref{tab:regimes-full} lists the 28 different settings of the regime study (Section~\ref{sec:results-regimes}), with the numerical 95\% Student-\(t\) intervals over the 50 replicates for the projected MPC and the gap-constrained MPC of Eq.~\eqref{eq:gap-constrained}, the mean \(|u|\) of the original and constrained controllers, and the share of steps where the multiplier search leaves the equality unresolved. These rates do not certify box infeasibility; Eq.~\eqref{eq:gap-box-range} gives the separate test that is needed. Table~\ref{tab:budget} gives the budget check. We report \(\rho\le5\). In the first, ten-replicate version of this check, the multiplier at \(\rho=8\) exceeded the bisection interval of the uncertified solver; the certified solver of Appendix~\ref{app:solver}, used for the reported values, was not run at \(\rho=8\).

	Table~\ref{tab:groupwise} gives the paired effects on each group for the projected, gap-penalty and gap-constrained MPC in the settings of Table~\ref{tab:regimes}, and Table~\ref{tab:beta-gap} gives the log-scale grid of the gap-penalty weight, with the control budget and the multipliers of the constrained controller.

	\begin{table}[htbp]
		\centering\footnotesize
		\caption{Budget check of the gap-constrained MPC (rank 4, depth 3, \(T=25\), \(\beta_{\mathrm{ind}}=8\) for the original MPC). A larger control cost \(\rho\) in Eq.~\eqref{eq:gap-constrained} lowers the control budget only when the constraint is not binding. Changes are paired with the original MPC, mean over 50 replicates with 95\% interval.}
		\label{tab:budget}
		\input{tab_budget.tex}
	\end{table}

	\begin{table}[htbp]
		\centering\small
		\caption{Log-scale grid of the gap-penalty weight (rank 4, depth 3, \(T=25\), \(\rho=1\)); ``constraint'' is the gap-constrained MPC. Paired change compared with the original MPC (\(\beta_{\mathrm{ind}}=8\)), mean over 50 replicates with 95\% interval. \(|u|\): mean absolute control of the original MPC and of the arm. Median \(|\lambda|\): median over steps of the multiplier that links the students (for the penalty, \(\lambda=2\beta_{\mathrm{gap}}\widehat\delta_{t+H\mid t}\)). Infeas.: share of steps where the constraint cannot be met.}
		\label{tab:beta-gap}
		\resizebox{\linewidth}{!}{%
			\input{tab_beta_gap.tex}}
	\end{table}

	\begin{table}[htbp]
		\centering\small
		\caption{Scenarios not used to design the method (\(\gamma=1\), \(\beta_{\mathrm{ind}}=8\), ten replicates). Paired changes, mean with 95\% Student-\(t\) interval: absolute final gap and final mean knowledge of each group. The projected and gap-constrained MPC are compared with the original MPC; the floor-constrained MPC (\(c=2\)) and the equal-budget top-up with the reference group-aware MPC.}
		\label{tab:scenarios}
		\resizebox{\linewidth}{!}{\input{tab_scenarios.tex}}
	\end{table}
	
	\begin{table}[htbp]
		\centering\small
		\caption{Budget frontier of Section~\ref{sec:results-pareto} (group-aware surrogate, rank 4, depth 3, \(T=25\), \(\beta_{\mathrm{ind}}=8\)). Paired change compared with the reference MPC (\(\rho=1\)), mean over ten replicates with 95\% interval. \(|u|\) ratio: mean absolute control relative to the reference. \(\Delta\delta^k_T\): change of the signed knowledge gap, computed from the unrounded group means; the frontier run stores group means only, so this column has no interval (Table~\ref{tab:knowledge-controller} gives intervals at \(c=2\)). Worst group (min): smallest change of group-mean knowledge over the ten replicates. The floor-constrained MPC with \(c=1\) keeps the reference plan at every step.}
		\label{tab:frontier}
		\resizebox{\linewidth}{!}{%
			\input{tab_frontier.tex}}
	\end{table}

	\begin{table}[htbp]
		\centering\small
		\caption{The 28 settings of the regime study with 50 replicates. Paired changes compared with the original MPC, mean with 95\% Student-\(t\) interval, and Holm-adjusted \(p\)-values over the 28 settings: absolute final gap for the projected MPC and for the gap-constrained MPC (certified solver), and final mean knowledge of the group whose knowledge falls more under the gap-constrained MPC.}
		\label{tab:replicates50}
		\resizebox{\linewidth}{!}{\input{tab_replicates50.tex}}
	\end{table}

	\begin{sidewaystable}
		\centering\small
		\caption{All 28 settings of the regime study. Paired changes compared with the original MPC, mean over 50 replicates with 95\% Student-\(t\) interval: size of the final gap \(|\delta_T|\) (positive: larger gap) and average knowledge \(\overline k_T\). \(|u|\): mean absolute control. Infeas.: share of steps at which no control in the box can make \(\widehat\delta_{t+H\mid t}=0\), certified with Eq.~\eqref{eq:gap-box-range}.}
		\label{tab:regimes-full}
		\resizebox{\textheight}{!}{%
			\input{tab_regimes_full.tex}}
	\end{sidewaystable}

	\begin{sidewaystable}
		\centering\small
		\caption{Effects on each group. Paired change compared with the original MPC (rank 4, depth 3, \(T=25\) unless stated), mean over 50 replicates with 95\% Student-\(t\) interval: size of the final gap, final simulated knowledge \(\overline k_{T,g}\) and final disparity \(\overline d_{T,g}\) of group \(g\). The gap-penalty MPC uses \(\beta_{\mathrm{gap}}=\beta_{\mathrm{ind}}\). The column \(\Delta\delta^k_T\) is the paired change of the signed knowledge gap, with its interval.}
		\label{tab:groupwise}
		\resizebox{\textheight}{!}{%
			\input{tab_groupwise.tex}}
	\end{sidewaystable}

	\FloatBarrier

	\begin{table}[htbp]
		\centering\small
		\caption{Regime study, selected settings (rank 4, depth 3, \(T=25\) unless stated). Each entry is the mean paired change compared with the original MPC, over 50 replicates, in mean final knowledge \(\overline k_T\), in the size of the knowledge gap \(|\delta^k_T|\) and in the size of the disparity gap \(|\delta_T|\) (positive: larger gap). Raw MPC: knowledge gap of the original MPC. An asterisk means a Holm-adjusted \(p<0.05\) over the 28 settings; Tables~\ref{tab:replicates50} and~\ref{tab:regimes-full} in Appendix~\ref{app:regimes} give the intervals.}
		\label{tab:regimes}
		\resizebox{\linewidth}{!}{%
			\begin{tabular}{@{}lrrrrrrrrrr@{}}
				\toprule
				& Raw MPC & \multicolumn{3}{c}{Projected MPC} & \multicolumn{3}{c}{Gap-penalty MPC} & \multicolumn{3}{c}{Gap-constrained MPC}\\
				\cmidrule(lr){3-5}\cmidrule(lr){6-8}\cmidrule(l){9-11}
				Configuration & \(\delta^k_T\) & \(\Delta\overline k_T\) & \(\Delta|\delta^k_T|\) & \(\Delta|\delta_T|\) & \(\Delta\overline k_T\) & \(\Delta|\delta^k_T|\) & \(\Delta|\delta_T|\) & \(\Delta\overline k_T\) & \(\Delta|\delta^k_T|\) & \(\Delta|\delta_T|\)\\
				\midrule
				\(\gamma=0\), \(\beta_{\mathrm{ind}}=8\) & \(-0.001\) & \(-0.133^*\) & \(0.000\) & \(+0.0005^*\) & \(+0.169^*\) & \(-0.002\) & \(+0.0007^*\) & \(+0.169^*\) & \(+0.008^*\) & \(-0.002^*\)\\
				\(\gamma=0.5\), \(\beta_{\mathrm{ind}}=8\) & \(+0.088\) & \(-0.129^*\) & \(+0.172^*\) & \(+0.068^*\) & \(-0.025^*\) & \(+0.171^*\) & \(+0.070^*\) & \(-0.049^*\) & \(+0.613^*\) & \(-0.132^*\)\\
				\(\gamma=1\), \(\beta_{\mathrm{ind}}=8\) (main setting) & \(+0.192\) & \(-0.158^*\) & \(+0.304^*\) & \(+0.136^*\) & \(-0.224^*\) & \(+0.331^*\) & \(+0.142^*\) & \(-0.172^*\) & \(+0.785^*\) & \(-0.228^*\)\\
				\(\gamma=2\), \(\beta_{\mathrm{ind}}=8\) & \(+0.343\) & \(-0.177^*\) & \(+0.524^*\) & \(+0.260^*\) & \(-0.374^*\) & \(+0.256^*\) & \(+0.276^*\) & \(-0.182^*\) & \(+0.642^*\) & \(-0.344^*\)\\
				\(\gamma=-1\), \(\beta_{\mathrm{ind}}=8\) & \(-0.181\) & \(+0.147^*\) & \(+0.271^*\) & \(+0.136^*\) & \(+0.393^*\) & \(-0.007\) & \(+0.144^*\) & \(+0.201^*\) & \(+0.380^*\) & \(-0.220^*\)\\
				\(\gamma=1\), \(\beta_{\mathrm{ind}}=0\) & \(-0.530\) & \(-0.079^*\) & \(-0.014^*\) & \(0.000\) & \(0.000\) & \(0.000\) & \(0.000\) & \(+0.053^*\) & \(+0.446^*\) & \(-0.372^*\)\\
				\(\gamma=1\), \(\beta_{\mathrm{ind}}=32\) & \(+0.583\) & \(+0.011^*\) & \(-0.164^*\) & \(+0.287^*\) & \(-0.257^*\) & \(-0.084^*\) & \(+0.284^*\) & \(-0.204^*\) & \(+0.394^*\) & \(-0.081^*\)\\
				\(\gamma=1\), \(\beta_{\mathrm{ind}}=8\), \(T=50\) & \(+0.286\) & \(-0.202^*\) & \(+0.242^*\) & \(+0.155^*\) & \(-0.264^*\) & \(+0.290^*\) & \(+0.164^*\) & \(-0.227^*\) & \(+0.693^*\) & \(-0.235^*\)\\
				\bottomrule
		\end{tabular}}
	\end{table}

	\subsection{Sensitivity to the ridge value}
	\label{app:ridge}
	
	\label{sec:results-ridge}
	We try six values of the relative ridge \(\eta/\lambda_{\max}(\Gram)\), from \(10^{-6}\) to \(10^{-1}\), one power of ten apart. Inside each replicate, all settings share the same training trajectories, scaling, dynamics, decoder, cohort and noise. Only the metric of the projection, the chosen subspace and the projected controller change.
	
	In all sixty runs, the applied rank is 4 and the candidate rank is 7. The mean gap with the original model is \(-0.316687\). With the projected model, the mean gap goes from \(-0.463989\) to \(-0.452999\), and the mean paired effect from \(-0.147302\) to \(-0.136312\). Both gaps stay negative, and the size of the gap grows in all sixty runs.
	
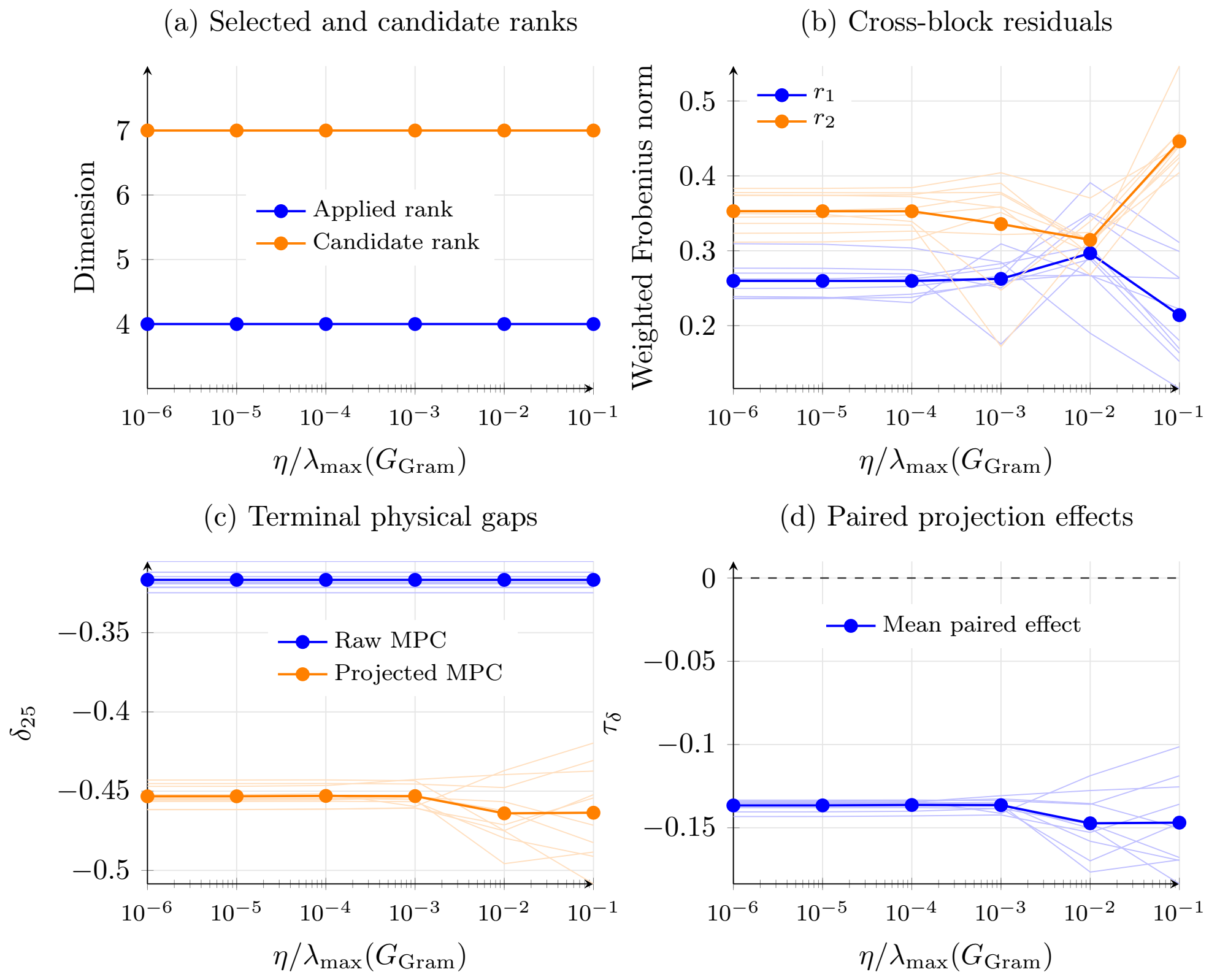
\begin{figure}[htbp]
		\centering
		\resizebox{0.82\textwidth}{!}{\begin{tikzpicture}[font=\small]
				\begin{groupplot}[group style={group size=2 by 2,horizontal sep=1.7cm,vertical sep=2.1cm},
					width=7.0cm,height=5.5cm,xmode=log,xmin=1e-6,xmax=1e-1,
					xtick={1e-6,1e-5,1e-4,1e-3,1e-2,1e-1},xticklabel style={font=\scriptsize},
					scaled y ticks=false,yticklabel style={/pgf/number format/fixed},
					xlabel={\(\eta/\lambda_{\max}(\Gram)\)},axis lines=left,grid=major,grid style={black!10},
					legend style={draw=none,font=\scriptsize},legend cell align=left]
					\nextgroupplot[title={(a) Selected and candidate ranks},ylabel={Dimension},ymin=3,ymax=8,ytick={4,5,6,7},legend style={at={(0.5,0.5)},anchor=center}]
					\addplot[blue,thick,mark=*] coordinates {(1e-06,4) (1e-05,4) (0.0001,4) (0.001,4) (0.01,4) (0.1,4)};
					\addlegendentry{Applied rank}
					\addplot[orange,thick,mark=*] coordinates {(1e-06,7) (1e-05,7) (0.0001,7) (0.001,7) (0.01,7) (0.1,7)};
					\addlegendentry{Candidate rank}
					\nextgroupplot[title={(b) Cross-block residuals},ylabel={Weighted Frobenius norm},legend pos=north west]
					\addplot[blue!25,thin,forget plot] coordinates {(1e-06,0.2701307032774322) (1e-05,0.2700690509181561) (0.0001,0.2690988731804883) (0.001,0.1754841432284407) (0.01,0.291964702036146) (0.1,0.1800935490048435)};
					\addplot[blue!25,thin,forget plot] coordinates {(1e-06,0.2590871227132236) (1e-05,0.2593416517575978) (0.0001,0.2622057201104814) (0.001,0.276956373055361) (0.01,0.3500262501159748) (0.1,0.2990738062346901)};
					\addplot[blue!25,thin,forget plot] coordinates {(1e-06,0.2767662253630676) (1e-05,0.276540634872421) (0.0001,0.2745672022819423) (0.001,0.2502785392504862) (0.01,0.3477407070896141) (0.1,0.2638703441111711)};
					\addplot[blue!25,thin,forget plot] coordinates {(1e-06,0.2389006966269516) (1e-05,0.2380719628666321) (0.0001,0.2306839519917452) (0.001,0.30922720565451) (0.01,0.2664629718531653) (0.1,0.2631349913322277)};
					\addplot[blue!25,thin,forget plot] coordinates {(1e-06,0.3094002564289488) (1e-05,0.3087113301400202) (0.0001,0.3037230985283109) (0.001,0.2848054266708513) (0.01,0.1898924219310026) (0.1,0.1160059156185061)};
					\addplot[blue!25,thin,forget plot] coordinates {(1e-06,0.2592697049636002) (1e-05,0.2592249225931562) (0.0001,0.2592970133699198) (0.001,0.2631444983857734) (0.01,0.3909494864399429) (0.1,0.3109268915255027)};
					\addplot[blue!25,thin,forget plot] coordinates {(1e-06,0.2622521511307675) (1e-05,0.2624848783644173) (0.0001,0.2654799077807344) (0.001,0.2828221333543486) (0.01,0.3059062361888818) (0.1,0.169083565086161)};
					\addplot[blue!25,thin,forget plot] coordinates {(1e-06,0.2496897579856128) (1e-05,0.2499005144896073) (0.0001,0.2523862756115621) (0.001,0.2664048891977882) (0.01,0.2670339523846858) (0.1,0.2221757225322807)};
					\addplot[blue!25,thin,forget plot] coordinates {(1e-06,0.2363693792507959) (1e-05,0.2368858936711142) (0.0001,0.2420521769807293) (0.001,0.2561463452366428) (0.01,0.2887177945944183) (0.1,0.1633808912207464)};
					\addplot[blue!25,thin,forget plot] coordinates {(1e-06,0.2361871897345009) (1e-05,0.236259044802714) (0.0001,0.2378418656364231) (0.001,0.2596877851494033) (0.01,0.2684025055159751) (0.1,0.1521966156341807)};
					\addplot[blue,thick,mark=*] coordinates {(1e-06,0.2598053187474901) (1e-05,0.2597489884475836) (0.0001,0.2597336085472337) (0.001,0.2624957339183606) (0.01,0.2967097028149807) (0.1,0.213994229230031)};
					\addlegendentry{\(r_1\)}
					\addplot[orange!25,thin,forget plot] coordinates {(1e-06,0.3365399405467049) (1e-05,0.336551824962214) (0.0001,0.3341089989951666) (0.001,0.1720825793415364) (0.01,0.3161724302896348) (0.1,0.4416354954482504)};
					\addplot[orange!25,thin,forget plot] coordinates {(1e-06,0.3832639467983542) (1e-05,0.3832998523891423) (0.0001,0.3842816415593711) (0.001,0.4042003342498747) (0.01,0.3704861982331197) (0.1,0.4397716820373407)};
					\addplot[orange!25,thin,forget plot] coordinates {(1e-06,0.373843214565027) (1e-05,0.373866238887969) (0.0001,0.3748408310517065) (0.001,0.3903154044314841) (0.01,0.2944385521250242) (0.1,0.43989312117298)};
					\addplot[orange!25,thin,forget plot] coordinates {(1e-06,0.3497284263585894) (1e-05,0.3486054856196679) (0.0001,0.3391261776894897) (0.001,0.2472641523268705) (0.01,0.3391590034200058) (0.1,0.4570355294926035)};
					\addplot[orange!25,thin,forget plot] coordinates {(1e-06,0.3742841280081346) (1e-05,0.3740619077488054) (0.0001,0.3723355891483536) (0.001,0.3576129698194548) (0.01,0.2676354544356326) (0.1,0.4187396027209501)};
					\addplot[orange!25,thin,forget plot] coordinates {(1e-06,0.3777100975700602) (1e-05,0.3775944132704608) (0.0001,0.3773384537744083) (0.001,0.3778935426879381) (0.01,0.3130962011586758) (0.1,0.4244498554501525)};
					\addplot[orange!25,thin,forget plot] coordinates {(1e-06,0.3450045083341156) (1e-05,0.3452494362612445) (0.0001,0.3478834196766857) (0.001,0.3587521608304551) (0.01,0.3141472476653182) (0.1,0.4044597893023)};
					\addplot[orange!25,thin,forget plot] coordinates {(1e-06,0.3115476317654631) (1e-05,0.3117876169605832) (0.0001,0.3145621740383164) (0.001,0.3512033104064943) (0.01,0.2968441466704798) (0.1,0.5467330479741959)};
					\addplot[orange!25,thin,forget plot] coordinates {(1e-06,0.3233614465983299) (1e-05,0.323654177652163) (0.0001,0.3262277682349807) (0.001,0.3216450018688893) (0.01,0.3243883858269308) (0.1,0.4586819599290141)};
					\addplot[orange!25,thin,forget plot] coordinates {(1e-06,0.3535040638600672) (1e-05,0.3538319319822683) (0.0001,0.3569859298901577) (0.001,0.375839405599486) (0.01,0.3092280442980396) (0.1,0.4292898283065864)};
					\addplot[orange,thick,mark=*] coordinates {(1e-06,0.3528787404404846) (1e-05,0.3528502885734518) (0.0001,0.3527690984058637) (0.001,0.3356808861562483) (0.01,0.3145595664122861) (0.1,0.4460689911834373)};
					\addlegendentry{\(r_2\)}
					\nextgroupplot[title={(c) Terminal physical gaps},ylabel={\(\delta_{25}\)},legend style={at={(0.55,0.7)},anchor=center}]
					\addplot[blue!25,thin,forget plot] coordinates {(1e-06,-0.3215526634954957) (1e-05,-0.3215526634954957) (0.0001,-0.3215526634954957) (0.001,-0.3215526634954957) (0.01,-0.3215526634954957) (0.1,-0.3215526634954957)};
					\addplot[blue!25,thin,forget plot] coordinates {(1e-06,-0.3119140755353739) (1e-05,-0.3119140755353739) (0.0001,-0.3119140755353739) (0.001,-0.3119140755353739) (0.01,-0.3119140755353739) (0.1,-0.3119140755353739)};
					\addplot[blue!25,thin,forget plot] coordinates {(1e-06,-0.3146135116884961) (1e-05,-0.3146135116884961) (0.0001,-0.3146135116884961) (0.001,-0.3146135116884961) (0.01,-0.3146135116884961) (0.1,-0.3146135116884961)};
					\addplot[blue!25,thin,forget plot] coordinates {(1e-06,-0.3182813117207202) (1e-05,-0.3182813117207202) (0.0001,-0.3182813117207202) (0.001,-0.3182813117207202) (0.01,-0.3182813117207202) (0.1,-0.3182813117207202)};
					\addplot[blue!25,thin,forget plot] coordinates {(1e-06,-0.319166814054615) (1e-05,-0.319166814054615) (0.0001,-0.319166814054615) (0.001,-0.319166814054615) (0.01,-0.319166814054615) (0.1,-0.319166814054615)};
					\addplot[blue!25,thin,forget plot] coordinates {(1e-06,-0.3118188005343359) (1e-05,-0.3118188005343359) (0.0001,-0.3118188005343359) (0.001,-0.3118188005343359) (0.01,-0.3118188005343359) (0.1,-0.3118188005343359)};
					\addplot[blue!25,thin,forget plot] coordinates {(1e-06,-0.3050159521947666) (1e-05,-0.3050159521947666) (0.0001,-0.3050159521947666) (0.001,-0.3050159521947666) (0.01,-0.3050159521947666) (0.1,-0.3050159521947666)};
					\addplot[blue!25,thin,forget plot] coordinates {(1e-06,-0.3212492067748964) (1e-05,-0.3212492067748964) (0.0001,-0.3212492067748964) (0.001,-0.3212492067748964) (0.01,-0.3212492067748964) (0.1,-0.3212492067748964)};
					\addplot[blue!25,thin,forget plot] coordinates {(1e-06,-0.3184481508277692) (1e-05,-0.3184481508277692) (0.0001,-0.3184481508277692) (0.001,-0.3184481508277692) (0.01,-0.3184481508277692) (0.1,-0.3184481508277692)};
					\addplot[blue!25,thin,forget plot] coordinates {(1e-06,-0.3248115685334658) (1e-05,-0.3248115685334658) (0.0001,-0.3248115685334658) (0.001,-0.3248115685334658) (0.01,-0.3248115685334658) (0.1,-0.3248115685334658)};
					\addplot[blue,thick,mark=*] coordinates {(1e-06,-0.3166872055359934) (1e-05,-0.3166872055359934) (0.0001,-0.3166872055359934) (0.001,-0.3166872055359934) (0.01,-0.3166872055359934) (0.1,-0.3166872055359934)};
					\addlegendentry{Raw MPC}
					\addplot[orange!25,thin,forget plot] coordinates {(1e-06,-0.4564852144760429) (1e-05,-0.456465288081728) (0.0001,-0.4563938913352206) (0.001,-0.4566328656182748) (0.01,-0.479629615118626) (0.1,-0.4911332116391768)};
					\addplot[orange!25,thin,forget plot] coordinates {(1e-06,-0.4469795649379003) (1e-05,-0.4469300841103228) (0.0001,-0.4464606820132737) (0.001,-0.4426556036697226) (0.01,-0.4395421453052668) (0.1,-0.4373286445934312)};
					\addplot[orange!25,thin,forget plot] coordinates {(1e-06,-0.4550718393134187) (1e-05,-0.4550319964995845) (0.0001,-0.4546741308962458) (0.001,-0.4529959601981056) (0.01,-0.4623893753973942) (0.1,-0.4824737588795402)};
					\addplot[orange!25,thin,forget plot] coordinates {(1e-06,-0.4523308765595329) (1e-05,-0.4522694097493927) (0.0001,-0.4516808361053682) (0.001,-0.4596537311751581) (0.01,-0.4370116438694561) (0.1,-0.4196263307553466)};
					\addplot[orange!25,thin,forget plot] coordinates {(1e-06,-0.4546438869453216) (1e-05,-0.4546298284520796) (0.0001,-0.4545500276224547) (0.001,-0.4552248955477525) (0.01,-0.4957501698107427) (0.1,-0.488419542931109)};
					\addplot[orange!25,thin,forget plot] coordinates {(1e-06,-0.4452186122194798) (1e-05,-0.4452110031197807) (0.0001,-0.4451619886401844) (0.001,-0.4454840160585023) (0.01,-0.4477489153743122) (0.1,-0.4306033790350823)};
					\addplot[orange!25,thin,forget plot] coordinates {(1e-06,-0.4429687801342815) (1e-05,-0.442961004959011) (0.0001,-0.4428982453506184) (0.001,-0.4433951248312599) (0.01,-0.4748705420845728) (0.1,-0.4522259315667744)};
					\addplot[orange!25,thin,forget plot] coordinates {(1e-06,-0.4556930165423745) (1e-05,-0.4556707871521991) (0.0001,-0.4554800266698301) (0.001,-0.4538874498061128) (0.01,-0.4566892147191574) (0.1,-0.4714874950164284)};
					\addplot[orange!25,thin,forget plot] coordinates {(1e-06,-0.4617004511260982) (1e-05,-0.4616656323391737) (0.0001,-0.461344160859089) (0.001,-0.4607277193248033) (0.01,-0.4712277371330185) (0.1,-0.4542817120522058)};
					\addplot[orange!25,thin,forget plot] coordinates {(1e-06,-0.4615885227778329) (1e-05,-0.4615634657199158) (0.0001,-0.4613458757787576) (0.001,-0.4604899220084206) (0.01,-0.4750336708950422) (0.1,-0.5085010893406431)};
					\addplot[orange,thick,mark=*] coordinates {(1e-06,-0.4532680765032284) (1e-05,-0.4532398500183188) (0.0001,-0.4529989865271043) (0.001,-0.4531147288238112) (0.01,-0.4639893029707589) (0.1,-0.4636081095809738)};
					\addlegendentry{Projected MPC}
					\nextgroupplot[title={(d) Paired projection effects},ylabel={\(\tau_\delta\)},ymax=0.01,legend style={at={(0.5,0.8)},anchor=center}]
					\addplot[blue!25,thin,forget plot] coordinates {(1e-06,-0.1349325509805472) (1e-05,-0.1349126245862324) (0.0001,-0.134841227839725) (0.001,-0.1350802021227792) (0.01,-0.1580769516231303) (0.1,-0.1695805481436812)};
					\addplot[blue!25,thin,forget plot] coordinates {(1e-06,-0.1350654894025264) (1e-05,-0.1350160085749489) (0.0001,-0.1345466064778998) (0.001,-0.1307415281343488) (0.01,-0.1276280697698929) (0.1,-0.1254145690580574)};
					\addplot[blue!25,thin,forget plot] coordinates {(1e-06,-0.1404583276249226) (1e-05,-0.1404184848110885) (0.0001,-0.1400606192077497) (0.001,-0.1383824485096095) (0.01,-0.1477758637088982) (0.1,-0.1678602471910442)};
					\addplot[blue!25,thin,forget plot] coordinates {(1e-06,-0.1340495648388127) (1e-05,-0.1339880980286726) (0.0001,-0.1333995243846481) (0.001,-0.1413724194544379) (0.01,-0.118730332148736) (0.1,-0.1013450190346264)};
					\addplot[blue!25,thin,forget plot] coordinates {(1e-06,-0.1354770728907066) (1e-05,-0.1354630143974646) (0.0001,-0.1353832135678397) (0.001,-0.1360580814931375) (0.01,-0.1765833557561277) (0.1,-0.169252728876494)};
					\addplot[blue!25,thin,forget plot] coordinates {(1e-06,-0.133399811685144) (1e-05,-0.1333922025854448) (0.0001,-0.1333431881058486) (0.001,-0.1336652155241664) (0.01,-0.1359301148399764) (0.1,-0.1187845785007465)};
					\addplot[blue!25,thin,forget plot] coordinates {(1e-06,-0.137952827939515) (1e-05,-0.1379450527642445) (0.0001,-0.1378822931558518) (0.001,-0.1383791726364934) (0.01,-0.1698545898898062) (0.1,-0.1472099793720078)};
					\addplot[blue!25,thin,forget plot] coordinates {(1e-06,-0.1344438097674781) (1e-05,-0.1344215803773027) (0.0001,-0.1342308198949337) (0.001,-0.1326382430312165) (0.01,-0.135440007944261) (0.1,-0.1502382882415321)};
					\addplot[blue!25,thin,forget plot] coordinates {(1e-06,-0.143252300298329) (1e-05,-0.1432174815114045) (0.0001,-0.1428960100313199) (0.001,-0.1422795684970341) (0.01,-0.1527795863052493) (0.1,-0.1358335612244366)};
					\addplot[blue!25,thin,forget plot] coordinates {(1e-06,-0.1367769542443671) (1e-05,-0.13675189718645) (0.0001,-0.1365343072452918) (0.001,-0.1356783534749548) (0.01,-0.1502221023615763) (0.1,-0.1836895208071773)};
					\addplot[blue,thick,mark=*] coordinates {(1e-06,-0.1365808709672349) (1e-05,-0.1365526444823253) (0.0001,-0.1363117809911108) (0.001,-0.1364275232878178) (0.01,-0.1473020974347655) (0.1,-0.1469209040449803)};
					\addlegendentry{Mean paired effect}
					\addplot[black,dashed,forget plot] coordinates {(1e-6,0) (1e-1,0)};
				\end{groupplot}
		\end{tikzpicture}}
		\caption{Effect of the ridge value on ten paired replicates. Panel (a): applied and candidate ranks. Panel (b): cross-block residuals. Panel (c): final gaps. Panel (d): paired effects of the projection; the dashed line is zero. Thin lines: single replicates; thick lines: means. The same cohorts are reused for every ridge value, so these are paired comparisons, not 60 independent repeats.}
		\label{fig:metric-sensitivity}
	\end{figure}

	Figure~\ref{fig:metric-sensitivity} shows four checks. Panel (a): the ranks do not change. Panel (b): the residuals \(r_1\) and \(r_2\), each in its own metric. Panel (c): the final gaps. Panel (d): the paired effects \(\tau_\delta\), with a dashed line at zero. Thin lines are single replicates and thick lines are means. For every ridge value, the projection makes things worse.
	
	\begin{table}[t]
		\centering\small
		\resizebox{0.9\textwidth}{!}{\begin{tabular}{@{}lrrrrr@{}}
				\toprule
				Ridge ratio & Mean \(r_1\) & Mean \(r_2\) & Mean \(\delta^{\mathrm{proj}}\) & Mean \(\tau_\delta\) & 95\% interval \\
				\midrule
				\(10^{-6}\) & 0.259805 & 0.352879 & -0.453268 & -0.136581 & \([-0.138828,-0.134334]\) \\
				\(10^{-5}\) & 0.259749 & 0.352850 & -0.453240 & -0.136553 & \([-0.138798,-0.134307]\) \\
				\(10^{-4}\) & 0.259734 & 0.352769 & -0.452999 & -0.136312 & \([-0.138545,-0.134079]\) \\
				\(10^{-3}\) & 0.262496 & 0.335681 & -0.453115 & -0.136428 & \([-0.139074,-0.133781]\) \\
				\(10^{-2}\) & 0.296710 & 0.314560 & -0.463989 & -0.147302 & \([-0.160350,-0.134254]\) \\
				\(10^{-1}\) & 0.213994 & 0.446069 & -0.463608 & -0.146921 & \([-0.165799,-0.128043]\) \\
				\bottomrule
		\end{tabular}}
		\caption{Mean paired effect for each ridge value, over ten replicates.}
		\label{tab:metric-sensitivity}
	\end{table}

	\subsection{Policy-matched conformal check}
	\label{app:conformal}
	\subsubsection{Method}
	\label{sec:calibration}
	Before calibration, we freeze the trained model and the projected MPC policy. For each cohort, the model predicts \(\widehat\delta_b\) from the starting information only, while the real policy still acts on new observations at every step. Calibration and test cohorts have the same balanced design, the same policy, the same horizon and the same outcome.
	
	With \(n\) calibration cohorts, scores \(s_b=|\delta_b-\widehat\delta_b|\) and a miscoverage level \(\alpha\), we set
	\begin{equation}
		\ell_\alpha=\lceil(n+1)(1-\alpha)\rceil,\qquad
		\widehat q=\begin{cases}s_{(\ell_\alpha)},&\ell_\alpha\leq n,\\
			+\infty,&\ell_\alpha=n+1.
		\end{cases}
	\end{equation}
	The interval for a new cohort is \(C_{n+1}=[\widehat\delta_{n+1}-\widehat q,\widehat\delta_{n+1}+\widehat q]\). If the training data are independent and the calibration and new cohorts are exchangeable, this interval contains the realized gap \(\delta_{n+1}\) of the next cohort, under the same frozen policy, with probability at least \(\ell_\alpha/(n+1)\geq1-\alpha\). The target is the random gap of a new cohort, not a fixed population value such as \(\Delta_D\), and the guarantee holds on average over calibration and test cohorts. This is the standard split-conformal argument \citep{lei2018distribution,angelopoulos2023conformal}. We use 100 calibration cohorts and 100 test cohorts, with \(\alpha=0.05\).
	
	\paragraph{Policy updates and calibration validity.}
	The guarantee needs the model, the prediction rule and the policy to stay fixed before calibration. If the policy changes, the outcomes and the errors usually change too, so scores from different policies cannot be mixed. After any update, the new policy must be frozen and calibrated again on new cohorts.
	\subsubsection{Results: coverage is high, but the intervals are wide}
	\label{sec:results-conformal}
	The conformal order is \(\ell_\alpha=96\) and the calibrated radius is \(\widehat q=0.467798\). Ninety-eight of the one hundred test gaps fall inside their interval, so the observed coverage is 0.980, with a Wilson 95\% interval of \([0.930,0.994]\). This sounds very good, but the intervals are about 0.936 wide, which is large compared with the gaps themselves. So coverage can look excellent while the intervals tell us almost nothing about whether the gap is close to zero. This is a check of average calibration for the next cohort under one fixed policy; it says nothing about how useful the prediction is.
	
\begin{figure}[t]
		\centering
		\resizebox{0.9\textwidth}{!}{\input{journal_conformal_diagnostics.tex}}
		\caption{Conformal check under one fixed policy. Left: distributions of the absolute errors and the calibrated radius. Right: intervals, predictions from the starting information only, real test gaps, and the two misses.}
		\label{fig:conformal-diagnostics}
	\end{figure}
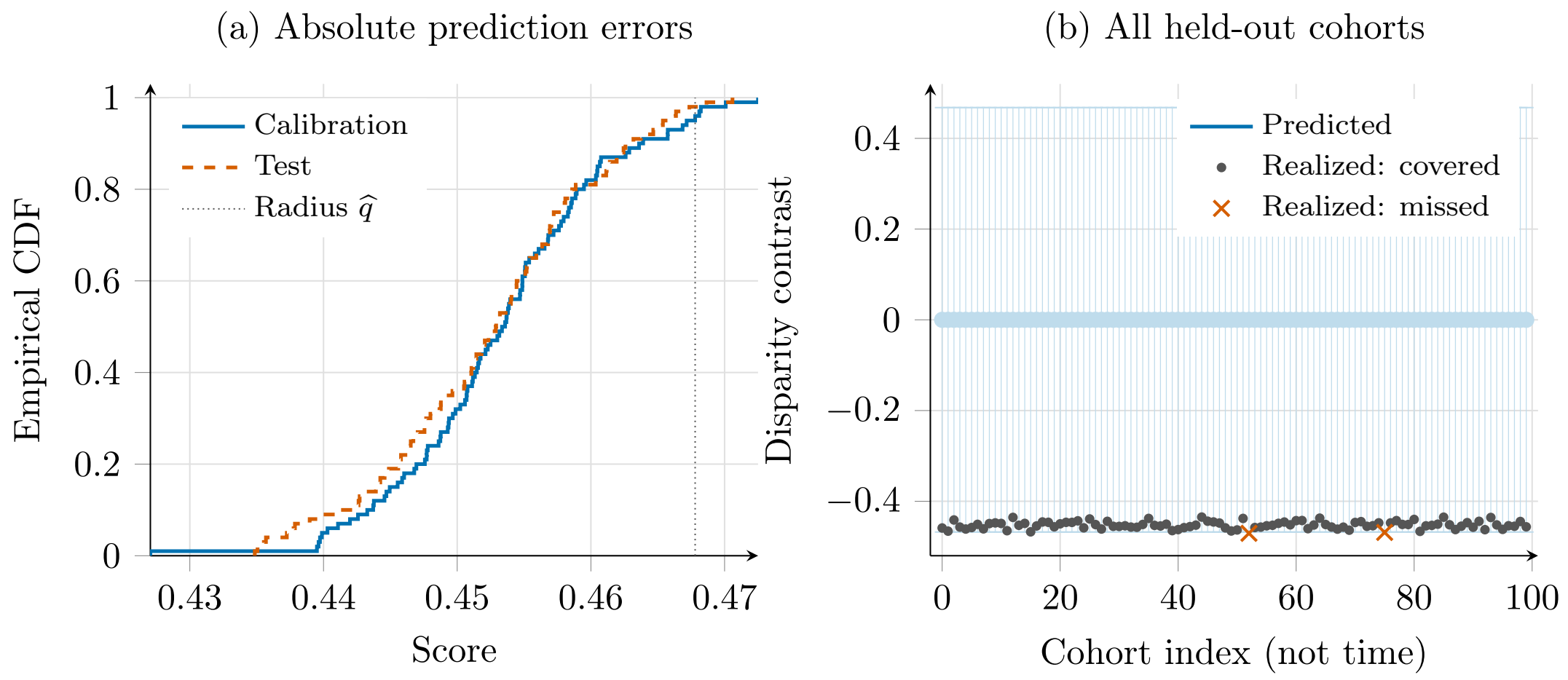

	Figure~\ref{fig:conformal-diagnostics} shows both sides of this result. On the left are the distributions of the absolute errors for calibration and test cohorts, with the radius \(\widehat q\). On the right are all 100 test cohorts: the thin vertical lines are the intervals, the blue curve is the prediction from the starting information only, and the black points are the real gaps; the two orange crosses are the misses. No test outcome was used to choose the radius.
	
	\section{Randomized experiments}
	\label{app:block-rct}
	\subsection{A complementary illustration: 30 randomized experiments}
	\label{app:rct}\label{sec:results-rct}
	The policy results of the paper come from a simulator. This appendix reports an exploratory analysis of a different and narrower question with real students: does choosing between two existing conditions of an experiment by group narrow a real completion gap, and does any group lose? In randomized experiments the assignment probability is known by design, so the value of a new assignment rule can be estimated from past data. These data contain one decision per student, not learning paths. The analysis therefore tests neither the EDMD projection, nor a sequential controller, nor the effect of extra help; it illustrates the decision level of the audit and does not confirm the mechanism found in the simulator.

	\paragraph{Data and rules.} We use the randomized experiments released with \citet{sales2023auxiliary}. Two selection rules were set in the data audit, which uses no outcome: at least 56 students in each arm (the rule of the data authors), and a treated share compatible with assignment probability \(1/2\) (two-sided binomial test, \(p\ge0.10\)). The probability \(1/2\) comes from the design: the data authors report that the platform assigned students independently to the conditions with equal probability, but that a few experiments probably merged two conditions, giving \(1/3\) or \(2/3\); the test, which they also use, removes such experiments. A test that is not significant does not prove that the probability was \(1/2\). Table~\ref{tab:randomization} documents the assignment in each retained experiment: the treated share lies between 0.47 and 0.54 (median 0.50), and within the low and high groups 2 of the 60 shares differ from \(1/2\) at the 5\% level, about the number expected by chance. Thirty experiments pass both rules; the 26 that pass the first rule but not the second enter only a sensitivity analysis, with their observed assignment probabilities. The 30 experiments contain 30,640 records of 21,557 students in 1,788 classes (125 to 2,909 records per experiment). In each experiment, students who worked on one problem set were assigned at random to a control or a treatment condition, for example a hint, feedback or another presentation. The outcome is whether the student completed the assignment. The groups are terciles of the student's prior correctness, measured before randomization; we compare the bottom (``low'') and the top (``high'') terciles. These are achievement groups, not demographic groups. We compare six rules that choose a condition for each student: control for all (P0); treatment for all (P1); the condition with the higher estimated overall completion (P2, the group-blind reference); the better condition within each group (P3); a personalized, group-blind rule (P4, a T-learner with gradient boosting on the baseline covariates); and a rule with floors (P5). P5 is like P4, but a low achiever gets the condition that is better for the low group, as long as the estimated completion of each group stays at or above its value under P2. It is the one-decision counterpart of the floor-constrained controller.

	\paragraph{Estimation.} In each experiment, classes are split into five folds. The rules are learned on four folds and applied to the fifth, so that every student is evaluated once, by a rule that did not see them. The value of a rule for a group is estimated with a doubly robust estimator that uses the known assignment probability. Values are pooled over experiments with student weights. Because 31\% of the classes and 27\% of the students appear in more than one experiment, the 95\% intervals come from a cluster bootstrap that resamples classes \emph{jointly across experiments}, so that a class keeps all its records in every experiment (2,000 draws); this is the primary analysis, and it is listed among the changes to the plan. Resampling classes within experiments, as the plan first specified, or students jointly gives the same intervals to within 0.1 point (Appendix~\ref{app:rct-dependence}). A rule meets the floor if, for each group, the one-sided 5\% lower limit of its difference with P2 is above \(-2\) points: under the assumptions of the analysis, this excludes a loss larger than 2 points, not every loss. The analysis plan sets the selection rules, the groups, the rules for choosing a condition, the margin and four sensitivity analyses. It was not registered, and its original text was not preserved: it was later extended with the list of changes that it requires. What can be checked is the code. The analysis code was hashed before any outcome file was written (SHA-256, 12 October 2026, 13:04 UTC, two minutes before the first outcome file), and the code included with the paper matches this hash (\texttt{3f77e48d}\dots). Every later change is listed in the plan.

	\begin{table}[t]
		\centering\small
		\caption{Randomized experiments: completion under each rule minus completion under the group-blind reference P2, in percentage points, pooled over 30 experiments, with 95\% intervals from a bootstrap that resamples classes jointly across experiments. Floor: the one-sided 5\% lower limit is above \(-2\) points for both groups. Under P2, completion is 63.8\% for low and 82.0\% for high achievers, a gap of 18.1 points.}
		\label{tab:rct}
		\resizebox{\linewidth}{!}{%
		\begin{tabular}{@{}llllll@{}}
			\toprule
			Rule & Uses group & Low achievers & High achievers & Gap (high \(-\) low) & Floor met\\
			\midrule
			P0 control for all & no & \(-0.5\) \([-1.6,+0.7]\) & \(+0.8\) \([-0.1,+1.7]\) & \(+1.3\) \([-0.1,+2.7]\) & yes\\
			P1 treatment for all & no & \(-1.5\) \([-2.9,-0.3]\) & \(-0.5\) \([-1.6,+0.5]\) & \(+1.0\) \([-0.5,+2.5]\) & no (low)\\
			P3 better condition per group & yes & \(+0.6\) \([-0.1,+1.3]\) & \(+0.4\) \([-0.3,+1.1]\) & \(-0.2\) \([-1.2,+0.8]\) & yes\\
			P4 personalized & no & \(-0.7\) \([-1.8,+0.5]\) & \(+0.2\) \([-0.7,+1.1]\) & \(+0.8\) \([-0.6,+2.2]\) & yes\\
			P5 with floors & yes & \(+0.6\) \([-0.1,+1.3]\) & \(+0.2\) \([-0.7,+1.1]\) & \(-0.5\) \([-1.5,+0.7]\) & yes\\
			\bottomrule
		\end{tabular}}
	\end{table}

	\paragraph{Results.} No rule narrows the 18-point gap significantly (Table~\ref{tab:rct}). This does not show that the effect is zero: the rule with floors changes the gap by \(-0.5\) points, with an interval from \(-1.5\) to \(+0.7\), and it meets the floor for both groups; choosing the better condition within each group gives \(-0.2\) \([-1.2,+0.8]\). Giving the treatment to every student lowers the completion of low achievers by \(1.5\) points \([-2.9,-0.3]\); it fails the floor for this group (lower limit \(-2.6\)), and it lowers them in every sensitivity analysis (by \(1.5\) to \(3.5\) points). The other results do not depend on the analysis choices. In the four sensitivity analyses (26 more experiments with observed assignment probabilities, groups defined by prior completion, inverse-probability values, and only experiments with at least ten classes), the gap change of P5 lies between \(-0.2\) and \(-1.0\) points, and with the joint bootstrap none of the intervals excludes zero (with 56 experiments: \(-1.0\) \([-1.9,+0.0]\)). Tests in each experiment, with a Holm correction across the 30 experiments, find one experiment where choosing by group narrows the gap (by \(4.2\) points).

	\paragraph{Why so little?} The estimated treatment effects are not small: for low achievers, the middle half runs from \(-6\) to \(+3\) points, and in 43\% of the experiments the estimated effect is larger than 5 points in size. But each estimate is noisy, and with only two conditions the room to move a gap is limited. Even an in-sample oracle that picks, in each experiment, the condition for each group that makes the gap smallest, with no floor and with estimation noise in its favour, moves the gap by only \(3.6\) points on average, and partly by giving the high achievers the worse condition. (This spread and this oracle are descriptions added after the outcomes were seen; they are not part of the plan.)

	\paragraph{What this shows, and what it does not.} With real students and one decision each, choosing between two existing conditions by group gives no detectable narrowing of an 18-point gap: the data are compatible with changes from about \(-1.5\) to \(+0.7\) points. This is a limited conclusion: the average improvement is not established in this analysis, which does not show that reallocating interventions is ineffective in general. With floors, the analysis excludes a loss larger than 2 points for either group, and an untargeted change, treatment for all, lowers the completion of the group behind. The experiments are small and their effects imprecise, as the data authors also note, and the plan was not registered, so the analysis is exploratory. These data cannot test whether extra help would narrow the gap.

	\begin{table}[htbp]
		\centering\footnotesize
		\caption{Assignment in the 30 retained experiments, ordered by size (no outcome is used). \(n\): students; share: treated share; \(p\): two-sided binomial test against \(1/2\) (the selection rule keeps \(p\ge0.10\)); low, high: treated share in the low and high prior-achievement groups. The assignment probability itself is not recorded in the data.}
		\label{tab:randomization}
		\begin{tabular}{@{}rrrrrr@{\hspace{1.2em}}rrrrrr@{}}
			\toprule
			Exp. & \(n\) & Share & \(p\) & Low & High & Exp. & \(n\) & Share & \(p\) & Low & High\\
			\midrule
			1 & 2909 & 0.498 & 0.88 & 0.49 & 0.51 & 16 & 901 & 0.497 & 0.89 & 0.47 & 0.56\\
			2 & 2371 & 0.483 & 0.10 & 0.48 & 0.47 & 17 & 749 & 0.477 & 0.21 & 0.43 & 0.55\\
			3 & 2023 & 0.499 & 0.96 & 0.49 & 0.51 & 18 & 746 & 0.483 & 0.36 & 0.47 & 0.49\\
			4 & 1903 & 0.481 & 0.11 & 0.49 & 0.46 & 19 & 688 & 0.528 & 0.16 & 0.54 & 0.54\\
			5 & 1881 & 0.509 & 0.46 & 0.51 & 0.52 & 20 & 467 & 0.518 & 0.46 & 0.54 & 0.51\\
			6 & 1835 & 0.490 & 0.40 & 0.53 & 0.48 & 21 & 431 & 0.522 & 0.39 & 0.52 & 0.49\\
			7 & 1669 & 0.512 & 0.35 & 0.48 & 0.53 & 22 & 349 & 0.513 & 0.67 & 0.52 & 0.48\\
			8 & 1598 & 0.496 & 0.78 & 0.48 & 0.51 & 23 & 333 & 0.492 & 0.83 & 0.57 & 0.41\\
			9 & 1540 & 0.485 & 0.25 & 0.44 & 0.54 & 24 & 276 & 0.493 & 0.86 & 0.52 & 0.49\\
			10 & 1355 & 0.511 & 0.42 & 0.50 & 0.58 & 25 & 269 & 0.532 & 0.33 & 0.48 & 0.46\\
			11 & 1189 & 0.516 & 0.30 & 0.51 & 0.53 & 26 & 252 & 0.524 & 0.49 & 0.53 & 0.52\\
			12 & 1186 & 0.506 & 0.71 & 0.49 & 0.53 & 27 & 181 & 0.470 & 0.46 & 0.51 & 0.48\\
			13 & 1125 & 0.480 & 0.19 & 0.50 & 0.50 & 28 & 172 & 0.494 & 0.94 & 0.55 & 0.52\\
			14 & 1038 & 0.509 & 0.60 & 0.53 & 0.52 & 29 & 126 & 0.540 & 0.42 & 0.61 & 0.56\\
			15 & 953 & 0.486 & 0.40 & 0.46 & 0.49 & 30 & 125 & 0.512 & 0.86 & 0.47 & 0.50\\
			\bottomrule
		\end{tabular}
	\end{table}

	\subsubsection{Dependence between the randomized experiments}
	\label{app:rct-dependence}
	Of the 21,557 students, 5,848 (27\%) take part in more than one experiment, and they account for 49\% of the 30,640 records; 547 of the 1,788 classes (31\%) appear in more than one experiment, with 58\% of the records. The bootstrap first specified in the plan resamples classes within each experiment, independently across experiments, so it ignores this dependence. The primary analysis of Appendix~\ref{app:rct} therefore resamples classes jointly across experiments. As a check, a second implementation recomputes the pooled differences with P2 with classes, and with students, resampled jointly across experiments, from the same doubly robust scores (2,000 draws each). The intervals agree to within 0.1 point: for the rule with floors, the gap change is \(-0.5\) points with \([-1.6,+0.6]\) (classes) and \([-1.6,+0.7]\) (students), against \([-1.5,+0.7]\) in Table~\ref{tab:rct} and \([-1.6,+0.7]\) with the within-experiment bootstrap; for treatment for all, the change for low achievers is \(-1.5\) with \([-2.9,-0.3]\) and \([-2.8,-0.3]\), and its one-sided lower limit stays at \(-2.6\). These checks were added after the outcomes were seen and are listed in the plan's changes.

	\FloatBarrier
	\end{document}

%% file: tab_controlled.tex
\textbf{(a) Price and announced cost}\\[2pt]
\begin{tabular}{@{}llrrrrr@{}}
\toprule
 & & & \multicolumn{2}{c}{Population cost} & \multicolumn{2}{c}{Median announced}\\
\cmidrule(lr){4-5}\cmidrule(l){6-7}
Group signal & \(a\) & \(R^2_{\grp}\) & minimum & built & v1 & quadratic\\
\midrule
Useless & 0.2 & 0.010 & 0.000 & 0.000 & 0.002 & 0.002\\
Useless & 0.7 & 0.107 & 0.000 & 0.000 & 0.000 & 0.000\\
Useless & 1.5 & 0.375 & 0.000 & 0.000 & 0.000 & 0.000\\
\addlinespace
Partly aligned & 0.2 & 0.010 & 0.252 & 0.252 & 0.253 & 0.248\\
Partly aligned & 0.7 & 0.107 & 0.276 & 0.276 & 0.277 & 0.275\\
Partly aligned & 1.5 & 0.375 & 0.360 & 0.360 & 0.359 & 0.362\\
\addlinespace
Aligned & 0.2 & 0.010 & 1.010 & 1.010 & 0.981 & 1.035\\
Aligned & 0.7 & 0.106 & 1.104 & 1.104 & 1.100 & 1.111\\
Aligned & 1.5 & 0.375 & 1.441 & 1.441 & 1.442 & 1.439\\
\addlinespace
Shift: utility lost & 0.7 & 0.106 & 1.104 & 1.104 & 1.102 & 1.105\\
Shift: utility gained & 0.7 & 0.107 & 0.000 & 0.000 & 0.000 & 0.000\\
\addlinespace
Calibrated, misspecified \(s\) & -- & 0.320 & 0.500 & 1.000 & 0.997 & 1.000\\
Same, correct \(s\) & -- & 0.320 & 0.500 & 0.500 & 0.501 & 0.500\\
\bottomrule
\end{tabular}

\bigskip

\textbf{(b) Audit with 1{,}000 bootstrap draws: acceptance, transfer, coverage and width}\\[2pt]
\begin{tabular}{@{}llrrrrrrrr@{}}
\toprule
 & & \multicolumn{4}{c}{\(n=2{,}000\)} & \multicolumn{4}{c}{\(n=20{,}000\)}\\
\cmidrule(lr){3-6}\cmidrule(l){7-10}
Group signal & \(a\) & acc. & transf. & cov. & width & acc. & transf. & cov. & width\\
\midrule
Useless & 0.2 & 0.00 & 0.00 & -- & -- & 0.00 & 0.00 & -- & --\\
Useless & 0.7 & 0.00 & 0.00 & -- & -- & 0.00 & 0.00 & -- & --\\
Useless & 1.5 & 0.00 & 0.00 & -- & -- & 0.00 & 0.00 & -- & --\\
\addlinespace
Partly aligned & 0.2 & 0.00 & 0.00 & -- & 12.27 & 1.00 & 0.72 & 1.00 & 1.04\\
Partly aligned & 0.7 & 1.00 & 0.80 & 1.00 & 0.98 & 1.00 & 1.00 & 1.00 & 0.30\\
Partly aligned & 1.5 & 1.00 & 0.99 & 1.00 & 0.52 & 1.00 & 1.00 & 1.00 & 0.16\\
\addlinespace
Aligned & 0.2 & 0.05 & 0.05 & 1.00 & 7.86 & 1.00 & 1.00 & 1.00 & 0.22\\
Aligned & 0.7 & 1.00 & 1.00 & 0.96 & 0.27 & 1.00 & 1.00 & 0.95 & 0.07\\
Aligned & 1.5 & 1.00 & 1.00 & 0.96 & 0.19 & 1.00 & 1.00 & 0.92 & 0.06\\
\addlinespace
Shift: utility lost & 0.7 & 1.00 & 0.02 & 0.99 & 0.28 & 1.00 & 0.01 & 0.95 & 0.07\\
Shift: utility gained & 0.7 & 0.00 & 0.00 & -- & -- & 0.00 & 0.00 & -- & --\\
\addlinespace
Calibrated, misspecified \(s\) & -- & 1.00 & 1.00 & 0.98 & 0.18 & 1.00 & 1.00 & 0.91 & 0.06\\
Same, correct \(s\) & -- & 1.00 & 1.00 & 1.00 & 0.40 & 1.00 & 1.00 & 0.97 & 0.13\\
\bottomrule
\end{tabular}

%% file: tab_next_item.tex
\begin{tabular}{@{}lrrrrrrrr@{}}
\toprule
 & \multicolumn{4}{c}{ASSISTments (school)} & \multicolumn{4}{c}{Eedi (PremiumPupil)}\\
\cmidrule(lr){2-5}\cmidrule(l){6-9}
Model & AUC & \(\Delta\)AUC & \(\Delta\)Brier & \(\widehat\tau\) & AUC & \(\Delta\)AUC & \(\Delta\)Brier & \(\widehat\tau\)\\
\midrule
EDMD, original & 0.659 & \(-0.017\) 2/5\(\downarrow\) & \(+0.0005\) & \(+0.83\) & 0.687 & \(-0.032\) 5/5\(\downarrow\) & \(+0.0088\) & \(+0.97\)\\
EDMD, projected & 0.518 & \(-0.158\) 5/5\(\downarrow\) & \(+0.0151\) & \(-0.00\) & 0.521 & \(-0.198\) 5/5\(\downarrow\) & \(+0.0320\) & \(+0.04\)\\
\addlinespace
DKT, selected on validation & 0.676 & ref. & ref. & 1 & 0.719 & ref. & ref. & 1\\
DKT, LEACE on the hidden state & 0.676 & \(-0.000\) 1/5\(\downarrow\) & \(+0.0000\) & \(+1.11\) & 0.693 & \(-0.026\) 5/5\(\downarrow\) & \(+0.0077\) & \(+0.61\)\\
DKT, LEACE within each next skill & 0.670 & \(-0.006\) 4/5\(\downarrow\) & \(+0.0018\) & \(+0.88\) & 0.685 & \(-0.034\) 5/5\(\downarrow\) & \(+0.0098\) & \(+0.43\)\\
DKT, mean-parity fine-tuning & 0.628 & \(-0.048\) 5/5\(\downarrow\) & \(+0.0125\) & \(+0.31\) & 0.623 & \(-0.096\) 5/5\(\downarrow\) & \(+0.0224\) & \(+0.03\)\\
DKT, output correction (Prop.~\ref{prop:built}), three-stage & 0.588 & \(-0.088\) 5/5\(\downarrow\) & \(+0.0628\) & \(+0.59\) & 0.642 & \(-0.077\) 5/5\(\downarrow\) & \(+0.0211\) & \(+0.03\)\\
\bottomrule
\end{tabular}

%% file: tab_next_item_compare.tex
\begin{tabular}{@{}lrrrrrrr@{}}
\toprule
 & \multicolumn{3}{c}{Method} & \multicolumn{4}{c}{Output correction at the same gap}\\
\cmidrule(lr){2-4}\cmidrule(l){5-8}
 & & & & \multicolumn{2}{c}{matched on test} & \multicolumn{2}{c}{set on validation}\\
\cmidrule(lr){5-6}\cmidrule(l){7-8}
Method & \(\widehat\tau\) & Cost & Cross & Cost & splits \(>0\) & \(\widehat\tau\) & Cost\\
\midrule
LEACE, hidden state & \(+0.62\) & 0.0077 & \(-0.0001\) & 0.0033 & 5/5 & \(+0.66\) & 0.0028\\
LEACE, within next skill & \(+0.40\) & 0.0098 & \(-0.0006\) & 0.0079 & 2/5 & \(+0.42\) & 0.0076\\
Mean-parity fine-tuning\(^{\dagger}\) & \(+0.04\) & 0.0224 & \(+0.0006\) & 0.0200 & 2/5 & \(+0.02\) & 0.0201\\
Output correction (Prop.~\ref{prop:built}) & \(-0.01\) & 0.0214 & \(+0.0015\) & -- & -- & -- & --\\
\bottomrule
\end{tabular}

%% file: tab_next_item_audit_main.tex
\begin{tabular}{@{}lrlrrll@{}}
\toprule
 & \multicolumn{3}{c}{Announced total cost} & Measured & Measured \(-\) & \\
\cmidrule(lr){2-4}
Split & estimate & 95\% interval & width & (test) & announced, 95\% & Decision\\
\midrule
\multicolumn{7}{@{}l}{\emph{Eedi, new classes}}\\
1 & 0.0209 & [0.0134, 0.0351] & 0.0217 & 0.0198 & [-0.0038, 0.0053] & accepted; transfers, \(\widehat\tau=-0.02\)\\
2 & 0.0228 & [0.0127, 0.0586] & 0.0460 & 0.0269 & [-0.0003, 0.0143] & accepted; does not transfer (T2), \(\widehat\tau=-0.21\)\\
3 & 0.0176 & [0.0030, 0.0321] & 0.0291 & 0.0172 & [-0.0033, 0.0036] & accepted; transfers, \(\widehat\tau=+0.17\)\\
4 & 0.0210 & [0.0124, 0.0396] & 0.0272 & 0.0210 & [-0.0051, 0.0052] & accepted; transfers, \(\widehat\tau=-0.01\)\\
5 & 0.0228 & [0.0138, 0.0472] & 0.0334 & 0.0192 & [-0.0092, 0.0014] & accepted; transfers, \(\widehat\tau=+0.09\)\\
\addlinespace
\multicolumn{7}{@{}l}{\emph{ASSISTments, new schools}}\\
1 & 0.0032 & [0.0000, 5.5114] & 5.5115 & 0.0021 & [-0.1996, 0.4422] & abstains (D1, D2, D3)\\
2 & 0.0001 & [-0.0008, 1.0482] & 1.0490 & 0.0017 & [-0.0110, 0.1311] & abstains (D1, D2, D3)\\
3 & 0.3026 & [0.0146, 91.5417] & 91.5272 & 0.2631 & [-10.2103, 0.5537] & abstains (D2, D3)\\
4 & 0.0172 & [0.0002, 4.1039] & 4.1038 & 0.0266 & [-0.0024, 1.6389] & abstains (D1, D2, D3)\\
5 & 0.0024 & [-0.0010, 3.4924] & 3.4933 & 0.0051 & [-0.1010, 0.2768] & abstains (D1, D2, D3)\\
\bottomrule
\end{tabular}

%% file: tab_oracle.tex
\begin{tabular}{@{}lrrrrrrr@{}}
\toprule
 & \multicolumn{5}{c}{Approximately equal budget (\(\mathcal U_1\) of the reference)} & \multicolumn{2}{c}{\(\rho=1\)}\\
\cmidrule(lr){2-6}\cmidrule(l){7-8}
Controller & \(\overline k_T\) & \(\overline k_{T,0}\) & \(\overline k_{T,1}\) & \(|\delta^k_T|\) & \(\rho\) & \(\overline k_T\) & \(\mathcal U_1\)\\
\midrule
\multicolumn{8}{@{}l}{Objective: mean knowledge, \(\gamma=1\); budget \(\mathcal U_1=0.300\)}\\
\quad Known equations, with the group & 0.667 & 0.921 & 0.414 & 0.507 & 1.00 & 0.667 & 0.300\\
\quad Known equations, group replaced by its mean & 0.667 & 0.921 & 0.413 & 0.508 & 1.01 & 0.667 & 0.300\\
\quad Learned, group-aware & 0.667 & 0.921 & 0.414 & 0.507 & 0.67 & 0.447 & 0.201\\
\quad Learned, original & 0.667 & 0.921 & 0.414 & 0.507 & 0.68 & 0.450 & 0.203\\
\quad Learned, group direction projected out & 0.667 & 0.921 & 0.414 & 0.507 & 0.55 & 0.369 & 0.165\\
\quad Learned, random rank-4 removal & 0.667 & 0.921 & 0.414 & 0.507 & 0.39 & 0.282 & 0.116\\
\addlinespace
\multicolumn{8}{@{}l}{Objective of the paper (\(\beta_{\mathrm{ind}}=8\)), \(\gamma=1\); budget \(\mathcal U_1=0.346\)}\\
\quad Known equations, with the group & 0.803 & 0.702 & 0.904 & 0.202 & 1.00 & 0.803 & 0.346\\
\quad Known equations, group replaced by its mean & 0.805 & 0.777 & 0.833 & 0.056 & 1.00 & 0.806 & 0.346\\
\quad Learned, group-aware & 0.792 & 0.633 & 0.952 & 0.319 & 0.78 & 0.680 & 0.289\\
\quad Learned, original & 0.794 & 0.649 & 0.939 & 0.290 & 0.75 & 0.676 & 0.287\\
\quad Learned, group direction projected out & 0.744 & 0.955 & 0.532 & 0.423 & 0.61 & 0.511 & 0.232\\
\quad Learned, random rank-4 removal & 0.675 & 0.754 & 0.596 & 0.327 & 0.42 & 0.422 & 0.182\\
\bottomrule
\end{tabular}

%% file: tab_structural.tex
\begin{tabular}{@{}lrrrrrr@{}}
\toprule
Quantity & Simulator & School-disj. & Other labels & Student split & Subsamples & Full label\\
\midrule
fits & 650 & 5 & 5 & 5 & 3 & 1\\
\(\max|K-A^\top|\) & 0 & 0 & 0 & 0 & 0 & 0\\
\(\max|P^2-P|\) & \(4.1\times10^{-13}\) & \(1.9\times10^{-14}\) & \(1.9\times10^{-14}\) & \(6.0\times10^{-14}\) & \(4.8\times10^{-14}\) & \(9.5\times10^{-15}\)\\
\(\max|\Gamma^{-1}P^\top\Gamma-P|\) & \(2.7\times10^{-12}\) & \(3.9\times10^{-13}\) & \(3.9\times10^{-13}\) & \(1.4\times10^{-12}\) & \(1.1\times10^{-12}\) & \(2.1\times10^{-13}\)\\
\(\max\epsilon_q\) & \(3.9\times10^{-13}\) & \(5.2\times10^{-15}\) & \(5.2\times10^{-15}\) & \(9.1\times10^{-15}\) & \(7.9\times10^{-15}\) & \(2.2\times10^{-15}\)\\
\(\max\|R^\top B_P\|\) & \(1.5\times10^{-13}\) & -- & -- & -- & -- & --\\
\(\max\|R^\top a_P\|\) & \(1.7\times10^{-14}\) & \(5.7\times10^{-16}\) & \(5.7\times10^{-16}\) & \(1.3\times10^{-15}\) & \(1.0\times10^{-15}\) & \(3.7\times10^{-16}\)\\
\(r_1\) (min--max) & 0.05--0.38 & 0.22--0.57 & 0.22--0.57 & 0.26--0.43 & 0.17--0.46 & 0.40--0.40\\
\(r_2\) (min--max) & 0.09--0.42 & 0.35--0.58 & 0.35--0.58 & 0.42--0.56 & 0.35--0.58 & 0.43--0.43\\
requested rank & 1--8 & 4 & 4 & 4 & 4 & 4\\
candidate rank & 1--10 & 13 & 13 & 13 & 13 & 13\\
applied rank & 1--7 & 4 & 4 & 4 & 4 & 4\\
numerical rank of \(\Gram\) & 10 & 13 & 13 & 13 & 13 & 13\\
relative SVD cutoff & \(1.0\times10^{-8}\) & \(1.0\times10^{-8}\) & \(1.0\times10^{-8}\) & \(1.0\times10^{-8}\) & \(1.0\times10^{-8}\) & \(1.0\times10^{-8}\)\\
\bottomrule
\end{tabular}

%% file: tab_flow.tex
\begin{tabular}{@{}lrrr@{}}
\toprule
Stage & Students & Schools & Rows\\
\midrule
Rows with a skill and a school & 29{,}006 & 402 & 2{,}711{,}753\\
At least 20 attempts & 18{,}366 & 243 & 2{,}622{,}847\\
At least 26 attempts (full window \(x_0,\ldots,x_{25}\)) & 16{,}668 & 216 & 2{,}584{,}780\\
School has a step after \(t=25\) (label window) & 16{,}661 & 211 & 2{,}584{,}598\\
\bottomrule
\end{tabular}

%% file: tab_eedi.tex
\begin{tabular}{@{}lrrrrrrl@{}}
\toprule
Split & Observed [95\% CI] & Raw & Recal. & Proj. & MSE ratio & \(\widehat\tau_\Delta\) [95\% CI] & Gap error \(\not\ni0\)\\
\midrule
1 & \(-0.093\) \([-0.113,-0.074]\) & \(-0.088\) & \(-0.095\) & \(+0.0003\) & 8.6 & \(+0.0069\) \([+0.0048,+0.0090]\) & Y/n/Y\\
2 & \(-0.079\) \([-0.098,-0.061]\) & \(-0.073\) & \(-0.080\) & \(+0.0020\) & 8.8 & \(+0.0062\) \([+0.0038,+0.0085]\) & Y/n/Y\\
3 & \(-0.086\) \([-0.107,-0.066]\) & \(-0.080\) & \(-0.087\) & \(+0.0001\) & 8.5 & \(+0.0066\) \([+0.0044,+0.0090]\) & Y/n/Y\\
4 & \(-0.074\) \([-0.098,-0.058]\) & \(-0.073\) & \(-0.080\) & \(+0.0002\) & 9.0 & \(+0.0069\) \([+0.0049,+0.0092]\) & n/Y/Y\\
5 & \(-0.092\) \([-0.109,-0.075]\) & \(-0.086\) & \(-0.093\) & \(+0.0000\) & 8.4 & \(+0.0092\) \([+0.0069,+0.0116]\) & Y/n/Y\\
\midrule
Mean & \(-0.085\) & \(-0.080\) & \(-0.087\) & \(+0.0005\) & 8.4--9.0 & \(+0.0072\) & 4/1/5 of 5\\
Strict (no shared class) & \(-0.080\) & \(-0.076\) & \(-0.083\) & \(+0.0016\) & 7.7--8.0 & \(+0.0072\) & 3/1/5 of 5\\
\bottomrule
\end{tabular}

%% file: tab_oulad.tex
\begin{tabular}{@{}llrrrrrrl@{}}
\toprule
Attribute & Split & Observed [95\% CI] & Raw & Recal. & Proj. & MSE ratio & \(\widehat\tau_\Delta\) & Gap error \(\not\ni0\)\\
\midrule
Deprivation (IMD 0--30\%) & 1 & \(-0.045\) \([-0.059,-0.032]\) & \(-0.044\) & \(-0.047\) & \(+0.0005\) & 10.8 & \(+0.0019\) & Y/Y/Y\\
 & 2 & \(-0.052\) \([-0.059,-0.045]\) & \(-0.050\) & \(-0.052\) & \(+0.0006\) & 10.6 & \(+0.0034\) & Y/Y/Y\\
 & 3 & \(-0.049\) \([-0.059,-0.039]\) & \(-0.048\) & \(-0.050\) & \(+0.0021\) & 10.9 & \(+0.0034\) & n/Y/Y\\
 & 4 & \(-0.042\) \([-0.051,-0.033]\) & \(-0.041\) & \(-0.043\) & \(+0.0005\) & 11.0 & \(+0.0029\) & Y/Y/Y\\
 & 5 & \(-0.050\) \([-0.057,-0.043]\) & \(-0.049\) & \(-0.051\) & \(+0.0026\) & 10.9 & \(+0.0054\) & Y/Y/Y\\
\midrule
Deprivation & mean & \(-0.048\) (5/5 \(\not\ni0\)) & \(-0.047\) & \(-0.049\) & \(+0.0013\) & 10.6--11.0 & \(+0.0034\) & 4/5/5 of 5\\
Disability & mean & \(-0.025\) (5/5 \(\not\ni0\)) & \(-0.024\) & \(-0.026\) & \(-0.0001\) & 8.4--8.9 & \(+0.0019\) & 0/4/5 of 5\\
Gender (female) & mean & \(-0.006\) (0/5 \(\not\ni0\)) & \(-0.007\) & \(-0.005\) & \(-0.0040\) & 7.5--9.3 & \(+0.0106\) & 0/1/0 of 5\\
\bottomrule
\end{tabular}

%% file: tab_paired_differences.tex
\begin{tabular}{@{}llrrrrrr@{}}
\toprule
Predictor minus raw EDMD & \(h\) & Split 1 & Split 2 & Split 3 & Split 4 & Split 5 & Mean [range]\\
\midrule
Persistence & 1 & +1.21 \scriptsize[+1.10,\,+1.37] & +1.12 \scriptsize[+0.90,\,+1.35] & +1.09 \scriptsize[+0.91,\,+1.33] & +1.06 \scriptsize[+0.82,\,+1.27] & +1.13 \scriptsize[+0.88,\,+1.35] & +1.12 [+1.06,\,+1.21]\\
Persistence & 25 & +196.87 \scriptsize[+183.41,\,+218.42] & +179.57 \scriptsize[+151.62,\,+204.65] & +179.40 \scriptsize[+152.67,\,+216.78] & +183.60 \scriptsize[+153.61,\,+209.65] & +196.32 \scriptsize[+178.14,\,+213.55] & +187.15 [+179.40,\,+196.87]\\
Affine LS & 1 & +0.12 \scriptsize[+0.09,\,+0.17] & +0.10 \scriptsize[+0.05,\,+0.16] & +0.13 \scriptsize[+0.09,\,+0.18] & +0.06 \scriptsize[-0.02,\,+0.13] & +0.09 \scriptsize[+0.01,\,+0.17] & +0.10 [+0.06,\,+0.13]\\
Affine LS & 25 & -0.24 \scriptsize[-0.73,\,+0.51] & +0.73 \scriptsize[+0.00,\,+1.63] & -0.27 \scriptsize[-0.94,\,+0.95] & +0.74 \scriptsize[+0.06,\,+1.53] & +0.84 \scriptsize[+0.11,\,+1.71] & +0.36 [-0.27,\,+0.84]\\
Gradient boosting & 1 & -0.47 \scriptsize[-0.56,\,-0.37] & -0.34 \scriptsize[-0.46,\,-0.23] & -0.39 \scriptsize[-0.49,\,-0.31] & -0.36 \scriptsize[-0.44,\,-0.28] & -0.43 \scriptsize[-0.54,\,-0.34] & -0.40 [-0.47,\,-0.34]\\
Gradient boosting & 25 & +87.66 \scriptsize[+80.16,\,+100.46] & +70.38 \scriptsize[+54.77,\,+85.05] & +91.08 \scriptsize[+77.22,\,+110.69] & +79.70 \scriptsize[+66.25,\,+90.86] & +95.56 \scriptsize[+84.33,\,+106.68] & +84.88 [+70.38,\,+95.56]\\
Recalibrated EDMD & 1 & -0.03 \scriptsize[-0.05,\,-0.01] & +0.01 \scriptsize[-0.04,\,+0.05] & -0.03 \scriptsize[-0.06,\,+0.01] & +0.03 \scriptsize[-0.02,\,+0.07] & +0.03 \scriptsize[-0.04,\,+0.08] & +0.00 [-0.03,\,+0.03]\\
Recalibrated EDMD & 25 & -0.24 \scriptsize[-0.37,\,-0.12] & -0.10 \scriptsize[-0.46,\,+0.09] & -0.28 \scriptsize[-0.47,\,-0.06] & -0.17 \scriptsize[-0.42,\,-0.01] & -0.13 \scriptsize[-0.49,\,+0.11] & -0.18 [-0.28,\,-0.10]\\
Projected EDMD & 1 & +47.82 \scriptsize[+44.39,\,+52.04] & +48.92 \scriptsize[+44.77,\,+53.65] & +49.23 \scriptsize[+46.44,\,+53.05] & +51.22 \scriptsize[+46.30,\,+57.82] & +53.57 \scriptsize[+48.64,\,+58.97] & +50.15 [+47.82,\,+53.57]\\
Projected EDMD & 25 & +1.35 \scriptsize[+0.86,\,+1.71] & +1.57 \scriptsize[+1.05,\,+2.00] & +1.45 \scriptsize[+0.92,\,+1.89] & +1.45 \scriptsize[+0.88,\,+2.06] & +1.06 \scriptsize[+0.35,\,+1.68] & +1.38 [+1.06,\,+1.57]\\
\midrule
\multicolumn{8}{@{}l}{\emph{Predicted minus observed final gap in \(k_{25}\) (one-step replay)}}\\
Raw EDMD & -- & -8.33 \scriptsize[-17.23,\,-0.26] & -0.71 \scriptsize[-12.74,\,+7.24] & -4.75 \scriptsize[-15.56,\,+7.30] & -6.26 \scriptsize[-14.93,\,-0.82] & -5.04 \scriptsize[-18.10,\,+4.15] & -5.02 [-8.33,\,-0.71]\\
Recalibrated EDMD & -- & -1.56 \scriptsize[-10.23,\,+6.55] & +9.94 \scriptsize[-1.39,\,+17.80] & +1.44 \scriptsize[-9.19,\,+13.88] & +2.59 \scriptsize[-6.33,\,+8.05] & +4.73 \scriptsize[-8.06,\,+13.92] & +3.43 [-1.56,\,+9.94]\\
Projected EDMD & -- & -96.98 \scriptsize[-172.89,\,-41.08] & -21.39 \scriptsize[-124.01,\,+20.06] & -81.91 \scriptsize[-171.38,\,-21.05] & -51.19 \scriptsize[-142.58,\,-6.81] & -31.95 \scriptsize[-132.97,\,+18.93] & -56.69 [-96.98,\,-21.39]\\
\bottomrule
\end{tabular}

%% file: tab_calibration.tex
\begin{tabular}{@{}lllrrrrr@{}}
\toprule
Model & Group & Quantity & Split 1 & Split 2 & Split 3 & Split 4 & Split 5\\
\midrule
Raw EDMD & 0 & CITL \(\times10^3\) & -4.5 \scriptsize[-11.8,\,+1.3] & -1.7 \scriptsize[-12.4,\,+2.0] & -0.7 \scriptsize[-10.6,\,+2.8] & -1.9 \scriptsize[-5.9,\,+0.5] & -1.4 \scriptsize[-13.2,\,+2.5]\\
 &  & Slope & 1.000 \scriptsize[0.978,\,1.023] & 1.008 \scriptsize[0.982,\,1.029] & 1.003 \scriptsize[0.982,\,1.018] & 1.004 \scriptsize[0.984,\,1.022] & 1.003 \scriptsize[0.983,\,1.023]\\
 & 1 & CITL \(\times10^3\) & +5.1 \scriptsize[+2.4,\,+7.9] & +1.4 \scriptsize[-2.0,\,+5.3] & +7.2 \scriptsize[+2.9,\,+10.3] & -4.0 \scriptsize[-7.9,\,+7.5] & -0.7 \scriptsize[-6.4,\,+6.9]\\
 &  & Slope & 0.977 \scriptsize[0.959,\,0.988] & 0.998 \scriptsize[0.977,\,1.010] & 0.979 \scriptsize[0.962,\,0.992] & 1.017 \scriptsize[0.982,\,1.032] & 1.011 \scriptsize[0.976,\,1.024]\\
\midrule
Recalibrated EDMD & 0 & CITL \(\times10^3\) & -0.5 \scriptsize[-7.6,\,+5.1] & +5.7 \scriptsize[-4.2,\,+9.3] & +2.4 \scriptsize[-6.9,\,+5.9] & +2.5 \scriptsize[-1.0,\,+4.6] & +4.3 \scriptsize[-7.0,\,+7.9]\\
 &  & Slope & 0.996 \scriptsize[0.974,\,1.019] & 0.999 \scriptsize[0.973,\,1.020] & 0.997 \scriptsize[0.976,\,1.012] & 0.991 \scriptsize[0.971,\,1.009] & 0.996 \scriptsize[0.976,\,1.016]\\
 & 1 & CITL \(\times10^3\) & +2.3 \scriptsize[-0.5,\,+5.3] & -1.7 \scriptsize[-5.2,\,+2.7] & +4.3 \scriptsize[-0.3,\,+7.6] & -9.3 \scriptsize[-13.2,\,+3.8] & -4.6 \scriptsize[-10.7,\,+3.6]\\
 &  & Slope & 0.984 \scriptsize[0.966,\,0.995] & 1.008 \scriptsize[0.986,\,1.020] & 0.989 \scriptsize[0.972,\,1.002] & 1.040 \scriptsize[1.004,\,1.056] & 1.024 \scriptsize[0.989,\,1.037]\\
\bottomrule
\end{tabular}

%% file: tab_controlled_full_ls.tex
\begin{tabular}{@{}lrrrrrrrrrrr@{}}
\toprule
 & & \multicolumn{2}{c}{Announced (median)} & \multicolumn{2}{c}{Realized (median)} & & & \multicolumn{3}{c}{Coverage of realized total} & \\
\cmidrule(lr){3-4}\cmidrule(lr){5-6}\cmidrule(lr){9-11}
Cell & \(n\) & quad. & total & quad. & total & acc. & transf. & audit & audit, acc. & cond. & cov.\ built, acc.\\
\midrule
Useless, \(a=0.2\) & 2000 & 0.034 & 0.027 & 0.035 & 0.030 & 0.00 & 0.00 & 1.00 & -- & 1.00 & --\\
Useless, \(a=0.2\) & 20000 & 0.002 & 0.002 & 0.002 & 0.002 & 0.00 & 0.00 & 0.98 & -- & 0.99 & --\\
\addlinespace
Useless, \(a=0.7\) & 2000 & 0.004 & 0.003 & 0.004 & 0.003 & 0.00 & 0.00 & 1.00 & -- & 1.00 & --\\
Useless, \(a=0.7\) & 20000 & 0.000 & 0.000 & 0.000 & 0.000 & 0.00 & 0.00 & 1.00 & -- & 1.00 & --\\
\addlinespace
Useless, \(a=1.5\) & 2000 & 0.001 & 0.000 & 0.001 & 0.000 & 0.00 & 0.00 & 1.00 & -- & 0.99 & --\\
Useless, \(a=1.5\) & 20000 & 0.000 & 0.000 & 0.000 & 0.000 & 0.00 & 0.00 & 1.00 & -- & 1.00 & --\\
\addlinespace
Partly aligned, \(a=0.2\) & 2000 & 0.291 & 0.313 & 0.292 & 0.303 & 0.01 & 0.00 & 1.00 & 1.00 & 1.00 & 1.00\\
Partly aligned, \(a=0.2\) & 20000 & 0.248 & 0.249 & 0.248 & 0.247 & 1.00 & 0.72 & 1.00 & 1.00 & 1.00 & 0.89\\
\addlinespace
Partly aligned, \(a=0.7\) & 2000 & 0.273 & 0.274 & 0.275 & 0.277 & 1.00 & 0.80 & 1.00 & 1.00 & 1.00 & 0.91\\
Partly aligned, \(a=0.7\) & 20000 & 0.275 & 0.276 & 0.277 & 0.275 & 1.00 & 1.00 & 1.00 & 1.00 & 1.00 & 0.91\\
\addlinespace
Partly aligned, \(a=1.5\) & 2000 & 0.364 & 0.363 & 0.363 & 0.361 & 1.00 & 0.99 & 1.00 & 1.00 & 1.00 & 0.95\\
Partly aligned, \(a=1.5\) & 20000 & 0.362 & 0.361 & 0.362 & 0.360 & 1.00 & 1.00 & 1.00 & 1.00 & 1.00 & 0.96\\
\addlinespace
Aligned, \(a=0.2\) & 2000 & 1.296 & 1.267 & 1.289 & 1.271 & 0.13 & 0.13 & 1.00 & 1.00 & 1.00 & 0.85\\
Aligned, \(a=0.2\) & 20000 & 1.035 & 1.033 & 1.034 & 1.033 & 1.00 & 1.00 & 0.99 & 0.99 & 0.99 & 0.93\\
\addlinespace
Aligned, \(a=0.7\) & 2000 & 1.127 & 1.126 & 1.125 & 1.122 & 1.00 & 1.00 & 0.96 & 0.96 & 0.99 & 0.89\\
Aligned, \(a=0.7\) & 20000 & 1.111 & 1.110 & 1.105 & 1.108 & 1.00 & 1.00 & 0.94 & 0.94 & 0.95 & 0.92\\
\addlinespace
Aligned, \(a=1.5\) & 2000 & 1.444 & 1.449 & 1.441 & 1.452 & 1.00 & 1.00 & 0.94 & 0.94 & 0.97 & 0.90\\
Aligned, \(a=1.5\) & 20000 & 1.439 & 1.439 & 1.438 & 1.439 & 1.00 & 1.00 & 0.88 & 0.88 & 0.92 & 0.86\\
\addlinespace
Shift: utility lost, \(a=0.7\) & 2000 & 1.132 & 1.133 & 1.137 & 1.134 & 1.00 & 0.04 & 0.99 & 0.99 & 0.97 & 0.88\\
Shift: utility lost, \(a=0.7\) & 20000 & 1.105 & 1.103 & 1.105 & 1.105 & 1.00 & 0.01 & 0.90 & 0.90 & 0.95 & 0.91\\
\addlinespace
Shift: utility gained, \(a=0.7\) & 2000 & 0.003 & 0.001 & 0.003 & 0.002 & 0.00 & 0.00 & 1.00 & -- & 1.00 & --\\
Shift: utility gained, \(a=0.7\) & 20000 & 0.000 & 0.000 & 0.000 & 0.000 & 0.00 & 0.00 & 1.00 & -- & 0.99 & --\\
\addlinespace
Calibrated, misspecified \(s\) & 2000 & 0.996 & 1.014 & 0.997 & 1.002 & 1.00 & 1.00 & 0.96 & 0.96 & 0.97 & 0.91\\
Calibrated, misspecified \(s\) & 20000 & 1.000 & 0.998 & 1.001 & 1.000 & 1.00 & 1.00 & 0.89 & 0.89 & 0.91 & 0.89\\
\addlinespace
Same, correct \(s\) & 2000 & 0.505 & 0.504 & 0.503 & 0.507 & 1.00 & 1.00 & 1.00 & 1.00 & 0.99 & 0.90\\
Same, correct \(s\) & 20000 & 0.500 & 0.498 & 0.501 & 0.500 & 1.00 & 1.00 & 0.96 & 0.96 & 0.98 & 0.87\\
\bottomrule
\end{tabular}

%% file: tab_controlled_full_gb.tex
\begin{tabular}{@{}lrrrrrrrrrrr@{}}
\toprule
 & & \multicolumn{2}{c}{Announced (median)} & \multicolumn{2}{c}{Realized (median)} & & & \multicolumn{3}{c}{Coverage of realized total} & \\
\cmidrule(lr){3-4}\cmidrule(lr){5-6}\cmidrule(lr){9-11}
Cell & \(n\) & quad. & total & quad. & total & acc. & transf. & audit & audit, acc. & cond. & cov.\ built, acc.\\
\midrule
Useless, \(a=0.2\) & 2000 & 0.043 & 0.045 & 0.043 & 0.041 & 0.00 & 0.00 & 1.00 & -- & 1.00 & --\\
Useless, \(a=0.2\) & 20000 & 0.002 & 0.001 & 0.002 & 0.002 & 0.00 & 0.00 & 1.00 & -- & 1.00 & --\\
\addlinespace
Useless, \(a=0.7\) & 2000 & 0.002 & 0.002 & 0.002 & 0.001 & 0.00 & 0.00 & 1.00 & -- & 1.00 & --\\
Useless, \(a=0.7\) & 20000 & 0.000 & 0.000 & 0.000 & 0.000 & 0.00 & 0.00 & 1.00 & -- & 1.00 & --\\
\addlinespace
Useless, \(a=1.5\) & 2000 & 0.001 & 0.000 & 0.001 & 0.000 & 0.02 & 0.00 & 0.94 & 0.00 & 0.96 & 0.00\\
Useless, \(a=1.5\) & 20000 & 0.000 & 0.000 & 0.000 & 0.000 & 0.00 & 0.00 & 0.98 & -- & 0.98 & --\\
\addlinespace
Partly aligned, \(a=0.2\) & 2000 & 0.388 & 0.431 & 0.387 & 0.434 & 0.00 & 0.00 & 1.00 & -- & 1.00 & --\\
Partly aligned, \(a=0.2\) & 20000 & 0.267 & 0.267 & 0.268 & 0.269 & 1.00 & 0.80 & 1.00 & 1.00 & 1.00 & 0.94\\
\addlinespace
Partly aligned, \(a=0.7\) & 2000 & 0.272 & 0.281 & 0.275 & 0.281 & 1.00 & 0.82 & 1.00 & 1.00 & 1.00 & 0.94\\
Partly aligned, \(a=0.7\) & 20000 & 0.269 & 0.280 & 0.269 & 0.275 & 1.00 & 1.00 & 1.00 & 1.00 & 1.00 & 0.82\\
\addlinespace
Partly aligned, \(a=1.5\) & 2000 & 0.352 & 0.350 & 0.347 & 0.354 & 1.00 & 0.98 & 1.00 & 1.00 & 1.00 & 0.76\\
Partly aligned, \(a=1.5\) & 20000 & 0.348 & 0.357 & 0.348 & 0.361 & 1.00 & 1.00 & 0.98 & 0.98 & 1.00 & 0.80\\
\addlinespace
Aligned, \(a=0.2\) & 2000 & 1.237 & 1.237 & 1.228 & 1.267 & 0.24 & 0.18 & 1.00 & 1.00 & 1.00 & 0.83\\
Aligned, \(a=0.2\) & 20000 & 0.999 & 1.023 & 1.000 & 1.028 & 1.00 & 1.00 & 1.00 & 1.00 & 1.00 & 0.98\\
\addlinespace
Aligned, \(a=0.7\) & 2000 & 1.106 & 1.116 & 1.110 & 1.135 & 1.00 & 1.00 & 0.96 & 0.96 & 0.98 & 0.88\\
Aligned, \(a=0.7\) & 20000 & 1.071 & 1.108 & 1.073 & 1.105 & 1.00 & 1.00 & 0.98 & 0.98 & 1.00 & 0.30\\
\addlinespace
Aligned, \(a=1.5\) & 2000 & 1.441 & 1.429 & 1.434 & 1.434 & 1.00 & 1.00 & 0.96 & 0.96 & 0.96 & 0.82\\
Aligned, \(a=1.5\) & 20000 & 1.413 & 1.443 & 1.415 & 1.441 & 1.00 & 1.00 & 0.90 & 0.90 & 0.92 & 0.50\\
\addlinespace
Shift: utility lost, \(a=0.7\) & 2000 & 1.110 & 1.115 & 1.101 & 1.122 & 1.00 & 0.04 & 0.98 & 0.98 & 1.00 & 0.98\\
Shift: utility lost, \(a=0.7\) & 20000 & 1.069 & 1.105 & 1.068 & 1.105 & 1.00 & 0.02 & 0.96 & 0.96 & 0.96 & 0.26\\
\addlinespace
Shift: utility gained, \(a=0.7\) & 2000 & 0.004 & 0.003 & 0.004 & 0.003 & 0.00 & 0.00 & 1.00 & -- & 1.00 & --\\
Shift: utility gained, \(a=0.7\) & 20000 & 0.000 & 0.000 & 0.000 & 0.000 & 0.00 & 0.00 & 0.98 & -- & 0.98 & --\\
\addlinespace
Calibrated, misspecified \(s\) & 2000 & 1.008 & 1.000 & 1.009 & 0.999 & 1.00 & 1.00 & 0.88 & 0.88 & 0.92 & 0.24\\
Calibrated, misspecified \(s\) & 20000 & 0.995 & 1.004 & 0.995 & 0.996 & 1.00 & 1.00 & 0.96 & 0.96 & 0.96 & 0.04\\
\addlinespace
Same, correct \(s\) & 2000 & 0.507 & 0.496 & 0.503 & 0.496 & 1.00 & 1.00 & 1.00 & 1.00 & 1.00 & 0.90\\
Same, correct \(s\) & 20000 & 0.500 & 0.506 & 0.498 & 0.499 & 1.00 & 1.00 & 0.98 & 0.98 & 1.00 & 0.78\\
\bottomrule
\end{tabular}

%% file: tab_controlled_stability.tex
\begin{tabular}{@{}llrrrrrrrr@{}}
\toprule
 & & \multicolumn{4}{c}{Coverage of realized total} & \multicolumn{4}{c}{End-point change vs.\ \(B=400\)}\\
\cmidrule(lr){3-6}\cmidrule(l){7-10}
Cell & Learner & \(B=40\) & 60 & 200 & 400 & 40 & 60 & 200 & 400\\
\midrule
Partly aligned, \(a=0.7\) & LS & 1.00 & 1.00 & 1.00 & 1.00 & 0.10 & 0.06 & 0.03 & 0.00\\
Partly aligned, \(a=0.7\) & GB & 1.00 & 1.00 & 1.00 & 1.00 & 0.10 & 0.08 & 0.03 & 0.00\\
Aligned, \(a=0.2\) & LS & 1.00 & 1.00 & 1.00 & 1.00 & 0.44 & 0.40 & 0.16 & 0.00\\
Aligned, \(a=0.2\) & GB & 1.00 & 1.00 & 1.00 & 1.00 & 0.46 & 0.29 & 0.13 & 0.00\\
Aligned, \(a=1.5\) & LS & 0.90 & 0.90 & 0.90 & 0.93 & 0.12 & 0.06 & 0.03 & 0.00\\
Aligned, \(a=1.5\) & GB & 0.90 & 0.93 & 0.93 & 0.97 & 0.11 & 0.09 & 0.03 & 0.00\\
Calibrated, misspecified \(s\) & LS & 0.93 & 0.97 & 0.97 & 0.97 & 0.09 & 0.07 & 0.04 & 0.00\\
Calibrated, misspecified \(s\) & GB & 0.87 & 0.90 & 0.93 & 0.90 & 0.10 & 0.07 & 0.04 & 0.00\\
\bottomrule
\end{tabular}

%% file: tab_controlled_robust.tex
\textbf{(a) Coverage of the realized total cost among accepted replicates}\\[2pt]
\begin{tabular}{@{}llrrrrrr@{}}
\toprule
 & & \multicolumn{3}{c}{\(n=2{,}000\)} & \multicolumn{3}{c}{\(n=20{,}000\)}\\
\cmidrule(lr){3-5}\cmidrule(l){6-8}
 & & \multicolumn{2}{c}{percentile} & basic & \multicolumn{2}{c}{percentile} & basic\\
Group signal & \(a\) & \(B=60\) & 1{,}000 & 1{,}000 & \(B=60\) & 1{,}000 & 1{,}000\\
\midrule
Partly aligned & 0.2 & -- & -- & -- & 1.00 & 1.00 & 1.00\\
Partly aligned & 0.7 & 1.00 & 1.00 & 1.00 & 1.00 & 1.00 & 1.00\\
Partly aligned & 1.5 & 1.00 & 1.00 & 1.00 & 1.00 & 1.00 & 1.00\\
\addlinespace
Aligned & 0.2 & 1.00 & 1.00 & 1.00 & 0.99 & 1.00 & 1.00\\
Aligned & 0.7 & 0.96 & 0.96 & 0.99 & 0.94 & 0.95 & 0.95\\
Aligned & 1.5 & 0.94 & 0.96 & 0.98 & 0.88 & 0.92 & 0.93\\
\addlinespace
Shift: utility lost & 0.7 & 0.99 & 0.99 & 0.98 & 0.90 & 0.95 & 0.97\\
Calibrated, misspecified \(s\) & -- & 0.96 & 0.98 & 0.98 & 0.89 & 0.91 & 0.92\\
Same, correct \(s\) & -- & 1.00 & 1.00 & 1.00 & 0.96 & 0.97 & 0.97\\
\bottomrule
\end{tabular}

\bigskip

\textbf{(b) Other thresholds: share accepted (D3) and share that transfers (T2)}\\[2pt]
\begin{tabular}{@{}llrrrrrrr@{}}
\toprule
 & & \multicolumn{4}{c}{Accepted} & \multicolumn{3}{c}{Transfers}\\
 & & \multicolumn{4}{c}{D3, \(n=2{,}000\)} & \multicolumn{3}{c}{T2, \(n=20{,}000\)}\\
\cmidrule(lr){3-6}\cmidrule(l){7-9}
Group signal & \(a\) & 0.25 & 0.5 & 1 & none & 0.1 & 0.2 & 0.3\\
\midrule
Useless & 0.2 & 0.00 & 0.00 & 0.00 & 0.07 & 0.00 & 0.00 & 0.00\\
Useless & 0.7 & 0.00 & 0.00 & 0.07 & 0.07 & 0.00 & 0.00 & 0.00\\
Useless & 1.5 & 0.00 & 0.00 & 0.07 & 0.07 & 0.00 & 0.00 & 0.00\\
\addlinespace
Partly aligned & 0.2 & 0.00 & 0.00 & 0.03 & 0.53 & 0.42 & 0.72 & 0.90\\
Partly aligned & 0.7 & 0.56 & 1.00 & 1.00 & 1.00 & 0.93 & 1.00 & 1.00\\
Partly aligned & 1.5 & 1.00 & 1.00 & 1.00 & 1.00 & 1.00 & 1.00 & 1.00\\
\addlinespace
Aligned & 0.2 & 0.00 & 0.05 & 0.13 & 0.89 & 1.00 & 1.00 & 1.00\\
Aligned & 0.7 & 1.00 & 1.00 & 1.00 & 1.00 & 1.00 & 1.00 & 1.00\\
Aligned & 1.5 & 1.00 & 1.00 & 1.00 & 1.00 & 1.00 & 1.00 & 1.00\\
\addlinespace
Shift: utility lost & 0.7 & 1.00 & 1.00 & 1.00 & 1.00 & 0.00 & 0.01 & 0.01\\
Shift: utility gained & 0.7 & 0.00 & 0.00 & 0.06 & 0.06 & 0.00 & 0.00 & 0.00\\
\addlinespace
Calibrated, misspecified \(s\) & -- & 1.00 & 1.00 & 1.00 & 1.00 & 1.00 & 1.00 & 1.00\\
Same, correct \(s\) & -- & 1.00 & 1.00 & 1.00 & 1.00 & 1.00 & 1.00 & 1.00\\
\bottomrule
\end{tabular}

%% file: tab_next_item_grid.tex
\begin{tabular}{@{}llrrrrl@{}}
\toprule
Data & Split & Hidden & Learning rate & Dropout & Epochs & Validation log-loss (best; range over grid)\\
\midrule
ASSISTments & 1 & 32 & 0.003 & 0.2 & 23 & 0.5558 (0.5558--0.5862)\\
 & 2 & 128 & 0.003 & 0.2 & 16 & 0.5998 (0.5998--0.6407)\\
 & 3 & 32 & 0.003 & 0.2 & 15 & 0.5740 (0.5740--0.6095)\\
 & 4 & 32 & 0.003 & 0.2 & 27 & 0.5713 (0.5713--0.6046)\\
 & 5 & 32 & 0.003 & 0.2 & 21 & 0.5624 (0.5624--0.5977)\\
Eedi & 1 & 64 & 0.003 & 0.2 & 21 & 0.5850 (0.5850--0.6012)\\
 & 2 & 64 & 0.003 & 0.2 & 17 & 0.6014 (0.6014--0.6163)\\
 & 3 & 32 & 0.003 & 0.2 & 20 & 0.5865 (0.5865--0.6015)\\
 & 4 & 32 & 0.003 & 0.2 & 22 & 0.5974 (0.5974--0.6130)\\
 & 5 & 32 & 0.003 & 0.2 & 23 & 0.6002 (0.6002--0.6153)\\
\bottomrule
\end{tabular}

%% file: tab_next_item_audit_eedi.tex
\begin{tabular}{@{}llrrrrrr@{}}
\toprule
 & & \multicolumn{2}{c}{Announced (training + validation)} & \multicolumn{2}{c}{Measured (test)} & \multicolumn{2}{c}{Measured \(-\) announced}\\
\cmidrule(lr){3-4}\cmidrule(lr){5-6}\cmidrule(l){7-8}
Split & & Quadratic \(\widehat\pi_s\) & Total & \(\E d^2\) & Total & Quadratic & Total\\
\midrule
1 & estimate & 0.0201 & 0.0227 & 0.0211 & 0.0232 & \(+0.0010\) & \(+0.0005\)\\
 & 95\% & [0.0160, 0.0322] & [0.0181, 0.0351] & [0.0200, 0.0221] & [0.0212, 0.0252] & [-0.0006, 0.0048] & [-0.0027, 0.0065]\\
\addlinespace
2 & estimate & 0.0252 & 0.0251 & 0.0267 & 0.0288 & \(+0.0015\) & \(+0.0037\)\\
 & 95\% & [0.0202, 0.0427] & [0.0203, 0.0427] & [0.0254, 0.0286] & [0.0262, 0.0325] & [-0.0010, 0.0057] & [0.0001, 0.0098]\\
\addlinespace
3 & estimate & 0.0179 & 0.0204 & 0.0175 & 0.0187 & \(-0.0004\) & \(-0.0017\)\\
 & 95\% & [0.0129, 0.0285] & [0.0146, 0.0309] & [0.0165, 0.0188] & [0.0169, 0.0208] & [-0.0022, 0.0017] & [-0.0043, 0.0012]\\
\addlinespace
4 & estimate & 0.0188 & 0.0197 & 0.0177 & 0.0194 & \(-0.0011\) & \(-0.0003\)\\
 & 95\% & [0.0144, 0.0284] & [0.0147, 0.0296] & [0.0163, 0.0187] & [0.0155, 0.0225] & [-0.0032, 0.0007] & [-0.0047, 0.0037]\\
\addlinespace
5 & estimate & 0.0175 & 0.0193 & 0.0162 & 0.0167 & \(-0.0013\) & \(-0.0026\)\\
 & 95\% & [0.0135, 0.0272] & [0.0152, 0.0287] & [0.0154, 0.0172] & [0.0151, 0.0188] & [-0.0031, 0.0005] & [-0.0057, 0.0007]\\
\bottomrule
\end{tabular}

%% file: tab_next_item_audit_assist.tex
\begin{tabular}{@{}llrrrrrr@{}}
\toprule
 & & \multicolumn{2}{c}{Announced (training + validation)} & \multicolumn{2}{c}{Measured (test)} & \multicolumn{2}{c}{Measured \(-\) announced}\\
\cmidrule(lr){3-4}\cmidrule(lr){5-6}\cmidrule(l){7-8}
Split & & Quadratic \(\widehat\pi_s\) & Total & \(\E d^2\) & Total & Quadratic & Total\\
\midrule
1 & estimate & 0.0484 & 0.0380 & 0.0535 & 0.0532 & \(+0.0051\) & \(+0.0152\)\\
 & 95\% & [0.0000, 28.6582] & [-0.0002, 28.4034] & [0.0444, 0.0621] & [0.0429, 0.0624] & [-0.1969, 3.0692] & [-0.1439, 2.8946]\\
\addlinespace
2 & estimate & 0.0236 & 0.0245 & 0.0244 & 0.0258 & \(+0.0008\) & \(+0.0014\)\\
 & 95\% & [0.0001, 26.1249] & [-0.0001, 26.1235] & [0.0194, 0.0284] & [0.0200, 0.0314] & [-0.1266, 2.1562] & [-0.1363, 2.1172]\\
\addlinespace
3 & estimate & 0.5116 & 0.5119 & 0.4850 & 0.4848 & \(-0.0266\) & \(-0.0271\)\\
 & 95\% & [0.0146, 59.1386] & [0.0146, 59.1411] & [0.3708, 0.6335] & [0.3715, 0.6260] & [-5.2012, 0.7046] & [-4.7629, 0.6729]\\
\addlinespace
4 & estimate & 0.0198 & 0.0253 & 0.0258 & 0.0270 & \(+0.0060\) & \(+0.0017\)\\
 & 95\% & [0.0005, 10.3030] & [0.0005, 10.1257] & [0.0191, 0.0315] & [0.0212, 0.0318] & [-0.0082, 0.7777] & [-0.0170, 0.7507]\\
\addlinespace
5 & estimate & 0.0026 & 0.0023 & 0.0028 & 0.0021 & \(+0.0002\) & \(-0.0002\)\\
 & 95\% & [0.0000, 0.7780] & [-0.0001, 0.7540] & [0.0022, 0.0032] & [0.0005, 0.0036] & [-0.0292, 0.0677] & [-0.0227, 0.0765]\\
\bottomrule
\end{tabular}

%% file: tab_next_item_rel.tex
\begin{tabular}{@{}llrrrrrrrl@{}}
\toprule
 & & \multicolumn{3}{c}{First version (training units)} & \multicolumn{2}{c}{Platt recalibrated} & \multicolumn{3}{c}{Three-stage audit}\\
\cmidrule(lr){3-5}\cmidrule(lr){6-7}\cmidrule(l){8-10}
Data & Split & Price & \(\widehat\tau\) & slope & \(\widehat\tau\) & slope & Total announced & \(\widehat\tau\) & Accepted / transfers (failed)\\
\midrule
ASSISTments & 1 & 0.0025 & \(+0.80\) & 0.32 & \(+0.88\) & 3.38 & 0.0380 & \(-0.34\) & no / -- (D1,D2,D3)\\
 & 2 & 0.0075 & \(+0.82\) & -0.15 & \(+0.78\) & -0.84 & 0.0245 & \(+0.59\) & no / -- (D2,D3)\\
 & 3 & 0.0044 & \(+1.15\) & -0.11 & \(+1.14\) & -0.98 & 0.5119 & \(+3.05\) & no / -- (D2,D3)\\
 & 4 & 0.0052 & \(+4.98\) & -0.11 & \(+4.77\) & -0.27 & 0.0253 & \(+11.12\) & no / -- (D1,D2,D3)\\
 & 5 & 0.0072 & \(+0.40\) & 0.18 & \(+0.16\) & 0.76 & 0.0023 & \(+0.59\) & no / -- (D1,D2,D3)\\
\addlinespace
Eedi & 1 & 0.0175 & \(+0.07\) & 0.97 & \(+0.07\) & 1.29 & 0.0227 & \(-0.10\) & yes / yes\\
 & 2 & 0.0178 & \(+0.09\) & 0.76 & \(+0.10\) & 1.18 & 0.0251 & \(-0.25\) & yes / no (T2)\\
 & 3 & 0.0177 & \(+0.19\) & 0.73 & \(+0.20\) & 0.92 & 0.0204 & \(+0.12\) & yes / yes\\
 & 4 & 0.0172 & \(+0.10\) & 0.75 & \(+0.11\) & 1.01 & 0.0197 & \(+0.03\) & yes / yes\\
 & 5 & 0.0165 & \(+0.25\) & 1.06 & \(+0.25\) & 1.05 & 0.0193 & \(+0.17\) & yes / yes\\
\bottomrule
\end{tabular}

%% file: tab_next_item_v4.tex
\textbf{(a) Three sets of units against four}\\[2pt]
\begin{tabular}{@{}llrrrrrrrrll@{}}
\toprule
 & & \multicolumn{2}{c}{AUC of DKT} & \multicolumn{2}{c}{Announced total} & \multicolumn{2}{c}{Interval width} & \multicolumn{2}{c}{Measured (test)} & \multicolumn{2}{c}{Decision}\\
\cmidrule(lr){3-4}\cmidrule(lr){5-6}\cmidrule(lr){7-8}\cmidrule(lr){9-10}\cmidrule(l){11-12}
Data & Split & three & four & three & four & three & four & three & four & three & four\\
\midrule
Eedi & 1 & 0.718 & 0.719 & 0.0227 & 0.0209 & 0.0171 & 0.0217 & 0.0232 & 0.0198 & tr. & tr.\\
 & 2 & 0.720 & 0.720 & 0.0251 & 0.0228 & 0.0224 & 0.0460 & 0.0288 & 0.0269 & acc. & acc.\\
 & 3 & 0.720 & 0.720 & 0.0204 & 0.0176 & 0.0162 & 0.0291 & 0.0187 & 0.0172 & tr. & tr.\\
 & 4 & 0.723 & 0.723 & 0.0197 & 0.0210 & 0.0149 & 0.0272 & 0.0194 & 0.0210 & tr. & tr.\\
 & 5 & 0.715 & 0.715 & 0.0193 & 0.0228 & 0.0136 & 0.0334 & 0.0167 & 0.0192 & tr. & tr.\\
\addlinespace
ASSISTments & 1 & 0.682 & 0.675 & 0.0380 & 0.0032 & 28.4036 & 5.5115 & 0.0532 & 0.0021 & abst. & abst.\\
 & 2 & 0.661 & 0.653 & 0.0245 & 0.0001 & 26.1236 & 1.0490 & 0.0258 & 0.0017 & abst. & abst.\\
 & 3 & 0.677 & 0.682 & 0.5119 & 0.3026 & 59.1265 & 91.5272 & 0.4848 & 0.2631 & abst. & abst.\\
 & 4 & 0.685 & 0.683 & 0.0253 & 0.0172 & 10.1252 & 4.1038 & 0.0270 & 0.0266 & abst. & abst.\\
 & 5 & 0.677 & 0.678 & 0.0023 & 0.0024 & 0.7540 & 3.4933 & 0.0021 & 0.0051 & abst. & abst.\\
\bottomrule
\end{tabular}

\bigskip

\textbf{(b) The output correction on test units, unclipped and clipped to \([0,1]\)}\\[2pt]
\begin{tabular}{@{}llrrrrrc@{}}
\toprule
 & & Outside & \multicolumn{2}{c}{Extra Brier score} & \multicolumn{2}{c}{\(\widehat\tau\)} & Clipped cost in\\
\cmidrule(lr){4-5}\cmidrule(lr){6-7}
Data & Split & \([0,1]\) & unclipped & clipped & unclipped & clipped & announced interval\\
\midrule
Eedi & 1 & 1.0\% & 0.0198 & 0.0195 & \(-0.02\) & \(-0.03\) & yes\\
 & 2 & 0.7\% & 0.0269 & 0.0264 & \(-0.21\) & \(-0.20\) & yes\\
 & 3 & 0.3\% & 0.0172 & 0.0171 & \(+0.17\) & \(+0.17\) & yes\\
 & 4 & 0.3\% & 0.0210 & 0.0209 & \(-0.01\) & \(-0.01\) & yes\\
 & 5 & 0.6\% & 0.0192 & 0.0189 & \(+0.09\) & \(+0.10\) & yes\\
\addlinespace
ASSISTments & 1 & 1.0\% & 0.0021 & 0.0021 & \(+1.31\) & \(+1.31\) & yes\\
 & 2 & 0.6\% & 0.0017 & 0.0016 & \(+0.95\) & \(+0.94\) & yes\\
 & 3 & 25.1\% & 0.2631 & 0.1618 & \(+1.06\) & \(+1.52\) & yes\\
 & 4 & 5.5\% & 0.0266 & 0.0241 & \(-1.61\) & \(-1.85\) & yes\\
 & 5 & 2.8\% & 0.0051 & 0.0048 & \(+1.18\) & \(+1.15\) & yes\\
\bottomrule
\end{tabular}

%% file: tab_next_item_long.tex
\begin{tabular}{@{}lrrrrrl@{}}
\toprule
History & DKT AUC & Removal \(\Delta\)AUC & Removal \(\Delta\)Brier & Price & Measured & Interpretable\\
\midrule
25 items & 0.711 & \(-0.064\) & \(+0.0189\) & 0.0236 & 0.0192 & 1/2\\
50 items & 0.740 & \(-0.077\) & \(+0.0243\) & 0.0275 & 0.0245 & 1/2\\
100 items & 0.746 & \(-0.079\) & \(+0.0257\) & 0.0259 & 0.0258 & 2/2\\
\bottomrule
\end{tabular}

%% file: tab_next_item_long_v2.tex
\begin{tabular}{@{}lrrrrrr@{}}
\toprule
History & Announced total & \(\widehat\tau\) & Measured total & In announced interval & Accepted & Gap reduction transfers\\
\midrule
25 items & 0.0313 & \(-0.05\) & 0.0307 & 2/2 & 2/2 & 2/2\\
50 items & 0.0347 & \(-0.05\) & 0.0344 & 2/2 & 2/2 & 2/2\\
100 items & 0.0344 & \(-0.06\) & 0.0351 & 2/2 & 2/2 & 2/2\\
\bottomrule
\end{tabular}

%% file: journal_edmd_results.tex
\begin{tikzpicture}[font=\small]
\definecolor{journalgray}{HTML}{555555}
\definecolor{journalorange}{HTML}{D55E00}
\definecolor{journalblue}{HTML}{0072B2}
\begin{groupplot}[
group style={group size=2 by 1,horizontal sep=1.6cm},
width=7.4cm,height=6.1cm,
xlabel={Turn},xmin=0,xmax=25,xtick={0,5,10,15,20,25},
axis lines=left,grid=major,grid style={black!12},
tick align=outside,scaled ticks=false,
legend style={draw=none,fill=white,font=\footnotesize},
legend cell align=left]
\nextgroupplot[title={(a) Realized group contrast},ylabel={$\delta_t$ (group 1 minus group 0)},legend pos=north east]
\addplot+[mark=none,line width=1pt,journalgray,densely dotted] coordinates {
(0,0)
(1,-0.054938751012795613)
(2,-0.099789368172897339)
(3,-0.14341020124658194)
(4,-0.17919756154135633)
(5,-0.21281552505461959)
(6,-0.24183989609173703)
(7,-0.27109255424734857)
(8,-0.2981351339032205)
(9,-0.32230819165780156)
(10,-0.33624130756065096)
(11,-0.35194197451819575)
(12,-0.36774914645867091)
(13,-0.38076122223100711)
(14,-0.39186084064305193)
(15,-0.4062668621628937)
(16,-0.41567180549044225)
(17,-0.42497849936896875)
(18,-0.43087157756852906)
(19,-0.44428000287414565)
(20,-0.4426140878917072)
(21,-0.44800169962747038)
(22,-0.45432130708543422)
(23,-0.45712979607032578)
(24,-0.46051133824421381)
(25,-0.46854434958263291)
};
\addlegendentry{No control}
\addplot+[mark=none,line width=1pt,journalorange,dashed] coordinates {
(0,0)
(1,-0.054807511748179834)
(2,-0.099502649723951286)
(3,-0.1429633916823656)
(4,-0.17852607879996482)
(5,-0.21179297784992912)
(6,-0.24041911807968752)
(7,-0.26918016109848264)
(8,-0.29563517640179438)
(9,-0.31924682581049257)
(10,-0.3325547897449036)
(11,-0.34764723843895945)
(12,-0.36281703169368335)
(13,-0.37508982860874163)
(14,-0.38547050695682283)
(15,-0.39923473000367082)
(16,-0.40806331172591948)
(17,-0.41680247390223074)
(18,-0.42214159558381459)
(19,-0.434969397647684)
(20,-0.43277033502015938)
(21,-0.43767825294340845)
(22,-0.44352704440932261)
(23,-0.44593852673100337)
(24,-0.44893730423581335)
(25,-0.45663286561830013)
};
\addlegendentry{Projected MPC}
\addplot+[mark=none,line width=1pt,journalblue,solid] coordinates {
(0,0)
(1,-0.0551480095767675)
(2,-0.096785399677376358)
(3,-0.13496985828019192)
(4,-0.16369214086173417)
(5,-0.18935359264569537)
(6,-0.20972189617994164)
(7,-0.23012826004241571)
(8,-0.24815865640127155)
(9,-0.26321792288332735)
(10,-0.26815433826953428)
(11,-0.27546786336022433)
(12,-0.28334902961753766)
(13,-0.28891646155474626)
(14,-0.29304248687334949)
(15,-0.30092999527920805)
(16,-0.30408617044003083)
(17,-0.30766898440969881)
(18,-0.30830760397630319)
(19,-0.31707093490565752)
(20,-0.31084748580556426)
(21,-0.31255729563398299)
(22,-0.31558879530559997)
(23,-0.3152988418767414)
(24,-0.31601297762734631)
(25,-0.32155266349549677)
};
\addlegendentry{Raw MPC}
\nextgroupplot[title={(b) Mean knowledge},ylabel={$\overline{k}_t$}]
\addplot+[mark=none,line width=1pt,journalgray,densely dotted] coordinates {
(0,0.29825448449473074)
(1,0.28009700122285658)
(2,0.25819309413316088)
(3,0.23268221989442292)
(4,0.21412694250710823)
(5,0.19187969201674682)
(6,0.17700288370390596)
(7,0.16244458740070272)
(8,0.15349331143372244)
(9,0.14379858447501814)
(10,0.14183188461788382)
(11,0.14096717698669123)
(12,0.13611699272410813)
(13,0.13485820970305148)
(14,0.1332298808564607)
(15,0.13192678052183773)
(16,0.12961237890184374)
(17,0.12597955699254235)
(18,0.12525097286830683)
(19,0.12321529259594395)
(20,0.12139885462952625)
(21,0.12047468116718313)
(22,0.12050863935778608)
(23,0.12157023164762568)
(24,0.12038194706701671)
(25,0.12011416094375271)
};
\addplot+[mark=none,line width=1pt,journalorange,dashed] coordinates {
(0,0.29825448449473074)
(1,0.33374953289758846)
(2,0.36235522687822874)
(3,0.3831201015223305)
(4,0.40678051748798655)
(5,0.4210726074750889)
(6,0.43807181086360469)
(7,0.4514171395003217)
(8,0.4656262827546887)
(9,0.47489178387901365)
(10,0.48812021517585275)
(11,0.49822090347056197)
(12,0.50269639555609014)
(13,0.51059200987266862)
(14,0.5171221160878211)
(15,0.52159546733862805)
(16,0.52692276027886265)
(17,0.53224837707642658)
(18,0.53826857011902374)
(19,0.54130933190437192)
(20,0.54357994059277281)
(21,0.54778489795353402)
(22,0.55122612497967238)
(23,0.55817597172097333)
(24,0.55526452145504668)
(25,0.55659891404260964)
};
\addplot+[mark=none,line width=1pt,journalblue,solid] coordinates {
(0,0.29825448449473074)
(1,0.32569552936614782)
(2,0.34891407262887014)
(3,0.36757013888681667)
(4,0.3916171484554038)
(5,0.40816230870437914)
(6,0.4288568976465521)
(7,0.44763358487410859)
(8,0.46760693571842998)
(9,0.4831496703520971)
(10,0.50271434529991632)
(11,0.51914000754272749)
(12,0.53059871035860318)
(13,0.54586483740164859)
(14,0.55952127174567579)
(15,0.57104747196573957)
(16,0.58360472196182533)
(17,0.59599693929480169)
(18,0.60869350757762752)
(19,0.61884973843638746)
(20,0.62808712022906343)
(21,0.63951616965030034)
(22,0.64960583827942409)
(23,0.66233730171577931)
(24,0.66485359771490704)
(25,0.67082080608255079)
};
\end{groupplot}
\end{tikzpicture}

%% file: journal_paired_effects.tex
\begin{tikzpicture}[font=\small]
\definecolor{journalgray}{HTML}{555555}
\definecolor{journalorange}{HTML}{D55E00}
\definecolor{journalblue}{HTML}{0072B2}
\definecolor{journalgreen}{HTML}{009E73}
\begin{groupplot}[
group style={group size=2 by 1,horizontal sep=1.65cm,vertical sep=2.3cm},
width=7.4cm,height=6.1cm,
axis lines=left,grid=major,grid style={black!12},tick align=outside,scaled ticks=false,
legend style={draw=none,fill=white,font=\scriptsize},legend cell align=left]
\nextgroupplot[ytick={-0.145,-0.140,-0.135,-0.130},ymin=-0.145,ymax=-0.128,yticklabel style={/pgf/number format/fixed,/pgf/number format/precision=3},title={(a) Effect on disparity contrast},ylabel={$\tau_\delta$},xlabel={Replicate / summary},xmin=-0.5,xmax=11.8,xtick={0,1,2,3,4,5,6,7,8,9,11},xticklabels={0,1,2,3,4,5,6,7,8,9,Mean}]
\addplot[only marks,mark=*,journalorange,mark options={draw=journalorange,fill=journalorange},mark size=2pt] coordinates {
(0,-0.13508020212280342)
(1,-0.13074152813431467)
(2,-0.13838244850960568)
(3,-0.14137241945442908)
(4,-0.13605808149312137)
(5,-0.13366521552417254)
(6,-0.13837917263649924)
(7,-0.13263824303121086)
(8,-0.14227956849702733)
(9,-0.13567835347494833)
};
\addplot[only marks,mark=diamond*,journalblue,mark options={draw=journalblue,fill=journalblue},mark size=3pt,error bars/.cd,y dir=both,y explicit] coordinates {
(11,-0.13642752328781327) +- (0,0.0026463011255004579)};
\nextgroupplot[title={(b) Effect on mean knowledge},ylabel={$\tau_k$},xlabel={Replicate / summary},xmin=-0.5,xmax=11.8,xtick={0,1,2,3,4,5,6,7,8,9,11},xticklabels={0,1,2,3,4,5,6,7,8,9,Mean}]
\addplot[only marks,mark=*,journalorange,mark options={draw=journalorange,fill=journalorange},mark size=2pt] coordinates {
(0,-0.11422189203994082)
(1,-0.16119826417187333)
(2,-0.16244394397158957)
(3,-0.084020933112200935)
(4,-0.1595474788349065)
(5,-0.20442037793350237)
(6,-0.18078997716712852)
(7,-0.15674980085196211)
(8,-0.21296730415388343)
(9,-0.16288514536359699)
};
\addplot[only marks,mark=diamond*,journalblue,mark options={draw=journalblue,fill=journalblue},mark size=3pt,error bars/.cd,y dir=both,y explicit] coordinates {
(11,-0.15992451176005845) +- (0,0.027298182162712622)};
\end{groupplot}
\end{tikzpicture}

%% file: journal_paired_summary.tex
\begin{tabular}{@{}lrrr@{}}
\toprule
Projected minus raw MPC & Mean effect & Sample SD & 95\% interval for mean \\
\midrule
Disparity contrast $\delta$ & $-0.136428$ & $0.003699$ & $[-0.139074,-0.133781]$ \\
Pooled disparity $\overline D_T$ & $-0.031113$ & $0.007852$ & $[-0.036730,-0.025495]$ \\
Pooled knowledge $\overline k_T$ & $-0.159925$ & $0.038160$ & $[-0.187223,-0.132626]$ \\
Knowledge, group 0 & $+0.184695$ & $0.038688$ & $[+0.157019,+0.212371]$ \\
Knowledge, group 1 & $-0.504544$ & $0.039338$ & $[-0.532685,-0.476403]$ \\
\bottomrule
\end{tabular}

%% file: tab_equal_budget.tex
\begin{tabular}{@{}llrrrrr@{}}
\toprule
 & Arm & \(\Delta|\delta_T|\) & \(\Delta\overline k_{T,0}\) & \(\Delta\overline k_{T,1}\) & \(\mathcal U_1\) ratio & \(\mathcal U_2\) ratio\\
\midrule
\(\gamma=0.5\) & Floor-constrained MPC (\(c=2\)) & \(-0.122\) \([-0.124,-0.120]\) & \(-0.000\) \([-0.000,-0.000]\) & \(+0.277\) \([+0.272,+0.283]\) & 1.61 & 3.09\\
 & Targeted top-up, equal \(\mathcal U_1\) & \(-0.126\) \([-0.129,-0.124]\) & \(+0.000\) \([+0.000,+0.000]\) & \(+0.278\) \([+0.272,+0.284]\) & 1.61 & 3.08\\
 & Targeted top-up, equal \(\mathcal U_2\) & \(-0.127\) \([-0.129,-0.125]\) & \(+0.000\) \([+0.000,+0.000]\) & \(+0.278\) \([+0.272,+0.284]\) & 1.61 & 3.09\\
 & Uniform increase, equal \(\mathcal U_1\) & \(-0.043\) \([-0.044,-0.042]\) & \(+0.196\) \([+0.193,+0.200]\) & \(+0.265\) \([+0.260,+0.269]\) & 1.61 & 2.60\\
\addlinespace
\(\gamma=1\) & Floor-constrained MPC (\(c=2\)) & \(-0.157\) \([-0.160,-0.154]\) & \(-0.000\) \([-0.000,-0.000]\) & \(+0.213\) \([+0.206,+0.221]\) & 1.61 & 3.02\\
 & Targeted top-up, equal \(\mathcal U_1\) & \(-0.155\) \([-0.158,-0.153]\) & \(+0.000\) \([+0.000,+0.000]\) & \(+0.213\) \([+0.206,+0.221]\) & 1.61 & 3.02\\
 & Targeted top-up, equal \(\mathcal U_2\) & \(-0.156\) \([-0.158,-0.153]\) & \(+0.000\) \([+0.000,+0.000]\) & \(+0.213\) \([+0.206,+0.221]\) & 1.61 & 3.02\\
 & Uniform increase, equal \(\mathcal U_1\) & \(-0.086\) \([-0.088,-0.084]\) & \(+0.182\) \([+0.178,+0.186]\) & \(+0.213\) \([+0.206,+0.220]\) & 1.61 & 2.60\\
\addlinespace
\(\gamma=2\) & Floor-constrained MPC (\(c=2\)) & \(-0.104\) \([-0.105,-0.102]\) & \(-0.000\) \([-0.000,-0.000]\) & \(+0.092\) \([+0.085,+0.100]\) & 1.36 & 1.91\\
 & Targeted top-up, equal \(\mathcal U_1\) & \(-0.111\) \([-0.113,-0.110]\) & \(+0.000\) \([+0.000,+0.000]\) & \(+0.093\) \([+0.085,+0.100]\) & 1.36 & 1.91\\
 & Targeted top-up, equal \(\mathcal U_2\) & \(-0.111\) \([-0.112,-0.109]\) & \(+0.000\) \([+0.000,+0.000]\) & \(+0.093\) \([+0.085,+0.100]\) & 1.35 & 1.91\\
 & Uniform increase, equal \(\mathcal U_1\) & \(-0.086\) \([-0.088,-0.085]\) & \(+0.098\) \([+0.096,+0.100]\) & \(+0.093\) \([+0.085,+0.100]\) & 1.36 & 1.80\\
\bottomrule
\end{tabular}

%% file: tab_oracle_diff.tex
\begin{tabular}{@{}llrrrrrr@{}}
\toprule
Objective & \(\gamma\) & Mode & Model error, with group & Model error, without group & Erasure & Random compression & Group information\\
\midrule
knowledge & 0.5 & equal \(\mathcal U_1\) & \(-0.000\) [\(-0.001\), \(+0.000\)] & \(+0.000\) [\(-0.000\), \(+0.001\)] & \(-0.000\) [\(-0.001\), \(+0.001\)] & \(-0.000\) [\(-0.001\), \(+0.001\)] & \(-0.000\) [\(-0.000\), \(-0.000\)]\\
 &  & \(\rho=1\) & \(+0.166\) [\(+0.162\), \(+0.169\)] & \(+0.163\) [\(+0.160\), \(+0.167\)] & \(+0.158\) [\(+0.135\), \(+0.181\)] & \(+0.232\) [\(+0.182\), \(+0.283\)] & \(-0.000\) [\(-0.000\), \(-0.000\)]\\
 & 1.0 & equal \(\mathcal U_1\) & \(+0.000\) [\(-0.000\), \(+0.001\)] & \(-0.000\) [\(-0.001\), \(+0.000\)] & \(+0.000\) [\(-0.001\), \(+0.001\)] & \(-0.000\) [\(-0.001\), \(+0.001\)] & \(+0.000\) [\(+0.000\), \(+0.000\)]\\
 &  & \(\rho=1\) & \(+0.220\) [\(+0.216\), \(+0.223\)] & \(+0.217\) [\(+0.214\), \(+0.221\)] & \(+0.081\) [\(+0.072\), \(+0.091\)] & \(+0.168\) [\(+0.132\), \(+0.204\)] & \(-0.000\) [\(-0.000\), \(-0.000\)]\\
 & 2.0 & equal \(\mathcal U_1\) & \(+0.132\) [\(+0.131\), \(+0.132\)] & \(+0.082\) [\(+0.082\), \(+0.083\)] & \(-0.000\) [\(-0.000\), \(+0.000\)] & \(+0.012\) [\(-0.013\), \(+0.038\)] & \(+0.049\) [\(+0.049\), \(+0.050\)]\\
 &  & \(\rho=1\) & \(+0.157\) [\(+0.156\), \(+0.159\)] & \(+0.161\) [\(+0.159\), \(+0.164\)] & \(+0.016\) [\(+0.015\), \(+0.018\)] & \(+0.081\) [\(+0.056\), \(+0.107\)] & \(-0.006\) [\(-0.007\), \(-0.006\)]\\
\addlinespace
paper & 0.5 & equal \(\mathcal U_1\) & \(+0.002\) [\(+0.001\), \(+0.002\)] & \(+0.001\) [\(-0.000\), \(+0.002\)] & \(+0.020\) [\(+0.016\), \(+0.024\)] & \(+0.012\) [\(-0.002\), \(+0.025\)] & \(+0.000\) [\(-0.000\), \(+0.001\)]\\
 &  & \(\rho=1\) & \(+0.086\) [\(+0.084\), \(+0.089\)] & \(+0.088\) [\(+0.086\), \(+0.090\)] & \(+0.128\) [\(+0.106\), \(+0.149\)] & \(+0.206\) [\(+0.150\), \(+0.261\)] & \(-0.001\) [\(-0.001\), \(-0.001\)]\\
 & 1.0 & equal \(\mathcal U_1\) & \(+0.011\) [\(+0.010\), \(+0.012\)] & \(+0.011\) [\(+0.010\), \(+0.013\)] & \(+0.050\) [\(+0.044\), \(+0.056\)] & \(+0.119\) [\(+0.040\), \(+0.198\)] & \(-0.002\) [\(-0.003\), \(-0.002\)]\\
 &  & \(\rho=1\) & \(+0.123\) [\(+0.120\), \(+0.125\)] & \(+0.130\) [\(+0.127\), \(+0.133\)] & \(+0.165\) [\(+0.150\), \(+0.180\)] & \(+0.254\) [\(+0.170\), \(+0.338\)] & \(-0.003\) [\(-0.004\), \(-0.003\)]\\
 & 2.0 & equal \(\mathcal U_1\) & \(+0.071\) [\(+0.070\), \(+0.073\)] & \(+0.089\) [\(+0.086\), \(+0.091\)] & \(+0.115\) [\(+0.107\), \(+0.122\)] & \(+0.262\) [\(+0.141\), \(+0.383\)] & \(-0.032\) [\(-0.034\), \(-0.031\)]\\
 &  & \(\rho=1\) & \(+0.140\) [\(+0.137\), \(+0.144\)] & \(+0.204\) [\(+0.200\), \(+0.208\)] & \(+0.179\) [\(+0.162\), \(+0.195\)] & \(+0.267\) [\(+0.149\), \(+0.385\)] & \(-0.038\) [\(-0.039\), \(-0.037\)]\\
\bottomrule
\end{tabular}

%% file: tab_budget.tex
\begin{tabular}{@{}rrrrll@{}}
\toprule
\(\gamma\) & \(\rho\) & \(|u|\) raw & \(|u|\) constr. & \(\Delta|\delta_T|\) & \(\Delta\overline k_T\)\\
\midrule
0.5 & 1 & 0.236 & 0.233 & -0.132 \scriptsize[-0.134,\,-0.129] & -0.049 \scriptsize[-0.052,\,-0.046]\\
0.5 & 1.5 & 0.236 & 0.217 & -0.136 \scriptsize[-0.138,\,-0.134] & -0.184 \scriptsize[-0.186,\,-0.182]\\
0.5 & 2 & 0.236 & 0.213 & -0.135 \scriptsize[-0.137,\,-0.134] & -0.252 \scriptsize[-0.254,\,-0.250]\\
0.5 & 3 & 0.236 & 0.208 & -0.133 \scriptsize[-0.134,\,-0.131] & -0.316 \scriptsize[-0.318,\,-0.313]\\
0.5 & 5 & 0.236 & 0.204 & -0.130 \scriptsize[-0.131,\,-0.128] & -0.364 \scriptsize[-0.367,\,-0.361]\\
1 & 1 & 0.287 & 0.399 & -0.228 \scriptsize[-0.232,\,-0.225] & -0.172 \scriptsize[-0.174,\,-0.169]\\
1 & 1.5 & 0.287 & 0.394 & -0.223 \scriptsize[-0.225,\,-0.220] & -0.203 \scriptsize[-0.206,\,-0.200]\\
1 & 2 & 0.287 & 0.393 & -0.224 \scriptsize[-0.226,\,-0.222] & -0.233 \scriptsize[-0.236,\,-0.229]\\
1 & 3 & 0.287 & 0.392 & -0.226 \scriptsize[-0.228,\,-0.223] & -0.267 \scriptsize[-0.270,\,-0.263]\\
1 & 5 & 0.287 & 0.391 & -0.227 \scriptsize[-0.229,\,-0.225] & -0.296 \scriptsize[-0.300,\,-0.292]\\
2 & 1 & 0.390 & 0.689 & -0.344 \scriptsize[-0.350,\,-0.338] & -0.182 \scriptsize[-0.185,\,-0.179]\\
2 & 1.5 & 0.390 & 0.691 & -0.345 \scriptsize[-0.349,\,-0.340] & -0.197 \scriptsize[-0.200,\,-0.194]\\
2 & 2 & 0.390 & 0.693 & -0.348 \scriptsize[-0.352,\,-0.344] & -0.210 \scriptsize[-0.213,\,-0.206]\\
2 & 3 & 0.390 & 0.696 & -0.353 \scriptsize[-0.356,\,-0.349] & -0.226 \scriptsize[-0.230,\,-0.222]\\
2 & 5 & 0.390 & 0.699 & -0.357 \scriptsize[-0.360,\,-0.353] & -0.240 \scriptsize[-0.244,\,-0.236]\\
\bottomrule
\end{tabular}

%% file: tab_beta_gap.tex
\begin{tabular}{@{}rrrrllllr@{}}
\toprule
\(\gamma\) & \(\beta_{\mathrm{gap}}\) & \(|u|\) raw & \(|u|\) arm & \(\Delta|\delta_T|\) & \(\Delta\overline k_{T,0}\) & \(\Delta\overline k_{T,1}\) & median \(|\lambda|\) & infeas.\\
\midrule
0.5 & 1 & 0.236 & 0.229 & +0.071 \scriptsize[+0.070,\,+0.071] & +0.150 \scriptsize[+0.148,\,+0.153] & -0.200 \scriptsize[-0.203,\,-0.198] & 0.4 & 0.0\%\\
0.5 & 8 & 0.236 & 0.229 & +0.070 \scriptsize[+0.069,\,+0.071] & +0.148 \scriptsize[+0.146,\,+0.151] & -0.199 \scriptsize[-0.201,\,-0.196] & 2.8 & 0.0\%\\
0.5 & 64 & 0.236 & 0.229 & +0.065 \scriptsize[+0.064,\,+0.065] & +0.136 \scriptsize[+0.134,\,+0.138] & -0.185 \scriptsize[-0.187,\,-0.182] & 22.1 & 0.0\%\\
0.5 & 512 & 0.236 & 0.229 & +0.029 \scriptsize[+0.029,\,+0.030] & +0.052 \scriptsize[+0.050,\,+0.054] & -0.095 \scriptsize[-0.097,\,-0.093] & 151.3 & 0.0\%\\
0.5 & 4096 & 0.236 & 0.229 & -0.068 \scriptsize[-0.069,\,-0.067] & -0.196 \scriptsize[-0.200,\,-0.192] & +0.145 \scriptsize[+0.143,\,+0.148] & 537.3 & 0.0\%\\
0.5 & 32768 & 0.236 & 0.230 & -0.121 \scriptsize[-0.123,\,-0.119] & -0.330 \scriptsize[-0.336,\,-0.324] & +0.245 \scriptsize[+0.243,\,+0.248] & 745.8 & 0.0\%\\
0.5 & constraint & 0.236 & 0.233 & -0.132 \scriptsize[-0.134,\,-0.129] & -0.356 \scriptsize[-0.362,\,-0.349] & +0.257 \scriptsize[+0.255,\,+0.260] & 789.1 & 0.0\%\\
\midrule
1 & 1 & 0.287 & 0.203 & +0.144 \scriptsize[+0.143,\,+0.145] & +0.136 \scriptsize[+0.134,\,+0.139] & -0.585 \scriptsize[-0.587,\,-0.582] & 0.8 & 0.0\%\\
1 & 8 & 0.287 & 0.203 & +0.142 \scriptsize[+0.142,\,+0.143] & +0.133 \scriptsize[+0.130,\,+0.136] & -0.581 \scriptsize[-0.584,\,-0.579] & 6.1 & 0.0\%\\
1 & 64 & 0.287 & 0.203 & +0.131 \scriptsize[+0.130,\,+0.132] & +0.105 \scriptsize[+0.102,\,+0.107] & -0.555 \scriptsize[-0.558,\,-0.552] & 48.0 & 0.0\%\\
1 & 512 & 0.287 & 0.203 & +0.056 \scriptsize[+0.055,\,+0.056] & -0.082 \scriptsize[-0.084,\,-0.080] & -0.373 \scriptsize[-0.376,\,-0.370] & 328.0 & 0.0\%\\
1 & 4096 & 0.287 & 0.291 & -0.137 \scriptsize[-0.138,\,-0.136] & -0.504 \scriptsize[-0.507,\,-0.501] & +0.120 \scriptsize[+0.117,\,+0.122] & 1103.2 & 0.0\%\\
1 & 32768 & 0.287 & 0.380 & -0.212 \scriptsize[-0.215,\,-0.209] & -0.559 \scriptsize[-0.563,\,-0.555] & +0.215 \scriptsize[+0.213,\,+0.218] & 1381.0 & 0.0\%\\
1 & constraint & 0.287 & 0.399 & -0.228 \scriptsize[-0.232,\,-0.225] & -0.564 \scriptsize[-0.568,\,-0.560] & +0.221 \scriptsize[+0.218,\,+0.224] & 1443.5 & 0.0\%\\
\midrule
2 & 1 & 0.390 & 0.159 & +0.279 \scriptsize[+0.278,\,+0.281] & +0.105 \scriptsize[+0.102,\,+0.108] & -0.845 \scriptsize[-0.852,\,-0.839] & 1.6 & 0.0\%\\
2 & 8 & 0.390 & 0.159 & +0.276 \scriptsize[+0.275,\,+0.278] & +0.097 \scriptsize[+0.094,\,+0.100] & -0.845 \scriptsize[-0.852,\,-0.839] & 12.8 & 0.0\%\\
2 & 64 & 0.390 & 0.159 & +0.252 \scriptsize[+0.251,\,+0.254] & +0.037 \scriptsize[+0.034,\,+0.040] & -0.845 \scriptsize[-0.851,\,-0.839] & 101.0 & 0.0\%\\
2 & 512 & 0.390 & 0.195 & +0.088 \scriptsize[+0.087,\,+0.090] & -0.349 \scriptsize[-0.352,\,-0.346] & -0.771 \scriptsize[-0.776,\,-0.766] & 719.8 & 0.0\%\\
2 & 4096 & 0.390 & 0.522 & -0.218 \scriptsize[-0.220,\,-0.216] & -0.502 \scriptsize[-0.506,\,-0.498] & -0.026 \scriptsize[-0.029,\,-0.023] & 1926.1 & 0.0\%\\
2 & 32768 & 0.390 & 0.660 & -0.321 \scriptsize[-0.326,\,-0.316] & -0.503 \scriptsize[-0.507,\,-0.499] & +0.130 \scriptsize[+0.126,\,+0.134] & 2274.1 & 0.0\%\\
2 & constraint & 0.390 & 0.689 & -0.344 \scriptsize[-0.350,\,-0.338] & -0.503 \scriptsize[-0.507,\,-0.499] & +0.139 \scriptsize[+0.135,\,+0.144] & 2352.6 & 6.4\%\\
\bottomrule
\end{tabular}

%% file: tab_scenarios.tex
\begin{tabular}{@{}llrrr@{}}
\toprule
Scenario & Comparison & \(\Delta|\delta_T|\) & \(\Delta\overline k_{T,0}\) & \(\Delta\overline k_{T,1}\)\\
\midrule
Group 1 responds half as much & Projected MPC vs.\ original & \(+0.080\) \([+0.078,+0.082]\) & \(+0.130\) \([+0.112,+0.149]\) & \(-0.173\) \([-0.179,-0.167]\)\\
 & Gap-constrained vs.\ original & \(-0.284\) \([-0.295,-0.272]\) & \(-0.525\) \([-0.538,-0.513]\) & \(+0.315\) \([+0.293,+0.337]\)\\
 & Floor-constrained vs.\ group-aware reference & \(-0.093\) \([-0.095,-0.092]\) & \(-0.000\) \([-0.000,-0.000]\) & \(+0.480\) \([+0.474,+0.487]\)\\
 & Top-up (equal \(\mathcal U_1\)) vs.\ reference & \(-0.089\) \([-0.090,-0.088]\) & \(+0.000\) \([+0.000,+0.000]\) & \(+0.463\) \([+0.457,+0.469]\)\\
\addlinespace
Slower learning, twice the noise & Projected MPC vs.\ original & \(+0.138\) \([+0.134,+0.142]\) & \(-0.018\) \([-0.064,+0.029]\) & \(-0.735\) \([-0.817,-0.653]\)\\
 & Gap-constrained vs.\ original & \(-0.217\) \([-0.230,-0.203]\) & \(-0.806\) \([-0.832,-0.780]\) & \(+0.019\) \([+0.014,+0.023]\)\\
 & Floor-constrained vs.\ group-aware reference & \(-0.166\) \([-0.169,-0.162]\) & \(-0.000\) \([-0.000,-0.000]\) & \(+0.013\) \([+0.010,+0.016]\)\\
 & Top-up (equal \(\mathcal U_1\)) vs.\ reference & \(-0.165\) \([-0.168,-0.162]\) & \(+0.000\) \([+0.000,+0.000]\) & \(+0.013\) \([+0.010,+0.016]\)\\
\addlinespace
Diminishing returns to help & Projected MPC vs.\ original & \(+0.139\) \([+0.137,+0.141]\) & \(+0.081\) \([+0.067,+0.095]\) & \(-0.256\) \([-0.278,-0.234]\)\\
 & Gap-constrained vs.\ original & \(-0.274\) \([-0.281,-0.266]\) & \(-0.609\) \([-0.618,-0.599]\) & \(+0.100\) \([+0.095,+0.104]\)\\
 & Floor-constrained vs.\ group-aware reference & \(-0.175\) \([-0.179,-0.170]\) & \(-0.000\) \([-0.000,-0.000]\) & \(+0.150\) \([+0.148,+0.152]\)\\
 & Top-up (equal \(\mathcal U_1\)) vs.\ reference & \(-0.175\) \([-0.179,-0.171]\) & \(+0.000\) \([+0.000,+0.000]\) & \(+0.151\) \([+0.149,+0.153]\)\\
\bottomrule
\end{tabular}

%% file: tab_frontier.tex
\begin{tabular}{@{}rlrlllll@{}}
\toprule
\(\gamma\) & Arm & \(|u|\) ratio & \(\Delta|\delta_T|\) & \(\Delta\overline k_{T,0}\) & \(\Delta\overline k_{T,1}\) & \(\Delta\delta^k_T\) & worst group (min)\\
\midrule
0.5 & more control, \(\rho=0.6\) & 1.41 & -0.029 \scriptsize[-0.030,\,-0.029] & +0.140 \scriptsize[+0.138,\,+0.143] & +0.232 \scriptsize[+0.229,\,+0.235] & +0.092 & +0.1326\\
0.5 & more control, \(\rho=0.4\) & 1.78 & -0.054 \scriptsize[-0.055,\,-0.053] & +0.240 \scriptsize[+0.236,\,+0.244] & +0.274 \scriptsize[+0.269,\,+0.279] & +0.034 & +0.2293\\
0.5 & more control, \(\rho=0.25\) & 2.20 & -0.081 \scriptsize[-0.084,\,-0.079] & +0.309 \scriptsize[+0.303,\,+0.315] & +0.278 \scriptsize[+0.272,\,+0.284] & -0.031 & +0.2677\\
0.5 & more control, \(\rho=0.15\) & 2.58 & -0.105 \scriptsize[-0.109,\,-0.102] & +0.339 \scriptsize[+0.330,\,+0.347] & +0.279 \scriptsize[+0.273,\,+0.285] & -0.060 & +0.2678\\
0.5 & more control, \(\rho=0.1\) & 2.79 & -0.120 \scriptsize[-0.124,\,-0.115] & +0.348 \scriptsize[+0.338,\,+0.358] & +0.279 \scriptsize[+0.273,\,+0.285] & -0.069 & +0.2678\\
0.5 & floor-constrained, \(c=1\) & 1.00 & +0.000 \scriptsize[+0.000,\,+0.000] & +0.000 \scriptsize[+0.000,\,+0.000] & +0.000 \scriptsize[+0.000,\,+0.000] & +0.000 & +0.0000\\
0.5 & floor-constrained, \(c=1.1\) & 1.06 & -0.012 \scriptsize[-0.012,\,-0.011] & +0.000 \scriptsize[+0.000,\,+0.000] & +0.059 \scriptsize[+0.058,\,+0.059] & +0.059 & +0.0000\\
0.5 & floor-constrained, \(c=1.25\) & 1.16 & -0.033 \scriptsize[-0.034,\,-0.033] & -0.000 \scriptsize[-0.000,\,-0.000] & +0.160 \scriptsize[+0.157,\,+0.162] & +0.160 & -0.0000\\
0.5 & floor-constrained, \(c=1.5\) & 1.32 & -0.067 \scriptsize[-0.068,\,-0.065] & -0.000 \scriptsize[-0.000,\,-0.000] & +0.254 \scriptsize[+0.250,\,+0.257] & +0.254 & -0.0000\\
0.5 & floor-constrained, \(c=1.75\) & 1.47 & -0.096 \scriptsize[-0.098,\,-0.094] & -0.000 \scriptsize[-0.000,\,-0.000] & +0.274 \scriptsize[+0.269,\,+0.279] & +0.274 & -0.0000\\
0.5 & floor-constrained, \(c=2\) & 1.61 & -0.122 \scriptsize[-0.124,\,-0.120] & -0.000 \scriptsize[-0.000,\,-0.000] & +0.277 \scriptsize[+0.272,\,+0.283] & +0.277 & -0.0000\\
\midrule
1 & more control, \(\rho=0.6\) & 1.42 & -0.059 \scriptsize[-0.060,\,-0.058] & +0.128 \scriptsize[+0.125,\,+0.131] & +0.209 \scriptsize[+0.203,\,+0.216] & +0.081 & +0.1210\\
1 & more control, \(\rho=0.4\) & 1.79 & -0.109 \scriptsize[-0.112,\,-0.107] & +0.225 \scriptsize[+0.220,\,+0.230] & +0.213 \scriptsize[+0.206,\,+0.220] & -0.011 & +0.2009\\
1 & more control, \(\rho=0.25\) & 2.19 & -0.158 \scriptsize[-0.161,\,-0.155] & +0.306 \scriptsize[+0.301,\,+0.311] & +0.213 \scriptsize[+0.206,\,+0.221] & -0.092 & +0.2009\\
1 & more control, \(\rho=0.15\) & 2.39 & -0.164 \scriptsize[-0.168,\,-0.160] & +0.349 \scriptsize[+0.343,\,+0.356] & +0.213 \scriptsize[+0.206,\,+0.221] & -0.136 & +0.2009\\
1 & more control, \(\rho=0.1\) & 2.45 & -0.153 \scriptsize[-0.158,\,-0.149] & +0.364 \scriptsize[+0.356,\,+0.372] & +0.213 \scriptsize[+0.206,\,+0.221] & -0.151 & +0.2009\\
1 & floor-constrained, \(c=1\) & 1.00 & +0.000 \scriptsize[+0.000,\,+0.000] & +0.000 \scriptsize[+0.000,\,+0.000] & +0.000 \scriptsize[+0.000,\,+0.000] & +0.000 & +0.0000\\
1 & floor-constrained, \(c=1.1\) & 1.06 & -0.015 \scriptsize[-0.016,\,-0.015] & +0.000 \scriptsize[+0.000,\,+0.000] & +0.074 \scriptsize[+0.073,\,+0.075] & +0.074 & +0.0000\\
1 & floor-constrained, \(c=1.25\) & 1.16 & -0.043 \scriptsize[-0.044,\,-0.043] & -0.000 \scriptsize[-0.000,\,-0.000] & +0.173 \scriptsize[+0.170,\,+0.177] & +0.173 & -0.0000\\
1 & floor-constrained, \(c=1.5\) & 1.33 & -0.086 \scriptsize[-0.087,\,-0.084] & -0.000 \scriptsize[-0.000,\,-0.000] & +0.210 \scriptsize[+0.204,\,+0.217] & +0.210 & -0.0000\\
1 & floor-constrained, \(c=1.75\) & 1.48 & -0.123 \scriptsize[-0.125,\,-0.121] & -0.000 \scriptsize[-0.000,\,-0.000] & +0.213 \scriptsize[+0.206,\,+0.220] & +0.213 & -0.0000\\
1 & floor-constrained, \(c=2\) & 1.61 & -0.157 \scriptsize[-0.160,\,-0.154] & -0.000 \scriptsize[-0.000,\,-0.000] & +0.213 \scriptsize[+0.206,\,+0.221] & +0.213 & -0.0000\\
\midrule
2 & more control, \(\rho=0.6\) & 1.37 & -0.087 \scriptsize[-0.088,\,-0.085] & +0.103 \scriptsize[+0.100,\,+0.105] & +0.093 \scriptsize[+0.085,\,+0.100] & -0.010 & +0.0801\\
2 & more control, \(\rho=0.4\) & 1.49 & -0.080 \scriptsize[-0.082,\,-0.078] & +0.185 \scriptsize[+0.180,\,+0.190] & +0.093 \scriptsize[+0.085,\,+0.100] & -0.092 & +0.0801\\
2 & more control, \(\rho=0.25\) & 1.58 & -0.065 \scriptsize[-0.068,\,-0.062] & +0.266 \scriptsize[+0.258,\,+0.274] & +0.093 \scriptsize[+0.085,\,+0.100] & -0.174 & +0.0801\\
2 & more control, \(\rho=0.15\) & 1.64 & -0.050 \scriptsize[-0.054,\,-0.047] & +0.326 \scriptsize[+0.317,\,+0.335] & +0.093 \scriptsize[+0.085,\,+0.100] & -0.233 & +0.0801\\
2 & more control, \(\rho=0.1\) & 1.67 & -0.041 \scriptsize[-0.046,\,-0.037] & +0.353 \scriptsize[+0.344,\,+0.361] & +0.093 \scriptsize[+0.085,\,+0.100] & -0.260 & +0.0801\\
2 & floor-constrained, \(c=1\) & 1.00 & +0.000 \scriptsize[+0.000,\,+0.000] & +0.000 \scriptsize[+0.000,\,+0.000] & +0.000 \scriptsize[+0.000,\,+0.000] & +0.000 & +0.0000\\
2 & floor-constrained, \(c=1.1\) & 1.06 & -0.023 \scriptsize[-0.023,\,-0.023] & -0.000 \scriptsize[-0.000,\,+0.000] & +0.066 \scriptsize[+0.062,\,+0.070] & +0.066 & -0.0000\\
2 & floor-constrained, \(c=1.25\) & 1.17 & -0.063 \scriptsize[-0.064,\,-0.063] & -0.000 \scriptsize[-0.000,\,-0.000] & +0.091 \scriptsize[+0.084,\,+0.098] & +0.091 & -0.0000\\
2 & floor-constrained, \(c=1.5\) & 1.29 & -0.096 \scriptsize[-0.098,\,-0.095] & +0.000 \scriptsize[-0.000,\,+0.000] & +0.092 \scriptsize[+0.085,\,+0.100] & +0.092 & -0.0000\\
2 & floor-constrained, \(c=1.75\) & 1.33 & -0.102 \scriptsize[-0.103,\,-0.100] & -0.000 \scriptsize[-0.000,\,+0.000] & +0.092 \scriptsize[+0.085,\,+0.100] & +0.092 & -0.0000\\
2 & floor-constrained, \(c=2\) & 1.36 & -0.104 \scriptsize[-0.105,\,-0.102] & -0.000 \scriptsize[-0.000,\,-0.000] & +0.092 \scriptsize[+0.085,\,+0.100] & +0.092 & -0.0000\\
\bottomrule
\end{tabular}

%% file: tab_replicates50.tex
\begin{tabular}{@{}lrrrrr@{}}
\toprule
Setting & Projected: \(\Delta|\delta_T|\) & Holm \(p\) & Constrained: \(\Delta|\delta_T|\) & Constrained: lower group \(\Delta\overline k_T\) & Holm \(p\)\\
\midrule
\(\gamma=0\), rank 4, depth 3, \(T=25\), \(\beta_{\mathrm{ind}}=8\) & \(+0.000\) \([+0.000,+0.001]\) & 1.4e-02 & \(-0.002\) \([-0.002,-0.001]\) & \(+0.169\) \([+0.166,+0.172]\) & 6.6e-62\\
\(\gamma=0.25\), rank 4, depth 3, \(T=25\), \(\beta_{\mathrm{ind}}=8\) & \(+0.031\) \([+0.030,+0.032]\) & 3.9e-52 & \(-0.059\) \([-0.061,-0.057]\) & \(-0.072\) \([-0.077,-0.067]\) & 1.2e-33\\
\(\gamma=0.5\), rank 4, depth 3, \(T=25\), \(\beta_{\mathrm{ind}}=8\) & \(+0.068\) \([+0.067,+0.069]\) & 6.7e-65 & \(-0.132\) \([-0.134,-0.129]\) & \(-0.356\) \([-0.362,-0.349]\) & 3.5e-60\\
\(\gamma=1\), rank 4, depth 3, \(T=25\), \(\beta_{\mathrm{ind}}=8\) & \(+0.136\) \([+0.135,+0.138]\) & 5.1e-75 & \(-0.228\) \([-0.232,-0.225]\) & \(-0.564\) \([-0.568,-0.560]\) & 2.9e-80\\
\(\gamma=2\), rank 4, depth 3, \(T=25\), \(\beta_{\mathrm{ind}}=8\) & \(+0.260\) \([+0.256,+0.263]\) & 4.1e-66 & \(-0.344\) \([-0.350,-0.338]\) & \(-0.503\) \([-0.507,-0.499]\) & 8.5e-76\\
\(\gamma=-1\), rank 4, depth 3, \(T=25\), \(\beta_{\mathrm{ind}}=8\) & \(+0.136\) \([+0.135,+0.137]\) & 7.0e-79 & \(-0.220\) \([-0.222,-0.217]\) & \(+0.012\) \([+0.004,+0.019]\) & 7.5e-03\\
\(\gamma=1\), rank 1, depth 3, \(T=25\), \(\beta_{\mathrm{ind}}=8\) & \(+0.141\) \([+0.138,+0.143]\) & 7.4e-64 & \(-0.228\) \([-0.232,-0.225]\) & \(-0.564\) \([-0.568,-0.560]\) & 2.9e-80\\
\(\gamma=1\), rank 2, depth 3, \(T=25\), \(\beta_{\mathrm{ind}}=8\) & \(+0.146\) \([+0.144,+0.148]\) & 2.9e-65 & \(-0.228\) \([-0.232,-0.225]\) & \(-0.564\) \([-0.568,-0.560]\) & 2.9e-80\\
\(\gamma=1\), rank 6, depth 3, \(T=25\), \(\beta_{\mathrm{ind}}=8\) & \(+0.137\) \([+0.136,+0.139]\) & 1.0e-74 & \(-0.228\) \([-0.232,-0.225]\) & \(-0.564\) \([-0.568,-0.560]\) & 2.9e-80\\
\(\gamma=1\), rank 8, depth 3, \(T=25\), \(\beta_{\mathrm{ind}}=8\) & \(+0.139\) \([+0.137,+0.140]\) & 3.8e-74 & \(-0.228\) \([-0.232,-0.225]\) & \(-0.564\) \([-0.568,-0.560]\) & 2.9e-80\\
\(\gamma=1\), rank 4, depth 1, \(T=25\), \(\beta_{\mathrm{ind}}=8\) & \(+0.141\) \([+0.138,+0.143]\) & 7.4e-64 & \(-0.228\) \([-0.232,-0.225]\) & \(-0.564\) \([-0.568,-0.560]\) & 2.9e-80\\
\(\gamma=1\), rank 4, depth 2, \(T=25\), \(\beta_{\mathrm{ind}}=8\) & \(+0.145\) \([+0.143,+0.147]\) & 3.1e-65 & \(-0.228\) \([-0.232,-0.225]\) & \(-0.564\) \([-0.568,-0.560]\) & 2.9e-80\\
\(\gamma=1\), rank 4, depth 5, \(T=25\), \(\beta_{\mathrm{ind}}=8\) & \(+0.136\) \([+0.135,+0.137]\) & 9.8e-76 & \(-0.228\) \([-0.232,-0.225]\) & \(-0.564\) \([-0.568,-0.560]\) & 2.9e-80\\
\(\gamma=1\), rank 4, depth 3, \(T=5\), \(\beta_{\mathrm{ind}}=8\) & \(+0.023\) \([+0.023,+0.023]\) & 2.0e-63 & \(-0.105\) \([-0.107,-0.104]\) & \(-0.255\) \([-0.258,-0.253]\) & 5.2e-71\\
\(\gamma=1\), rank 4, depth 3, \(T=10\), \(\beta_{\mathrm{ind}}=8\) & \(+0.064\) \([+0.064,+0.065]\) & 8.0e-74 & \(-0.202\) \([-0.204,-0.200]\) & \(-0.500\) \([-0.503,-0.497]\) & 1.6e-81\\
\(\gamma=1\), rank 4, depth 3, \(T=50\), \(\beta_{\mathrm{ind}}=8\) & \(+0.155\) \([+0.153,+0.156]\) & 8.6e-70 & \(-0.235\) \([-0.239,-0.232]\) & \(-0.573\) \([-0.577,-0.569]\) & 1.6e-78\\
\(\gamma=1\), rank 4, depth 3, \(T=25\), \(\beta_{\mathrm{ind}}=0\) & \(-0.000\) \([-0.000,+0.000]\) & 1.0e+00 & \(-0.372\) \([-0.376,-0.368]\) & \(-0.701\) \([-0.704,-0.698]\) & 1.4e-90\\
\(\gamma=1\), rank 4, depth 3, \(T=25\), \(\beta_{\mathrm{ind}}=1\) & \(+0.024\) \([+0.023,+0.024]\) & 1.6e-73 & \(-0.348\) \([-0.352,-0.344]\) & \(-0.679\) \([-0.682,-0.676]\) & 5.7e-90\\
\(\gamma=1\), rank 4, depth 3, \(T=25\), \(\beta_{\mathrm{ind}}=32\) & \(+0.287\) \([+0.278,+0.296]\) & 7.5e-48 & \(-0.081\) \([-0.083,-0.079]\) & \(-0.401\) \([-0.406,-0.395]\) & 6.1e-65\\
\(\gamma=0\), rank 4, depth 3, \(T=25\), \(\beta_{\mathrm{ind}}=0\) & \(-0.000\) \([-0.000,+0.000]\) & 1.0e+00 & \(-0.002\) \([-0.004,-0.001]\) & \(-0.001\) \([-0.004,+0.002]\) & 4.8e-01\\
\(\gamma=0\), rank 4, depth 3, \(T=25\), \(\beta_{\mathrm{ind}}=1\) & \(+0.000\) \([+0.000,+0.000]\) & 4.9e-02 & \(-0.002\) \([-0.003,-0.001]\) & \(+0.026\) \([+0.023,+0.029]\) & 1.8e-22\\
\(\gamma=0\), rank 4, depth 3, \(T=25\), \(\beta_{\mathrm{ind}}=32\) & \(+0.001\) \([+0.001,+0.002]\) & 4.5e-03 & \(-0.001\) \([-0.001,-0.000]\) & \(+0.372\) \([+0.369,+0.375]\) & 3.1e-78\\
\(\gamma=0.5\), rank 4, depth 3, \(T=25\), \(\beta_{\mathrm{ind}}=0\) & \(-0.000\) \([-0.000,+0.000]\) & 2.2e-01 & \(-0.203\) \([-0.206,-0.199]\) & \(-0.506\) \([-0.513,-0.499]\) & 2.3e-66\\
\(\gamma=0.5\), rank 4, depth 3, \(T=25\), \(\beta_{\mathrm{ind}}=1\) & \(+0.011\) \([+0.011,+0.012]\) & 3.1e-65 & \(-0.191\) \([-0.194,-0.188]\) & \(-0.482\) \([-0.488,-0.475]\) & 1.3e-65\\
\(\gamma=0.5\), rank 4, depth 3, \(T=25\), \(\beta_{\mathrm{ind}}=32\) & \(+0.142\) \([+0.139,+0.144]\) & 2.5e-61 & \(-0.055\) \([-0.057,-0.053]\) & \(-0.177\) \([-0.183,-0.170]\) & 5.9e-45\\
\(\gamma=2\), rank 4, depth 3, \(T=25\), \(\beta_{\mathrm{ind}}=0\) & \(-0.000\) \([-0.000,+0.000]\) & 2.2e-01 & \(-0.624\) \([-0.630,-0.617]\) & \(-0.609\) \([-0.612,-0.606]\) & 4.6e-87\\
\(\gamma=2\), rank 4, depth 3, \(T=25\), \(\beta_{\mathrm{ind}}=1\) & \(+0.049\) \([+0.049,+0.049]\) & 4.2e-76 & \(-0.573\) \([-0.580,-0.566]\) & \(-0.592\) \([-0.595,-0.589]\) & 1.6e-86\\
\(\gamma=2\), rank 4, depth 3, \(T=25\), \(\beta_{\mathrm{ind}}=32\) & \(+0.401\) \([+0.391,+0.410]\) & 1.5e-54 & \(-0.189\) \([-0.195,-0.182]\) & \(-0.376\) \([-0.382,-0.369]\) & 1.3e-60\\
\bottomrule
\end{tabular}

%% file: tab_regimes_full.tex
\begin{tabular}{@{}rrrrrllllrrr@{}}
\toprule
\(\gamma\) & \(\beta\) & rank & depth & \(T\) & Proj.: \(\Delta|\delta_T|\) & Proj.: \(\Delta\overline k_T\) & Constr.: \(\Delta|\delta_T|\) & Constr.: \(\Delta\overline k_T\) & \(|u|\) raw & \(|u|\) constr. & infeas.\\
\midrule
-1 & 8 & 4 & 3 & 25 & +0.136 \scriptsize[+0.135,\,+0.137] & +0.147 \scriptsize[+0.139,\,+0.155] & -0.220 \scriptsize[-0.222,\,-0.217] & +0.201 \scriptsize[+0.196,\,+0.207] & 0.154 & 0.398 & 0.0\%\\
0 & 0 & 4 & 3 & 25 & -0.000 \scriptsize[-0.000,\,+0.000] & -0.248 \scriptsize[-0.292,\,-0.204] & -0.002 \scriptsize[-0.004,\,-0.001] & -0.000 \scriptsize[-0.000,\,-0.000] & 0.247 & 0.247 & 0.0\%\\
0 & 1 & 4 & 3 & 25 & +0.000 \scriptsize[+0.000,\,+0.000] & -0.229 \scriptsize[-0.271,\,-0.188] & -0.002 \scriptsize[-0.003,\,-0.001] & +0.027 \scriptsize[+0.027,\,+0.027] & 0.237 & 0.247 & 0.0\%\\
0 & 8 & 4 & 3 & 25 & +0.000 \scriptsize[+0.000,\,+0.001] & -0.133 \scriptsize[-0.164,\,-0.101] & -0.002 \scriptsize[-0.002,\,-0.001] & +0.169 \scriptsize[+0.168,\,+0.171] & 0.181 & 0.247 & 0.0\%\\
0 & 32 & 4 & 3 & 25 & +0.001 \scriptsize[+0.001,\,+0.002] & -0.005 \scriptsize[-0.027,\,+0.018] & -0.001 \scriptsize[-0.001,\,-0.000] & +0.372 \scriptsize[+0.369,\,+0.375] & 0.109 & 0.247 & 0.0\%\\
0.25 & 8 & 4 & 3 & 25 & +0.031 \scriptsize[+0.030,\,+0.032] & -0.123 \scriptsize[-0.137,\,-0.109] & -0.059 \scriptsize[-0.061,\,-0.057] & +0.072 \scriptsize[+0.070,\,+0.074] & 0.209 & 0.239 & 0.0\%\\
0.5 & 0 & 4 & 3 & 25 & -0.000 \scriptsize[-0.000,\,+0.000] & -0.160 \scriptsize[-0.174,\,-0.147] & -0.203 \scriptsize[-0.206,\,-0.199] & -0.024 \scriptsize[-0.026,\,-0.022] & 0.229 & 0.233 & 0.0\%\\
0.5 & 1 & 4 & 3 & 25 & +0.011 \scriptsize[+0.011,\,+0.012] & -0.157 \scriptsize[-0.170,\,-0.144] & -0.191 \scriptsize[-0.194,\,-0.188] & -0.029 \scriptsize[-0.031,\,-0.027] & 0.230 & 0.233 & 0.0\%\\
0.5 & 8 & 4 & 3 & 25 & +0.068 \scriptsize[+0.067,\,+0.069] & -0.129 \scriptsize[-0.141,\,-0.117] & -0.132 \scriptsize[-0.134,\,-0.129] & -0.049 \scriptsize[-0.052,\,-0.046] & 0.236 & 0.233 & 0.0\%\\
0.5 & 32 & 4 & 3 & 25 & +0.142 \scriptsize[+0.139,\,+0.144] & -0.043 \scriptsize[-0.050,\,-0.035] & -0.055 \scriptsize[-0.057,\,-0.053] & -0.047 \scriptsize[-0.051,\,-0.043] & 0.254 & 0.233 & 0.0\%\\
1 & 0 & 4 & 3 & 25 & -0.000 \scriptsize[-0.000,\,+0.000] & -0.079 \scriptsize[-0.086,\,-0.073] & -0.372 \scriptsize[-0.376,\,-0.368] & +0.053 \scriptsize[+0.050,\,+0.055] & 0.203 & 0.399 & 0.0\%\\
1 & 1 & 4 & 3 & 25 & +0.024 \scriptsize[+0.023,\,+0.024] & -0.094 \scriptsize[-0.101,\,-0.087] & -0.348 \scriptsize[-0.352,\,-0.344] & +0.017 \scriptsize[+0.015,\,+0.019] & 0.217 & 0.399 & 0.0\%\\
1 & 8 & 1 & 3 & 25 & +0.141 \scriptsize[+0.138,\,+0.143] & -0.031 \scriptsize[-0.035,\,-0.027] & -0.228 \scriptsize[-0.232,\,-0.225] & -0.172 \scriptsize[-0.174,\,-0.169] & 0.287 & 0.399 & 0.0\%\\
1 & 8 & 2 & 3 & 25 & +0.146 \scriptsize[+0.144,\,+0.148] & -0.060 \scriptsize[-0.064,\,-0.056] & -0.228 \scriptsize[-0.232,\,-0.225] & -0.172 \scriptsize[-0.174,\,-0.169] & 0.287 & 0.399 & 0.0\%\\
1 & 8 & 4 & 1 & 25 & +0.141 \scriptsize[+0.138,\,+0.143] & -0.031 \scriptsize[-0.035,\,-0.027] & -0.228 \scriptsize[-0.232,\,-0.225] & -0.172 \scriptsize[-0.174,\,-0.169] & 0.287 & 0.399 & 0.0\%\\
1 & 8 & 4 & 2 & 25 & +0.145 \scriptsize[+0.143,\,+0.147] & -0.087 \scriptsize[-0.093,\,-0.081] & -0.228 \scriptsize[-0.232,\,-0.225] & -0.172 \scriptsize[-0.174,\,-0.169] & 0.287 & 0.399 & 0.0\%\\
1 & 8 & 4 & 3 & 5 & +0.023 \scriptsize[+0.023,\,+0.023] & -0.001 \scriptsize[-0.006,\,+0.003] & -0.105 \scriptsize[-0.107,\,-0.104] & -0.037 \scriptsize[-0.038,\,-0.036] & 0.246 & 0.332 & 0.0\%\\
1 & 8 & 4 & 3 & 10 & +0.064 \scriptsize[+0.064,\,+0.065] & -0.040 \scriptsize[-0.047,\,-0.033] & -0.202 \scriptsize[-0.204,\,-0.200] & -0.071 \scriptsize[-0.072,\,-0.069] & 0.264 & 0.401 & 0.0\%\\
1 & 8 & 4 & 3 & 25 & +0.136 \scriptsize[+0.135,\,+0.138] & -0.158 \scriptsize[-0.168,\,-0.147] & -0.228 \scriptsize[-0.232,\,-0.225] & -0.172 \scriptsize[-0.174,\,-0.169] & 0.287 & 0.399 & 0.0\%\\
1 & 8 & 4 & 3 & 50 & +0.155 \scriptsize[+0.153,\,+0.156] & -0.202 \scriptsize[-0.213,\,-0.190] & -0.235 \scriptsize[-0.239,\,-0.232] & -0.227 \scriptsize[-0.230,\,-0.224] & 0.295 & 0.402 & 0.0\%\\
1 & 8 & 4 & 5 & 25 & +0.136 \scriptsize[+0.135,\,+0.137] & -0.194 \scriptsize[-0.204,\,-0.185] & -0.228 \scriptsize[-0.232,\,-0.225] & -0.172 \scriptsize[-0.174,\,-0.169] & 0.287 & 0.399 & 0.0\%\\
1 & 8 & 6 & 3 & 25 & +0.137 \scriptsize[+0.136,\,+0.139] & -0.215 \scriptsize[-0.230,\,-0.200] & -0.228 \scriptsize[-0.232,\,-0.225] & -0.172 \scriptsize[-0.174,\,-0.169] & 0.287 & 0.399 & 0.0\%\\
1 & 8 & 8 & 3 & 25 & +0.139 \scriptsize[+0.137,\,+0.140] & -0.246 \scriptsize[-0.267,\,-0.226] & -0.228 \scriptsize[-0.232,\,-0.225] & -0.172 \scriptsize[-0.174,\,-0.169] & 0.287 & 0.399 & 0.0\%\\
1 & 32 & 4 & 3 & 25 & +0.287 \scriptsize[+0.278,\,+0.296] & +0.011 \scriptsize[+0.002,\,+0.020] & -0.081 \scriptsize[-0.083,\,-0.079] & -0.204 \scriptsize[-0.207,\,-0.201] & 0.392 & 0.399 & 0.0\%\\
2 & 0 & 4 & 3 & 25 & -0.000 \scriptsize[-0.000,\,+0.000] & -0.016 \scriptsize[-0.019,\,-0.014] & -0.624 \scriptsize[-0.630,\,-0.617] & +0.188 \scriptsize[+0.186,\,+0.190] & 0.159 & 0.689 & 6.4\%\\
2 & 1 & 4 & 3 & 25 & +0.049 \scriptsize[+0.049,\,+0.049] & +0.021 \scriptsize[+0.017,\,+0.024] & -0.573 \scriptsize[-0.580,\,-0.566] & +0.193 \scriptsize[+0.190,\,+0.195] & 0.199 & 0.689 & 6.4\%\\
2 & 8 & 4 & 3 & 25 & +0.260 \scriptsize[+0.256,\,+0.263] & -0.177 \scriptsize[-0.188,\,-0.166] & -0.344 \scriptsize[-0.350,\,-0.338] & -0.182 \scriptsize[-0.185,\,-0.179] & 0.390 & 0.689 & 6.4\%\\
2 & 32 & 4 & 3 & 25 & +0.401 \scriptsize[+0.391,\,+0.410] & +0.096 \scriptsize[+0.081,\,+0.111] & -0.189 \scriptsize[-0.195,\,-0.182] & -0.195 \scriptsize[-0.199,\,-0.191] & 0.543 & 0.689 & 6.4\%\\
\bottomrule
\end{tabular}

%% file: tab_groupwise.tex
\begin{tabular}{@{}llllllll@{}}
\toprule
Setting & Arm & \(\Delta|\delta_T|\) & \(\Delta\overline k_{T,0}\) & \(\Delta\overline k_{T,1}\) & \(\Delta\delta^k_T\) & \(\Delta\overline d_{T,0}\) & \(\Delta\overline d_{T,1}\)\\
\midrule
\(\gamma=0.5\), \(\beta_{\mathrm{ind}}=8\) & Projected MPC & +0.068 \scriptsize[+0.067,\,+0.069] & +0.045 \scriptsize[+0.034,\,+0.056] & -0.303 \scriptsize[-0.316,\,-0.289] & -0.348 \scriptsize[-0.353,\,-0.343] & +0.009 \scriptsize[+0.007,\,+0.012] & -0.059 \scriptsize[-0.062,\,-0.056]\\
 & Gap-penalty MPC & +0.070 \scriptsize[+0.069,\,+0.071] & +0.148 \scriptsize[+0.146,\,+0.151] & -0.199 \scriptsize[-0.201,\,-0.196] & -0.347 \scriptsize[-0.350,\,-0.344] & +0.031 \scriptsize[+0.030,\,+0.031] & -0.039 \scriptsize[-0.040,\,-0.039]\\
 & Gap-constrained MPC & -0.132 \scriptsize[-0.134,\,-0.129] & -0.356 \scriptsize[-0.362,\,-0.349] & +0.257 \scriptsize[+0.255,\,+0.260] & +0.613 \scriptsize[+0.605,\,+0.621] & -0.070 \scriptsize[-0.072,\,-0.069] & +0.062 \scriptsize[+0.060,\,+0.063]\\
\addlinespace
\(\gamma=1\), \(\beta_{\mathrm{ind}}=8\) & Projected MPC & +0.136 \scriptsize[+0.135,\,+0.138] & +0.186 \scriptsize[+0.176,\,+0.197] & -0.502 \scriptsize[-0.513,\,-0.491] & -0.688 \scriptsize[-0.693,\,-0.683] & +0.037 \scriptsize[+0.035,\,+0.040] & -0.099 \scriptsize[-0.101,\,-0.097]\\
 & Gap-penalty MPC & +0.142 \scriptsize[+0.142,\,+0.143] & +0.133 \scriptsize[+0.130,\,+0.136] & -0.581 \scriptsize[-0.584,\,-0.579] & -0.714 \scriptsize[-0.718,\,-0.710] & +0.027 \scriptsize[+0.026,\,+0.028] & -0.115 \scriptsize[-0.116,\,-0.115]\\
 & Gap-constrained MPC & -0.228 \scriptsize[-0.232,\,-0.225] & -0.564 \scriptsize[-0.568,\,-0.560] & +0.221 \scriptsize[+0.218,\,+0.224] & +0.785 \scriptsize[+0.780,\,+0.789] & -0.158 \scriptsize[-0.160,\,-0.156] & +0.070 \scriptsize[+0.068,\,+0.072]\\
\addlinespace
\(\gamma=2\), \(\beta_{\mathrm{ind}}=8\) & Projected MPC & +0.260 \scriptsize[+0.256,\,+0.263] & +0.428 \scriptsize[+0.414,\,+0.441] & -0.782 \scriptsize[-0.794,\,-0.771] & -1.210 \scriptsize[-1.223,\,-1.198] & +0.092 \scriptsize[+0.088,\,+0.096] & -0.167 \scriptsize[-0.172,\,-0.163]\\
 & Gap-penalty MPC & +0.276 \scriptsize[+0.275,\,+0.278] & +0.097 \scriptsize[+0.094,\,+0.100] & -0.845 \scriptsize[-0.852,\,-0.839] & -0.943 \scriptsize[-0.950,\,-0.935] & +0.020 \scriptsize[+0.019,\,+0.021] & -0.256 \scriptsize[-0.258,\,-0.255]\\
 & Gap-constrained MPC & -0.344 \scriptsize[-0.350,\,-0.338] & -0.503 \scriptsize[-0.507,\,-0.499] & +0.139 \scriptsize[+0.135,\,+0.144] & +0.642 \scriptsize[+0.635,\,+0.649] & -0.292 \scriptsize[-0.296,\,-0.289] & +0.052 \scriptsize[+0.049,\,+0.054]\\
\addlinespace
\(\gamma=-1\), \(\beta_{\mathrm{ind}}=8\) & Projected MPC & +0.136 \scriptsize[+0.135,\,+0.137] & -0.169 \scriptsize[-0.178,\,-0.160] & +0.463 \scriptsize[+0.456,\,+0.471] & +0.632 \scriptsize[+0.628,\,+0.637] & -0.031 \scriptsize[-0.033,\,-0.030] & +0.104 \scriptsize[+0.102,\,+0.106]\\
 & Gap-penalty MPC & +0.144 \scriptsize[+0.143,\,+0.145] & +0.217 \scriptsize[+0.214,\,+0.220] & +0.570 \scriptsize[+0.567,\,+0.574] & +0.354 \scriptsize[+0.350,\,+0.357] & +0.045 \scriptsize[+0.044,\,+0.045] & +0.189 \scriptsize[+0.188,\,+0.189]\\
 & Gap-constrained MPC & -0.220 \scriptsize[-0.222,\,-0.217] & +0.391 \scriptsize[+0.387,\,+0.395] & +0.012 \scriptsize[+0.004,\,+0.019] & -0.380 \scriptsize[-0.386,\,-0.373] & +0.226 \scriptsize[+0.225,\,+0.228] & +0.007 \scriptsize[+0.005,\,+0.009]\\
\addlinespace
\(\gamma=1\), \(\beta_{\mathrm{ind}}=0\) & Projected MPC & -0.000 \scriptsize[-0.000,\,+0.000] & -0.086 \scriptsize[-0.094,\,-0.079] & -0.072 \scriptsize[-0.078,\,-0.067] & +0.014 \scriptsize[+0.012,\,+0.016] & -0.017 \scriptsize[-0.018,\,-0.015] & -0.017 \scriptsize[-0.018,\,-0.015]\\
 & Gap-penalty MPC & +0.000 \scriptsize[+0.000,\,+0.000] & +0.000 \scriptsize[+0.000,\,+0.000] & +0.000 \scriptsize[+0.000,\,+0.000] & +0.000 \scriptsize[+0.000,\,+0.000] & +0.000 \scriptsize[+0.000,\,+0.000] & +0.000 \scriptsize[+0.000,\,+0.000]\\
 & Gap-constrained MPC & -0.372 \scriptsize[-0.376,\,-0.368] & -0.701 \scriptsize[-0.704,\,-0.698] & +0.806 \scriptsize[+0.803,\,+0.809] & +1.507 \scriptsize[+1.503,\,+1.510] & -0.186 \scriptsize[-0.188,\,-0.184] & +0.186 \scriptsize[+0.184,\,+0.188]\\
\addlinespace
\(\gamma=1\), \(\beta_{\mathrm{ind}}=32\) & Projected MPC & +0.287 \scriptsize[+0.278,\,+0.296] & +0.512 \scriptsize[+0.497,\,+0.526] & -0.490 \scriptsize[-0.502,\,-0.477] & -1.001 \scriptsize[-1.021,\,-0.981] & +0.116 \scriptsize[+0.109,\,+0.124] & -0.171 \scriptsize[-0.174,\,-0.167]\\
 & Gap-penalty MPC & +0.284 \scriptsize[+0.281,\,+0.288] & +0.284 \scriptsize[+0.279,\,+0.289] & -0.798 \scriptsize[-0.800,\,-0.795] & -1.081 \scriptsize[-1.087,\,-1.076] & +0.058 \scriptsize[+0.057,\,+0.059] & -0.227 \scriptsize[-0.230,\,-0.224]\\
 & Gap-constrained MPC & -0.081 \scriptsize[-0.083,\,-0.079] & -0.401 \scriptsize[-0.406,\,-0.395] & -0.007 \scriptsize[-0.008,\,-0.006] & +0.394 \scriptsize[+0.388,\,+0.399] & -0.125 \scriptsize[-0.127,\,-0.123] & -0.044 \scriptsize[-0.045,\,-0.042]\\
\addlinespace
\(\gamma=1\), \(\beta_{\mathrm{ind}}=8\), \(T=50\) & Projected MPC & +0.155 \scriptsize[+0.153,\,+0.156] & +0.205 \scriptsize[+0.193,\,+0.217] & -0.608 \scriptsize[-0.621,\,-0.596] & -0.814 \scriptsize[-0.821,\,-0.807] & +0.041 \scriptsize[+0.038,\,+0.044] & -0.114 \scriptsize[-0.116,\,-0.112]\\
 & Gap-penalty MPC & +0.164 \scriptsize[+0.163,\,+0.165] & +0.167 \scriptsize[+0.164,\,+0.170] & -0.695 \scriptsize[-0.698,\,-0.691] & -0.862 \scriptsize[-0.866,\,-0.857] & +0.032 \scriptsize[+0.032,\,+0.033] & -0.132 \scriptsize[-0.132,\,-0.131]\\
 & Gap-constrained MPC & -0.235 \scriptsize[-0.239,\,-0.232] & -0.573 \scriptsize[-0.577,\,-0.569] & +0.120 \scriptsize[+0.117,\,+0.123] & +0.693 \scriptsize[+0.689,\,+0.697] & -0.167 \scriptsize[-0.169,\,-0.165] & +0.068 \scriptsize[+0.066,\,+0.069]\\
\bottomrule
\end{tabular}

%% file: journal_conformal_diagnostics.tex
\begin{tikzpicture}[font=\small]
\definecolor{journalgray}{HTML}{555555}
\definecolor{journalorange}{HTML}{D55E00}
\definecolor{journalblue}{HTML}{0072B2}
\definecolor{journalgreen}{HTML}{009E73}
\begin{groupplot}[
group style={group size=2 by 1,horizontal sep=1.65cm,vertical sep=2.3cm},
width=7.4cm,height=6.1cm,
axis lines=left,grid=major,grid style={black!12},tick align=outside,scaled ticks=false,
legend style={draw=none,fill=white,font=\scriptsize},legend cell align=left]
\nextgroupplot[title={(a) Absolute prediction errors},xlabel={Score},ylabel={Empirical CDF},ymin=0,ymax=1.03,legend pos=north west]
\addplot[const plot,mark=none,line width=1pt,journalblue] coordinates {
(0.42704839710016329,0)
(0.42704839710016329,0.01)
(0.43950835715007636,0.02)
(0.43963051273192044,0.029999999999999999)
(0.43973542790684528,0.040000000000000001)
(0.43987924468013978,0.050000000000000003)
(0.44029637779007397,0.059999999999999998)
(0.4410831105792658,0.070000000000000007)
(0.44197644770877681,0.080000000000000002)
(0.44257741139207757,0.089999999999999997)
(0.44327100090201443,0.10000000000000001)
(0.44371565779084754,0.11)
(0.44376604375488704,0.12)
(0.44456103488978904,0.13)
(0.44470308891231125,0.14000000000000001)
(0.44495146974770894,0.14999999999999999)
(0.44554696074434041,0.16)
(0.44587967144507307,0.17000000000000001)
(0.44602957269628407,0.17999999999999999)
(0.44678779745927411,0.19)
(0.44694572353309969,0.20000000000000001)
(0.44758193413701042,0.20999999999999999)
(0.44771774377939977,0.22)
(0.44772238608413806,0.23000000000000001)
(0.44779668238709985,0.23999999999999999)
(0.4486103728715487,0.25)
(0.44873924404513238,0.26000000000000001)
(0.44876997230421911,0.27000000000000002)
(0.44929279472470102,0.28000000000000003)
(0.44935437987157867,0.28999999999999998)
(0.44938074215969659,0.29999999999999999)
(0.44966953637340484,0.31)
(0.44987352299149358,0.32000000000000001)
(0.45022974293878421,0.33000000000000002)
(0.45058096376967355,0.34000000000000002)
(0.45070135906020514,0.34999999999999998)
(0.45071863540911095,0.35999999999999999)
(0.45077241790808831,0.37)
(0.45111576999708664,0.38)
(0.4511734201878978,0.39000000000000001)
(0.45132309944563492,0.40000000000000002)
(0.45146629311367281,0.40999999999999998)
(0.45156095490500503,0.41999999999999998)
(0.45164215070452746,0.42999999999999999)
(0.45175410341347427,0.44)
(0.45211951412295476,0.45000000000000001)
(0.45228637845242292,0.46000000000000002)
(0.45249423982905723,0.46999999999999997)
(0.45300084628135401,0.47999999999999998)
(0.4531521999351647,0.48999999999999999)
(0.45335034233762839,0.5)
(0.45358226691199638,0.51000000000000001)
(0.45367776905376672,0.52000000000000002)
(0.45370595463277669,0.53000000000000003)
(0.4537965259909062,0.54000000000000004)
(0.45386708025439154,0.55000000000000004)
(0.45396280949341616,0.56000000000000005)
(0.45469406192044137,0.56999999999999995)
(0.4547075685964701,0.57999999999999996)
(0.45486807546081126,0.58999999999999997)
(0.45487648931309277,0.59999999999999998)
(0.45487832918334448,0.60999999999999999)
(0.4550661289322131,0.62)
(0.45507069191299138,0.63)
(0.45511837773012653,0.64000000000000001)
(0.45538317359830016,0.65000000000000002)
(0.4558036557606005,0.66000000000000003)
(0.45608474844836522,0.67000000000000004)
(0.45653582236356516,0.68000000000000005)
(0.45678203840250436,0.68999999999999995)
(0.45680992324484843,0.69999999999999996)
(0.45720824488886558,0.70999999999999996)
(0.45760816306276969,0.71999999999999997)
(0.45776923977129658,0.72999999999999998)
(0.45796342499132986,0.73999999999999999)
(0.45828083762876309,0.75)
(0.45835283819005507,0.76000000000000001)
(0.45851518128856977,0.77000000000000002)
(0.45858167685623119,0.78000000000000003)
(0.45885439067419348,0.79000000000000004)
(0.45893980174697346,0.80000000000000004)
(0.45948410022897201,0.81000000000000005)
(0.45965032836025133,0.81999999999999995)
(0.4603729201347953,0.82999999999999996)
(0.46046699653251133,0.83999999999999997)
(0.46049384490437029,0.84999999999999998)
(0.46066941711058118,0.85999999999999999)
(0.46075595621606613,0.87)
(0.46260194038567681,0.88)
(0.46287081580732203,0.89000000000000001)
(0.46360113906091033,0.90000000000000002)
(0.46390977954688517,0.91000000000000003)
(0.46574434074659471,0.92000000000000004)
(0.46575015506641365,0.93000000000000005)
(0.46685968906628195,0.93999999999999995)
(0.46716303727114777,0.94999999999999996)
(0.46779824936163961,0.95999999999999996)
(0.46810778625900074,0.96999999999999997)
(0.46822343928034366,0.97999999999999998)
(0.47008604441081919,0.98999999999999999)
(0.47250294671225052,1)
};
\addlegendentry{Calibration}
\addplot[const plot,mark=none,line width=1pt,journalorange,dashed] coordinates {
(0.43485794834916008,0)
(0.43485794834916008,0.01)
(0.43506632704756665,0.02)
(0.4355358177395483,0.029999999999999999)
(0.43569342390814841,0.040000000000000001)
(0.43725448002032374,0.050000000000000003)
(0.43775815240544957,0.059999999999999998)
(0.43786154905522501,0.070000000000000007)
(0.43896305511427403,0.080000000000000002)
(0.43996628585792097,0.089999999999999997)
(0.44126474877139948,0.10000000000000001)
(0.44240150025977137,0.11)
(0.44262105290634918,0.12)
(0.44270365486451185,0.13)
(0.44313936355495115,0.14000000000000001)
(0.44391824005480479,0.14999999999999999)
(0.44415912500118648,0.16)
(0.44424644741992503,0.17000000000000001)
(0.44457114022774602,0.17999999999999999)
(0.44488827241165785,0.19)
(0.44559264161926515,0.20000000000000001)
(0.44562530766741759,0.20999999999999999)
(0.44579543262998778,0.22)
(0.4463526824715735,0.23000000000000001)
(0.44637683678839973,0.23999999999999999)
(0.44652611410029841,0.25)
(0.44672155042498446,0.26000000000000001)
(0.44704041835676711,0.27000000000000002)
(0.44755410536772661,0.28000000000000003)
(0.44766807990843682,0.28999999999999998)
(0.4476749073692356,0.29999999999999999)
(0.44799632731138594,0.31)
(0.44867143917676539,0.32000000000000001)
(0.44877254377571069,0.33000000000000002)
(0.44877972823024942,0.34000000000000002)
(0.44916228556322513,0.34999999999999998)
(0.44964328841097478,0.35999999999999999)
(0.45050672657753144,0.37)
(0.4505481360637672,0.38)
(0.45090737925391505,0.39000000000000001)
(0.45104985316440405,0.40000000000000002)
(0.45106543311825142,0.40999999999999998)
(0.45121620338269586,0.41999999999999998)
(0.45125845960993072,0.42999999999999999)
(0.45142979512452536,0.44)
(0.45182105246706716,0.45000000000000001)
(0.45190973334477957,0.46000000000000002)
(0.45206377296447575,0.46999999999999997)
(0.45234364859387988,0.47999999999999998)
(0.45258106314657731,0.48999999999999999)
(0.45288572425438239,0.5)
(0.45291355165559277,0.51000000000000001)
(0.45312362823927022,0.52000000000000002)
(0.45319196779603554,0.53000000000000003)
(0.4539022900561383,0.54000000000000004)
(0.45394294073519642,0.55000000000000004)
(0.45401649022888829,0.56000000000000005)
(0.45406051709136019,0.56999999999999995)
(0.45412782583236222,0.57999999999999996)
(0.45443798630154908,0.58999999999999997)
(0.45444580364815063,0.59999999999999998)
(0.45502457767279203,0.60999999999999999)
(0.45509441221891656,0.62)
(0.45519240792854621,0.63)
(0.45521441719132283,0.64000000000000001)
(0.45542518159632006,0.65000000000000002)
(0.45591627264656759,0.66000000000000003)
(0.45631555737081425,0.67000000000000004)
(0.45635314282938211,0.68000000000000005)
(0.45680022527056768,0.68999999999999995)
(0.45688006720264396,0.69999999999999996)
(0.45693076100412422,0.70999999999999996)
(0.45693492368563737,0.71999999999999997)
(0.45704642999709799,0.72999999999999998)
(0.45720866622072115,0.73999999999999999)
(0.45722035378653841,0.75)
(0.45778492170455054,0.76000000000000001)
(0.45787103546476904,0.77000000000000002)
(0.4580846369424984,0.78000000000000003)
(0.45851600379365431,0.79000000000000004)
(0.45859814854485781,0.80000000000000004)
(0.45886431236740666,0.81000000000000005)
(0.46036500803800595,0.81999999999999995)
(0.46076387785488426,0.82999999999999996)
(0.46114563389403618,0.83999999999999997)
(0.46118529503633787,0.84999999999999998)
(0.46141315811538153,0.85999999999999999)
(0.46193268937194631,0.87)
(0.46233194692354745,0.88)
(0.46244225813862028,0.89000000000000001)
(0.46258666980054075,0.90000000000000002)
(0.46317994987006239,0.91000000000000003)
(0.46373436917814748,0.92000000000000004)
(0.46466624742272455,0.93000000000000005)
(0.46487558839183207,0.93999999999999995)
(0.46538637107799552,0.94999999999999996)
(0.46585754547801872,0.95999999999999996)
(0.46636688924397696,0.96999999999999997)
(0.46736370694997026,0.97999999999999998)
(0.46864678132154375,0.98999999999999999)
(0.47059136872004081,1)
};
\addlegendentry{Test}
\addplot[mark=none,journalgray,densely dotted] coordinates {
(0.46779824936163961,0)
(0.46779824936163961,1)
};
\addlegendentry{Radius $\widehat q$}
\nextgroupplot[title={(b) All held-out cohorts},xlabel={Cohort index (not time)},ylabel={Disparity contrast},xmin=-2,xmax=101,ymin=-0.52,ymax=0.52,legend pos=north east]
\addplot[only marks,mark=none,journalblue!25,forget plot,error bars/.cd,y dir=both,y explicit,error bar style={line width=0.25pt}] coordinates {
(0,-1.5114766465457175e-05) +- (0,0.46779824936163961)
(1,1.8390637412296629e-05) +- (0,0.46779824936163961)
(2,4.4323861913220242e-05) +- (0,0.46779824936163961)
(3,6.3842108289663413e-05) +- (0,0.46779824936163961)
(4,-1.1242748184439599e-05) +- (0,0.46779824936163961)
(5,-6.6676696183709527e-05) +- (0,0.46779824936163961)
(6,4.0134250341217204e-05) +- (0,0.46779824936163961)
(7,-2.6768582762293214e-05) +- (0,0.46779824936163961)
(8,1.1594146262408622e-05) +- (0,0.46779824936163961)
(9,1.855575477315019e-05) +- (0,0.46779824936163961)
(10,-8.2113956566112756e-05) +- (0,0.46779824936163961)
(11,2.2588839271336925e-05) +- (0,0.46779824936163961)
(12,-2.0997010460682697e-05) +- (0,0.46779824936163961)
(13,-4.7960809322414311e-05) +- (0,0.46779824936163961)
(14,3.6589011325846577e-05) +- (0,0.46779824936163961)
(15,-7.5978811586302486e-06) +- (0,0.46779824936163961)
(16,3.3462709007142655e-05) +- (0,0.46779824936163961)
(17,-1.9221229327431866e-05) +- (0,0.46779824936163961)
(18,-1.5940482777240805e-05) +- (0,0.46779824936163961)
(19,-1.1224901476966642e-05) +- (0,0.46779824936163961)
(20,-3.6038463626680839e-05) +- (0,0.46779824936163961)
(21,-4.4312392271594891e-05) +- (0,0.46779824936163961)
(22,-2.8388302157206224e-05) +- (0,0.46779824936163961)
(23,3.1314566643111941e-05) +- (0,0.46779824936163961)
(24,-1.1158133773364609e-05) +- (0,0.46779824936163961)
(25,1.1387423437952959e-06) +- (0,0.46779824936163961)
(26,5.8437923455209595e-05) +- (0,0.46779824936163961)
(27,-1.7333201967895118e-05) +- (0,0.46779824936163961)
(28,2.05691636902261e-06) +- (0,0.46779824936163961)
(29,3.7027008234533088e-05) +- (0,0.46779824936163961)
(30,3.6958462483471788e-06) +- (0,0.46779824936163961)
(31,-8.0368534880609355e-06) +- (0,0.46779824936163961)
(32,-2.2121144476985877e-05) +- (0,0.46779824936163961)
(33,-7.0869672418080731e-05) +- (0,0.46779824936163961)
(34,-3.3925967723669181e-05) +- (0,0.46779824936163961)
(35,-4.2969875313145223e-05) +- (0,0.46779824936163961)
(36,4.0797132991043927e-06) +- (0,0.46779824936163961)
(37,-2.4306240462179662e-05) +- (0,0.46779824936163961)
(38,2.9883971310665225e-05) +- (0,0.46779824936163961)
(39,-1.1149149226147292e-05) +- (0,0.46779824936163961)
(40,5.4584386103589955e-05) +- (0,0.46779824936163961)
(41,-1.3782643961235408e-05) +- (0,0.46779824936163961)
(42,4.1083594602031592e-05) +- (0,0.46779824936163961)
(43,-5.1660150830190599e-06) +- (0,0.46779824936163961)
(44,1.883830039574319e-05) +- (0,0.46779824936163961)
(45,3.8489772503091935e-06) +- (0,0.46779824936163961)
(46,1.954459792719021e-05) +- (0,0.46779824936163961)
(47,4.4521206557734949e-05) +- (0,0.46779824936163961)
(48,1.3381449016733327e-05) +- (0,0.46779824936163961)
(49,-5.9333797298180557e-05) +- (0,0.46779824936163961)
(50,1.0097869021188921e-05) +- (0,0.46779824936163961)
(51,5.0434722125236586e-05) +- (0,0.46779824936163961)
(52,-4.0542258915601814e-05) +- (0,0.46779824936163961)
(53,2.2721634498498733e-05) +- (0,0.46779824936163961)
(54,-1.5576000422803937e-05) +- (0,0.46779824936163961)
(55,1.7202648535788456e-05) +- (0,0.46779824936163961)
(56,-4.158803445258874e-06) +- (0,0.46779824936163961)
(57,2.2684952921347978e-05) +- (0,0.46779824936163961)
(58,1.143053046669118e-05) +- (0,0.46779824936163961)
(59,-2.6008009287675771e-05) +- (0,0.46779824936163961)
(60,-2.7997303628235493e-08) +- (0,0.46779824936163961)
(61,-3.7649508673284759e-05) +- (0,0.46779824936163961)
(62,-3.0662936581590805e-05) +- (0,0.46779824936163961)
(63,3.8858133605254586e-05) +- (0,0.46779824936163961)
(64,4.2049474665253883e-05) +- (0,0.46779824936163961)
(65,-3.1345149687223017e-06) +- (0,0.46779824936163961)
(66,1.0565315541610776e-05) +- (0,0.46779824936163961)
(67,8.3249318244826842e-06) +- (0,0.46779824936163961)
(68,5.3644420817539062e-06) +- (0,0.46779824936163961)
(69,3.0045364464187374e-05) +- (0,0.46779824936163961)
(70,1.9016791791698351e-06) +- (0,0.46779824936163961)
(71,4.6650276776125099e-05) +- (0,0.46779824936163961)
(72,6.4108475991697667e-06) +- (0,0.46779824936163961)
(73,1.2888514587319766e-05) +- (0,0.46779824936163961)
(74,-1.5558794205644544e-06) +- (0,0.46779824936163961)
(75,-1.4210366808833919e-05) +- (0,0.46779824936163961)
(76,2.0672734690266314e-05) +- (0,0.46779824936163961)
(77,-7.3068167125017736e-06) +- (0,0.46779824936163961)
(78,2.9927634257187474e-05) +- (0,0.46779824936163961)
(79,8.0789162023170435e-06) +- (0,0.46779824936163961)
(80,-4.8483859137013474e-05) +- (0,0.46779824936163961)
(81,7.2948058675276206e-05) +- (0,0.46779824936163961)
(82,-4.9956389781002919e-06) +- (0,0.46779824936163961)
(83,5.1748671263021628e-05) +- (0,0.46779824936163961)
(84,2.5595323580610008e-05) +- (0,0.46779824936163961)
(85,-1.9335216364976815e-05) +- (0,0.46779824936163961)
(86,5.135675959175301e-05) +- (0,0.46779824936163961)
(87,-1.5159737358813596e-05) +- (0,0.46779824936163961)
(88,6.1085071616395759e-05) +- (0,0.46779824936163961)
(89,-2.9033060865990556e-05) +- (0,0.46779824936163961)
(90,3.0451080021004495e-06) +- (0,0.46779824936163961)
(91,-4.0885754137007235e-05) +- (0,0.46779824936163961)
(92,2.0976095148700358e-05) +- (0,0.46779824936163961)
(93,-4.0386811247891674e-05) +- (0,0.46779824936163961)
(94,-2.2118983066518605e-05) +- (0,0.46779824936163961)
(95,3.8787516145361023e-05) +- (0,0.46779824936163961)
(96,4.095258927583334e-06) +- (0,0.46779824936163961)
(97,2.2689563685299174e-05) +- (0,0.46779824936163961)
(98,-7.5319003299545439e-07) +- (0,0.46779824936163961)
(99,3.8948034505909002e-05) +- (0,0.46779824936163961)
};
\addplot[mark=none,journalblue,line width=1pt] coordinates {
(0,-1.5114766465457175e-05)
(1,1.8390637412296629e-05)
(2,4.4323861913220242e-05)
(3,6.3842108289663413e-05)
(4,-1.1242748184439599e-05)
(5,-6.6676696183709527e-05)
(6,4.0134250341217204e-05)
(7,-2.6768582762293214e-05)
(8,1.1594146262408622e-05)
(9,1.855575477315019e-05)
(10,-8.2113956566112756e-05)
(11,2.2588839271336925e-05)
(12,-2.0997010460682697e-05)
(13,-4.7960809322414311e-05)
(14,3.6589011325846577e-05)
(15,-7.5978811586302486e-06)
(16,3.3462709007142655e-05)
(17,-1.9221229327431866e-05)
(18,-1.5940482777240805e-05)
(19,-1.1224901476966642e-05)
(20,-3.6038463626680839e-05)
(21,-4.4312392271594891e-05)
(22,-2.8388302157206224e-05)
(23,3.1314566643111941e-05)
(24,-1.1158133773364609e-05)
(25,1.1387423437952959e-06)
(26,5.8437923455209595e-05)
(27,-1.7333201967895118e-05)
(28,2.05691636902261e-06)
(29,3.7027008234533088e-05)
(30,3.6958462483471788e-06)
(31,-8.0368534880609355e-06)
(32,-2.2121144476985877e-05)
(33,-7.0869672418080731e-05)
(34,-3.3925967723669181e-05)
(35,-4.2969875313145223e-05)
(36,4.0797132991043927e-06)
(37,-2.4306240462179662e-05)
(38,2.9883971310665225e-05)
(39,-1.1149149226147292e-05)
(40,5.4584386103589955e-05)
(41,-1.3782643961235408e-05)
(42,4.1083594602031592e-05)
(43,-5.1660150830190599e-06)
(44,1.883830039574319e-05)
(45,3.8489772503091935e-06)
(46,1.954459792719021e-05)
(47,4.4521206557734949e-05)
(48,1.3381449016733327e-05)
(49,-5.9333797298180557e-05)
(50,1.0097869021188921e-05)
(51,5.0434722125236586e-05)
(52,-4.0542258915601814e-05)
(53,2.2721634498498733e-05)
(54,-1.5576000422803937e-05)
(55,1.7202648535788456e-05)
(56,-4.158803445258874e-06)
(57,2.2684952921347978e-05)
(58,1.143053046669118e-05)
(59,-2.6008009287675771e-05)
(60,-2.7997303628235493e-08)
(61,-3.7649508673284759e-05)
(62,-3.0662936581590805e-05)
(63,3.8858133605254586e-05)
(64,4.2049474665253883e-05)
(65,-3.1345149687223017e-06)
(66,1.0565315541610776e-05)
(67,8.3249318244826842e-06)
(68,5.3644420817539062e-06)
(69,3.0045364464187374e-05)
(70,1.9016791791698351e-06)
(71,4.6650276776125099e-05)
(72,6.4108475991697667e-06)
(73,1.2888514587319766e-05)
(74,-1.5558794205644544e-06)
(75,-1.4210366808833919e-05)
(76,2.0672734690266314e-05)
(77,-7.3068167125017736e-06)
(78,2.9927634257187474e-05)
(79,8.0789162023170435e-06)
(80,-4.8483859137013474e-05)
(81,7.2948058675276206e-05)
(82,-4.9956389781002919e-06)
(83,5.1748671263021628e-05)
(84,2.5595323580610008e-05)
(85,-1.9335216364976815e-05)
(86,5.135675959175301e-05)
(87,-1.5159737358813596e-05)
(88,6.1085071616395759e-05)
(89,-2.9033060865990556e-05)
(90,3.0451080021004495e-06)
(91,-4.0885754137007235e-05)
(92,2.0976095148700358e-05)
(93,-4.0386811247891674e-05)
(94,-2.2118983066518605e-05)
(95,3.8787516145361023e-05)
(96,4.095258927583334e-06)
(97,2.2689563685299174e-05)
(98,-7.5319003299545439e-07)
(99,3.8948034505909002e-05)
};
\addlegendentry{Predicted}
\addplot[only marks,mark=*,mark size=1.1pt,journalgray] coordinates {
(0,-0.45887942713387214)
(1,-0.4658391548406064)
(2,-0.44122042490948626)
(3,-0.45687108157734774)
(4,-0.4611965377845223)
(5,-0.45785159840073425)
(6,-0.45086724500357384)
(7,-0.46079064643764656)
(8,-0.44915069141696273)
(9,-0.44765635161446243)
(10,-0.44886184218681552)
(11,-0.46485299955256076)
(12,-0.43555681475000896)
(13,-0.45293368506370479)
(14,-0.44873595476438483)
(15,-0.46737130483112888)
(16,-0.45440452359254191)
(17,-0.44561186284859255)
(18,-0.44636862295435076)
(19,-0.45632678227229123)
(20,-0.44967932687460144)
(21,-0.44642114918067133)
(22,-0.44655450240245564)
(23,-0.44310804898830802)
(24,-0.45852716192742765)
(25,-0.43896191637193022)
(26,-0.45137135720107013)
(27,-0.46116296709600407)
(28,-0.44424439050355602)
(29,-0.45517739018308828)
(30,-0.45542148575007169)
(31,-0.45402452708237634)
(32,-0.45690218834712093)
(33,-0.45729122345895651)
(34,-0.45109935908597509)
(35,-0.43780112228076273)
(36,-0.45311954852597108)
(37,-0.45447010988861281)
(38,-0.45051825209245655)
(39,-0.46467739657195073)
(40,-0.46238767375251671)
(41,-0.45809841958645964)
(42,-0.45587518905196556)
(43,-0.45258622916166036)
(44,-0.43483911004876435)
(45,-0.44391439107755448)
(46,-0.4456057630694904)
(47,-0.44795180610482821)
(48,-0.45858476709584106)
(49,-0.46544570487529369)
(50,-0.46316985200104122)
(51,-0.43781111433309977)
(53,-0.45784831383027053)
(54,-0.45694633700454701)
(55,-0.45404331444282442)
(56,-0.45291771045903806)
(57,-0.44864875422384404)
(58,-0.44578400209952107)
(59,-0.45193574135406728)
(60,-0.4427036828618155)
(61,-0.44243914976844467)
(62,-0.46039567097458756)
(63,-0.45230479046027461)
(64,-0.4372124305456585)
(65,-0.45121933789766455)
(66,-0.45678965995502607)
(67,-0.46140483318355702)
(68,-0.45704106555501622)
(69,-0.46370432381368332)
(70,-0.44671964874580528)
(71,-0.44484162213488171)
(72,-0.45518599708094704)
(73,-0.453889401541551)
(74,-0.44755566124714719)
(76,-0.44764740717374657)
(77,-0.44262835972306169)
(78,-0.45101992553014686)
(79,-0.45125038069372841)
(80,-0.440014769717058)
(81,-0.46629394118530171)
(82,-0.45413282147134032)
(83,-0.45314021912477254)
(84,-0.45048113125395084)
(85,-0.43508566226393164)
(86,-0.45176969570747538)
(87,-0.46234710666090628)
(88,-0.45496349260117563)
(89,-0.4470694514176331)
(90,-0.45720562111271906)
(91,-0.4442000107553235)
(92,-0.46256569370539202)
(93,-0.43573381071939632)
(94,-0.45208589194754228)
(95,-0.46189390185580093)
(96,-0.45393884547626884)
(97,-0.45507172265523127)
(98,-0.44457189341777903)
(99,-0.45631419479487623)
};
\addlegendentry{Realized: covered}
\addplot[only marks,mark=x,mark size=3pt,journalorange,thick] coordinates {
(52,-0.4706319109789564)
(75,-0.46866099168835257)
};
\addlegendentry{Realized: missed}
\end{groupplot}
\end{tikzpicture}